%% file: main-cikm.tex
\documentclass[sigconf]{acmart}

\pdfoutput=1
\usepackage{stackengine}
\usepackage{subcaption}
\usepackage{booktabs}
\usepackage{xspace}
\usepackage{balance}
\usepackage{array}
\usepackage{multirow}
\usepackage{bbm}
\usepackage{amsmath}
\usepackage{cleveref} 
\usepackage{url}
\usepackage{amsfonts}
\usepackage[utf8]{inputenc}
\usepackage{graphicx}
\usepackage{mathtools}

\usepackage{colortbl}
\usepackage{pgfplots}

\usepackage[export]{adjustbox}

\usepackage{float}
\usepackage{comment}
\usepackage{amsmath}
\usepackage{algorithmic}
\usepackage{graphicx}
\usepackage{textcomp}
\usepackage{xcolor}

\def\BibTeX{{\rm B\kern-.05em{\sc i\kern-.025em b}\kern-.08em
    T\kern-.1667em\lower.7ex\hbox{E}\kern-.125emX}}

\input{macros}

\copyrightyear{2023}
\acmYear{2023}
\setcopyright{acmlicensed}\acmConference[CIKM '23]{Proceedings of the 32nd ACM International Conference on Information and Knowledge Management}{October 21--25, 2023}{Birmingham, United Kingdom}
\acmBooktitle{Proceedings of the 32nd ACM International Conference on Information and Knowledge Management (CIKM '23), October 21--25, 2023, Birmingham, United Kingdom}
\acmPrice{15.00}
\acmDOI{10.1145/3583780.3615052}
\acmISBN{979-8-4007-0124-5/23/10}

\begin{document}

\title{Selecting Walk Schemes for Database Embedding}

\author{Yuval Lev Lubarsky}
\affiliation{%
  \institution{Technion}
  \city{Haifa}
  \country{Israel}
}
\email{lubarsky@cs.technion.ac.il}

\author{Jan T\"onshoff}
\affiliation{%
  \institution{RWTH Aachen University}
  \city{Aachen}
  \country{Germany}
}
\email{toenshoff@informatik.rwth-aachen.de}

\author{Martin Grohe}
\affiliation{%
  \institution{RWTH Aachen University}
  \city{Aachen}
  \country{Germany}
}
\email{grohe@informatik.rwth-aachen.de}

\author{Benny Kimelfeld}
\affiliation{%
  \institution{Technion}
  \city{Haifa}
  \country{Israel}
}
\email{bennyk@cs.technion.ac.il}




\begin{abstract}
Machinery for data analysis often requires a numeric representation of the input.
Towards that, a common practice is to embed components of structured data into a high-dimensional
vector space. We study the embedding of the tuples of a relational database, where existing techniques are often based on optimization tasks over a collection of random walks from the database. The focus of this paper is on the recent \forward algorithm
 that is designed for dynamic databases,  
where walks are sampled by following foreign keys between tuples.
Importantly,  different walks have different schemas, or ``walk schemes,'' that are derived by listing the relations and attributes along the walk. Also importantly, different walk schemes describe relationships of different natures in the database. 

We show that by focusing on a few informative walk schemes, we can obtain tuple embedding significantly faster, while retaining the quality. 
We define the problem of scheme selection for tuple embedding, devise
several approaches and strategies for scheme selection, and conduct a thorough empirical study of the performance over a collection of
downstream tasks. Our results confirm that with effective strategies
for scheme selection, we can obtain high-quality embeddings considerably
(e.g., three times) faster, preserve the extensibility to newly
inserted tuples, and even achieve an increase in the precision of some tasks.
\end{abstract}

%
%




\maketitle

\input{section_intro}


\input{section_preliminaries}

\input{section_problem_def}

\input{section_strategies}
\input{section_exp}

\input{section_conclusions}

\newpage



\bibliographystyle{ACM-Reference-Format}
\bibliography{bibliography}


\end{document}

%% file: macros.tex
\renewcommand{\phi}{\varphi}

\newcommand{\mgmargindone}[1]{}

\newcommand{\mginlinedone}[1]{}

\usepackage{mathtools,bm}
\newcommand{\defeq}{\mathrel{\vcentcolon=}}

\newcommand{\eat}[1]{}

\newcommand{\wordtovec}{\textsc{Word2Vec}\xspace}
\newcommand{\nodetovec}{\textsc{Node2Vec}\xspace}
\newcommand{\moltovec}{\textsc{Mol2Vec}}

\newcommand{\transe}{\textsc{TransE}}

\newcommand{\forward}{\textsc{FoRWaRD}\xspace}
\newcommand{\embdi}{\textsc{EmbDI}\xspace}

\def\tws{\mathsf{TWS}}

\renewcommand{\arraystretch}{1.1}

\DeclareMathOperator*{\mean}{mean}
\DeclareMathOperator*{\var}{\mathrm{Var}}

\def\scs{\sigma}
\def\e#1{\emph{#1}}
\def\key{\mathsf{key}}
\def\dom{\mathsf{dom}}
\def\adom{\mathsf{adom}}

\def\set#1{\mathord{\{#1\}}}

\def\attseq#1{\mathord{\boldsymbol{#1}}}

\def\W{\mathcal{W}}

\def\prob{\mathrm{Pr}}

\def\rel#1{\textit{\emph{\textsc{#1}}}}
\def\att#1{\textit{\emph{\textsf{#1}}}}
\def\val#1{\textit{\emph{\texttt{#1}}}}
\def\qed{\hfill$\Diamondblack$}

\newcommand{\authorx}[1]{}
\newcommand{\emailx}[1]{}
\newcommand{\affiliationx}[1]{}

\def\score{\mathord{\mathsf{score}}}

%% file: section_intro.tex
\section{Introduction}

Machine-learning algorithms are conventionally designed to generalize observations about numerical vectors, and hence, their application to non-numeric data requires \emph{embeddings} of these data into a numerical vector space. The embedding should faithfully reflect the semantics of data in the sense that 
similar entities are to be mapped to vectors that are close geometrically and vice versa. Instantiations of this practice include models like \wordtovec~\cite{miksutche+13} and BERT (Bidirectional Encoder Representations from Transformers)~\cite{DBLP:conf/naacl/DevlinCLT19} that map words (or tokens of words) in natural language~\cite{DBLP:conf/icml/LeM14}, \nodetovec that maps nodes of a graph~\cite{10.1145/2939672.2939754}, \transe~\cite{borusugar+13} that map entities of a knowledge graph, \moltovec~\cite{DBLP:journals/jcisd/JaegerFT18} that maps molecule structures, and \embdi~\cite{DBLP:conf/sigmod/CappuzzoPT20} and \forward~\cite{icde2023dynamic} that map database tuples. Database embeddings have enabled the deployment of machine-learning architectures to traditional database tasks such as record similarity~\cite{DBLP:journals/corr/abs-1712-07199, 10.1145/3076246.3076251, DBLP:conf/cidr/BordawekarS19, DBLP:conf/sigmod/Gunther18, DBLP:conf/edbt/0002TNL20}, record linking~\cite{DBLP:conf/sigmod/MudgalLRDPKDAR18,DBLP:journals/pvldb/EbraheemTJOT18} integration tasks such as schema, token and record matching (entity resolution)~\cite{DBLP:conf/sigmod/CappuzzoPT20}, and column prediction~\cite{icde2023dynamic}.
The embedded entities are typically either tuples or attribute values. 
Here we focus on tuple embeddings.


Embedding techniques are often based on the analysis of \e{sequences} obtained from the data. In word embedding, the data is naturally organized in sequences (e.g., sentences or sliding windows in the text)~\cite{miksutche+13,DBLP:conf/naacl/DevlinCLT19}; in node embedding, the sequences are paths obtained from random walks in the graph~\cite{10.1145/2939672.2939754}; and in tuple embedding, the sequences consist of database components (cells and tuples) that one can reach through natural joins~\cite{DBLP:conf/sigmod/CappuzzoPT20} or foreign-key references~\cite{icde2023dynamic}. The analysis is typically done by learning to predict masked parts of the sequence from other parts of the sequence~\cite{miksutche+13,DBLP:conf/naacl/DevlinCLT19,10.1145/2939672.2939754,DBLP:conf/sigmod/CappuzzoPT20}. We focus on \forward that is designed for producing stable embeddings in dynamic databases, as we explain next. \forward analyzes \e{walks}, which are sequences of tuples connected via foreign-key references, and it does so differently from masking. Roughly speaking, training aims for the distance between two (vector representations of) tuples to capture the distance between the \e{distributions} of values reachable from the tuples through foreign-key references, starting from the corresponding tuples. 
(We recall the exact definition of \forward in Section~\ref{sec:prelims}.)

The \forward algorithm has been designed to solve the \e{stable embedding problem}, where the goal is to infer the embedding of the new tuples \e{without recomputing the embedding over the entire database} and \e{without changing the embeddings of existing tuples} (that downstream tasks might already be using and rely on past decisions thereupon)~\cite{icde2023dynamic}. A technique for performing this task has been proposed along with the \forward framework~\cite{icde2023dynamic}.

\begin{figure}
\scalebox{0.8}{\input{story.pspdftex}}
\caption{\forward  vs.~\forward with scheme selection\label{fig:story}}
  \end{figure}
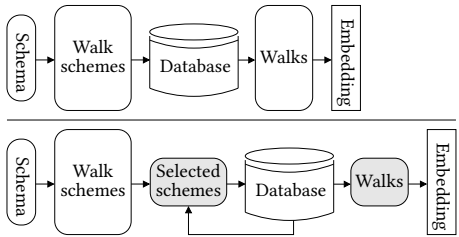

A walk in a database is naturally associated with a meta-data \e{pattern}, which is formed by taking the names of the relations of the tuples along the sequence, as well as the names of the attributes that are used for the  (outgoing and incoming) references. This pattern is called a \e{walk scheme}~\cite{icde2023dynamic}, and examples of these are depicted in Figure~\ref{fig:walks} in the context of a geographical database. Sequence-based embedding algorithms for other modalities do not encounter (and do not account for)  such meta-information in the training phase.
Our premise is that the walk scheme determines, to a large extent, the contribution of a walk to the quality of the learned embedding. Hence, unlike word and node embedding, in tuple embedding, we can introduce important a-priori bias over the training sequences.

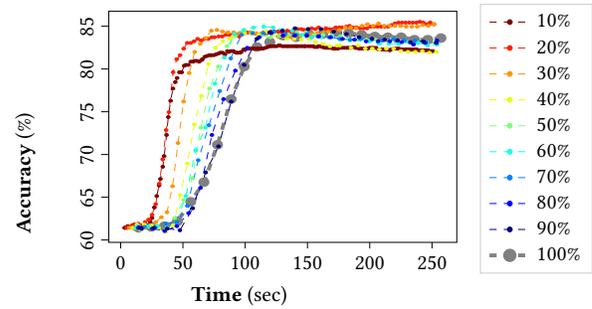
\begin{figure}
\ifx\QUICK\undefined
{\input{plot-example-mondial-religion-k-var}}
\else\QUICK
\fi
\vskip-1em
\caption{Religion prediction over the Mondial dataset with \forward and scheme selection via \e{kernel variance}. With a fifth of the walk schemes, we get to full equality in about one-third of the embedding time and eventually even outperform the embedding with the entire set of walk schemes.\label{fig:intro-chart}}
\end{figure}

We claim and prove empirically that one can considerably reduce the number of training walks by restricting the learning phase to the walks of the most effective walk schemes, with a mild (or no) reduction in quality. Moreover, the embedding quality might improve by filtering out walk schemes that contribute more noise than beneficial information. We devise techniques for the selection of walk schemes within \forward, as illustrated in Figure~\ref{fig:story}. 
To illustrate the importance of scheme selection, a sample of our experiments is shown in Figure~\ref{fig:intro-chart}. Here, we are using the learned embedding in the Mondial database~\cite{10.5555/2073876.2073934} to predict the religion of a country based on the database's information. Each curve corresponds to a selected percentage of the walk schemes and shows the quality of the prediction as a function of the embedding time (where each epoch contributes a point). The actual way of selecting the walk schemes is discussed in the next paragraphs. As the chart showed, selecting a fifth of the walk schemes fully preserves the quality
in about one-third of the embedding time and eventually even outperforms the embedding with the entire set of walk schemes.

The main question then is \e{how to select the best walk schemes for learning an embedding?} This is the challenge that we focus on in this paper. The goodness of a  selection method is reflected in two measures:
    \e{(1)} \textbf{Efficiency}---the choice should be considerably faster than the embedding itself, \e{and (2)} \textbf{Quality}---the choice should be such that we can select just a small number of walk schemes and retain the quality of the embedding (e.g., on downstream tasks). 
Regarding the efficiency measure, we already said that an important (but not the only) use case for walk-scheme selection is that the combined time it takes to select the walk schemes and then learn the embedding is considerably faster than learning the embedding on the initial full collection of walk schemes. As for the quality measure, if we are to select $\alpha\%$ of the walk schemes and have downstream success of $\beta\%$ compared to 100\% of the walk schemes, then we would like $\alpha$ to be as small as possible (e.g., $10$) and $\beta$ 
as large as possible (e.g., $95$). Actually, our experiments show cases where $\beta$ exceeds $100$ for the reason discussed above. 
Importantly, the quality of the selection strategy should also apply to the stable embedding problem in the dynamic setting.
In particular, we would like the performance of the learned embedding to reach (or even exceed) that of the full set of walk schemes when the embedding is extended to newly arriving tuples.

The choice can be made by \e{ranking} the walk schemes by some scoring function and selecting the top candidates. This scoring function can reflect properties that are hypothetically important to the embedding but do not require seeing in action the embedding algorithm, namely \forward; we refer to this approach as \e{\forward-less}. Alternatively, one can execute the embedding algorithm in some limited (light) manner and infer the walks from that execution (or, more precisely, from the internal state of the  model); we refer to this approach as \e{light training}. Finally, one can also eliminate walk schemes gradually \e{during} the embedding process, where in each epoch, we estimate the importance of a walk scheme (similarly to the way it is done in light training) and leave a strict subset for the next epochs, until we decide that none can be further eliminated; we refer to this approach as \e{online scheme elimination}. In summary, 
we devise and study strategies for selecting walk-schemes in three different approaches:
\e{(1)} \forward-less, \e{(2} light training, \e{and (3)} online scheme elimination.

In each approach, several different strategies can be proposed. In the \forward-less approach, we look at measures that inspect the probability distribution that one establishes by following random walks guided by the walk scheme (in addition to simple baselines such as eliminating the longest schemes and random schemes). The strategy that stands out is what we call \e{kernel variance}: what is the variance among the \e{differences} that one observes by selecting two random starting points for the walks? 
In the light-training approach, we look at two types of training restrictions: a single epoch (out of all epochs) and a full training on a \e{sample} of the database. In the online elimination approach, we apply the single-epoch selection (of light training) after every epoch.

In the experiments, we follow the convention of evaluating the embedding quality on \e{downstream tasks} and use column-prediction tasks in the bio-medical and geography domains. Our study (\Cref{sec:experiments}) has three parts. First, we test the performance of each selection strategy as a function of the time and number of selected schemes. Second, we conduct a comparison among the strategies in a technique that we devise. Third, we study the performance with the selected walk schemes in a dynamic setting. Our conclusion is that kernel-variance performs best.

In summary, our contributions are as follows.
First, we introduce the problem of \e{scheme selection} for tuple embedding. We do so in the context of \forward, yet the problem applies to every sequence-based database embedding. 
Second, we propose three approaches to scheme selection and devise several specific strategies within each approach.
Third, we conduct a thorough experimental evaluation that investigates the empirical effectiveness of the strategies, compares them, and studies their performance in the dynamic setting. 


%% file: story.pspdftex
\begin{picture}(0,0)%
\includegraphics{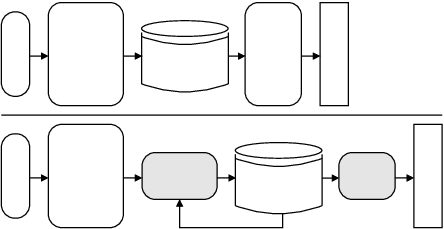}%
\end{picture}%
\setlength{\unitlength}{3947sp}%
\begingroup\makeatletter\ifx\SetFigFont\undefined%
\gdef\SetFigFont#1#2#3#4#5{%
  \reset@font\fontsize{#1}{#2pt}%
  \fontfamily{#3}\fontseries{#4}\fontshape{#5}%
  \selectfont}%
\fi\endgroup%
\begin{picture}(3549,1824)(1414,-1798)
\put(1501,-436){\rotatebox{270.0}{\makebox(0,0)[b]{\smash{{\SetFigFont{9}{10.8}{\familydefault}{\mddefault}{\updefault}{\color[rgb]{0,0,0}Schema}%
}}}}}
\put(2101,-361){\makebox(0,0)[b]{\smash{{\SetFigFont{9}{10.8}{\familydefault}{\mddefault}{\updefault}{\color[rgb]{0,0,0}Walk}%
}}}}
\put(2101,-511){\makebox(0,0)[b]{\smash{{\SetFigFont{9}{10.8}{\familydefault}{\mddefault}{\updefault}{\color[rgb]{0,0,0}schemes}%
}}}}
\put(2906,-510){\makebox(0,0)[b]{\smash{{\SetFigFont{9}{10.8}{\familydefault}{\mddefault}{\updefault}{\color[rgb]{0,0,0}Database}%
}}}}
\put(3601,-436){\makebox(0,0)[b]{\smash{{\SetFigFont{9}{10.8}{\familydefault}{\mddefault}{\updefault}{\color[rgb]{0,0,0}Walks}%
}}}}
\put(4051,-436){\rotatebox{270.0}{\makebox(0,0)[b]{\smash{{\SetFigFont{9}{10.8}{\familydefault}{\mddefault}{\updefault}{\color[rgb]{0,0,0}Embedding}%
}}}}}
\put(3656,-1485){\makebox(0,0)[b]{\smash{{\SetFigFont{9}{10.8}{\familydefault}{\mddefault}{\updefault}{\color[rgb]{0,0,0}Database}%
}}}}
\put(4351,-1411){\makebox(0,0)[b]{\smash{{\SetFigFont{9}{10.8}{\familydefault}{\mddefault}{\updefault}{\color[rgb]{0,0,0}Walks}%
}}}}
\put(4801,-1411){\rotatebox{270.0}{\makebox(0,0)[b]{\smash{{\SetFigFont{9}{10.8}{\familydefault}{\mddefault}{\updefault}{\color[rgb]{0,0,0}Embedding}%
}}}}}
\put(2101,-1336){\makebox(0,0)[b]{\smash{{\SetFigFont{9}{10.8}{\familydefault}{\mddefault}{\updefault}{\color[rgb]{0,0,0}Walk}%
}}}}
\put(2101,-1486){\makebox(0,0)[b]{\smash{{\SetFigFont{9}{10.8}{\familydefault}{\mddefault}{\updefault}{\color[rgb]{0,0,0}schemes}%
}}}}
\put(2851,-1336){\makebox(0,0)[b]{\smash{{\SetFigFont{9}{10.8}{\familydefault}{\mddefault}{\updefault}{\color[rgb]{0,0,0}Selected}%
}}}}
\put(2851,-1486){\makebox(0,0)[b]{\smash{{\SetFigFont{9}{10.8}{\familydefault}{\mddefault}{\updefault}{\color[rgb]{0,0,0}schemes}%
}}}}
\put(1501,-1411){\rotatebox{270.0}{\makebox(0,0)[b]{\smash{{\SetFigFont{9}{10.8}{\familydefault}{\mddefault}{\updefault}{\color[rgb]{0,0,0}Schema}%
}}}}}
\end{picture}%

%% file: plot-example-mondial-religion-k-var.tex
\begin{tikzpicture}[yscale=0.9, xscale=0.9]

\definecolor{blue}{RGB}{0,0,255}
\definecolor{cyan21255225}{RGB}{21,255,225}
\definecolor{darkgray176}{RGB}{176,176,176}
\definecolor{darkorange2551480}{RGB}{255,148,0}
\definecolor{dodgerblue0128255}{RGB}{0,128,255}
\definecolor{gray}{RGB}{128,128,128}
\definecolor{lightgray204}{RGB}{204,204,204}
\definecolor{lightgreen124255121}{RGB}{124,255,121}
\definecolor{maroon12700}{RGB}{127,0,0}
\definecolor{navy00127}{RGB}{0,0,127}
\definecolor{red255290}{RGB}{255,29,0}
\definecolor{yellow22825518}{RGB}{228,255,18}

\begin{axis}[
legend cell align={left},
legend style={draw=none, fill=none,
  xlabel={\textbf{Time} (sec)},
  ylabel={\textbf{Accuracy} (\%)},
},
yscale=0.6, xscale=0.75,
tick align=outside,
tick pos=left,
x grid style={darkgray176},
xmin=-9.49949369192124, xmax=269.891679618359,
xtick style={color=black},
y grid style={darkgray176},
ymin=59.7238095238095, ymax=86.7523809523809,
ytick style={color=black},
yticklabel style={rotate=90.0}
]
\addplot [ultra thick, gray, dashed, mark=*, mark size=2, mark options={solid}]
table {%
14.2887599468231 61.4285714285714
24.6740455150604 61.5238095238095
35.141822719574 61.6190476190476
45.638599729538 62.1904761904762
56.2690583229065 64.4761904761905
66.8843245983124 66.7619047619048
77.7658805847168 71.1428571428571
88.6925711154938 76.4761904761905
99.2472709655762 80.2857142857143
109.508946180344 82.5714285714286
119.790633487701 83.1428571428571
130.337510061264 83.9047619047619
140.955986309052 83.7142857142857
151.609069061279 84
162.209997320175 84.0952380952381
172.600471687317 84.2857142857143
183.052301597595 84.1904761904762
193.526771879196 84
204.086597251892 83.8095238095238
214.457344055176 83.7142857142857
225.259791564941 83.7142857142857
235.826626968384 83.4285714285714
246.619962596893 83.4285714285714
257.192080831528 83.6190476190476
};
\addplot [maroon12700, dashed, mark=*, mark size=1, mark options={solid}]
table {%
3.20010509490967 61.4285714285714
3.54124207496643 61.4285714285714
5.61002097129822 61.6190476190476
7.54065475463867 61.5238095238095
9.43085970878601 61.7142857142857
10.8351606369019 61.8095238095238
12.7455938339233 61.8095238095238
14.6470879077911 61.7142857142857
16.5664762020111 61.7142857142857
17.9700488090515 61.4285714285714
19.8871760368347 61.5238095238095
21.877098274231 61.7142857142857
23.7437811851501 62.0952380952381
25.1904420852661 62.5714285714286
27.0815311431885 63.8095238095238
28.9699927330017 65.1428571428572
30.8776287078857 66.3809523809524
31.9371017456055 67.1428571428572
33.8280087947845 68.5714285714286
34.9592000484467 69.8095238095238
36.8460198402405 72.2857142857143
38.2621229171753 74.3809523809524
40.1294054508209 76.2857142857143
42.0015071392059 77.7142857142857
43.859371805191 78.4761904761905
45.3107107639313 78.9523809523809
47.2096750736237 79.6190476190476
49.0738535404205 79.6190476190476
50.9947299480438 80.3809523809524
52.4271988391876 80.4761904761905
54.3406090736389 80.4761904761905
56.2781112194061 80.8571428571429
58.1644431591034 80.8571428571429
59.2299629688263 80.952380952381
61.141882276535 80.952380952381
63.0120406150818 80.8571428571428
64.9232442378998 81.0476190476191
65.6415172100067 81.1428571428571
67.5749418735504 81.4285714285714
69.4717048168182 81.5238095238095
71.4279217243195 81.6190476190476
72.8913562774658 81.6190476190476
74.7667907714844 81.6190476190476
76.6311904907227 81.7142857142857
78.5480662822723 81.8095238095238
79.9855391025543 81.9047619047619
81.9133882522583 81.9047619047619
83.8275811672211 81.7142857142857
85.6988813877106 82.0952380952381
86.830659198761 82
88.725977563858 82.0952380952381
90.5986225605011 82.0952380952381
92.4963326931 81.9047619047619
93.9224049091339 81.9047619047619
95.7779573917389 81.9047619047619
96.9180153369904 82
98.7893461704254 82.2857142857143
100.231992483139 82.2857142857143
102.19532251358 82.3809523809524
104.113616085052 82.3809523809524
106.024155092239 82.3809523809524
107.463712310791 82.3809523809524
109.421467590332 82.3809523809524
111.328037548065 82.3809523809524
113.196814489365 82.3809523809524
114.243938016891 82.3809523809524
116.189148759842 82.4761904761905
118.051090955734 82.5714285714286
120.01628575325 82.6666666666667
121.496128416061 82.6666666666667
123.365881919861 82.6666666666667
125.226609134674 82.6666666666667
127.164329195023 82.7619047619048
127.84545750618 82.7619047619048
129.715088319778 82.6666666666667
131.553492164612 82.6666666666667
133.417690324783 82.6666666666667
134.821751117706 82.6666666666667
136.687398958206 82.6666666666667
138.569668245316 82.6666666666667
140.477064561844 82.6666666666667
141.527513122559 82.6666666666667
143.397725486755 82.6666666666667
145.331127786636 82.6666666666667
147.255424690247 82.6666666666667
148.359981775284 82.6666666666667
150.621307468414 82.6666666666667
152.587673854828 82.6666666666667
154.516018867493 82.6666666666667
155.896261835098 82.5714285714286
157.768652439117 82.5714285714286
158.90481672287 82.6666666666667
160.789396858215 82.6666666666667
161.852890443802 82.5714285714286
164.040152215958 82.4761904761905
165.928826379776 82.4761904761905
167.792389297485 82.4761904761905
168.869053173065 82.4761904761905
170.760714578629 82.3809523809524
172.621715784073 82.3809523809524
174.446911478043 82.3809523809524
175.510928726196 82.4761904761905
177.720569038391 82.4761904761905
179.600006961823 82.4761904761905
181.518051862717 82.4761904761905
182.967929506302 82.4761904761905
184.91376748085 82.4761904761905
186.79220957756 82.5714285714286
188.730482053757 82.4761904761905
189.100766420364 82.4761904761905
191.322060823441 82.3809523809524
193.248282814026 82.3809523809524
195.105732488632 82.3809523809524
196.145779943466 82.3809523809524
198.035275030136 82.3809523809524
199.898157310486 82.4761904761905
201.783017683029 82.4761904761905
202.885470676422 82.4761904761905
205.103067970276 82.4761904761905
207.009948825836 82.2857142857143
208.879050779343 82.2857142857143
210.345342588425 82.2857142857143
212.246946001053 82.2857142857143
214.103614854813 82.2857142857143
215.993298721313 82.1904761904762
217.073727798462 82.2857142857143
219.289979600906 82.3809523809524
220.515578556061 82.3809523809524
222.450581216812 82.3809523809524
223.521905422211 82.3809523809524
225.376316261291 82.2857142857143
227.221725654602 82.2857142857143
229.131285047531 82.2857142857143
230.548656892776 82.2857142857143
232.405953359604 82.2857142857143
234.316825580597 82.2857142857143
236.256581306458 82.2857142857143
237.718823194504 82.2857142857143
239.551573991776 82.1904761904762
241.490371227264 82.1904761904762
243.419442129135 82.1904761904762
244.89688038826 82.1904761904762
246.761911678314 82.1904761904762
248.7189930439 82.0952380952381
250.635789203644 82.0952380952381
};
\addplot [red255290, dashed, mark=*, mark size=1, mark options={solid}]
table {%
4.89611015319824 61.4285714285714
5.50591926574707 61.4285714285714
8.40583052635193 61.6190476190476
11.2301819801331 61.8095238095238
14.0538328647614 61.9047619047619
16.8491012096405 61.7142857142857
19.6522707462311 62.0952380952381
22.4260688304901 62.3809523809524
25.2263686180115 62.952380952381
28.1268455505371 64.1904761904762
30.9646714687347 66.5714285714286
33.7516958236694 69.4285714285714
36.5469467639923 72.8571428571428
38.7606802463531 74.9523809523809
42.1619637012482 79.6190476190476
44.9626482486725 81.1428571428571
47.7161621570587 81.6190476190476
50.5871609687805 82.3809523809524
53.4217401027679 83.047619047619
56.2359489440918 83.047619047619
58.9948250293732 83.2380952380952
61.8181144714355 83.5238095238095
64.6322612285614 83.3333333333333
67.4561078548431 83.5238095238095
70.3708894729614 83.5238095238095
73.1669690132141 83.7142857142857
75.4113099098206 83.7142857142857
78.7237940311432 83.8095238095238
81.5368499755859 83.8095238095238
84.4678694725037 84
87.2067606925964 84
89.4838761806488 84
92.3313529968262 84
95.0961734294891 83.8095238095238
97.9719235897064 84.0952380952381
100.770885181427 84.2857142857143
103.611705446243 84.2857142857143
106.524723958969 84.1904761904762
109.359414434433 84.0952380952381
112.22438788414 84
115.024213314056 84.2857142857143
117.835305929184 84.2857142857143
120.667639017105 84.3809523809524
123.451947259903 84.3809523809524
126.368460321426 84.3809523809524
129.306394577026 84.3809523809524
131.019387340546 84.3809523809524
133.787702131271 84.4761904761905
136.674323987961 84.3809523809524
138.920525407791 84.3809523809524
142.192518997192 84.6666666666667
145.058482837677 84.6666666666667
147.244238710403 84.5714285714286
150.601484966278 84.6666666666667
153.402342987061 84.7619047619048
156.194509220123 84.6666666666667
159.027980613709 84.8571428571428
161.93503985405 84.5714285714286
164.817477893829 84.7619047619047
167.594429206848 84.6666666666666
170.412575435638 84.6666666666666
172.683518028259 84.8571428571428
175.480949354172 84.8571428571428
177.686825704575 84.8571428571428
180.981029891968 84.9523809523809
183.669411993027 85.0476190476191
186.626335430145 84.8571428571428
189.454277515411 84.8571428571428
192.243817901611 84.7619047619048
195.032463741302 84.7619047619048
197.826151752472 84.7619047619048
200.680549669266 84.952380952381
203.437863492966 84.952380952381
206.279640102386 84.952380952381
209.111816215515 84.952380952381
212.008021259308 84.952380952381
214.802338981628 84.952380952381
217.534843635559 85.1428571428571
220.277496337891 85.3333333333333
223.13670372963 85.4285714285714
225.913457250595 85.4285714285714
228.766126966476 85.4285714285714
231.626363372803 85.4285714285714
234.518062829971 85.4285714285714
237.382259464264 85.4285714285714
240.191265201569 85.5238095238095
242.974007129669 85.4285714285714
245.759115076065 85.2380952380952
248.516789627075 85.4285714285714
251.581876850128 85.2380952380952
};
\addplot [darkorange2551480, dashed, mark=*, mark size=1, mark options={solid}]
table {%
6.92584609985352 61.4285714285714
10.5151941776276 61.3333333333333
14.9225340366364 61.6190476190476
19.432484960556 61.4285714285714
23.8808061122894 61.7142857142857
28.2863722324371 62.0952380952381
32.7410010814667 62.6666666666667
37.1872175216675 63.2380952380952
41.6591190814972 66.2857142857143
46.0753615856171 71.3333333333333
50.5141795635223 76.1904761904762
55.0714073181152 79.5238095238095
59.4996929168701 82
63.9830903530121 82.9523809523809
68.4037806510925 83.6190476190476
72.8186039447784 84.4761904761905
77.372851228714 84.5714285714286
81.737136554718 84.3809523809524
86.2496625900269 84.2857142857143
90.7466389656067 84.2857142857143
95.3029866218567 84.1904761904762
99.7566875934601 84.1904761904762
104.216941642761 84.2857142857143
108.747588825226 84.1904761904762
113.138658905029 84.1904761904762
117.520470285416 84.2857142857143
121.822071170807 84.0952380952381
126.299953460693 83.8095238095238
130.839840459824 83.8095238095238
135.458820438385 83.7142857142857
139.948149776459 83.8095238095238
144.445042514801 83.9047619047619
148.903579235077 84.0952380952381
153.31237578392 84.0952380952381
157.801959228516 84.1904761904762
162.18702750206 84.3809523809524
166.630939292908 84.3809523809524
171.228766918182 84.4761904761905
175.681766700745 84.5714285714286
180.188776111603 84.8571428571428
184.772325611114 84.8571428571428
189.164321517944 84.7619047619048
193.70549864769 85.0476190476191
198.194571638107 85.2380952380952
202.677945423126 85.0476190476191
207.055950737 85.3333333333333
211.567883110046 85.0476190476191
215.983783245087 85.0476190476191
220.309953975678 85.0476190476191
224.766782188416 85.0476190476191
229.259447479248 84.952380952381
233.753171157837 84.952380952381
238.182022476196 84.952380952381
242.594867181778 85.1428571428571
247.101494312286 85.0476190476191
251.633205747604 85.2380952380952
};
\addplot [yellow22825518, dashed, mark=*, mark size=1, mark options={solid}]
table {%
9.57371940612793 61.4285714285714
13.1079522132874 61.8095238095238
18.9799912452698 61.8095238095238
24.864520740509 61.7142857142857
30.5035231590271 61.4285714285714
36.3322012424469 60.9523809523809
42.1520233631134 61.7142857142857
47.8357749938965 65.2380952380952
53.5323059082031 68.9523809523809
59.1573795318604 73.5238095238095
64.7749289989471 76.8571428571428
70.4675362110138 80.5714285714286
76.1357436180115 82.4761904761905
82.0169853687286 83.2380952380952
87.7845330238342 83.5238095238095
90.0811986923218 83.8095238095238
99.3793427467346 84.1904761904762
105.092013454437 84.0952380952381
110.909537553787 84
116.442808151245 84.3809523809524
122.085288715363 84
127.719506072998 83.9047619047619
133.487216186523 83.5238095238095
139.401715803146 83.5238095238095
145.085897159576 83.4285714285714
150.842044115067 83.5238095238095
156.521224927902 83.1428571428571
162.231451511383 82.9523809523809
168.091255760193 82.6666666666667
173.735045862198 82.7619047619047
176.006289052963 82.7619047619047
185.002396869659 82.4761904761905
190.684195756912 82.7619047619048
196.440318346024 82.3809523809524
202.094352912903 82.3809523809524
207.750426006317 82.2857142857143
213.344737482071 82.3809523809524
219.161036205292 82.3809523809524
224.973760128021 82.0952380952381
230.65581946373 82
236.569512224197 82
242.276014280319 82.0952380952381
247.822927427292 81.9047619047619
253.552560138702 82
};
\addplot [lightgreen124255121, dashed, mark=*, mark size=1, mark options={solid}]
table {%
11.4135353088379 61.4285714285714
15.5311346530914 61.2380952380952
22.4894075870514 61.6190476190476
29.4456149101257 61.6190476190476
36.4712969779968 61.2380952380952
43.3996119976044 62.5714285714286
50.5486529827118 64.6666666666667
57.4055013656616 67.2380952380952
64.4163696289063 73.1428571428571
71.3131937980652 77.7142857142857
78.3837872505188 80.2857142857143
85.3414153575897 82.6666666666667
92.3515394210815 83.9047619047619
99.1986594200134 84.4761904761905
106.142526197433 84.2857142857143
113.226497459412 83.9047619047619
120.25051817894 83.7142857142857
127.311418056488 83.9047619047619
134.648930692673 84
141.648421621323 84.2857142857143
148.70078663826 84.2857142857143
155.741444826126 84
162.620490646362 84
169.58930888176 83.9047619047619
176.591585969925 83.8095238095238
183.553468132019 83.5238095238095
190.580014848709 83.4285714285714
197.400158786774 83.4285714285714
204.451155567169 83.4285714285714
211.35576634407 83.3333333333333
218.166674613953 83.5238095238095
225.166023206711 83.2380952380952
232.062416553497 83.1428571428571
238.846868038177 83.047619047619
245.703594636917 83.1428571428571
252.495811891556 83.1428571428571
};
\addplot [cyan21255225, dashed, mark=*, mark size=1, mark options={solid}]
table {%
12.502206325531 61.4285714285714
13.9938461780548 61.6190476190476
21.8628584384918 61.6190476190476
29.4199700832367 61.6190476190476
37.168030166626 61.4285714285714
44.7855299472809 62.0952380952381
52.3668079853058 65.1428571428572
60.1366874217987 70.4761904761905
68.0455518722534 73.1428571428571
75.6139791488647 78.2857142857143
83.3375648498535 81.3333333333333
91.2836173534393 84.0952380952381
98.9658998012543 84.5714285714286
106.752843809128 84.8571428571428
114.770917844772 84.952380952381
122.527950239182 84.8571428571428
130.141804409027 84.5714285714286
137.856242847443 84.4761904761905
145.501202344894 84
153.203642654419 83.7142857142857
161.185816955566 83.7142857142857
169.034679079056 83.8095238095238
176.768963241577 83.4285714285714
184.248024320602 83.3333333333333
192.098141384125 83.4285714285714
199.799188995361 83.2380952380952
207.515574073791 82.9523809523809
215.375749826431 82.9523809523809
223.242349767685 82.9523809523809
231.003547668457 83.1428571428572
238.773049736023 83.0476190476191
246.451361513138 82.7619047619047
254.053384065628 82.8571428571428
};
\addplot [dodgerblue0128255, dashed, mark=*, mark size=1, mark options={solid}]
table {%
13.9964662075043 61.4285714285714
19.183938741684 61.5238095238095
27.8976254463196 61.4285714285714
36.407507610321 61.9047619047619
44.9960407733917 62.0952380952381
53.5448992729187 64
62.2820516586304 69.3333333333333
70.9445230484009 73.4285714285714
79.5700613021851 77.4285714285714
88.4025691986084 81.2380952380952
96.9813408374786 84
105.861007356644 84.1904761904762
114.605151224136 84.1904761904762
123.361117506027 84.3809523809524
132.156996059418 84.0952380952381
140.996408891678 84
149.584428596497 84.0952380952381
158.385422420502 83.8095238095238
167.226110839844 83.7142857142857
175.736115694046 83.5238095238095
184.136945199966 83.4285714285714
192.81523566246 83.5238095238095
201.673490333557 83.5238095238095
210.30283241272 83.3333333333333
219.013444948196 83.5238095238095
227.552308511734 83.3333333333333
236.277521705627 83.3333333333333
245.070337247848 83.1428571428572
253.787868070602 83.0476190476191
};
\addplot [blue, dashed, mark=*, mark size=1, mark options={solid}]
table {%
14.4601809024811 61.4285714285714
16.2947095870972 61.4285714285714
25.593458366394 61.3333333333333
35.3110456943512 61.047619047619
45.1214558124542 61.6190476190476
54.4904423236847 63.1428571428571
64.0754709720612 66.0952380952381
73.269922208786 72.2857142857143
82.6146405220032 76.4761904761905
92.1212346076965 79.8095238095238
101.601522350311 82.4761904761905
111.306623458862 83.7142857142857
120.863529348373 84.1904761904762
130.321436452866 83.9047619047619
139.759170436859 83.9047619047619
149.467693424225 83.9047619047619
159.048106098175 83.8095238095238
168.587756443024 83.7142857142857
178.139263105392 83.5238095238095
187.792557525635 83.4285714285714
197.125254583359 83.2380952380952
206.697101211548 83.2380952380952
216.147174406052 83.2380952380952
225.589801216125 83.1428571428571
235.112144136429 82.8571428571428
244.606186389923 83.047619047619
253.971092796326 83.3333333333333
};
\addplot [navy00127, dashed, mark=*, mark size=1, mark options={solid}]
table {%
14.7284090518951 61.4285714285714
16.7130358695984 61.3333333333333
26.852290725708 61.4285714285714
37.3170840740204 61.3333333333333
47.5842183113098 61.1428571428571
57.9010594367981 63.7142857142857
68.2653579235077 67.9047619047619
78.5202262401581 70.9523809523809
88.7467342853546 76.1904761904762
99.1922794342041 80.4761904761905
109.397260046005 83.5238095238095
119.729198551178 84.2857142857143
129.86603884697 84.4761904761905
140.067887639999 84.7619047619047
150.523748445511 84.6666666666666
160.799932003021 84.6666666666666
171.143469524384 84.3809523809524
181.311746358871 84.0952380952381
191.403445100784 83.7142857142857
201.77010974884 83.5238095238095
211.950331068039 83.4285714285714
222.369553995132 83.3333333333333
232.605009555817 83.1428571428571
242.918841075897 83.2380952380952
253.293868350983 83.047619047619
};
\end{axis}


\definecolor{blue}{RGB}{0,0,255}
\definecolor{cyan21255225}{RGB}{21,255,225}
\definecolor{darkgray176}{RGB}{176,176,176}
\definecolor{darkorange2551480}{RGB}{255,148,0}
\definecolor{dodgerblue0128255}{RGB}{0,128,255}
\definecolor{gray}{RGB}{128,128,128}
\definecolor{lightgray204}{RGB}{204,204,204}
\definecolor{lightgreen124255121}{RGB}{124,255,121}
\definecolor{maroon12700}{RGB}{127,0,0}
\definecolor{navy00127}{RGB}{0,0,127}
\definecolor{red255290}{RGB}{255,29,0}
\definecolor{yellow22825518}{RGB}{228,255,18}

\begin{axis}[
hide axis,
legend cell align={left},
legend style={
yscale=0.45, 
%
  draw opacity=1,
  text opacity=1,
  at={(1.25,-0.4)},
  anchor=south west,
  font= \normalsize,
  draw=lightgray204
}
,
yscale=0.45, xscale=0.8,
tick align=outside,
tick pos=left,
x grid style={darkgray176},
xmin=-7.62244165658951, xmax=267.859734280109,
xtick style={color=black},
y grid style={darkgray176},
ymin=58.5708333333333, ymax=84.5125,
ytick style={color=black},
yticklabel style={rotate=90.0}
]
\addlegendimage{maroon12700, dashed, mark=*, mark size=1, mark options={solid}}
\addlegendentry{10\%}
\addlegendimage{red255290, dashed, mark=*, mark size=1, mark options={solid}}
\addlegendentry{20\%}
\addlegendimage{darkorange2551480, dashed, mark=*, mark size=1, mark options={solid}}
\addlegendentry{30\%}
\addlegendimage{yellow22825518, dashed, mark=*, mark size=1, mark options={solid}}
\addlegendentry{40\%}
\addlegendimage{lightgreen124255121, dashed, mark=*, mark size=1, mark options={solid}}
\addlegendentry{50\%}
\addlegendimage{cyan21255225, dashed, mark=*, mark size=1, mark options={solid}}
\addlegendentry{60\%}
\addlegendimage{dodgerblue0128255, dashed, mark=*, mark size=1, mark options={solid}}
\addlegendentry{70\%}
\addlegendimage{blue, dashed, mark=*, mark size=1, mark options={solid}}
\addlegendentry{80\%}
\addlegendimage{navy00127, dashed, mark=*, mark size=1, mark options={solid}}
\addlegendentry{90\%}
\addlegendimage{ultra thick, gray, dashed, mark=*, mark size=2, mark options={solid}}
\addlegendentry{100\%}

\end{axis}

\end{tikzpicture}

%% file: section_preliminaries.tex
\section{Preliminaries}\label{sec:prelims}

\paragraph*{Relational Model}


A \e{database schema} $\scs$ consists of a finite collection of \e{relation schemas} $R(A_1,\dots,A_k)$ where $R$ is a distinct \e{relation name} and each $A_i$ is a distinct \e{attribute name}. 
Each attribute $A$ is associated with a \e{domain}, denoted $\dom(A)$. Each relation schema $R(A_1,\dots,A_k)$ has a unique \e{key}, denoted $\key(R)$, such that $\key(R)\subseteq\set{A_1,\dots,A_k}$. 
A \e{fact} over a relation schema $R(A_1,\dots,A_k)$ has the form
$R(a_1,\dots,a_k)$ where $a_i\in\dom(A_i)$ for all $i=1,\dots,k$.
A \e{database} $D$ over the schema $\scs$ is a finite set of facts over the relation schemas of $\scs$. In addition, such an $a_i$ can be missing and given as a distinguished \e{null} value.  The fact $R(a_1,\dots,a_k)$ is also called an \e{$R$-fact} and a \e{$\scs$-fact}. We denote by $R(D)$ the restriction of $D$ to its $R$-facts.  For a fact $f=R(a_1,\dots,a_k)$ over $R(A_1,\dots,A_k)$, we denote by $f[A_i]$ the value $a_i$, and by $f[B_1,\dots,B_\ell]$ the tuple $(f[B_1],\dots,f[B_\ell])$. 
The \e{active domain} of an attribute $A$ (\e{w.r.t.~to the database $D$}), denoted $\adom_D(A)$,
is the set 
$\set{f[A]\mid f\in R(D)}$.

A \e{foreign-key constraint} (FK) is an inclusion dependency of the form $R[\attseq B]\subseteq S[\attseq C]$ where $R$ and $S$ are relation names, $\attseq B=B_1,\dots, B_\ell$ and $\attseq C=C_1,\dots,C_\ell$ are sequences of distinct attributes of $R$ and $S$, respectively, and $\key(S)=\set{C_1,\dots,C_\ell}$.
For every FK $R[\attseq B]\subseteq S[\attseq C]$ and $R$-fact $f\in D$ 
there exists an $S$-fact $g\in D$ such that $f[\attseq B]=g[\attseq C]$. 
We then say that \e{$f$ references $g$}.


\paragraph*{\forward{}}
The goal of \forward{} 
is to derive an embedding function $\gamma: D \rightarrow \mathbb{R}^{k}$ for the tuples in a database. 
Here, the dimension $k>0$ is a hyperparameter.
The general objective is to compute an embedding $\gamma$ that represents the data in a way that makes it accessible for data analysis and machine learning algorithms. 
To this end, \forward{} learns embeddings that encode the distribution of values seen along random walks through the database.
Next, we introduce \forward{}'s notion of random walks through databases formally and recap how these walks are used to produce an embedding.
A \emph{walk scheme} $s$ has the form
\begin{eqnarray}
\label{eq:walkscheme}
R_0[\attseq A^0]\mbox{---}[\attseq B^1]R_1[\attseq A^1]
\mbox{---}[\attseq B^2]R_2[\attseq A^2]\mbox{---}\cdots\mbox{---}
[\attseq B^{\ell}]R_{\ell}
\end{eqnarray}
such that for all $k=1,\dots,\ell$, either
$R_{k-1}[\attseq A^{k-1}] \subseteq R_{k}[\attseq B^k]$ is an FK or
$R_{k}[\attseq B^k] \subseteq R_{k-1}[\attseq A^{k-1}]$
is an FK.
We say that $s$ has \e{length} $\ell$, that it \e{starts from} $R_0$, and that it \e{ends with} $R_\ell$.

A \emph{walk} with the scheme $s$ is a sequence $(f_0,\dots,f_{\ell})$ of facts such that $f_k$ is an $R_k$-fact and $f_{k-1}[\attseq A^{k-1}]=f_{k}[\attseq B^{k}]$ for all $k=1,\dots,\ell$.  
We say that $(f_0,\dots,f_\ell)$ \e{starts from}, or has the \e{source}, $f_0$, and that it \e{ends with}, or has the \e{destination}, $f_\ell$.
\forward{} allows walk schemes and walks of length zero;
the walks of this scheme have the form $(f_0)$ and consist of the fact $f_0$.
%
A \emph{random walk} with $s$ defines a distribution over the destinations.
Formally, let $f_0=f$ be an $R_0$-fact. 
We denote by $\W(f,s)$ the distribution over the walks with the walk scheme $s$ where each walk is sampled by starting from $f_0$ and then iteratively selecting $f_{k}$, for $k=1,\dots,\ell$, randomly and uniformly from the set $\{f\in R_{k}\mid f[\attseq B^k]=f_{k-1}[\attseq A^{k-1}]\}]$.
We denote by $d_{f,s}$ the random element that maps each walk in $\W(f,s)$ to its destination---the last fact in the walk. For $g\in R^k(D)$, the probability that a walk sampled from $\W(f,s)$ ends with $g$ is $\prob(d_{f,s}=g)$.

\paragraph{Targeted Walk Schemes}
A \e{targeted walk scheme} is a pair $(s,A)$ such that $s$ is a walk scheme from a relation $R$ to a relation $R'$, and $A$ is an attribute of $R'$.
Given a start fact $f$ in the start relation $R$ of the walk scheme $s$,
a targeted walk scheme $(s,A)$ defines the random variable $d_{f,s}[A]$ that forms the value of the random walk's destination in the attribute $A$.
We denote by $\tws(R, \ell_\text{max})$ is the set of all targeted walk schemes $(s,A)$ such that $s$ is a walk scheme of length at most $\ell_{\text{max}}$ starting from the relation $R$ (and ending in any relation that includes $A$).
For example, 
\Cref{fig:walks} shows several targeted walk schemes over the schema of the database $D$ of \Cref{fig:dbexample}. To illustrate, $(s_7,A_7)$ is the targeted walk scheme with $A_7\defeq \att{name}$ and
$
s_7\defeq \rel{Country}[\att{code}]\mbox{---}[\att{country}]\rel{Member}[\att{org}]\mbox{---}[\att{abbrev}]\rel{Org}
$.
Walks of $s_7$ include the sequences $(c_1,m_1,o_1)$ and $(c_2,m_3,o_2)$. The distribution 
$d_{c_2,s_7}[A_7]$ is uniform between \val{European Union} and \val{Nordic Council}.
The walk schemes $(s_1,A_1)$, \dots,  $(s_7,A_7)$
are in the set $\tws(\rel{Country},3)$ since they all start with $\rel{Country}$ and have length at most three, 
yet $(s_8,A_8)$ is not in $\tws(\rel{Country},3)$ but rather in
$\tws(\rel{Country},4)$.

Recall that databases 
may have nulls.
A random walk starting at $f$ might end at an $R_\ell$-fact $g$ with null on $A$.
As a convention, we define the probability distribution of $d_{f,s}[A]$ by ignoring the nulls (and normalizing).
With this modification, we enforce $d_{f,s}[A] \in \dom(A)$.
This will be crucial in Section~\ref{sec:kernels}, where we define similarity measures for $d_{f,s}[A]$ based on $\dom(A)$.


\paragraph*{Kernelized Domains}
\label{sec:kernels}
\forward{} assumes that every attribute $A$ is associated with a kernel function $\kappa_{A}$
that maps all pairs of elements from $\dom(A)$ to the nonnegative reals. 
Intuitively, $\kappa_{A}(a,b)$ measures the similarity between elements $a,b\in\dom(A)$. 
Kernel functions offer a straightforward way of encoding domain knowledge by modeling the similarity of the domain values.
Kernels are also helpful when dealing with noisy data.
For example, kernels based on the edit distance can be used to smooth out random typos in text.
\forward{} uses these kernel functions to define similarity measures for the random variables $d_{f,s}[A]$.

Let $s$ be a walk scheme of length $\ell$ from $R$ to $R'$. 
Let $A$ be an attribute of $R'$ and let $f$ and $f'$ be two distinct $R$-facts. 
Then $d_{f,s}[A]$ and $d_{f,s'}[A]$ are random variables over a shared kernelized domain $\dom(A)$.  
The \emph{Expected Kernel Distance} \text{KD} is  the expected distance between two random values selected independently at random:
\begin{align}
    \text{KD}(d_{s,f}[A],\!d_{s,f'}[A]) \label{eq:KEK}
    \!&=\! \underset{\W(f,s)\times\W(f',s)}{\mathbb{E}}\hspace{-2em}[\kappa_A(d_{s,f}[A],\!d_{s,f'}[A])]
\end{align}
\forward{} uses the Expected Kernel Distance to quantify the similarity between $d_{f,s}[A]$ and $d_{f',s}[A]$ with respect to the  kernel $\kappa_{A}$.

\input{example_db.tex}
\input{walks.tex}

\paragraph*{Embedding \label{sec:forward_embeddings}}
Intuitively, \forward{} aims to encode the kernel $\text{KD}(d_{s,f}[A],\!d_{s,f'}[A])$.
The primary output is the embedding $\varphi: D \rightarrow \mathbb{R}^k$.
Additionally, the \forward{} algorithm learns an auxiliary embedding $\psi:\tws(R, \ell_\text{max}) \rightarrow \mathbb{R}^{d \times d}$ that maps each targeted walk scheme $(s,A)$ to a symmetric matrix $\psi(s,A)$.
(Recall that $\tws(R, \ell_\text{max})$ is the set of targeted walk schemes of length at most $k$.) 
The objective is to find $\varphi$ and $\psi$ satisfying
$\varphi(f)^{\top}\psi(s,A)\varphi(f') = \text{KD}(d_{s,f}[A], d_{s,f'}[A])$ 
for all $f,f'\in R(D)$ and $(s,A)\in \tws(R, \ell_\text{max})$.
Effectively, \forward{} minimizes
\begin{equation}
    \label{eq:objective-min}
    \big|\varphi(f)^{\top}\psi(s,A)\varphi(f') - \text{KD}(d_{s,f}[A], d_{s,f'}[A])\big|
\end{equation}
for all $f$, $f'$, $s$ and $A$, via   Stochastic Gradient Descent (SGD).

Intuitively, the algorithm learns an inner product $\langle \cdot,\cdot\rangle_{S,A}$ on the latent space of the embedding $\phi$ defined by 
$\langle x,y\rangle_{S,A}=x^\top \psi(s,A) y$ 
for all $s$ and $A$ such that the similarity of facts $f$ and $f'$ with respect to this inner product matches the similarity between the random variables $d_{f,s}[A]$ and $d_{f',s}[A]$ with respect to the kernel $\kappa_{R_\ell.A}$.


\forward{} uses gradient descent to learn $\phi$ and $\psi$.
During training, tuples of the form $(f, f', s, A, g, g')$ are sampled, where $f$ and $f'$ are $R$ facts from the database and $(s,A) \in \tws(R, \ell_\text{max})$.
The $R_\ell$ facts $g$ and $g'$ are the destinations of random walks with scheme $s$ sampled for $f$ and $f'$, respectively.
We  use a hyperparameter $n_\text{samples} \in \mathbb{N}$.
For each $R$-fact $f$ and scheme $(s,A) \in \tws(R, \ell_\text{max})$, we sample $n_\text{samples}$  tuples $(f, f', s, A, g, g')$ with $f' \neq f$.
The following loss is then minimized for each sample using SGD:
\begin{equation}
    \label{eq:loss} 
    \mathcal{L} = \frac{1}{2}|\varphi(f)^{\top}\psi(s,A)\varphi(f') - \kappa_{R_\ell.A}(g[A], g'[A])|^2.
\end{equation}
This objective uses $\kappa_{R_\ell.A}(g[A], g'[A])$ as a (stochastic) estimate of $\text{KD}(d_{s,f}[A], d_{s,f'}[A])$.

%% file: example_db.tex
\newcommand{\cc}{\cellcolor{lightgray}}

\begin{figure}[t]
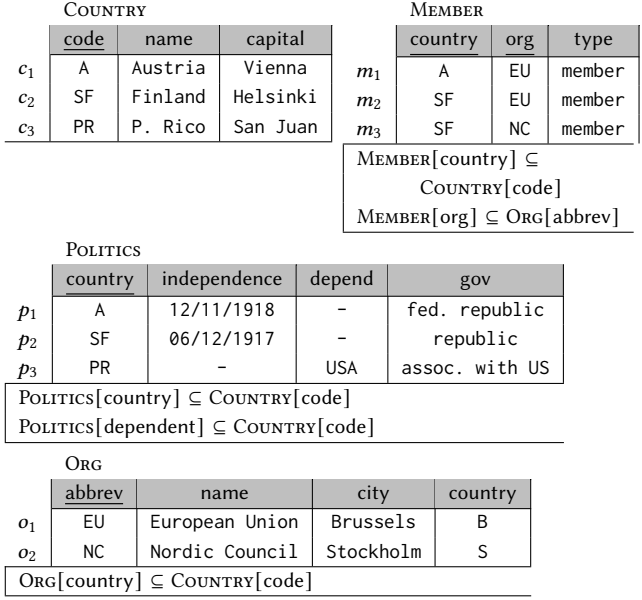

\small
 \begin{flushleft}
  \begin{tabular}[t]{l|c|c|c|}
         \multicolumn{1}{c}{} & \multicolumn{3}{l}{\rel{Country}}\\  \cline{2-4}
         &\cc \att{\underline{code}} &\cc \att{name} &\cc
                                \att{capital} \\
        \cline{2-4}
        $c_1$ & \val{A} & \val{Austria} & \val{Vienna} \\
        $c_2$ & \val{SF} & \val{Finland} & \val{Helsinki}\\
        $c_3$ & \val{PR} & \val{P.~Rico} & \val{San Juan}\\
       \hline
  \end{tabular}
  \hskip0.5em
\begin{tabular}[t]{l|c|c|c|c}
        \multicolumn{1}{c}{} & \multicolumn{3}{l}{\rel{Member}}\\
        \cline{2-4}
        & \cc \att{\underline{country}} & \cc \att{\underline{org}} & \cc \att{type} \\
        \cline{2-4}
        $m_1$ & \val{A} & \val{EU} & \val{member} \\
        $m_2$ & \val{SF} & \val{EU} & \val{member} \\
        $m_3$ & \val{SF} & \val{NC} & \val{member} \\
        \hline
       \multicolumn{5}{|l}{$\rel{Member}[\att{country}]\subseteq$}\\
       \multicolumn{5}{|c}{~~$\rel{Country}[\att{code}]$}\\
       \multicolumn{5}{|l}{$\rel{Member}[\att{org}]\subseteq \rel{Org}[\att{abbrev}]$}\\
       \hline
    \end{tabular}
      \vskip0em
    \begin{tabular}[t]{l|c|c|c|c|}
        \multicolumn{1}{c}{} &   \multicolumn{4}{l}{\rel{Politics}}\\ \cline{2-5}
        &\cc \underline{\att{country}} & \cc \att{independence} & \cc \att{depend} & \cc \att{gov} \\
         \cline{2-5}
        $p_1$ & \val{A} & \val{12/11/1918} & \val{-} & \val{fed.~republic} \\
        $p_2$ & \val{SF} & \val{06/12/1917} &\val{-} & \val{republic} \\
        $p_3$ & \val{PR} & \val{-} &\val{USA} & \val{assoc. with US} \\
       \hline
       \multicolumn{5}{|l}{$\rel{Politics}[\att{country}]\subseteq \rel{Country}[\att{code}]$}\\
       \multicolumn{5}{|l}{$\rel{Politics}[\att{dependent}]\subseteq \rel{Country}[\att{code}]$}\\\hline
    \end{tabular}
    \vskip0.2em
    \begin{tabular}[t]{l|c|c|c|c|}
        \multicolumn{1}{c}{} &\multicolumn{4}{l}{\rel{Org}}\\
        \cline{2-5}
         & \cc \att{\underline{abbrev}} & \cc \att{name} & \cc \att{city} & \cc \att{country} \\
       \cline{2-5}
        $o_1$ & \val{EU} & \val{European Union} & \val{Brussels} & \val{B} \\
        $o_2$ & \val{NC} & \val{Nordic Council} & \val{Stockholm} & \val{S} \\ \cline{1-5}
       \multicolumn{5}{|l}{$\rel{Org}[\att{country}]\subseteq\rel{Country}[\att{code}]$}\\
       \hline
    \end{tabular}
     \end{flushleft}
     \vskip-1em
    \caption{\label{fig:dbexample}Example of a database, with foreign-key constraints, taken from the Mondial dataset.}
\end{figure}

%% file: walks.tex
\begin{figure}

  \def\CO{\rel{Country}}
  \def\PO{\rel{Politics}}
  \def\ME{\rel{Member}}
  \def\OR{\rel{Org}}

  \def\a#1{\att{#1}}

  \scalebox{0.8}{\input{walks.pspdftex}}
  \caption{\label{fig:walks} Examples of targeted walk schemes of length one to four, for the database schema of Figure~\ref{fig:dbexample}. All walk schemes start at the \rel{Country} relation. The figure of the walk scheme for $(s,A)$ shows $s$ as a path of rectangles and $A$ (e.g.~\att{name}) as an attribute under the rightmost (last) rectangle.}
\end{figure}
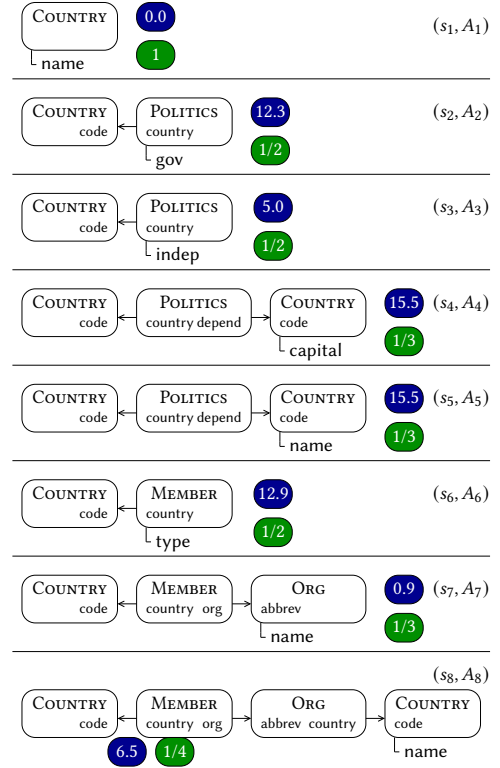

%% file: walks.pspdftex
\begin{picture}(0,0)%
\includegraphics{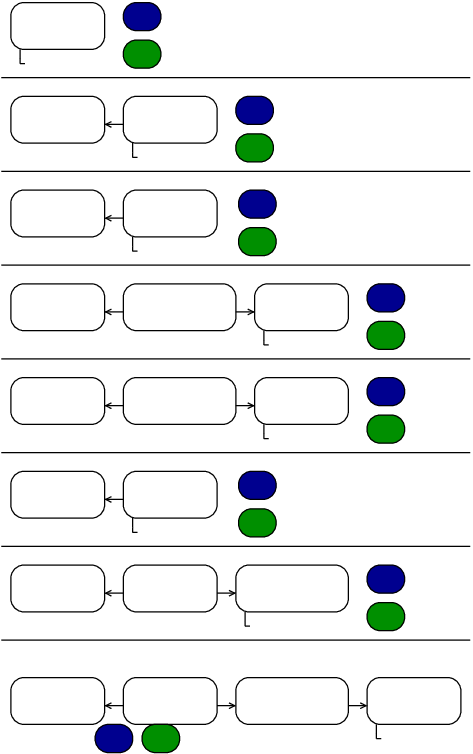}%
\end{picture}%
\setlength{\unitlength}{3947sp}%
\begingroup\makeatletter\ifx\SetFigFont\undefined%
\gdef\SetFigFont#1#2#3#4#5{%
  \reset@font\fontsize{#1}{#2pt}%
  \fontfamily{#3}\fontseries{#4}\fontshape{#5}%
  \selectfont}%
\fi\endgroup%
\begin{picture}(3777,6024)(289,-5323)
\put(3376,-2761){\makebox(0,0)[b]{\smash{{\SetFigFont{9}{10.8}{\familydefault}{\mddefault}{\updefault}{\color[rgb]{1,1,1}1/3}%
}}}}
\put(3376,-2461){\makebox(0,0)[b]{\smash{{\SetFigFont{9}{10.8}{\familydefault}{\mddefault}{\updefault}{\color[rgb]{1,1,1}15.5}%
}}}}
\put(2348,-3511){\makebox(0,0)[b]{\smash{{\SetFigFont{9}{10.8}{\familydefault}{\mddefault}{\updefault}{\color[rgb]{1,1,1}1/2}%
}}}}
\put(2348,-3211){\makebox(0,0)[b]{\smash{{\SetFigFont{9}{10.8}{\familydefault}{\mddefault}{\updefault}{\color[rgb]{1,1,1}12.9}%
}}}}
\put(751,-3211){\makebox(0,0)[b]{\smash{{\SetFigFont{9}{10.8}{\familydefault}{\mddefault}{\updefault}{\color[rgb]{0,0,0}\CO}%
}}}}
\put(1651,-3211){\makebox(0,0)[b]{\smash{{\SetFigFont{9}{10.8}{\familydefault}{\mddefault}{\updefault}{\color[rgb]{0,0,0}\ME}%
}}}}
\put(1051,-3361){\makebox(0,0)[rb]{\smash{{\SetFigFont{7}{8.4}{\familydefault}{\mddefault}{\updefault}{\color[rgb]{0,0,0}\a{code}}%
}}}}
\put(1351,-3361){\makebox(0,0)[lb]{\smash{{\SetFigFont{7}{8.4}{\familydefault}{\mddefault}{\updefault}{\color[rgb]{0,0,0}\a{country}}%
}}}}
\put(1426,-3586){\makebox(0,0)[lb]{\smash{{\SetFigFont{9}{10.8}{\familydefault}{\mddefault}{\updefault}{\color[rgb]{0,0,0}\a{type}}%
}}}}
\put(3376,-3961){\makebox(0,0)[b]{\smash{{\SetFigFont{9}{10.8}{\familydefault}{\mddefault}{\updefault}{\color[rgb]{1,1,1}0.9}%
}}}}
\put(3376,-4261){\makebox(0,0)[b]{\smash{{\SetFigFont{9}{10.8}{\familydefault}{\mddefault}{\updefault}{\color[rgb]{1,1,1}1/3}%
}}}}
\put(751,-3961){\makebox(0,0)[b]{\smash{{\SetFigFont{9}{10.8}{\familydefault}{\mddefault}{\updefault}{\color[rgb]{0,0,0}\CO}%
}}}}
\put(1651,-3961){\makebox(0,0)[b]{\smash{{\SetFigFont{9}{10.8}{\familydefault}{\mddefault}{\updefault}{\color[rgb]{0,0,0}\ME}%
}}}}
\put(1051,-4111){\makebox(0,0)[rb]{\smash{{\SetFigFont{7}{8.4}{\familydefault}{\mddefault}{\updefault}{\color[rgb]{0,0,0}\a{code}}%
}}}}
\put(1351,-4111){\makebox(0,0)[lb]{\smash{{\SetFigFont{7}{8.4}{\familydefault}{\mddefault}{\updefault}{\color[rgb]{0,0,0}\a{country}}%
}}}}
\put(1951,-4111){\makebox(0,0)[rb]{\smash{{\SetFigFont{7}{8.4}{\familydefault}{\mddefault}{\updefault}{\color[rgb]{0,0,0}\a{org}}%
}}}}
\put(2626,-3961){\makebox(0,0)[b]{\smash{{\SetFigFont{9}{10.8}{\familydefault}{\mddefault}{\updefault}{\color[rgb]{0,0,0}\OR}%
}}}}
\put(2251,-4111){\makebox(0,0)[lb]{\smash{{\SetFigFont{7}{8.4}{\familydefault}{\mddefault}{\updefault}{\color[rgb]{0,0,0}\a{abbrev}}%
}}}}
\put(2326,-4336){\makebox(0,0)[lb]{\smash{{\SetFigFont{9}{10.8}{\familydefault}{\mddefault}{\updefault}{\color[rgb]{0,0,0}\a{name}}%
}}}}
\put(2626,-4861){\makebox(0,0)[b]{\smash{{\SetFigFont{9}{10.8}{\familydefault}{\mddefault}{\updefault}{\color[rgb]{0,0,0}\OR}%
}}}}
\put(3001,-5011){\makebox(0,0)[rb]{\smash{{\SetFigFont{7}{8.4}{\familydefault}{\mddefault}{\updefault}{\color[rgb]{0,0,0}\a{country}}%
}}}}
\put(2251,-5011){\makebox(0,0)[lb]{\smash{{\SetFigFont{7}{8.4}{\familydefault}{\mddefault}{\updefault}{\color[rgb]{0,0,0}\a{abbrev}}%
}}}}
\put(3601,-4861){\makebox(0,0)[b]{\smash{{\SetFigFont{9}{10.8}{\familydefault}{\mddefault}{\updefault}{\color[rgb]{0,0,0}\CO}%
}}}}
\put(3376,-5236){\makebox(0,0)[lb]{\smash{{\SetFigFont{9}{10.8}{\familydefault}{\mddefault}{\updefault}{\color[rgb]{0,0,0}\a{name}}%
}}}}
\put(3301,-5011){\makebox(0,0)[lb]{\smash{{\SetFigFont{7}{8.4}{\familydefault}{\mddefault}{\updefault}{\color[rgb]{0,0,0}\a{code}}%
}}}}
\put(1201,-5236){\makebox(0,0)[b]{\smash{{\SetFigFont{9}{10.8}{\familydefault}{\mddefault}{\updefault}{\color[rgb]{1,1,1}6.5}%
}}}}
\put(1576,-5236){\makebox(0,0)[b]{\smash{{\SetFigFont{9}{10.8}{\familydefault}{\mddefault}{\updefault}{\color[rgb]{1,1,1}1/4}%
}}}}
\put(751,-4861){\makebox(0,0)[b]{\smash{{\SetFigFont{9}{10.8}{\familydefault}{\mddefault}{\updefault}{\color[rgb]{0,0,0}\CO}%
}}}}
\put(1651,-4861){\makebox(0,0)[b]{\smash{{\SetFigFont{9}{10.8}{\familydefault}{\mddefault}{\updefault}{\color[rgb]{0,0,0}\ME}%
}}}}
\put(1051,-5011){\makebox(0,0)[rb]{\smash{{\SetFigFont{7}{8.4}{\familydefault}{\mddefault}{\updefault}{\color[rgb]{0,0,0}\a{code}}%
}}}}
\put(1351,-5011){\makebox(0,0)[lb]{\smash{{\SetFigFont{7}{8.4}{\familydefault}{\mddefault}{\updefault}{\color[rgb]{0,0,0}\a{country}}%
}}}}
\put(1951,-5011){\makebox(0,0)[rb]{\smash{{\SetFigFont{7}{8.4}{\familydefault}{\mddefault}{\updefault}{\color[rgb]{0,0,0}\a{org}}%
}}}}
\put(1426,239){\makebox(0,0)[b]{\smash{{\SetFigFont{9}{10.8}{\familydefault}{\mddefault}{\updefault}{\color[rgb]{1,1,1}1}%
}}}}
\put(751,-211){\makebox(0,0)[b]{\smash{{\SetFigFont{9}{10.8}{\familydefault}{\mddefault}{\updefault}{\color[rgb]{0,0,0}\CO}%
}}}}
\put(1051,-361){\makebox(0,0)[rb]{\smash{{\SetFigFont{7}{8.4}{\familydefault}{\mddefault}{\updefault}{\color[rgb]{0,0,0}\a{code}}%
}}}}
\put(1426,-586){\makebox(0,0)[lb]{\smash{{\SetFigFont{9}{10.8}{\familydefault}{\mddefault}{\updefault}{\color[rgb]{0,0,0}\a{gov}}%
}}}}
\put(1651,-211){\makebox(0,0)[b]{\smash{{\SetFigFont{9}{10.8}{\familydefault}{\mddefault}{\updefault}{\color[rgb]{0,0,0}\PO}%
}}}}
\put(1351,-361){\makebox(0,0)[lb]{\smash{{\SetFigFont{7}{8.4}{\familydefault}{\mddefault}{\updefault}{\color[rgb]{0,0,0}\a{country}}%
}}}}
\put(2326,-511){\makebox(0,0)[b]{\smash{{\SetFigFont{9}{10.8}{\familydefault}{\mddefault}{\updefault}{\color[rgb]{1,1,1}1/2}%
}}}}
\put(526,164){\makebox(0,0)[lb]{\smash{{\SetFigFont{9}{10.8}{\familydefault}{\mddefault}{\updefault}{\color[rgb]{0,0,0}\a{name}}%
}}}}
\put(751,539){\makebox(0,0)[b]{\smash{{\SetFigFont{9}{10.8}{\familydefault}{\mddefault}{\updefault}{\color[rgb]{0,0,0}\CO}%
}}}}
\put(4051,464){\makebox(0,0)[rb]{\smash{{\SetFigFont{9}{10.8}{\familydefault}{\mddefault}{\updefault}{\color[rgb]{0,0,0}$(s_1,A_1)$}%
}}}}
\put(1426,539){\makebox(0,0)[b]{\smash{{\SetFigFont{9}{10.8}{\familydefault}{\mddefault}{\updefault}{\color[rgb]{1,1,1}0.0}%
}}}}
\put(4051,-211){\makebox(0,0)[rb]{\smash{{\SetFigFont{9}{10.8}{\familydefault}{\mddefault}{\updefault}{\color[rgb]{0,0,0}$(s_2,A_2)$}%
}}}}
\put(2326,-211){\makebox(0,0)[b]{\smash{{\SetFigFont{9}{10.8}{\familydefault}{\mddefault}{\updefault}{\color[rgb]{1,1,1}12.3}%
}}}}
\put(4051,-961){\makebox(0,0)[rb]{\smash{{\SetFigFont{9}{10.8}{\familydefault}{\mddefault}{\updefault}{\color[rgb]{0,0,0}$(s_3,A_3)$}%
}}}}
\put(751,-1711){\makebox(0,0)[b]{\smash{{\SetFigFont{9}{10.8}{\familydefault}{\mddefault}{\updefault}{\color[rgb]{0,0,0}\CO}%
}}}}
\put(1051,-1861){\makebox(0,0)[rb]{\smash{{\SetFigFont{7}{8.4}{\familydefault}{\mddefault}{\updefault}{\color[rgb]{0,0,0}\a{code}}%
}}}}
\put(1351,-1861){\makebox(0,0)[lb]{\smash{{\SetFigFont{7}{8.4}{\familydefault}{\mddefault}{\updefault}{\color[rgb]{0,0,0}\a{country}}%
}}}}
\put(1726,-1711){\makebox(0,0)[b]{\smash{{\SetFigFont{9}{10.8}{\familydefault}{\mddefault}{\updefault}{\color[rgb]{0,0,0}\PO}%
}}}}
\put(2101,-1861){\makebox(0,0)[rb]{\smash{{\SetFigFont{7}{8.4}{\familydefault}{\mddefault}{\updefault}{\color[rgb]{0,0,0}\a{depend}}%
}}}}
\put(2701,-1711){\makebox(0,0)[b]{\smash{{\SetFigFont{9}{10.8}{\familydefault}{\mddefault}{\updefault}{\color[rgb]{0,0,0}\CO}%
}}}}
\put(2401,-1861){\makebox(0,0)[lb]{\smash{{\SetFigFont{7}{8.4}{\familydefault}{\mddefault}{\updefault}{\color[rgb]{0,0,0}\a{code}}%
}}}}
\put(4051,-1711){\makebox(0,0)[rb]{\smash{{\SetFigFont{9}{10.8}{\familydefault}{\mddefault}{\updefault}{\color[rgb]{0,0,0}$(s_4,A_4)$}%
}}}}
\put(751,-2461){\makebox(0,0)[b]{\smash{{\SetFigFont{9}{10.8}{\familydefault}{\mddefault}{\updefault}{\color[rgb]{0,0,0}\CO}%
}}}}
\put(1051,-2611){\makebox(0,0)[rb]{\smash{{\SetFigFont{7}{8.4}{\familydefault}{\mddefault}{\updefault}{\color[rgb]{0,0,0}\a{code}}%
}}}}
\put(1351,-2611){\makebox(0,0)[lb]{\smash{{\SetFigFont{7}{8.4}{\familydefault}{\mddefault}{\updefault}{\color[rgb]{0,0,0}\a{country}}%
}}}}
\put(2101,-2611){\makebox(0,0)[rb]{\smash{{\SetFigFont{7}{8.4}{\familydefault}{\mddefault}{\updefault}{\color[rgb]{0,0,0}\a{depend}}%
}}}}
\put(2701,-2461){\makebox(0,0)[b]{\smash{{\SetFigFont{9}{10.8}{\familydefault}{\mddefault}{\updefault}{\color[rgb]{0,0,0}\CO}%
}}}}
\put(2476,-2836){\makebox(0,0)[lb]{\smash{{\SetFigFont{9}{10.8}{\familydefault}{\mddefault}{\updefault}{\color[rgb]{0,0,0}\a{name}}%
}}}}
\put(2401,-2611){\makebox(0,0)[lb]{\smash{{\SetFigFont{7}{8.4}{\familydefault}{\mddefault}{\updefault}{\color[rgb]{0,0,0}\a{code}}%
}}}}
\put(1726,-2461){\makebox(0,0)[b]{\smash{{\SetFigFont{9}{10.8}{\familydefault}{\mddefault}{\updefault}{\color[rgb]{0,0,0}\PO}%
}}}}
\put(4051,-2461){\makebox(0,0)[rb]{\smash{{\SetFigFont{9}{10.8}{\familydefault}{\mddefault}{\updefault}{\color[rgb]{0,0,0}$(s_5,A_5)$}%
}}}}
\put(4051,-3211){\makebox(0,0)[rb]{\smash{{\SetFigFont{9}{10.8}{\familydefault}{\mddefault}{\updefault}{\color[rgb]{0,0,0}$(s_6,A_6)$}%
}}}}
\put(4051,-3961){\makebox(0,0)[rb]{\smash{{\SetFigFont{9}{10.8}{\familydefault}{\mddefault}{\updefault}{\color[rgb]{0,0,0}$(s_7,A_7)$}%
}}}}
\put(4051,-4636){\makebox(0,0)[rb]{\smash{{\SetFigFont{9}{10.8}{\familydefault}{\mddefault}{\updefault}{\color[rgb]{0,0,0}$(s_8,A_8)$}%
}}}}
\put(2476,-2086){\makebox(0,0)[lb]{\smash{{\SetFigFont{9}{10.8}{\familydefault}{\mddefault}{\updefault}{\color[rgb]{0,0,0}\a{capital}}%
}}}}
\put(751,-961){\makebox(0,0)[b]{\smash{{\SetFigFont{9}{10.8}{\familydefault}{\mddefault}{\updefault}{\color[rgb]{0,0,0}\CO}%
}}}}
\put(1051,-1111){\makebox(0,0)[rb]{\smash{{\SetFigFont{7}{8.4}{\familydefault}{\mddefault}{\updefault}{\color[rgb]{0,0,0}\a{code}}%
}}}}
\put(1426,-1336){\makebox(0,0)[lb]{\smash{{\SetFigFont{9}{10.8}{\familydefault}{\mddefault}{\updefault}{\color[rgb]{0,0,0}\a{indep}}%
}}}}
\put(1651,-961){\makebox(0,0)[b]{\smash{{\SetFigFont{9}{10.8}{\familydefault}{\mddefault}{\updefault}{\color[rgb]{0,0,0}\PO}%
}}}}
\put(1351,-1111){\makebox(0,0)[lb]{\smash{{\SetFigFont{7}{8.4}{\familydefault}{\mddefault}{\updefault}{\color[rgb]{0,0,0}\a{country}}%
}}}}
\put(2348,-1261){\makebox(0,0)[b]{\smash{{\SetFigFont{9}{10.8}{\familydefault}{\mddefault}{\updefault}{\color[rgb]{1,1,1}1/2}%
}}}}
\put(2348,-961){\makebox(0,0)[b]{\smash{{\SetFigFont{9}{10.8}{\familydefault}{\mddefault}{\updefault}{\color[rgb]{1,1,1}5.0}%
}}}}
\put(3376,-1711){\makebox(0,0)[b]{\smash{{\SetFigFont{9}{10.8}{\familydefault}{\mddefault}{\updefault}{\color[rgb]{1,1,1}15.5}%
}}}}
\put(3376,-2011){\makebox(0,0)[b]{\smash{{\SetFigFont{9}{10.8}{\familydefault}{\mddefault}{\updefault}{\color[rgb]{1,1,1}1/3}%
}}}}
\end{picture}%

%% file: section_problem_def.tex
\section{Problem Definition}\label{sec:problem}
Recall that \forward uses all targeted walk schemes of length at most 
$\ell_\text{max}$ for training its embedding. That is, all of $\tws(R,\ell_{\text{max}})$. Our conjecture in this work is that some targeted walk schemes are considerably more useful than others for the task of embedding, and furthermore, that a small subset of $\tws(R, \ell_\text{max})$ suffices for achieving the quality of the full set. If so, then we can considerably reduce the training time by focusing on just a few walk schemes. It is also perceivable that, with a small yet valuable collection of targeted walk schemes, we can surpass the quality of the original set for the same training time. 

Consequently, our goal in this work is to find a subset $\mathcal{T}'$
of $\tws(R, \ell_\text{max})$ such that the training of 
\forward on $\mathcal{T}'$
rather than $\tws(R, \ell_\text{max})$ is more effective. In particular, we would like to find a $\mathcal{T}'$ such that:
 \e{(1)} $\mathcal{T}'$ is small compared to $\tws(R, \ell_\text{max})$, and importantly an epoch of training with
    $\mathcal{T}'$ is considerably faster than training with $\tws(R, \ell_\text{max})$; \e{and (2)} the quality (on downstream tasks) of the embedding resulting  from $\mathcal{T}'$ is high compared to $\tws(R, \ell_\text{max})$. 
Hence, we would like to find a 
$\mathcal{T}'$ that would enable us to train considerably faster without a penalty of loss in quality.

\begin{example}
One way of selecting a subset of $\tws(R, \ell_\text{max})$ is to score each and take the top-$k$ for a desired number $k$ of schemes. For illustration, 
Figure~\ref{fig:walks} depicts two scores (written in filled ellipses) beside each targeted walk scheme. The first (blue) score is what we later define as the \e{kernel variance} score. The second (green) is a simplistic score that we use as a baseline; this is the reciprocal of the scheme's length (e.g., it is $1/3$ for $(s_7,A_7)$ since $s_7$ is of length three). If we use the second scoring function, then we take the shortest of the targeted walk schemes (and apply tie-breaking if needed).
\qed
\end{example}

%% file: section_strategies.tex
\section{Strategies For Scheme Selection}\label{sec:strategies}
 Recall that our goal is to study how walk schemes should be selected in order to establish a proper balance between the execution cost and the quality of the embedding. 
 Our general approach is to start with a large collection of walk schemes (i.e., the initial one of \forward) and eliminate walk schemes one by one. 
In this section, we propose several strategies for such  elimination. 

Technically, a \e{strategy} $T$ assigns to every $(s,A) \in \tws(R, \ell_\text{max})$ a number
$\score_T(s,A,D)$ for the database $D$, where a higher score means that the targeted walk scheme is considered more valuable. When we select $k$ schemes to eliminate, we select the bottom $k$ according to the score.
We propose strategies that fall into three categories: 
\e{(1)} 
The \emph{\forward-less strategies} determine  $\score_T(s,A,D)$ based on an evaluation that does not require the actual execution of \forward.
\e{(2)} 
In the \emph{light training} strategies, we run  \forward in some limited and light fashion in order to determine $\score_T(s,A,D)$. 
\e{(3)} The strategy of \e{online scheme elimination} incorporates the scheme selection in the actual embedding phase (using \forward) while targeted walk schemes are eliminated during the epochs of the training; hence, the values $\score_T(s,A,D)$ are computed multiple times during the embedding phase.

\def\len{\mathsf{len}}
\subsection{\forward-Less Strategies}
This category includes simplistic baseline strategies such as the \e{length} of the scheme,
which is illustrated in Figure~\ref{fig:walks} in the green ellipses. Next, we describe two more involved strategies: \e{mutual information} and \e{kernel variance}.

\subsubsection*{Mutual Information} \label{sec:Mutual Information}
A walk scheme $s$, as defined in~\eqref{eq:walkscheme}, induces a probability distribution over random walks, which are  sequences $(f_1,\dots,f_\ell)$ of facts in $R_1,\dots,R_\ell$, respectively.
For $i=1,\dots,\ell$, let $X_i$ be the random variable that takes the 
random fact $f_i$. 
%
%
Let $p(f_i,f_{i+1})\in[0,1]$ denote the marginal joint distribution of the variables $X_i$ and $X_{i+1}$. Recall that the \e{mutual information} of $X_i$ and $X_{i+1}$ is given by:
\begin{equation}\label{eq:mutual_inf}
    I(X_i;X_{i+1})\defeq \displaystyle\sum_{\substack{f_i\in X_i \,,\, f_{i+1}\in X_{i+1}}} p(f_i,f_{i+1})\log{\frac{p(f_i,f_{i+1})}{p(f_i)p(f_{i+1})}}
\end{equation}
We estimate the probabilities $p(f_i,f_{i+1})$ as the empirical probabilities, that is, their probability in our samples. 

\def\scoremi{\score_{\mathsf{mi}}}
We score a targeted walk scheme by the minimal mutual information along the walk.  
Formally, we have the following score:
\begin{equation}\label{eq:cond_ent_score}
    \scoremi(s,A,D) \defeq -\min_{0 \leq i < \ell+1} I(X_i|X_{i+1})
\end{equation}
Note that the minus sign means that we favor schemes with small mutual information. The rationale is that small mutual information encourages the embedding to capture less predicted distributions. 
As we show later, this rationale is consistent with our experiments.

\subsubsection*{Kernel Variance} \label{{Kernel Variance}}
\def\scorekv{\score_{\mathsf{kvar}}}

The measure \e{kernel variance} is, intuitively, one that favours targeted walk schemes where different start tuples are associated with varied distributions, and so, the embedding of \forward is encouraged to distinguish between these starting tuples.
Formally, for a targeted walk scheme $(s,A)$
we define the \e{kernel variance} score, denoted $\scorekv(s,A,D)$, as the variance of the expected kernel distance between the distributions induced by $s$ and $A$ when starting with random $f$ and $f'$ in the source of $s$.
\begin{equation}
    \label{eq:kvar_score}
    \scorekv(s,A,D) \defeq \var_{f,f'}\big(\:\text{KD}(d_{s,f}[A], d_{s,f'}[A])\:\big)
\end{equation}
We can estimate $\scorekv(s,A,D)$ from a pool of samples of the form $(f,f',g,g')$, where $g$ and $g'$ are destinations of random walks of $s$ starting at $f$ and $f'$, respectively. The number of samples is a hyper-parameter, and in our experiments, we used 10\% of the number of samples that \forward uses for computing its embedding. 


\subsection{Light Training}
If we run the full embedding over all schemes, then we can track the experience of the embedding algorithm with respect to the different schemes. We can then score the contribution of the targeted walk scheme based on the accumulated loss incurred by instances of the targeted walk scheme. While this approach works well empirically, it beats an important purpose of the scheme reduction---to reduce the execution cost of the embedding. So, our approach is to apply this idea lightly, that is, we run the training phase but either stop it early (``Early Termination''), run it only a small sample of the data (``Sample''), or combine between the two. In what follows, we present a precise materialization of these alternatives. 

\def\meanloss{\mathsf{\overline{L_{s,A,D,e}}}}
\def\scoreloss{\score_{\mathsf{loss}}}


For the next strategies, we need some notation. Recall that the training of \forward involves several steps:
\e{(1)} For each $R$-fact $f$ and $(s,A) \in \tws(R, \ell_\text{max})$, we sample tuples of the form $(f, f', s, A, g, g')$.
    \e{(2)} For each $(s,A) \in \tws(R, \ell_\text{max})$, the $R_\ell$ facts $g$ and $g'$ are the destinations of random walks with scheme $s$ sampled for $f$ and $f'$, respectively.
    \e{(3)} Using SGD, we minimize the loss $L_{f, f', s, A, g, g'}$ according to Equation~\eqref{eq:loss}.
Note that the loss $L_{f, f', s, A, g, g'}$ is computed for each choice of $(s,A)$, $f, f', g$ and $g'$. 
We denote by $L_i(s,A,D)$ the mean
of the losses $L_{f, f', s, A, g, g'}$ for each combination of $s$, $A$ and $D$, until 
the $i$th epoch.


\subsubsection*{Single-Epoch Training} \label{Stop and Restart}
\def\scoreoneep{\score_{\mathsf{1ep}}}



This score is the mean loss after an epoch: 
\begin{equation}\label{eq:1ep_score}
    \scoreoneep(s,A,D) \defeq 
    L_1(s,A,D)
\end{equation}
Note that the computation of this score is disconnected from the training phase, where we train from scratch without accounting for the epoch we spend for computing. 


\subsubsection*{Sampling} \label{sec:sampling}  
\def\scoresampling{\score_{\mathsf{smpl}}}

The second type of light training uses a  sample of the database. 
It is crucial to have a sample where the walk schemes materialize, so we need to carefully select the sample. We create a sample $D'$ of the database $D$ in two steps:
    \e{(1)} randomly select a small set $F$ of facts from $D$;
    \e{(2)} insert to $D'$ all of the facts of $D$ that are reachable from $F$ through paths of foreign keys (in both directions).
Yet, with this approach, we still suffer from walk schemes with too few instances in $D'$. Thus, we construct $F$ by considering every scheme $s$ and selecting random facts from those that  participate in paths of $s$. Finally, we run ten epochs and take the average loss:
\begin{equation}\label{eq:sampling_score}
    \scoresampling(s,A,D) \defeq L_{10}(s,A,D')
\end{equation}

\subsection{Online Scheme Elimination} \label{sec:online}
This strategy is similar to the single-epoch training, except that we apply it iteratively during training, where in each epoch we remove the bottom-$k$ schemes according to the score computed for that epoch according to Equation~\eqref{eq:1ep_score}. This way, we are potentially allowing to account for schemes that become more important once other schemes are removed (in earlier epochs). 
Note that online scheme elimination is different from the light-training approach in the sense that the latter is performed as a pre-processing step that takes place \e{before} the embedding computation, while the former is performed \e{as part} of the embedding computation. 

%% file: section_exp.tex
\newcommand{\random}{\textsf{Random}\xspace}
\newcommand{\length}{\textsf{Length}\xspace}
\newcommand{\kvar}{\textsf{KVar}\xspace}
\newcommand{\mi}{\textsf{MI}\xspace}
\newcommand{\oneepoch}{\textsf{1epoch}\xspace}
\newcommand{\sampling}{\textsf{Sampling}\xspace}
\newcommand{\onlineschemeelimination}{\textsf{Online}\xspace}
\def\T{\mathcal{T}}



\begin{table}[t]
\small
 \caption{\label{tab:Datasets}Datasets used in the experiments. ``\#TWS'' is the cardinality of $\tws(R, \ell_{\max})$, that is, the number of targeted walk schemes of length up to $\ell_{\max}$. ``Avglen'' is the average length of a walk scheme in $\tws(R, \ell_{\max})$.
 }
\centering
\begin{tabular}{|c|c|c|c|c|c|c|}
 \hline
\cc {Dataset}  &\cc {\#Rel.} &\cc {\#Tuples} &\cc {\#Attr.} &\cc {\#TWS} &\cc {$\ell_{\max}$} &\cc {Avglen} \\ \hline
{\textbf{Mondial}} & $40$ & $21497$ & $167$ & $63$ & $3$ & $2.44$ \\ \hline
{\textbf{World}} & $3$ & $5411$ & $24$ & $60$ & $3$ & $1.68$ \\ \hline
{\textbf{Hepatitis}} & $7$ & $12927$ & $26$ & $21$ & $3$ & $1.73$ \\ \hline
{\textbf{Genes}} & $3$ & $6063$ & $15$ & $32$ & $3$  & $2.25$ \\ \hline
{\textbf{Mutagen.}} & $3$ & $10324$ & $14$ & $58$ & $4$ & $3.10$ \\ \hline

\end{tabular}
\end{table}

\section{Experimental Evaluation}\label{sec:experiments}
The goal of our experimental study is threefold. First, we evaluate the effectiveness of the scheme-selection strategies in terms of the quality of the embedding and the execution cost of its computation. 
Second, we compare the strategies. 
Third, we study the impact of the strategy on the performance in a dynamic setting where new tuples are repeatedly inserted, and their embedding is computed without changing the embedding of existing tuples.

To evaluate an embedding, we
adopt the common approach of measuring the accuracy on downstream predictions. Hence, we measure the running time of the embedding algorithm, namely \forward, and the quality of a learned model for the downstream task. 
We focus on multi-relational data and use the same databases and tasks that were used for the evaluation of \forward~\cite{icde2023dynamic}.

\subsection{Experimental Setup}

\subsubsection{Datasets and Tasks}\label{sec:exp:tasks}
Information about the datasets and downstream tasks of our experiments is given in Tables~\ref{tab:Datasets} and~\ref{tab:Downstream_tasks}, respectively. 
Each dataset is a database of multiple relations (with the number of relations given in the ``\#Rel.'' column of \Cref{tab:Datasets}), and the task is to predict the content of an attribute of one of the relations. Hereafter, we refer to this relation as the \e{prediction relation}. In different downstream tasks on the same dataset, the \e{prediction attribute} is changed in the prediction relation to the one we aim to predict. 
Importantly, the predicted attribute is excluded from the database throughout the entire embedding phase, and it is seen by neither \forward nor the walk-scheme selector.

\underline{Mondial}
contains information from multiple geographical resources~\cite{mondial}. We used multiple attributes for prediction tasks on this dataset: \att{religion} (Christian or not), 
\att{continent}, \att{infant mortality g40} (whether the rate is lower than forty per thousand), \att{gdp g8e3} (whether GDP is lower than \$8000M), and \att{inflation g6}(whether the inflation rate is lower than 6\%). 

\underline{World}
has geographical data on states and their cities~\cite{World}. The task is to predict the \att{continent} of a country.
The dataset contains 40 different relations with a total of 167 attributes and 21,497 tuples. We use the whole database and use the \rel{Target} relation as the prediction relation as previously done by Bina et al.~\cite{DBLP:journals/dss/BinaSCQX13}.

{\begin{table}
\small
 \caption{\label{tab:Downstream_tasks}Downstream tasks. ``CC'' (Common Class) is the frequency of the common value of the predicted bit.
 }
\centering
\renewcommand{\arraystretch}{1.2}
\begin{tabular}{|c|c|c|c|c|}
 \hline
\cc {Downstream task} & \cc {Pred.~Rel.} &\cc {Pred.~Attr.} &\cc {\#Samples} &\cc {CC} \\ \hline
{\textbf{M.-Religion}} & \multirow{5}{*}{Target} & religion & $206$ & {$54.8\%$} \\ \cline{1-1} \cline{3-5} 
{\textbf{M.-Continent}} & & continent & $242$ & {$22.7\%$} \\ \cline{1-1} \cline{3-5} 
{\textbf{M.-Infant Mort.}} & & infant g40 & 238 & {$60.5\%$} \\ \cline{1-1} \cline{3-5} 
{\textbf{M.-GDP}} & & gdp g8e3 & $238$ & {$50.0\%$} \\ \cline{1-1} \cline{3-5} 
{\textbf{M.-Inflation}} & & inflation g6 & $238$ & {$50.8\%$} \\ \hline
{\textbf{World}} & Country & continent & $239$ & $24.2\%$ \\ \hline
{\textbf{Hepatitis}}  & Dispat & type & $500$ & $58.8\%$ \\ \hline
{\textbf{Genes}} & Classific. & localization & $862$ & $42.5\%$ \\ \hline
{\textbf{Mutagenesis}} & Molecule & mutagenic & $188$ & $66.4\%$ \\ \hline
\end{tabular}
\end{table}}

\underline{Hepatitis} is from the 2002 ECML/PKDD Discovery Challenge.\footnote{\url{https://sorry.vse.cz/~berka/challenge/PAST/}. We use the modified version of~\cite{DBLP:conf/kdd/NevilleJFH03}.} The task is to predict the \att{type} column, which is either \e{Hepatitis B} or \e{Hepatitis C} based on medical examinations. There are in total 206 instances of the former and 484 cases of the latter. The relation with the predicted column contains, in addition to the type classification, the age, gender and identifier of the patient.  The other relations contain the rest of the medical data. The dataset contains 7 relations with a total of 26 attributes and 12,927 tuples.   

\underline{Genes}~\cite{10.1145/507515.507523} contains genomic and drug-design data. The task is to predict the \att{localization} of the gene, based on biological data, with 15 different labels. The prediction relation contains only the class and an identifier for the gene, while the rest contain biological data such as the function, gene type, cellular location, and the expression correlation between genes. The dataset contains 3 relations with a total of 15 attributes and 6,063 tuples. (We deleted two tuples with a unique class to prevent split in-balances during cross-validation.)
%

\input{plots/subset_of_mega_figure}

\underline{Mutagenesis}
contains data on the mutagenicity of molecules on Salmonella Typhimurium~\cite{doi:10.1021/jm00106a046}. The task is to predict the mutagenicity 
of molecules, based on chemical properties of the molecule, with 122 positive samples and 63 negative ones. The prediction relation contains the binary class, molecule ID, and some of the chemical data, while the other relations contain more chemical data and information about the relations between the molecules. The dataset contains 3 relations with a total of 14 attributes and 10,324 tuples. 



\subsubsection{Compared Strategies}\label{sec:methods}
We compare the following strategies.
%
\underline{\random} eliminates random schemes.
\underline{\length} eliminates the schemes with the longest length.
%
\underline{\kvar} eliminates the schemes with the least $\scorekv$ (kernel variance), as defined in Equation~\eqref{eq:kvar_score}.
\underline{\mi} eliminates the schemes with the least $\scoremi$ (mutual information) as defined in  Equation~\eqref{eq:mutual_inf}.
%
\underline{\oneepoch}
eliminate the schemes with the least $\scoreoneep$ (one epoch), as defined in Equation~\eqref{eq:1ep_score}. 
%
\underline{\sampling} eliminate the schemes with the least  $\scoresampling$ (sampling), as defined in Equation~\eqref{eq:sampling_score}. 
Recall that in this strategy, we run the algorithm for ten epochs on the sample $D'$ of the database $D$; afterward, we run \forward from scratch on the remaining schemes. 
%
\underline{\onlineschemeelimination} is the online scheme-elimination approach is described in Section~\ref{sec:online}. 

We have a full separation between the embedding process and the downstream task: 
we generate the embedding independently from the task (as opposed to training for the task) and then use these embeddings as the input to a downstream classifier (that sees only the embeddings and none of the other database information).


\subsubsection{Programming}\label{sec:exp:setup}
For a tuple embedding $\varphi$, we denote by $\alpha(\varphi) \in [0, 1]$ the mean accuracy achieved by an SVM (Scikit-learn's SVC implementation) trained as a classifier that takes the tuple embeddings in $\varphi$ as input, and learns to predict the target of the downstream task. 
The results are given for a ten-fold cross validation.
We fix one 
split for each dataset and task where this cross validation is performed, and we do so for every evaluated embedding.

The value of $\alpha(\varphi)$ is our primary metric for the quality of an embedding $\varphi$.
We usually train five embeddings $\Phi=\{\varphi_{1},\dots,\varphi_{5}\}$ for each configuration, each with a different seed.
We use the mean cross-validated accuracy across the five as a measure of the expected embedding quality:
$\alpha(\Phi) = \mean_{1 \leq i \leq 5} a(\varphi_i)$.
We study how this expectation develops over time for different strategies and selection ratios. More formally, let $\mathcal{T}$ be a scheme selection strategy  
and let $r \in [0,1]$ be the ratio of schemes  used for training.
By $\Phi(\mathcal{T}, r, t)$ we denote the set of embeddings obtained after training five \forward embeddings for $t$ seconds on the targeted walk schemes reduced using $\mathcal{T}$ by ratio $1-r$.
We evaluate the expected accuracy after each epoch, thus the set of times $t$ where we record $\alpha(\Phi(\mathcal{T}, r, t))$ is determined by the time it takes to complete an epoch.

We run all the experiments 
a server with two Intel Xeon Gold 6130 processors, 512 MB RAM, and an NVIDIA QUADRO RTX 6000 GPU with 24 GB memory. 

\eat{ 
\subsection{Performance of Individual Strategies}\label{sec:exp:results-individual}
\begin{figure*}
\ifx\QUICK\undefined
\centering
\rotatebox[origin=t]{90}{\small~\textbf{\emph{Genes}}}
     \begin{subfigure}{0.30\textwidth}
          \centering
          \caption*{\textbf{\kvar}}
          \resizebox{\linewidth}{!}{\input{plots/pgf_plots/genes_k_var.tex}} 
          \label{fig:A_with_a_star}
     \end{subfigure}
     \begin{subfigure}{0.30\textwidth}
          \centering
          \caption*{\textbf{\oneepoch}} 
          \resizebox{\linewidth}{!}{\input{plots/pgf_plots/genes_1ep.tex}}  
          \label{fig:B}
     \end{subfigure}
     \begin{subfigure}{0.30\textwidth}
          \centering
          \caption*{\textbf{\random}}
          \resizebox{\linewidth}{!}{\input{plots/pgf_plots/genes_random.tex}}  
          \label{fig:C}
     \end{subfigure}
     
\centering
\rotatebox[origin=t]{90}{\small~\textbf{\emph{Mondial-Religion}}}
     \begin{subfigure}{0.30\textwidth}
          \centering
          \resizebox{\linewidth}{!}{\input{plots/pgf_plots/mondial_religion_k_var.tex}}  
          \caption*{\textbf{Time} (sec)}
          \label{fig:D}
     \end{subfigure}
     \begin{subfigure}{0.30\textwidth}
          \centering
          \resizebox{\linewidth}{!}{\input{plots/pgf_plots/mondial_religion_1ep.tex}} 
          \caption*{\textbf{Time} (sec)}
          \label{fig:E}
     \end{subfigure}
     \begin{subfigure}{0.30\textwidth}
          \centering
          \resizebox{\linewidth}{!}{\input{plots/pgf_plots/mondial_religion_random.tex}} 
          \caption*{\textbf{Time} (sec)}
          \label{fig:F}
     \end{subfigure}
\else\QUICK
\fi
\caption{
Performance on the downstream task as a function of time for different ratios of schemes used for training. At the end of each training epoch of \forward we record the time (x-axis) and  accuracy of the downstream task (y-axis).
}
\label{fig:subset_of_mega_figure}
 \end{figure*}
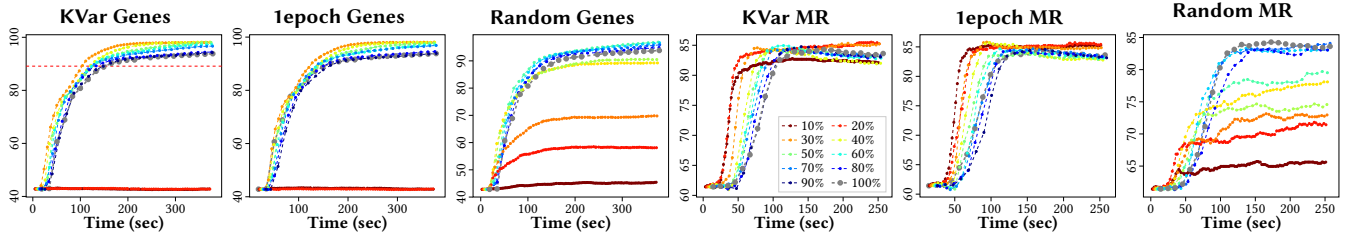
}

\newcommand{\cikmshortenscaley}{0.8} 
\eat{ 
\begin{figure}
\ifx\QUICK\undefined
\centering
     \begin{subfigure}{0.22\textwidth}
          \centering
          \caption*{\textbf{\emph{Mondial-Religion}}}
          \resizebox{\linewidth}{!}{\input{plots/short_pgf_plots_for_cikm/mondial_religion_k_var_short.tex}}
          \label{fig:D}
     \end{subfigure}
\vspace{-1.2em}
     \begin{subfigure}{0.22\textwidth}
          \centering
          \caption*{\textbf{\emph{Genes}}}\vspace{-0.6em}
          \resizebox{\linewidth}{!}{\input{plots/short_pgf_plots_for_cikm/genes_k_var_short.tex}} 
          \label{fig:A_with_a_star}
     \end{subfigure}
\vspace{-1.2em}
\centering
     \begin{subfigure}{0.22\textwidth}
          \centering
          \resizebox{\linewidth}{!}{\input{plots/short_pgf_plots_for_cikm/mondial_religion_1ep_short.tex}}
          \label{fig:E}
     \end{subfigure}
     \begin{subfigure}{0.22\textwidth}
          \centering
          \resizebox{\linewidth}{!}{\input{plots/short_pgf_plots_for_cikm/genes_1ep_short.tex}}  
          \label{fig:B}
     \end{subfigure}
\centering
     \begin{subfigure}{0.22\textwidth}
          \centering
          \resizebox{\linewidth}{!}{\input{plots/short_pgf_plots_for_cikm/mondial_religion_random_short.tex}} 
          \caption*{\textbf{Time} (sec)}
          \label{fig:F}
     \end{subfigure}
\vspace{-1.2em}
     \begin{subfigure}{0.22\textwidth}
          \centering
          \resizebox{\linewidth}{!}{\input{plots/short_pgf_plots_for_cikm/genes_random_short.tex}}  
          \caption*{\textbf{Time} (sec)}  
          \label{fig:C}
     \end{subfigure}
\else\QUICK
\fi
\caption{
Performance on the downstream task as a function of time for different ratios of schemes for training. At the end of each epoch of \forward, we record the time and  accuracy.
}
\label{fig:subset_of_mega_figure}
 \end{figure}
}


\subsection{Performance of Individual Strategies}\label{sec:exp:results-individual}

We first study how the embedding quality develops throughout the training of the embedding. We record the training time and the quality on the downstream task at the end of each epoch.
We study the progress when training on different subsets of the targeted walk schemes selected by the different strategies.
In Figure~\ref{fig:subset_of_mega_figure}, we provide results for the \emph{Mondial-Religion} (MR) and \emph{Genes} downstream tasks using the strategies \kvar (kernel variance), \oneepoch (one epoch), and \random.

\input{table-of-winners.tex}

In each sub-figure of Figure~\ref{fig:subset_of_mega_figure}, there are nine colored curves and one gray curve. 
Each colored curve represents one ratio of removed schemes from 10\% to 90\%. 
The gray curve shows the original \forward run when training on all targeted walk schemes.
The x-axis provides the training time $t$ in seconds and we plot the value of $\alpha(\Phi(\mathcal{T}, r, t))$ (i.e. the evaluation of the expected accuracy $t$ seconds) on the y-axis.
Intuitively, this shows how the embedding quality develops throughout the training of the embedding.


Across the experiments, we observe a range of behaviors.
First, the selection strategy has a significant influence on the result.
For example, the \random elimination strategy yields significantly worse performance when a large percentage of schemes is removed.
This justifies the design of more sophisticated strategies  that are able to yield embeddings of high quality even when a larger ratio of $\tws(R, \ell_\text{max})$ is removed.
One example of this is the \emph{Mondial-Religion} downstream task. 
Vanilla \forward with all schemes reaches $83\%$ accuracy on the downstream task.
When using \random to train on only $50\%$ of the  walk schemes, accuracy drops to $74\%$ while taking longer to converge.
Furthermore, when using the \random strategy to train on $20\%$ of the targeted walk schemes, we observe a more dramatic decrease in accuracy to $71\%$.
In contrast, when training with the \kvar strategy on $50\%$ or $20\%$ of the targeted walk schemes, the accuracy over the downstream task does not differ much from the original $83\%$.


For most of the combinations of tasks and strategies, a significant portion of the targeted walk schemes can be removed with a negligible decrease in quality. 
The speed of convergence naturally increases with the ratio of removed schemes.
An example of this behavior is the \emph{Mondial-Religion} task. 
Here, when we train on fewer schemes (using the \kvar and \oneepoch strategies), we reach the same $0.83$ accuracy with less training time. For instance, by selecting a fifth of the walk schemes using \kvar, we get to the full quality in about a third of the embedding time.
Another example is the \emph{Genes} downstream task. Here, the original \forward with all walk schemes takes $300$ seconds to reach $92.9\%$ accuracy over the downstream task. 
This time improves to $124.5$ and $148$ seconds when training on just $30\%$ of the targeted walk schemes selected by the \kvar and \oneepoch strategies, respectively.

There are several instances where the removal of schemes not only accelerates training but actually \e{increases} the downstream performance.
An example of this is the \emph{Genes} dataset.
Here, the accuracy achieved with all scheme selection strategies increases with the percentage of removed schemes until about 70\% of all schemes are discarded. 
More specifically, the accuracy on the downstream task is $93.8\%$
for the original \forward run with the full set of walk schemes; it goes up to $98\%$ when $70\%$ of the schemes are discarded using the \kvar and \oneepoch strategies.
Similar results are obtained on the \emph{Mondial-Inflation} and \emph{World} tasks, although the margin of improvement is different on these tasks.



\subsection{Comparison between Strategies}

We aim to better quantify the difference between strategies.
We first introduce additional metrics.
We establish a \e{high-performance threshold} by training five standard \forward{} embeddings $\Phi_{\text{FWD}}$ with all
targeted walk schemes.
We set 95\% of the expected cross-validated accuracy as the performance target on each downstream task:
    $\alpha^* = 0.95 \cdot \alpha(\Phi_{\text{FWD}})$.
In Figure~\ref{fig:subset_of_mega_figure} described next, the threshold is shown as a dashed red line.
We wish to study how each considered strategy for scheme selection affects the compute time needed to obtain an embedding of high quality.
To this end, we will measure how much training time is needed to reach the threshold $\alpha^*$ with each strategy. We will use the following definitions.

For a strategy $\T$ and a ratio $r$, we denote by $t^*(\mathcal{T},r)$ the earliest time it takes to reach the performance threshold:
$t^{*}(\mathcal{T},r) = \min \{t \: \mid \: \alpha(\Phi(\mathcal{T}, r, t)) \geq \alpha^* \}$.
For the strategy $\mathcal{T}$, we can further define a metric that is the shortest time it takes to reach the target quality over all tested ratios $r$:  
    $t^{*}(\mathcal{T}) = \min_r t^{*}(\mathcal{T},r)$.
This is the primary metric that we use for measuring the effectiveness of a strategy.





\begin{figure*}
\ifx\QUICK\undefined
\centering
\begin{subfigure}{0.18\textwidth}
    \caption*{\textbf{Genes}}\vspace{-0.4em}
    \resizebox{\linewidth}{!}{\input{plots/pgf_winners_croped/winners_genes.tex}} 
    \label{fig:winner1}
\end{subfigure}
\hfill
\begin{subfigure}{0.18\textwidth}
    \caption*{\textbf{World}}
    \resizebox{\linewidth}{!}{\input{plots/pgf_winners_croped/winners_world_B}} 
    \label{fig:winner2}
\end{subfigure}
\hfill
\begin{subfigure}{0.18\textwidth}
    \caption*{\textbf{M-Religion}}
    \resizebox{\linewidth}{!}{\input{plots/pgf_winners_croped/winners_mondial_original_target}} 
    \label{fig:winner3}
\end{subfigure}
\hfill
\begin{subfigure}{0.18\textwidth}
    \caption*{\textbf{M-Continent}}\vspace{-0.4em}
    \resizebox{\linewidth}{!}{\input{plots/pgf_winners_croped/winners_mondial_target_continent}} 
    \label{fig:winner4}
\end{subfigure}
\hfill
\begin{subfigure}{0.18\textwidth}
    \caption*{\textbf{M-GDP}}
    \resizebox{\linewidth}{!}{\input{plots/pgf_winners_croped/winners_mondial_target_GDP_g8e3}} 
    \label{fig:winner5}
\end{subfigure}
\hfill
\begin{subfigure}{0.18\textwidth}
    \caption*{\textbf{Legend}}
    \resizebox{\linewidth}{!}{\input{plots/pgf_winners/legend_winners}}
    \label{fig:winner6}
\end{subfigure}
\hfill
\begin{subfigure}{0.18\textwidth}
    \caption*{\textbf{M-Infant Mort.}}
    \resizebox{\linewidth}{!}{\input{plots/pgf_winners_croped/winners_mondial_target_infant_mortality_g40}} 
    \label{fig:winner6_5}
\end{subfigure}
\begin{subfigure}{0.18\textwidth}
    \caption*{\textbf{M-Inflation}}
    \resizebox{\linewidth}{!}{\input{plots/pgf_winners_croped/winners_mondial_target_Inflation_g6}} 
    \label{fig:winner7}
\end{subfigure}
\hfill
\begin{subfigure}{0.18\textwidth}
    \caption*{\textbf{Mutagenesis}}\vskip-0.25em
    \resizebox{\linewidth}{!}{\input{plots/pgf_winners_croped/winners_mutagenesis}} 
    \label{fig:winner8}
\end{subfigure}
\hfill
\begin{subfigure}{0.18\textwidth}
    \caption*{\textbf{Hepatitis}}
    \resizebox{\linewidth}{!}{\input{plots/pgf_winners_croped/winners_hepatitis}} 
    \label{fig:winner9}
\end{subfigure}
\fi
\vskip-1em
\caption {Comparison between different  strategies. For each downstream task, we provide the best learning curve of each strategy, that is, the first curve that reaches an accuracy of $\alpha^*$. Note that the color now denotes the selection strategy.
}
\label{fig:winners_fig}
\end{figure*}
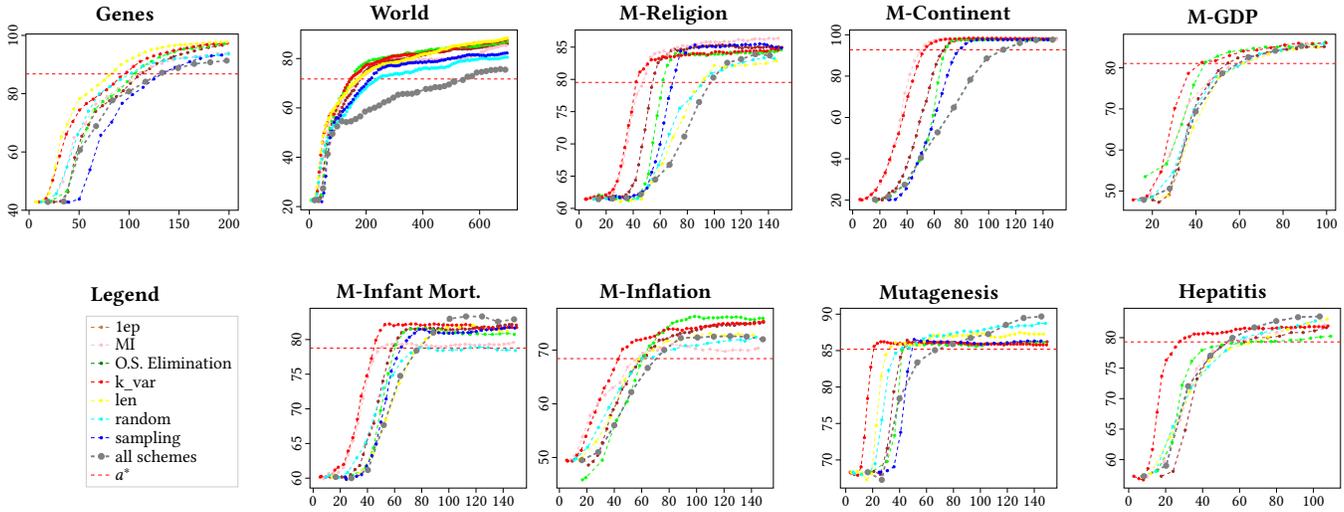

Table~\ref{tab:table_of_winners} provides the value of $t^{*}(\mathcal{T})$ for each strategy $\mathcal{T}$ and task.
Overall, \kvar yields the fastest training times on most tasks.
However, the exact performance of each strategy depends on the data. 
The \length strategy yields the best overall results on the Genes dataset, but \kvar performs substantially better than \length on all other tasks.
The \onlineschemeelimination strategy also yields the fastest training times on the \emph{World} and \emph{Modial-GDP} tasks.


To further study training different selection strategies, in Figure~\ref{fig:winners_fig} we provide the ``best'' learning curve of each strategy on each task.
More specifically, we plot the value of $\alpha(\Phi(\mathcal{T}, r^*(\mathcal{T}), t))$, where the ratio $r^*(\mathcal{T}) \in [0, 1]$ is $r^*(\mathcal{T}) = \arg\min_r t^{*}(\mathcal{T},r)$.
Note that we also provide the training curve of \forward{} when using all targeted walk schemes (gray).
When training with reduced scheme sets the training converges consistently speeds up.
As observed previously, the margin of the speedup depends on the scheme selection strategy.
The \kvar (kernel Variance) strategy (red) yields the fastest convergence on most downstream tasks.

The strong overall performance of the \kvar (kernel Variance) method suggests that it strikes the best balance between running efficiently and determining a good subset of schemes to remove.
Thus, it enables us to train on less targeted walk schemes and reach the same accuracy faster than simpler selection strategies.

\subsection{Embedding for Dynamic Databases}
Finally, we investigate how the reduction in the set of schemes affects the dynamic-database setting, where new tuples arrive and we need to compute an embedding of the new tuples without changing the embedding of existing tuples. (Note that tuple \emph{deletion} is not an issue in this setting since we simply leave intact the embedding of the remaining tuples; see \citet{icde2023dynamic} for a discussion on tuple insertion and its possible subtleties.)
For this experiment, we will use the dynamic extension of \forward~\cite{icde2023dynamic}.
This extension is one of the key motivations for the design of \forward{} as it is able to compute embeddings of new tuples unseen during training by solving a single linear system of equations.

Naturally, a useful scheme-selection strategy should not impair the quality of the new tuple embeddings.
Here, we will conduct a brief experiment to verify that this is indeed the case.
More specifically, we adopt the exact experimental setup of the dynamic experiments conducted by \citet{icde2023dynamic}.
This setup first deletes a part of the database and trains an embedding on the remaining data.
This embedding is then used to train a classifier for the downstream task.
Only after this, the removed tuples are inserted back into the database and the embedding is inductively extended to the new data.
These new embeddings can then be evaluated by measuring the accuracy of the downstream classifier on the inserted data.
By varying the ratio of data that is initially removed we can test the robustness of this dynamic embedding extension.

\begin{figure}[b]
\ifx\QUICK\undefined
    \centering
        \input{plots/dynamic/GenesDynamic.tex}
        \input{plots/dynamic/WorldDynamic.tex} 
        \\
     \hskip1em\input{plots/dynamic/MondialDynamic.tex}
        \input{plots/dynamic/MutagenesisDynamic.tex}
    
\else\QUICK
\fi
\vskip-1.5em
  \caption{Experiments on the dynamic setting: accuracy as a function of the percentage of inserted tuples for Node2Vec, \forward, and \forward with 60\% of the schemes selected by \kvar. The black line is the accuracy of the common class. }
  \label{fig:dynamic_results}
\end{figure}
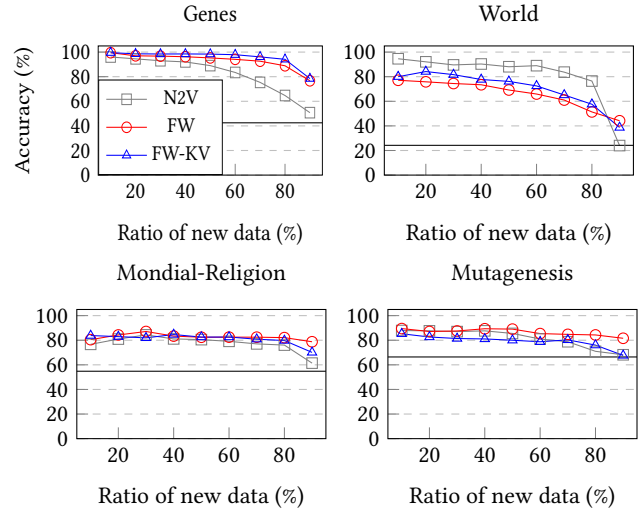

We train and then extend \forward{} where 60\% of all targeted walk schemes have been removed according to the \kvar strategy.
As baselines, we include the results of \forward{} with no schemes removed as well as the dynamic extension of node2vec, which was also proposed by T{\"{o}}nshoff et al.~\cite{icde2023dynamic}.
Figure~\ref{fig:dynamic_results} provides the results on four tasks: Genes, World, Mondial Religion and Mutagenesis.
We observe that the \forward{} version with the reduced set of schemes performs as well as the original \forward{} with all schemes.
On the Genes and World tasks the results actually \e{improve} slightly when only the selected 40\% of targeted walk schemes are used. There is, though, a slight decrease in the accuracy in the case of the  Mutagenesis dataset. 

Overall, we conclude that reduced scheme sets with a strong performance on a static database are also well-suited for a dynamic setting where new tuples are inserted over time.

%% file: plots/subset_of_mega_figure.tex
\begin{figure*}
\ifx\QUICK\undefined
\centering
     \begin{subfigure}{0.1625\textwidth}
          \centering
          \caption*{\textbf{\kvar Genes}}
          \vspace{-0.4em}
          \resizebox{\linewidth}{!}{\input{plots/pgf_plots/genes_k_var.tex}} 
          \label{fig:A_with_a_star}
     \end{subfigure}
     \begin{subfigure}{0.1625\textwidth}
          \centering
          \caption*{\textbf{\oneepoch Genes}}
          \vspace{-0.4em} 
          \resizebox{\linewidth}{!}{\input{plots/pgf_plots/genes_1ep.tex}}  
          \label{fig:B}
     \end{subfigure}
     \begin{subfigure}{0.1625\textwidth}
          \centering
          \caption*{\textbf{\random Genes}}
          \resizebox{\linewidth}{!}{\input{plots/pgf_plots/genes_random.tex}}  
          \label{fig:C}
     \end{subfigure}
     \begin{subfigure}{0.1625\textwidth}
          \centering
          \caption*{\textbf{\kvar MR}}
          \resizebox{\linewidth}{!}{\input{plots/pgf_plots/mondial_religion_k_var.tex}}  
          \label{fig:D}
     \end{subfigure}
     \begin{subfigure}{0.1625\textwidth}
          \centering
          \caption*{\textbf{\oneepoch MR}} 
          \resizebox{\linewidth}{!}{\input{plots/pgf_plots/mondial_religion_1ep.tex}} 
          \label{fig:E}
     \end{subfigure}
     \begin{subfigure}{0.1625\textwidth}
          \centering
          \caption*{\textbf{\random MR}}
          \resizebox{\linewidth}{!}{\input{plots/pgf_plots/mondial_religion_random.tex}} 
          \label{fig:F}
     \end{subfigure}
\else\QUICK
\fi
\\
\vspace{-2em}
\caption{
Performance on the downstream task as a function of time for different ratios of schemes used for training. At the end of each training epoch of \forward we record the time (x-axis) and  accuracy of the downstream task (y-axis).
}
\label{fig:subset_of_mega_figure}
\end{figure*}

%% file: plots/pgf_plots/genes_k_var.tex
\begin{tikzpicture}

\definecolor{blue}{RGB}{0,0,255}
\definecolor{cyan21255225}{RGB}{21,255,225}
\definecolor{darkgray176}{RGB}{176,176,176}
\definecolor{darkorange2551480}{RGB}{255,148,0}
\definecolor{dodgerblue0128255}{RGB}{0,128,255}
\definecolor{gray}{RGB}{128,128,128}
\definecolor{lightgreen124255121}{RGB}{124,255,121}
\definecolor{maroon12700}{RGB}{127,0,0}
\definecolor{navy00127}{RGB}{0,0,127}
\definecolor{red255290}{RGB}{255,29,0}
\definecolor{yellow22825518}{RGB}{228,255,18}

\begin{axis}[
tick align=outside,
tick pos=left,
x grid style={darkgray176},
xmin=-14.0328826880455, xmax=397.140934417248,
xtick style={color=black},
y grid style={darkgray176},
ymin=39.7767441860465, ymax=100.967441860465,
ytick style={color=black},
label style={font=\Huge},
xlabel={\textbf{Time (sec)}},
yticklabel style={rotate=90.0},
tick label style={font=\huge}
]
\addplot [semithick, red, dashed]
table {%
-14.0328826880455 89.1232558139535
397.140934417248 89.1232558139535
};
\addplot [ultra thick, gray, dashed, mark=*, mark size=2, mark options={solid}]
table {%
18.7418682098389 42.8372093023256
34.6578793048859 43
50.8279933929443 60.6744186046512
66.9683317661285 68.8604651162791
83.4322623729706 77.7209302325581
99.9514236450195 80.7674418604651
116.412863349915 84.1162790697674
132.916092252731 87.1860465116279
149.050437068939 88.7906976744186
165.219602060318 90.2325581395349
181.240503978729 90.7674418604651
197.671772670746 91.3488372093023
213.76312084198 91.5581395348837
213.76312084198 91.5581395348837
247.107247304916 92.3255813953488
263.459987735748 92.6744186046512
279.685011863708 92.7906976744186
296.458326721191 92.906976744186
313.223739433289 93.1860465116279
329.267318487167 93.2790697674419
345.646366643906 93.4883720930233
362.122578382492 93.6279069767442
378.451215457916 93.8139534883721
};
\addplot [maroon12700, dashed, mark=*, mark size=1, mark options={solid}]
table {%
4.65683627128601 42.8372093023256
5.37564210891724 42.8139534883721
8.91585721969605 42.8372093023256
12.4967307567596 42.8604651162791
15.9560532569885 42.8837209302325
19.551619386673 43.0930232558139
23.0817546367645 42.9302325581395
26.6246088027954 42.8604651162791
30.1970772743225 42.8837209302326
33.7174451828003 42.9302325581395
37.2591628551483 42.953488372093
40.814814043045 43.046511627907
44.3660357475281 43.0232558139535
47.9197825431824 43.0930232558139
51.4746498584747 43.1860465116279
55.0717413902283 43.1395348837209
58.6753057003021 43.0697674418605
62.1731298446655 43.0232558139535
65.6773921489716 42.9767441860465
69.1741401195526 43.0232558139535
72.7188216209412 42.953488372093
76.2816942691803 42.9302325581395
79.9515723705292 42.906976744186
83.5253832817078 42.9302325581395
87.0392501354218 42.906976744186
90.6432150363922 42.9302325581395
94.2452067375183 42.906976744186
97.85221824646 42.906976744186
101.323241043091 42.9302325581395
104.896925735474 42.9302325581395
108.417454004288 42.953488372093
111.898036813736 42.9767441860465
115.492794942856 42.953488372093
119.063938570023 42.953488372093
122.594718790054 42.906976744186
126.185467386246 42.8604651162791
128.262730312347 42.906976744186
131.819737052917 42.9302325581395
135.352217721939 42.906976744186
139.057763338089 42.8604651162791
142.556604480743 42.906976744186
146.071690750122 42.8604651162791
149.706871986389 42.8604651162791
153.451170778275 42.8372093023256
157.015243673325 42.8604651162791
160.587218236923 42.8372093023256
164.185776901245 42.7906976744186
167.747001647949 42.8139534883721
171.219807004929 42.8372093023256
174.822407150269 42.7906976744186
178.27899055481 42.7674418604651
181.816708230972 42.7441860465116
185.359631633759 42.7674418604651
188.922224235535 42.7906976744186
191.744671964645 42.7209302325581
195.269308423996 42.7674418604651
198.83927898407 42.6976744186046
202.472204637527 42.7209302325581
205.982473993301 42.6976744186046
209.514427852631 42.7441860465116
213.068478393555 42.7441860465116
216.612233924866 42.7441860465116
220.245177173615 42.6976744186046
223.819584178925 42.6744186046512
227.341230487823 42.6976744186046
230.87093834877 42.6976744186046
234.380801534653 42.6976744186046
237.93546833992 42.6046511627907
241.455200338364 42.6046511627907
244.989575481415 42.6046511627907
248.477643156052 42.6279069767442
250.643164587021 42.5581395348837
254.180692768097 42.6511627906977
257.737179899216 42.6511627906977
261.278067207336 42.6511627906977
264.736353397369 42.5813953488372
268.291567468643 42.6511627906977
271.880365753174 42.6046511627907
275.366758537292 42.6046511627907
278.936541700363 42.6279069767442
282.509337854385 42.6744186046512
286.101557779312 42.6279069767442
289.669760608673 42.6744186046512
293.162963056564 42.6511627906977
296.718877315521 42.6511627906977
300.332331085205 42.6976744186046
303.897773694992 42.6279069767442
307.447214746475 42.6279069767442
310.955476951599 42.6279069767442
314.468853092194 42.6511627906977
318.006241703033 42.6046511627907
321.483834218979 42.6511627906977
325.052871179581 42.6046511627907
328.637032651901 42.6511627906977
332.166634941101 42.6511627906977
335.807176160812 42.5813953488372
339.340653181076 42.6511627906977
342.966777467728 42.6511627906977
346.55522351265 42.6744186046512
350.102083492279 42.7441860465116
353.705636024475 42.7906976744186
357.173271512985 42.8139534883721
360.7540122509 42.8139534883721
364.287714147568 42.7441860465116
367.819392967224 42.7674418604651
371.293722724915 42.8372093023256
};
\addplot [red255290, dashed, mark=*, mark size=1, mark options={solid}]
table {%
5.8619592666626 42.8372093023256
8.83043675422669 42.8372093023256
13.4851257801056 42.953488372093
18.2548281669617 42.8139534883721
22.8845983505249 43.0697674418605
27.610197019577 43.046511627907
32.4035225391388 42.906976744186
37.1483335494995 42.9767441860465
41.8503079414368 42.953488372093
46.5348022460937 43.0232558139535
51.1765160560608 42.906976744186
55.7661699295044 42.9302325581395
60.4162755012512 42.9767441860465
65.2413229942322 42.9767441860465
69.9641365528107 42.9767441860465
74.665619468689 42.9302325581395
79.4792321205139 42.953488372093
84.1686374664307 42.953488372093
88.8887523651123 42.953488372093
93.5945940494537 43
98.3115624427795 43
100.149202537537 43
107.847269201279 43.046511627907
112.504111671448 43
117.196786260605 43.046511627907
121.898981952667 43.0232558139535
126.654943418503 43.0232558139535
131.268739509583 43
135.966292572021 43.0232558139535
140.734322404861 43.0232558139535
145.385327100754 43.046511627907
150.083221912384 43.0232558139535
154.832230901718 42.9767441860465
159.578937339783 43.0232558139535
164.371395874023 43.046511627907
169.034504890442 43.046511627907
173.791693115234 43.0232558139535
178.619537830353 42.9302325581395
183.284975099564 42.9767441860465
188.020134305954 42.906976744186
192.773446083069 42.9302325581395
197.530314588547 42.906976744186
199.454279088974 42.906976744186
206.991030931473 42.8372093023256
211.747831726074 42.8604651162791
216.570797872543 42.8139534883721
221.350337505341 42.8372093023256
223.29087562561 42.8139534883721
230.770955324173 42.7674418604651
235.409031772614 42.8139534883721
240.145981407166 42.7441860465116
244.911959981918 42.7674418604651
249.545217990875 42.7674418604651
254.283104991913 42.7906976744186
259.094871330261 42.7674418604651
263.896987056732 42.7674418604651
268.650044107437 42.6744186046512
273.263445138931 42.7441860465116
277.873847484589 42.7209302325581
282.575718069077 42.7674418604651
287.43803153038 42.7441860465116
292.239601612091 42.7674418604651
296.866827964783 42.6976744186046
301.614987516403 42.7209302325581
306.281884479523 42.6976744186046
310.951313257217 42.6976744186046
315.695355272293 42.7674418604651
320.416744709015 42.8139534883721
325.180294704437 42.8372093023256
329.846116876602 42.7441860465116
334.59952545166 42.7674418604651
339.293500423431 42.7674418604651
344.025556850433 42.7441860465116
348.844466352463 42.7441860465116
353.576533746719 42.7906976744186
358.250085544586 42.7674418604651
362.891864776611 42.7441860465116
367.611008501053 42.7906976744186
372.298185968399 42.7674418604651
};
\addplot [darkorange2551480, dashed, mark=*, mark size=1, mark options={solid}]
table {%
7.77462520599365 42.8372093023256
10.3594027996063 42.8372093023256
17.0372488498688 44.046511627907
23.6163023948669 50.3023255813953
30.5647964000702 58.6279069767442
37.1894445896149 66.093023255814
43.7793123245239 70.4418604651163
50.426499414444 74.3720930232558
57.2948344707489 76.6279069767442
64.0319117546082 78.1162790697674
70.7514793872833 79.7906976744186
77.4866584300995 81.3255813953488
84.1124987125397 83.3023255813954
91.0373836040497 85.6744186046512
97.7558959007263 87.3488372093023
104.498023176193 89.3488372093023
111.217102861404 91.046511627907
117.962502384186 92.1395348837209
124.576713180542 92.9302325581396
131.256309652328 93.6046511627907
138.036812067032 94.1627906976744
144.685187721252 94.8372093023256
151.476530122757 95.3023255813954
158.025437402725 95.6744186046512
164.564543676376 96.1162790697674
171.235788345337 96.4651162790698
177.908112955093 96.8372093023256
184.817539691925 97.0232558139535
191.605309295654 97.2093023255814
198.254035568237 97.3255813953488
205.066015100479 97.4883720930233
211.92867231369 97.5813953488372
218.698678779602 97.6511627906977
225.515848731995 97.7441860465116
232.231773948669 97.7441860465116
238.957235527039 97.8139534883721
245.502389526367 97.8139534883721
252.208363819122 97.8372093023256
258.937342977524 97.9069767441861
265.718951702118 97.9069767441861
272.650874090195 97.9302325581395
279.35715637207 97.9069767441861
286.229867362976 97.953488372093
293.087826156616 98
299.715262126923 98
306.462542247772 97.9767441860465
313.180337810516 98.0232558139535
319.794782066345 98
326.666536712646 97.9767441860465
333.430054664612 98.046511627907
340.202891206741 98.0232558139535
346.966923999786 98.0232558139535
353.612877941132 98.0232558139535
360.458914375305 98.046511627907
367.062369346619 98.046511627907
373.500980949402 98.0697674418605
};
\addplot [yellow22825518, dashed, mark=*, mark size=1, mark options={solid}]
table {%
10.40737657547 42.8372093023256
17.3273522853851 42.7441860465116
25.8241305351257 46.1627906976744
34.6712962627411 59.9069767441861
43.4067992210388 65.7674418604651
52.1807224750519 71.4883720930233
60.912023973465 75.6511627906977
69.5744720935822 77.6976744186046
78.3085381507874 79.4883720930233
86.9004789829254 81.6511627906977
95.6131331920624 83.5581395348837
104.238210248947 86.0697674418604
112.978817415237 87.6046511627907
121.688319826126 89.4418604651163
130.231570529938 90.8837209302326
138.825943422317 92.093023255814
147.63374710083 92.6976744186047
156.290187168121 93.8139534883721
165.108127307892 94.3023255813954
173.69300160408 95.046511627907
182.359554576874 95.3720930232558
190.964942979813 95.9767441860465
199.609686183929 96.2558139534884
201.315172386169 96.3488372093023
216.645715999603 96.8604651162791
225.219941282272 97.093023255814
233.830698204041 97.1860465116279
242.429631519318 97.4418604651163
251.18292016983 97.6046511627907
259.96048283577 97.7441860465116
268.723565626144 97.7674418604651
270.412981414795 97.8139534883721
285.801660728455 97.8837209302326
294.406510734558 97.9767441860465
303.145503473282 97.953488372093
311.684198999405 98.0232558139535
320.370187568665 98.046511627907
329.206597328186 98.0697674418604
338.016325139999 98.1395348837209
346.597557878494 98.1860465116279
355.155255889893 98.1860465116279
363.874028539658 98.1627906976744
372.466823911667 98.1860465116279
};
\addplot [lightgreen124255121, dashed, mark=*, mark size=1, mark options={solid}]
table {%
11.6167194843292 42.8372093023256
21.4230082035065 42.7209302325581
31.1771524429321 47.0697674418605
40.9508226394653 62.8837209302326
50.72467045784 67.5348837209302
60.6231926441193 74.1395348837209
70.459054851532 77.8604651162791
80.2118445873261 79.2093023255814
90.2599890232086 81.2558139534884
99.9358073234558 83.8604651162791
109.803356361389 85.8604651162791
119.704927062988 87.7674418604651
129.339247512817 89.5116279069767
139.3801217556 90.6279069767442
149.253162240982 91.7674418604651
159.16281619072 92.5581395348837
168.793564367294 93.2790697674419
178.570427846909 93.7674418604651
188.584294223785 94.093023255814
198.47213177681 94.6744186046512
208.519085884094 94.953488372093
218.44115447998 95.1860465116279
228.100399160385 95.5116279069768
237.983974170685 95.8139534883721
247.827902650833 96.046511627907
257.645413303375 96.2790697674419
267.340274095535 96.4883720930233
277.045525836945 96.6976744186046
286.991406965256 96.906976744186
296.618041324615 97
306.453501224518 97.2558139534884
316.341722917557 97.3720930232558
326.171481990814 97.5348837209302
335.904342508316 97.6744186046512
345.487187862396 97.8372093023256
355.089465332031 97.953488372093
365.051227140427 98
374.886465454102 98.046511627907
};
\addplot [cyan21255225, dashed, mark=*, mark size=1, mark options={solid}]
table {%
12.8278096199036 42.8372093023256
23.9056611537933 42.7209302325581
34.9654524326324 51.0232558139535
45.8735581874847 64.953488372093
56.8881539821625 70.906976744186
67.9489386081696 77.4418604651163
79.2024352550507 79.5116279069767
90.3599349021912 81.8604651162791
101.474826955795 84.4418604651163
112.739647722244 86.9767441860465
123.46970076561 89.0232558139535
134.505759572983 90.6511627906977
145.39651517868 91.7441860465116
156.315843629837 92.5116279069768
167.191132307053 92.9767441860465
178.139766407013 93.3255813953488
189.267197561264 93.6744186046512
200.245384550095 94.1395348837209
211.256330966949 94.3488372093023
222.113379573822 94.6279069767442
233.155984592438 94.8837209302326
244.123579359055 95.1162790697674
255.0471326828 95.5116279069768
265.943395090103 95.7906976744186
276.951868247986 96
288.071962785721 96.2790697674419
299.112960720062 96.5581395348837
310.233203363419 96.6511627906977
321.367013168335 96.8604651162791
332.439546966553 97.0232558139535
343.47875289917 97.1395348837209
354.42607755661 97.1627906976744
365.546039295197 97.2093023255814
376.482745170593 97.2558139534884
};
\addplot [dodgerblue0128255, dashed, mark=*, mark size=1, mark options={solid}]
table {%
14.0578108787537 42.8372093023256
16.4970048427582 42.7674418604651
28.8455300807953 44.8604651162791
41.5622821807861 56.1627906976744
53.974516916275 68.7209302325581
66.1024549484253 75.6046511627907
78.7087979793549 79.5348837209302
91.0352302551269 82.5116279069767
103.605426263809 85.6279069767442
116.197887086868 88.0930232558139
128.872345733643 89.7674418604651
141.310158967972 90.8837209302326
153.756464242935 91.6279069767442
166.112067699432 92.093023255814
178.533482265472 92.5348837209302
191.277584028244 93.1860465116279
203.78862156868 93.5813953488372
216.1148042202 93.8372093023256
228.694708538055 94.093023255814
241.420179891586 94.4186046511628
254.206185913086 94.6744186046512
266.621324396133 95.0232558139535
279.122977304459 95.2093023255814
291.650282478333 95.3720930232558
304.013698863983 95.7209302325581
316.240296220779 95.9302325581395
328.683960151672 96.0697674418605
341.199859571457 96.3720930232558
353.537830448151 96.5348837209302
365.783430528641 96.6046511627907
377.970103502274 96.6511627906977
};
\addplot [blue, dashed, mark=*, mark size=1, mark options={solid}]
table {%
16.2069649219513 42.8372093023256
19.1170386791229 42.8139534883721
33.4171431064606 45.5116279069767
47.5830177307129 57.953488372093
61.7885466575623 68.3488372093023
75.9906610965729 76.4651162790698
90.5197649478912 81.5813953488372
104.660490226746 84.906976744186
118.999130249023 87.7674418604651
133.264083862305 89.2093023255814
147.397602033615 90.3023255813954
161.62582244873 91.1395348837209
176.016419315338 91.7674418604651
190.228950023651 92.0697674418605
204.404938268662 92.3953488372093
218.615711641312 92.5581395348837
232.972533464432 92.5813953488372
247.361030054092 92.953488372093
261.514213466644 93.093023255814
276.089876413345 93.3255813953488
290.378377485275 93.4418604651163
304.786791658402 93.6511627906977
318.911885595322 93.8604651162791
333.294992542267 94.1395348837209
347.604949045181 94.3023255813954
361.80433049202 94.3720930232558
376.183323574066 94.4418604651163
};
\addplot [navy00127, dashed, mark=*, mark size=1, mark options={solid}]
table {%
19.5081378936768 42.8372093023256
25.5806555747986 42.7209302325581
40.711599445343 48.5116279069767
55.7829661846161 62.046511627907
71.1365592956543 71.1627906976744
86.6871074199676 78.906976744186
102.120712900162 82.3953488372093
117.229339170456 85.2790697674419
132.718991947174 87.7906976744186
148.240753793716 89.093023255814
163.750403213501 90.2790697674419
178.749170875549 91.1395348837209
194.18596663475 91.6744186046512
209.468977069855 91.9767441860465
224.935243225098 92.3023255813953
240.210892868042 92.5581395348837
255.259968519211 92.6976744186046
270.607349967957 92.8837209302326
286.050008821487 93.0232558139535
301.236810064316 93.1627906976744
316.243044519424 93.3953488372093
331.427279663086 93.6046511627907
346.685082769394 93.8837209302326
362.011796092987 93.953488372093
377.346285772324 94.2093023255814
};
\end{axis}

\end{tikzpicture}

%% file: plots/pgf_plots/genes_1ep.tex
\begin{tikzpicture}

\definecolor{blue}{RGB}{0,0,255}
\definecolor{cyan21255225}{RGB}{21,255,225}
\definecolor{darkgray176}{RGB}{176,176,176}
\definecolor{darkorange2551480}{RGB}{255,148,0}
\definecolor{dodgerblue0128255}{RGB}{0,128,255}
\definecolor{gray}{RGB}{128,128,128}
\definecolor{lightgreen124255121}{RGB}{124,255,121}
\definecolor{maroon12700}{RGB}{127,0,0}
\definecolor{navy00127}{RGB}{0,0,127}
\definecolor{red255290}{RGB}{255,29,0}
\definecolor{yellow22825518}{RGB}{228,255,18}

\begin{axis}[
tick align=outside,
tick pos=left,
x grid style={darkgray176},
xmin=0.750904121398925, xmax=396.552114067078,
xtick style={color=black},
y grid style={darkgray176},
ymin=39.9023255813953, ymax=100.888372093023,
ytick style={color=black},
yticklabel style={rotate=90.0},
label style={font=\Huge},
xlabel={\textbf{Time (sec)}},
tick label style={font=\huge}
]
\addplot [ultra thick, gray, dashed, mark=*, mark size=2, mark options={solid}]
table {%
18.7418682098389 42.8372093023256
34.6578793048859 43
50.8279933929443 60.6744186046512
66.9683317661285 68.8604651162791
83.4322623729706 77.7209302325581
99.9514236450195 80.7674418604651
116.412863349915 84.1162790697674
132.916092252731 87.1860465116279
149.050437068939 88.7906976744186
165.219602060318 90.2325581395349
181.240503978729 90.7674418604651
197.671772670746 91.3488372093023
213.76312084198 91.5581395348837
213.76312084198 91.5581395348837
247.107247304916 92.3255813953488
263.459987735748 92.6744186046512
279.685011863708 92.7906976744186
296.458326721191 92.906976744186
313.223739433289 93.1860465116279
329.267318487167 93.2790697674419
345.646366643906 93.4883720930233
362.122578382492 93.6279069767442
378.451215457916 93.8139534883721
};
\addplot [maroon12700, dashed, mark=*, mark size=1, mark options={solid}]
table {%
19.839794588089 42.8372093023256
22.695743560791 42.8372093023256
26.140824508667 42.8372093023256
29.6172603607178 42.8604651162791
33.1739074230194 42.906976744186
36.5923642635345 43
40.2112493515015 43.1627906976744
43.5286998748779 43.1627906976744
47.1236826896668 43.2093023255814
50.6501331806183 43.1860465116279
54.181188249588 43.1395348837209
57.7141266822815 43.1860465116279
61.4184632301331 43.2325581395349
64.8179613590241 43.2093023255814
68.1556833744049 43.2093023255814
71.5325571537018 43.2325581395349
74.9547048568726 43.2790697674419
78.4973690986633 43.3023255813954
82.005387210846 43.2790697674419
85.4736286640167 43.2558139534884
88.9436983108521 43.2558139534884
92.4008068561554 43.2790697674419
95.8825394630432 43.2790697674419
99.2770111560822 43.3488372093023
102.775295352936 43.3488372093023
106.258724164963 43.3720930232558
109.835197734833 43.3720930232558
113.182870197296 43.3023255813953
116.74207406044 43.3255813953488
120.157164096832 43.3023255813953
123.531704521179 43.3023255813953
124.211724281311 43.3023255813953
130.573982858658 43.2790697674419
134.063445043564 43.2558139534884
137.629412317276 43.2790697674419
141.012291955948 43.2790697674419
144.515233135223 43.2325581395349
148.021944713593 43.2093023255814
151.652167701721 43.2325581395349
155.16945438385 43.2325581395349
158.684369754791 43.2093023255814
162.306138706207 43.1162790697674
165.846583271027 43.1162790697674
169.344696712494 43.1395348837209
172.848862314224 43.1860465116279
176.319840335846 43.2093023255814
179.633705663681 43.1860465116279
183.051858663559 43.2093023255814
186.4593272686 43.2790697674419
189.947561788559 43.2558139534884
193.432874631882 43.2325581395349
196.963741779327 43.2093023255814
200.398335552216 43.2325581395349
203.893950891495 43.2790697674419
207.540527200699 43.2093023255814
211.081966257095 43.2093023255814
214.636774682999 43.2093023255814
218.044996023178 43.2558139534884
221.57808804512 43.2093023255814
225.069985675812 43.1627906976744
228.582210302353 43.1627906976744
231.959030056 43.1860465116279
232.637424373627 43.1627906976744
239.06225605011 43.2093023255814
242.475042152405 43.1627906976744
246.118146800995 43.1395348837209
249.620413064957 43.1627906976744
252.975922298431 43.2325581395349
256.391177415848 43.2093023255814
259.921007966995 43.1860465116279
263.527158498764 43.2325581395349
267.116750669479 43.2325581395349
270.749297904968 43.1395348837209
274.11722612381 43.1627906976744
277.585965251923 43.1162790697674
281.04747800827 43.0930232558139
284.579418706894 43.1162790697674
288.106624746323 43.046511627907
291.662030553818 43
295.08782954216 42.9767441860465
298.554833745956 43
302.082434654236 43
305.807660865784 42.9302325581395
309.420103883743 42.9767441860465
313.030579662323 42.9302325581395
316.524045085907 42.9302325581395
319.987852954864 42.906976744186
323.530163192749 42.8837209302325
327.029419565201 42.8604651162791
330.574849891663 42.906976744186
333.971498918533 42.8837209302325
337.546960830688 42.9302325581395
341.145388269424 42.9767441860465
344.716472911835 42.953488372093
348.287282657623 42.9302325581395
351.762807321548 42.8837209302325
355.309735155106 42.8372093023256
358.74199719429 42.8604651162791
362.309473609924 42.8604651162791
365.712177848816 42.8372093023256
369.192029953003 42.8604651162791
372.597342777252 42.8372093023256
};
\addplot [red255290, dashed, mark=*, mark size=1, mark options={solid}]
table {%
21.1624392032623 42.8372093023256
22.1191129684448 42.8604651162791
26.7930511951447 42.8837209302326
31.495298576355 42.8372093023256
36.1612645149231 42.953488372093
39.8364887237549 42.9302325581395
45.5284746646881 42.8837209302326
50.1638140678406 42.8139534883721
54.799205160141 42.7441860465116
59.4528113365173 42.6976744186046
63.274574804306 42.8837209302326
68.7310170650482 42.9767441860465
73.4314738750458 42.9302325581395
78.1291780948639 42.9302325581395
82.7930714130402 42.9302325581395
87.425718164444 43
92.0786502838135 43
96.7106272697449 43.0232558139535
101.428462028503 43
106.145950841904 43
109.840386390686 42.9302325581395
115.515019130707 42.9302325581395
120.257042598724 42.906976744186
124.959957170486 42.8837209302326
129.660809803009 42.8372093023256
134.381789779663 42.8837209302325
139.058727455139 42.906976744186
143.82794919014 42.953488372093
148.56606259346 42.953488372093
153.152679872513 42.8837209302325
157.925745916367 42.9302325581395
162.54202914238 42.9302325581395
167.16804728508 42.953488372093
171.848989200592 42.9302325581395
176.580663490295 43
181.356801462173 42.9767441860465
186.061311912537 42.906976744186
190.803492307663 42.906976744186
195.5181016922 42.8837209302326
198.397731494904 42.906976744186
203.259736442566 42.8837209302326
208.710161209106 42.8604651162791
213.401228380203 42.906976744186
218.09307808876 42.906976744186
222.765782833099 42.9302325581395
227.406156349182 42.8372093023256
231.93954744339 42.9302325581395
236.645421695709 42.8837209302326
241.249228048325 42.8837209302326
245.890829801559 42.906976744186
250.615105056763 42.8604651162791
255.290913963318 42.8372093023256
259.997003793716 42.8372093023256
264.664121246338 42.7906976744186
269.231747293472 42.8139534883721
274.004154825211 42.7906976744186
278.745925045013 42.7441860465116
283.427136611938 42.7209302325581
288.008933782578 42.6976744186046
292.710395002365 42.6976744186046
297.435066223145 42.7674418604651
302.143979454041 42.6976744186046
306.751320314407 42.6976744186046
311.452672433853 42.6976744186046
316.046512174606 42.7674418604651
320.762447071075 42.7209302325581
325.348928880692 42.6976744186046
330.115785741806 42.7441860465116
334.805218791962 42.7209302325581
339.523593187332 42.7441860465116
344.164177274704 42.7906976744186
348.807863235474 42.8139534883721
353.369349002838 42.8139534883721
358.029272842407 42.8837209302325
362.732591342926 42.8604651162791
367.486305856705 42.8837209302325
372.189680814743 42.9302325581395
};
\addplot [darkorange2551480, dashed, mark=*, mark size=1, mark options={solid}]
table {%
23.2494355678558 42.8372093023256
24.6247612476349 42.8139534883721
31.688454914093 43.0697674418605
38.7999643325806 46.3255813953488
45.5842258453369 58.046511627907
52.6717593669891 65.4186046511628
59.4605291366577 69.3255813953488
66.5535765171051 73.8837209302326
73.4266484260559 75.8837209302326
80.4255704402924 77.5581395348837
87.2797524929047 79.0232558139535
94.3607317447662 81.046511627907
101.28918337822 83.5348837209302
108.223071718216 85.7674418604651
113.709525680542 87.1627906976744
120.675100851059 88.6046511627907
127.683014583588 90.3488372093023
134.6981985569 91.4418604651163
141.929196023941 92.3023255813954
144.791168260574 92.7209302325581
151.725949192047 93.5116279069767
158.576783466339 94.0930232558139
165.444598293304 94.6744186046512
172.264963245392 95.3488372093023
179.365720272064 95.8372093023256
186.254044055939 96.3255813953488
193.238543987274 96.6279069767442
200.147317695618 96.906976744186
205.694056367874 97.1860465116279
212.749541091919 97.2558139534884
219.809916687012 97.4883720930232
226.875987434387 97.5581395348837
233.630612373352 97.6511627906977
240.506379747391 97.7674418604651
247.517276620865 97.7906976744186
254.556716299057 97.8372093023256
261.595568275452 97.860465116279
264.391700983047 97.8837209302326
271.376812410355 98
278.408281898499 97.9767441860465
285.361020803452 98.0232558139535
290.917532205582 98.0232558139535
297.954526567459 97.9767441860465
304.899319505692 98
312.014249753952 98.046511627907
319.016141843796 98.046511627907
325.873650836945 98.0697674418604
332.810785293579 98.0930232558139
339.741583681107 98.0697674418604
346.572565937042 98.0232558139535
353.387074804306 98.0930232558139
360.36987285614 98.046511627907
367.343076944351 98
374.325678443909 98.1162790697674
};
\addplot [yellow22825518, dashed, mark=*, mark size=1, mark options={solid}]
table {%
25.5751091003418 42.8372093023256
32.6352754116058 42.6744186046512
41.414736032486 47.3720930232558
50.1587089061737 59.6279069767442
58.924261713028 65.8139534883721
67.8646119594574 71.906976744186
76.7183005809784 75.7209302325581
85.658029127121 77.3953488372093
94.5503008842468 79.1627906976744
103.408923721313 81.0232558139535
112.431806135178 83.3953488372093
121.302973127365 86.0232558139535
130.075528383255 87.8139534883721
138.916523456573 89.4883720930233
147.776653575897 90.9302325581396
156.663361692429 91.9069767441861
165.351282262802 93.1162790697674
174.166581344604 93.9767441860465
182.95574297905 94.8372093023256
191.668877506256 95.3488372093023
200.540460586548 95.7441860465116
209.048454618454 96.2790697674419
217.874329328537 96.5581395348837
226.574269151688 96.906976744186
235.62039141655 97.1627906976744
244.31933016777 97.3023255813954
253.044112920761 97.3953488372093
261.648012924194 97.4651162790698
270.268446016312 97.6279069767442
278.926194953918 97.7674418604651
287.469729423523 97.8139534883721
296.200722789764 97.8837209302326
305.138183736801 97.9767441860465
313.898196840286 98
322.64971909523 98
331.51318230629 98
340.261712169647 98.0232558139535
349.075561618805 97.9767441860465
358.003320884705 98
366.658710575104 98.0232558139535
375.39258351326 98.046511627907
};
\addplot [lightgreen124255121, dashed, mark=*, mark size=1, mark options={solid}]
table {%
27.0188595294952 42.8372093023256
28.9375638961792 42.7674418604651
38.589860534668 43.3953488372093
48.3599937438965 50.953488372093
58.1882607460022 64.6511627906977
67.9366462230682 68.7906976744186
77.8537021160126 74.6744186046512
87.7641509056091 78.4418604651163
97.7038267612457 80
107.59081363678 82.046511627907
117.463397789001 84.5581395348837
127.378373813629 86.5581395348837
137.233432912827 88.4418604651163
146.905843114853 90.093023255814
156.678805541992 91.2093023255814
166.719143199921 92.2790697674419
176.601769542694 93.0697674418605
186.606962013245 93.7441860465116
196.556894540787 94.2325581395349
206.473974609375 94.5813953488372
216.355820512772 94.8837209302326
226.382388544083 95.2558139534884
236.305454540253 95.5813953488372
245.924613237381 95.8604651162791
255.748436641693 96.2790697674418
265.697275209427 96.5116279069768
275.646078443527 96.6744186046512
285.56091337204 96.8604651162791
295.377829122543 97.0232558139535
305.236872291565 97.0930232558139
315.118064212799 97.2093023255814
324.935134077072 97.3255813953488
334.945643615723 97.3488372093023
344.433732175827 97.4186046511628
354.280577850342 97.5116279069767
364.183051204681 97.6279069767442
374.104853916168 97.6744186046512
};
\addplot [cyan21255225, dashed, mark=*, mark size=1, mark options={solid}]
table {%
28.5189386367798 42.8372093023256
35.0869171619415 42.8372093023256
46.1204494476318 47.4186046511628
56.7810942173004 59.6046511627907
67.7864831924439 67.9767441860465
78.8079293251038 73.8139534883721
89.7747990131378 78.6976744186047
100.6577709198 81.2790697674419
111.745693016052 83.6976744186046
122.623010921478 86.1860465116279
133.713986539841 88.6511627906977
144.931122970581 90
156.04733543396 91
167.066186189651 91.8372093023256
177.929360628128 92.5813953488372
188.936727428436 93.0697674418605
200.054201841354 93.3488372093023
211.079790592194 93.7209302325581
222.060373735428 94.046511627907
233.153118896484 94.3488372093023
244.310883903503 94.5348837209303
255.27470369339 94.9069767441861
266.401179790497 95.1860465116279
277.441615104675 95.5348837209302
288.375249290466 95.8604651162791
299.662703752518 96.093023255814
310.81183180809 96.2093023255814
321.707679605484 96.3720930232558
332.616712522507 96.5348837209302
343.537173748016 96.6976744186047
354.730137586594 96.9767441860465
365.643888902664 97
376.828820085526 97.046511627907
};
\addplot [dodgerblue0128255, dashed, mark=*, mark size=1, mark options={solid}]
table {%
29.5265080928802 42.8372093023256
39.387158203125 42.7441860465116
51.9450975894928 50.7906976744186
64.5559819221497 63.7209302325582
77.2169281959534 72.7906976744186
90.0200599193573 78.3953488372093
102.569763660431 82.0697674418604
115.037800645828 85.2558139534884
127.189331245422 87.8604651162791
139.890969276428 89.953488372093
152.144149065018 91.0232558139535
165.161415433884 91.7209302325581
177.900431251526 92.4186046511628
190.234754276276 93.1395348837209
202.930086803436 93.5116279069768
215.539416122437 94.046511627907
228.186646556854 94.4418604651163
240.967716026306 94.8139534883721
253.519105815887 94.9767441860465
265.921100759506 95.2325581395349
278.324674320221 95.4651162790698
290.947765922546 95.7209302325581
303.367177724838 95.9767441860465
315.960980367661 96.1395348837209
328.456915378571 96.3255813953488
340.9063808918 96.3720930232558
353.386622571945 96.4883720930233
366.053413009644 96.7209302325582
378.522457504272 96.8604651162791
};
\addplot [blue, dashed, mark=*, mark size=1, mark options={solid}]
table {%
31.6754893302917 42.8372093023256
37.1994411945343 42.7441860465116
50.9459073543549 47.8139534883721
65.0815687656403 60.8372093023256
79.332327413559 70.6744186046512
93.7815219402313 77.3488372093023
107.636353445053 81.1860465116279
121.567006111145 84.8604651162791
135.696373558044 87.2325581395349
150.027084684372 89.1162790697675
163.966705131531 90.3953488372093
177.983021068573 91.1162790697675
192.375245952606 91.8139534883721
206.447863674164 92.2790697674419
220.531658411026 92.4418604651163
229.05703997612 92.5813953488372
248.590930366516 93
262.865787649155 93.3023255813954
276.851116323471 93.3023255813954
291.143612003326 93.5581395348838
305.163221311569 93.6744186046512
319.334906387329 93.8372093023256
333.346614360809 94
347.724591732025 94.3720930232558
361.675174474716 94.4651162790698
375.746467590332 94.5581395348837
};
\addplot [navy00127, dashed, mark=*, mark size=1, mark options={solid}]
table {%
34.408545589447 42.8372093023256
43.4481165409088 42.7906976744186
58.3664643764496 51.3953488372093
73.6694990634918 64.6511627906977
89.0846886634827 73.4651162790698
104.326226854324 79.8372093023256
119.404794454575 82.7441860465116
134.601596212387 85.3720930232558
149.725624275208 87.7209302325581
164.673254394531 89.3953488372093
180.043390035629 90.5813953488372
195.613752698898 91.2325581395349
210.634994411469 91.5581395348837
225.747588396072 91.8372093023256
241.040662097931 92.093023255814
256.214940834045 92.2790697674419
271.644025754929 92.4651162790698
286.859267234802 92.6511627906977
302.125139904022 92.7674418604651
317.108119297028 92.953488372093
332.299588346481 93.1162790697675
347.792752122879 93.4651162790698
363.27049908638 93.6279069767442
378.561149978638 93.8604651162791
};
\end{axis}

\end{tikzpicture}

%% file: plots/pgf_plots/genes_random.tex
\begin{tikzpicture}

\definecolor{blue00254}{RGB}{0,0,254}
\definecolor{darkgray176}{RGB}{176,176,176}
\definecolor{darkorange2551220}{RGB}{255,122,0}
\definecolor{deepskyblue0212255}{RGB}{0,212,255}
\definecolor{dodgerblue096255}{RGB}{0,96,255}
\definecolor{gold2552290}{RGB}{255,229,0}
\definecolor{gray}{RGB}{128,128,128}
\definecolor{greenyellow17025576}{RGB}{170,255,76}
\definecolor{maroon12700}{RGB}{127,0,0}
\definecolor{red254180}{RGB}{254,18,0}
\definecolor{turquoise76255170}{RGB}{76,255,170}

\begin{axis}[
tick align=outside,
tick pos=left,
x grid style={darkgray176},
xmin=-16.2479414820671, xmax=398.137098157406,
xtick style={color=black},
y grid style={darkgray176},
ymin=40.0127906976744, ymax=99.5918604651163,
ytick style={color=black},
yticklabel style={rotate=90.0},
label style={font=\Huge},
xlabel={\textbf{Time (sec)}},
tick label style={font=\huge}
]
\addplot [maroon12700, dashed, mark=*, mark size=1, mark options={solid}]
table {%
2.58774213790894 42.8372093023256
3.64498281478882 42.8372093023256
6.65758190155029 42.8372093023256
9.66098575592041 42.8372093023256
12.7042511463165 42.8604651162791
15.0632170200348 42.8837209302325
17.3694919586182 42.906976744186
20.3698163032532 42.953488372093
23.3625504970551 43.0232558139535
26.4224194526672 43.0232558139535
28.7044408321381 43.046511627907
30.97461977005 43.046511627907
33.9751330375671 43.046511627907
36.9684103965759 43.0697674418605
40.0352935791016 43.1395348837209
42.3032980918884 43.1860465116279
45.2747196674347 43.2325581395349
47.6180274009705 43.2790697674419
50.7034152507782 43.3488372093023
52.9804457187653 43.4186046511628
55.9396831035614 43.4186046511628
58.9492003917694 43.5116279069767
61.3004956245422 43.5581395348837
64.3602819442749 43.7209302325581
66.6317977905273 43.7441860465116
69.6478846549988 43.8604651162791
72.6586328983307 43.9302325581395
75.682994556427 44.046511627907
77.3960463047028 44.1162790697674
79.6655840396881 44.1395348837209
82.706064081192 44.2325581395349
85.72474360466 44.2558139534884
88.7378216266632 44.2558139534884
91.0521395683289 44.1860465116279
93.3226320266724 44.1162790697674
96.2821929454803 44.1860465116279
99.3078223705292 44.1627906976744
102.451764154434 44.2093023255814
104.829879426956 44.2790697674419
107.090959453583 44.3255813953488
110.166672420502 44.3488372093023
113.060292053223 44.4186046511628
116.059549665451 44.4883720930233
118.309755897522 44.4651162790698
120.67734246254 44.6744186046512
123.603910827637 44.6279069767442
126.613899803162 44.6279069767442
129.677856588364 44.6511627906977
132.052360200882 44.6744186046512
135.132677316666 44.7209302325581
137.490012598038 44.6744186046512
140.477062177658 44.6744186046512
142.790997982025 44.6976744186046
145.706832170486 44.6744186046512
148.644187927246 44.7209302325581
150.428758239746 44.6976744186046
153.375205612183 44.8139534883721
155.601852321625 44.8372093023256
158.658036088943 44.953488372093
160.538588237762 45.0232558139535
164.658755493164 44.9767441860465
166.876250362396 44.953488372093
169.112927436829 45
172.15791888237 45
175.133863115311 44.906976744186
178.096272468567 44.906976744186
180.407681179047 45.0232558139535
182.747528648376 44.9767441860465
185.695875549316 45
188.660253953934 44.953488372093
191.630499887466 44.9302325581395
194.051074123383 44.9767441860465
196.328502082825 45.093023255814
199.198698568344 45.0697674418605
202.212542486191 45.1627906976744
205.187919473648 45.1395348837209
207.541822004318 45.1860465116279
209.817450618744 45.2325581395349
212.914886236191 45.2325581395349
215.935506343842 45.3023255813953
218.924695301056 45.3255813953488
221.333780431747 45.3023255813953
224.284006977081 45.3255813953488
226.083477735519 45.3488372093023
229.1890604496 45.3255813953488
232.142998075485 45.3023255813953
234.505363464355 45.2790697674419
237.427761173248 45.2093023255814
239.728816080093 45.1627906976744
242.646643686295 45.0930232558139
244.950968551636 45.1395348837209
247.907439184189 45.0930232558139
250.981411457062 45.0930232558139
253.981238222122 45.0930232558139
256.326464653015 45.0697674418605
258.655218982697 45.046511627907
261.645161008835 45.1162790697674
264.59021821022 44.9767441860465
267.558640384674 45
269.929532194138 45.046511627907
272.244231939316 45.0697674418605
275.205262041092 45.0697674418605
278.353735256195 45.2325581395349
281.479959154129 45.1627906976744
283.80009765625 45.1860465116279
286.105051183701 45.2558139534884
289.126295232773 45.1860465116279
292.189613389969 45.2558139534884
295.167724084854 45.2093023255814
297.485000085831 45.2093023255814
299.236741542816 45.2325581395349
302.271394777298 45.1395348837209
305.261975765228 45.2093023255814
308.329113769531 45.2325581395349
310.668033123016 45.2325581395349
313.663542079926 45.1395348837209
316.018733930588 45.1162790697674
318.945441436768 45.0930232558139
320.93944568634 45.1162790697674
324.390257787704 45.1162790697674
327.37824587822 45.1395348837209
329.773576545715 45.1627906976744
332.784627532959 45.1395348837209
335.240753030777 45.2093023255814
338.334758234024 45.1162790697674
341.36905207634 45.1395348837209
344.408077049255 45.1860465116279
346.763212108612 45.1627906976744
349.098349142075 45.2558139534884
352.169553661346 45.2790697674419
355.051040744782 45.2558139534884
358.07454624176 45.2790697674418
360.35591173172 45.2558139534884
362.630840682983 45.3023255813953
365.715178632736 45.2790697674419
368.823685121536 45.3953488372093
371.84010720253 45.4418604651163
};
\addplot [red254180, dashed, mark=*, mark size=1, mark options={solid}]
table {%
4.31169328689575 42.8372093023256
4.31169328689575 42.8372093023256
8.790345287323 42.8372093023256
14.0260390758514 43
16.6623085021973 42.9302325581395
21.9681106567383 44.4651162790698
25.6406538486481 47.0232558139535
30.1031998634338 48.046511627907
31.9441672801971 48.1860465116279
36.2950868606567 49.4651162790698
40.8391095161438 50.3255813953488
44.2206384181976 50.4883720930233
47.7593249320984 50.8372093023256
52.2320488452911 51.4651162790698
56.6488260746002 52.2790697674419
60.197688627243 52.4418604651163
62.9905045509338 52.8372093023256
68.2740552902222 53.2790697674419
69.893551158905 53.3488372093023
74.3738564968109 53.8372093023256
78.9175070285797 54.4186046511628
83.3931062698364 55.1627906976744
86.7473275184631 55.3720930232558
90.4086434364319 55.5581395348837
93.9096077442169 55.6976744186046
97.3201712608337 55.6976744186046
101.788083934784 55.7441860465116
104.670553350449 55.8604651162791
109.935978794098 56.1627906976744
112.414102697372 56.2558139534884
116.955051422119 56.3953488372093
120.493772602081 56.6511627906977
123.166401672363 56.6744186046512
128.486864280701 56.9302325581395
132.922188043594 57.2558139534884
136.542898368835 57.3720930232558
139.133341693878 57.3720930232558
143.645131731033 57.5348837209302
147.264328193665 57.6046511627907
150.6314848423 57.6976744186046
155.019815063477 57.8139534883721
158.591538381577 57.7906976744186
163.164888715744 57.7209302325581
166.527259922028 57.8372093023256
170.972443294525 57.9302325581395
174.603324317932 57.9302325581395
177.170546150208 57.953488372093
180.629717826843 58.1162790697674
185.035067081451 58.1162790697674
189.506503582001 58.1860465116279
192.973140859604 58.3255813953488
197.377596998215 58.2790697674419
200.908441925049 58.3255813953488
203.526812553406 58.3255813953488
206.393932342529 58.2325581395349
210.919243431091 58.3255813953488
216.160555076599 58.2558139534884
219.661838722229 58.2558139534884
223.347539901733 58.2790697674419
227.803232049942 58.3720930232558
231.302397012711 58.3720930232558
234.860852193832 58.3720930232558
239.310798072815 58.4883720930232
242.936102581024 58.3255813953488
244.698364496231 58.3255813953488
249.950592422485 58.3488372093023
254.408914804459 58.2325581395349
257.750686168671 58.2325581395349
261.289889764786 58.2790697674419
265.73624124527 58.2790697674418
269.268650960922 58.3255813953488
272.592414808273 58.2790697674418
276.241024208069 58.3488372093023
280.669624757767 58.3023255813954
284.071642208099 58.2325581395349
287.664269685745 58.1395348837209
291.204878425598 58.1162790697674
295.725933456421 58.0930232558139
299.117264795303 58.2093023255814
303.583247661591 58.1627906976744
308.085458278656 58.2093023255814
309.706330490112 58.2093023255814
314.089363956451 58.1162790697674
317.653959274292 58.1162790697674
322.119074678421 58.2558139534884
325.565437936783 58.2325581395349
329.973973321915 58.2325581395349
333.43696770668 58.2558139534884
336.905438804626 58.3023255813953
341.343982553482 58.2325581395349
345.04150800705 58.2325581395349
348.570084810257 58.2093023255814
352.097009325027 58.1395348837209
355.585381889343 58.1162790697674
360.10271730423 58.1395348837209
363.547417211533 58.0697674418605
367.875452756882 58.093023255814
372.390941524506 58.1162790697674
};
\addplot [darkorange2551220, dashed, mark=*, mark size=1, mark options={solid}]
table {%
6.51910810470581 42.8372093023256
7.59494256973267 42.8372093023256
13.6851147651672 42.953488372093
19.813981962204 44.0232558139535
25.6875759601593 48.953488372093
30.4115665912628 50.9767441860465
36.5832704544067 52.9069767441861
41.3642321586609 55.3953488372093
47.2997756958008 56.7441860465116
53.2122789382935 57.9302325581395
57.9082548618317 58.5581395348837
62.8889686584473 59.6511627906977
68.8552669048309 60.4651162790698
73.5144555568695 61.4418604651163
78.1426040172577 62.093023255814
84.2341351509094 62.7906976744186
90.2226552009583 63.4418604651163
96.2571826934814 64.3488372093023
102.173807239532 65.3255813953488
105.595280885696 65.6046511627907
111.545814085007 66.3023255813954
117.575326251984 66.6976744186047
122.611799192429 67
129.611014175415 67.5813953488372
133.036630630493 67.8837209302326
139.110080385208 68
143.823471212387 68.1860465116279
149.981896543503 68.3953488372093
154.581298971176 68.5581395348837
160.530321788788 68.4883720930233
166.488547420502 68.5581395348837
172.495368909836 68.7209302325581
177.355223655701 68.9767441860465
181.987901258469 68.8837209302326
188.041476631165 68.9767441860465
192.819005632401 69.2325581395349
198.862381982803 69.3255813953488
203.633615970612 69.1860465116279
208.476520681381 69.2790697674419
214.606227874756 69.2558139534884
220.722453832626 69.2790697674419
226.810481929779 69.2790697674419
231.465577125549 69.2790697674419
237.464864253998 69.1627906976744
241.185885572433 69.1162790697674
248.371618175507 69.2325581395349
254.387973642349 69.2790697674419
257.853048086166 69.3023255813954
263.831109333038 69.3023255813954
269.920245981216 69.2558139534884
276.05698390007 69.3255813953488
279.453358745575 69.2790697674419
285.50254406929 69.1860465116279
291.464717769623 69.2093023255814
297.318070554733 69.2790697674419
303.392170000076 69.3255813953488
307.995660400391 69.3488372093023
312.802172374725 69.3255813953488
317.598172569275 69.3488372093023
323.630449914932 69.4186046511628
328.197847604752 69.3953488372093
334.092807102203 69.4418604651163
340.099442243576 69.6046511627907
343.862395429611 69.6279069767442
350.91166434288 69.5813953488372
355.566873979568 69.7209302325581
361.661051607132 69.7209302325581
367.612751865387 69.8372093023256
373.555086135864 69.7674418604651
};
\addplot [gold2552290, dashed, mark=*, mark size=1, mark options={solid}]
table {%
8.50979175567627 42.8372093023256
11.23190741539 42.7209302325581
18.7447402000427 43.6744186046512
26.3842206954956 50.046511627907
33.9271439552307 59.6279069767442
41.4747193813324 64.3023255813954
47.3051511287689 68.3255813953488
53.0441835403442 70.7906976744186
60.4205767154694 73
68.0285848140716 74.6511627906977
75.6040594577789 76.7441860465116
81.3027666091919 78
88.9091298103333 79.4651162790698
96.3846137523651 81.3488372093023
102.141714811325 82.6976744186046
109.621267986298 83.6511627906977
117.088499069214 84.7209302325581
122.791181468964 85.4651162790698
130.210542678833 85.9069767441861
137.741696166992 86.2093023255814
145.158782291412 86.7441860465116
151.071596288681 86.8837209302326
156.830593109131 87.1627906976744
164.221179628372 87.4186046511628
171.540661334991 87.6046511627907
178.960459899902 87.8372093023256
186.385471343994 88.2093023255814
193.839096069336 88.3255813953488
197.992307424545 88.4186046511628
205.607534313202 88.6046511627907
213.352861785889 88.7906976744186
220.835208320618 88.7906976744186
228.433389234543 88.8372093023256
234.346502733231 88.8604651162791
240.226408529282 88.8837209302326
247.806152439117 88.906976744186
255.137610435486 89.0232558139535
262.717171049118 89
268.62842502594 89.046511627907
276.344967794418 89.1162790697674
283.823961591721 89.0697674418605
289.724031543732 89.0930232558139
297.086873435974 89.2093023255814
304.429228448868 89.1395348837209
310.350488853455 89.1860465116279
317.877024841309 89.0930232558139
325.235259246826 89.1860465116279
332.849966001511 89.1627906976744
338.770459938049 89.1162790697674
344.489685201645 89.1395348837209
352.064213275909 89.2093023255814
359.677355384827 89.2325581395349
367.235768079758 89.2093023255814
374.575281190872 89.2325581395349
};
\addplot [greenyellow17025576, dashed, mark=*, mark size=1, mark options={solid}]
table {%
10.2594799041748 42.8372093023256
13.7322043418884 42.7674418604651
22.8115735054016 45.6511627906977
31.6335004329681 53.5348837209302
40.5786554813385 63.4883720930233
49.5313712120056 67.7441860465116
58.7491654872894 72.8139534883721
67.7819414138794 75.2325581395349
76.9264497756958 77.4883720930232
86.079205083847 79.2790697674419
95.0152848243713 81.3953488372093
102.028309202194 82.6511627906977
110.931581878662 84.1162790697674
119.999487686157 84.8604651162791
129.192391014099 85.6511627906977
136.306196308136 86.0232558139535
145.164340877533 86.5581395348837
154.186092185974 86.9069767441861
163.321058797836 87.3720930232558
172.367743682861 87.8604651162791
181.207101345062 88.3023255813954
190.086663532257 88.5813953488372
197.171494436264 88.8139534883721
202.565614557266 88.953488372093
215.18542971611 89.3023255813954
224.267865514755 89.3023255813954
233.182832241058 89.6976744186047
242.172231674194 89.8837209302326
251.137376308441 90.1627906976744
258.355054283142 90.1860465116279
267.370028495789 90.3255813953488
272.73792347908 90.2790697674419
284.969505214691 90.3720930232558
291.924668645859 90.3023255813954
300.778187131882 90.3720930232558
309.815341043472 90.3953488372093
318.767434215546 90.5581395348837
327.877022790909 90.5348837209302
336.772830581665 90.5116279069767
345.854109334946 90.5116279069767
354.906508255005 90.6046511627907
364.019502019882 90.4651162790698
373.131201553345 90.4883720930232
};
\addplot [turquoise76255170, dashed, mark=*, mark size=1, mark options={solid}]
table {%
12.0210280895233 42.8372093023256
16.2015858650208 42.8604651162791
27.1489161014557 45.6744186046512
37.8110353469849 54.8139534883721
48.6037202835083 65.953488372093
59.370680141449 74.046511627907
70.1103183269501 79.953488372093
80.9176777362823 82.4418604651163
91.5143751621246 84.9069767441861
102.522165346146 87.0697674418605
113.208915996552 89.3953488372093
124.217329454422 90.6046511627907
135.128449106216 91.3488372093023
145.768268489838 91.9767441860465
156.587456130981 92.3023255813954
167.393191576004 92.5813953488372
178.112707996368 92.906976744186
189.023208332062 93.3255813953488
199.573137712479 93.6279069767442
203.634694719315 93.7674418604651
214.446725225449 93.953488372093
225.127882051468 94.3023255813954
236.066826295853 94.7209302325582
247.011710548401 94.8837209302326
258.107643985748 95.093023255814
269.031563520432 95.2093023255814
279.677290296555 95.4186046511628
290.383556890488 95.6279069767442
301.287638998032 95.906976744186
312.244945907593 96.0697674418605
322.754330205917 96.1860465116279
333.528727769852 96.3255813953488
344.277459192276 96.4651162790698
354.933006858826 96.6511627906977
365.47378783226 96.7209302325581
376.293748235703 96.8837209302326
};
\addplot [deepskyblue0212255, dashed, mark=*, mark size=1, mark options={solid}]
table {%
14.1317253112793 42.8372093023256
16.4813459396362 42.9302325581395
28.4612557411194 45.0697674418605
40.8850030422211 54.7906976744186
52.8570613861084 67.1627906976744
64.8925700187683 74.3023255813954
76.9088493347168 79.6511627906977
89.0400539875031 82.953488372093
100.839324808121 85.7906976744186
113.097235822678 88.2325581395349
125.002539348602 90
136.701075935364 90.953488372093
146.393066596985 91.7209302325581
158.796195745468 92.4186046511628
170.941131496429 92.6046511627907
182.89168047905 93.1627906976744
194.988002443314 93.2093023255814
199.768937397003 93.4418604651163
212.123265266418 93.5581395348837
224.317221879959 93.7209302325581
236.564394903183 94.046511627907
248.746862459183 94.3023255813954
260.505202007294 94.7209302325581
270.123868227005 94.8837209302326
282.293724012375 95.2325581395349
294.423954772949 95.3023255813953
306.410960102081 95.3953488372093
318.351197052002 95.6511627906977
330.363576984406 95.8604651162791
342.551256227493 96.1395348837209
354.892612504959 96.1860465116279
367.188641881943 96.3953488372093
379.026525783539 96.6511627906977
};
\addplot [dodgerblue096255, dashed, mark=*, mark size=1, mark options={solid}]
table {%
15.6186882972717 42.8372093023256
18.3695953845978 42.8372093023256
31.8037138462067 45.0930232558139
45.3939785957336 55.7209302325582
58.8721303462982 69.0697674418605
72.2492671966553 76.3720930232558
85.7724052429199 80.5348837209303
99.1154541015625 83.7674418604651
112.914837026596 86.4418604651163
127.014417505264 88.7674418604651
140.299557065964 90.5813953488372
153.457708692551 91.5116279069768
167.044228458405 92.2790697674419
180.277895402908 92.7906976744186
193.655517482758 93.0697674418605
204.278023147583 93.3255813953488
218.102642202377 93.6744186046512
231.230952072144 93.8837209302326
244.536957073212 94.0930232558139
257.928685188293 94.1860465116279
271.307204675674 94.3023255813954
284.767873191833 94.3953488372093
298.107170581818 94.6046511627907
311.483020067215 94.8139534883721
325.1706407547 94.9302325581395
338.601550626755 95.2325581395349
352.200178956985 95.5116279069768
365.538585329056 95.6046511627907
379.30141453743 95.906976744186
};
\addplot [blue00254, dashed, mark=*, mark size=1, mark options={solid}]
table {%
17.5018193721771 42.8372093023256
20.3336221694946 42.8372093023256
35.1589440822601 45.4883720930233
50.1571625709534 58.8837209302326
64.844313287735 69.8372093023256
79.8479808807373 78.2790697674419
94.9318290710449 82.0697674418604
110.028542280197 85.6046511627907
124.789586877823 88.1627906976744
139.798922109604 89.6976744186047
154.779183673859 90.8837209302325
169.743154478073 91.6744186046512
184.546446561813 92.1627906976744
199.318388652802 92.4418604651163
214.39347114563 92.6744186046512
226.604488134384 92.8372093023256
244.461021757126 93
259.059178161621 93.2790697674419
273.91031794548 93.4418604651163
288.811427164078 93.6046511627907
303.677708530426 93.9069767441861
318.653675031662 94.1162790697674
333.550012636185 94.4186046511628
348.366209888458 94.5581395348837
363.153579187393 94.6511627906977
377.866205739975 94.8837209302326
};
\addplot [ultra thick, gray, dashed, mark=*, mark size=2, mark options={solid}]
table {%
18.7418682098389 42.8372093023256
34.6578793048859 43
50.8279933929443 60.6744186046512
66.9683317661285 68.8604651162791
83.4322623729706 77.7209302325581
99.9514236450195 80.7674418604651
116.412863349915 84.1162790697674
132.916092252731 87.1860465116279
149.050437068939 88.7906976744186
165.219602060318 90.2325581395349
181.240503978729 90.7674418604651
197.671772670746 91.3488372093023
213.76312084198 91.5581395348837
213.76312084198 91.5581395348837
247.107247304916 92.3255813953488
263.459987735748 92.6744186046512
279.685011863708 92.7906976744186
296.458326721191 92.906976744186
313.223739433289 93.1860465116279
329.267318487167 93.2790697674419
345.646366643906 93.4883720930233
362.122578382492 93.6279069767442
378.451215457916 93.8139534883721
};
\end{axis}

\end{tikzpicture}

%% file: plots/pgf_plots/mondial_religion_k_var.tex
\begin{tikzpicture}

\definecolor{blue}{RGB}{0,0,255}
\definecolor{cyan21255225}{RGB}{21,255,225}
\definecolor{darkgray176}{RGB}{176,176,176}
\definecolor{darkorange2551480}{RGB}{255,148,0}
\definecolor{dodgerblue0128255}{RGB}{0,128,255}
\definecolor{gray}{RGB}{128,128,128}
\definecolor{lightgray204}{RGB}{204,204,204}
\definecolor{lightgreen124255121}{RGB}{124,255,121}
\definecolor{maroon12700}{RGB}{127,0,0}
\definecolor{navy00127}{RGB}{0,0,127}
\definecolor{red255290}{RGB}{255,29,0}
\definecolor{yellow22825518}{RGB}{228,255,18}

\begin{axis}[
legend cell align={left},
legend columns=2, 
legend style={
  /tikz/column 2/.style={column sep=5pt,},
  font=\huge,
  fill opacity=0.8,
  draw opacity=1,
  text opacity=1,
  at={(0.97,0.03)},
  anchor=south east,
  draw=lightgray204
},
tick align=outside,
tick pos=left,
x grid style={darkgray176},
xmin=-9.49949369192124, xmax=269.891679618359,
xtick style={color=black},
y grid style={darkgray176},
ymin=59.7238095238095, ymax=86.7523809523809,
ytick style={color=black},
yticklabel style={rotate=90.0},
label style={font=\Huge},
xlabel={\textbf{Time (sec)}},
tick label style={font=\huge}
]

\addplot [maroon12700, dashed, mark=*, mark size=1, mark options={solid}]
table {%
3.20010509490967 61.4285714285714
3.54124207496643 61.4285714285714
5.61002097129822 61.6190476190476
7.54065475463867 61.5238095238095
9.43085970878601 61.7142857142857
10.8351606369019 61.8095238095238
12.7455938339233 61.8095238095238
14.6470879077911 61.7142857142857
16.5664762020111 61.7142857142857
17.9700488090515 61.4285714285714
19.8871760368347 61.5238095238095
21.877098274231 61.7142857142857
23.7437811851501 62.0952380952381
25.1904420852661 62.5714285714286
27.0815311431885 63.8095238095238
28.9699927330017 65.1428571428572
30.8776287078857 66.3809523809524
31.9371017456055 67.1428571428572
33.8280087947845 68.5714285714286
34.9592000484467 69.8095238095238
36.8460198402405 72.2857142857143
38.2621229171753 74.3809523809524
40.1294054508209 76.2857142857143
42.0015071392059 77.7142857142857
43.859371805191 78.4761904761905
45.3107107639313 78.9523809523809
47.2096750736237 79.6190476190476
49.0738535404205 79.6190476190476
50.9947299480438 80.3809523809524
52.4271988391876 80.4761904761905
54.3406090736389 80.4761904761905
56.2781112194061 80.8571428571429
58.1644431591034 80.8571428571429
59.2299629688263 80.952380952381
61.141882276535 80.952380952381
63.0120406150818 80.8571428571428
64.9232442378998 81.0476190476191
65.6415172100067 81.1428571428571
67.5749418735504 81.4285714285714
69.4717048168182 81.5238095238095
71.4279217243195 81.6190476190476
72.8913562774658 81.6190476190476
74.7667907714844 81.6190476190476
76.6311904907227 81.7142857142857
78.5480662822723 81.8095238095238
79.9855391025543 81.9047619047619
81.9133882522583 81.9047619047619
83.8275811672211 81.7142857142857
85.6988813877106 82.0952380952381
86.830659198761 82
88.725977563858 82.0952380952381
90.5986225605011 82.0952380952381
92.4963326931 81.9047619047619
93.9224049091339 81.9047619047619
95.7779573917389 81.9047619047619
96.9180153369904 82
98.7893461704254 82.2857142857143
100.231992483139 82.2857142857143
102.19532251358 82.3809523809524
104.113616085052 82.3809523809524
106.024155092239 82.3809523809524
107.463712310791 82.3809523809524
109.421467590332 82.3809523809524
111.328037548065 82.3809523809524
113.196814489365 82.3809523809524
114.243938016891 82.3809523809524
116.189148759842 82.4761904761905
118.051090955734 82.5714285714286
120.01628575325 82.6666666666667
121.496128416061 82.6666666666667
123.365881919861 82.6666666666667
125.226609134674 82.6666666666667
127.164329195023 82.7619047619048
127.84545750618 82.7619047619048
129.715088319778 82.6666666666667
131.553492164612 82.6666666666667
133.417690324783 82.6666666666667
134.821751117706 82.6666666666667
136.687398958206 82.6666666666667
138.569668245316 82.6666666666667
140.477064561844 82.6666666666667
141.527513122559 82.6666666666667
143.397725486755 82.6666666666667
145.331127786636 82.6666666666667
147.255424690247 82.6666666666667
148.359981775284 82.6666666666667
150.621307468414 82.6666666666667
152.587673854828 82.6666666666667
154.516018867493 82.6666666666667
155.896261835098 82.5714285714286
157.768652439117 82.5714285714286
158.90481672287 82.6666666666667
160.789396858215 82.6666666666667
161.852890443802 82.5714285714286
164.040152215958 82.4761904761905
165.928826379776 82.4761904761905
167.792389297485 82.4761904761905
168.869053173065 82.4761904761905
170.760714578629 82.3809523809524
172.621715784073 82.3809523809524
174.446911478043 82.3809523809524
175.510928726196 82.4761904761905
177.720569038391 82.4761904761905
179.600006961823 82.4761904761905
181.518051862717 82.4761904761905
182.967929506302 82.4761904761905
184.91376748085 82.4761904761905
186.79220957756 82.5714285714286
188.730482053757 82.4761904761905
189.100766420364 82.4761904761905
191.322060823441 82.3809523809524
193.248282814026 82.3809523809524
195.105732488632 82.3809523809524
196.145779943466 82.3809523809524
198.035275030136 82.3809523809524
199.898157310486 82.4761904761905
201.783017683029 82.4761904761905
202.885470676422 82.4761904761905
205.103067970276 82.4761904761905
207.009948825836 82.2857142857143
208.879050779343 82.2857142857143
210.345342588425 82.2857142857143
212.246946001053 82.2857142857143
214.103614854813 82.2857142857143
215.993298721313 82.1904761904762
217.073727798462 82.2857142857143
219.289979600906 82.3809523809524
220.515578556061 82.3809523809524
222.450581216812 82.3809523809524
223.521905422211 82.3809523809524
225.376316261291 82.2857142857143
227.221725654602 82.2857142857143
229.131285047531 82.2857142857143
230.548656892776 82.2857142857143
232.405953359604 82.2857142857143
234.316825580597 82.2857142857143
236.256581306458 82.2857142857143
237.718823194504 82.2857142857143
239.551573991776 82.1904761904762
241.490371227264 82.1904761904762
243.419442129135 82.1904761904762
244.89688038826 82.1904761904762
246.761911678314 82.1904761904762
248.7189930439 82.0952380952381
250.635789203644 82.0952380952381
};
\addlegendentry{10\%}
\addplot [red255290, dashed, mark=*, mark size=1, mark options={solid}]
table {%
4.89611015319824 61.4285714285714
5.50591926574707 61.4285714285714
8.40583052635193 61.6190476190476
11.2301819801331 61.8095238095238
14.0538328647614 61.9047619047619
16.8491012096405 61.7142857142857
19.6522707462311 62.0952380952381
22.4260688304901 62.3809523809524
25.2263686180115 62.952380952381
28.1268455505371 64.1904761904762
30.9646714687347 66.5714285714286
33.7516958236694 69.4285714285714
36.5469467639923 72.8571428571428
38.7606802463531 74.9523809523809
42.1619637012482 79.6190476190476
44.9626482486725 81.1428571428571
47.7161621570587 81.6190476190476
50.5871609687805 82.3809523809524
53.4217401027679 83.047619047619
56.2359489440918 83.047619047619
58.9948250293732 83.2380952380952
61.8181144714355 83.5238095238095
64.6322612285614 83.3333333333333
67.4561078548431 83.5238095238095
70.3708894729614 83.5238095238095
73.1669690132141 83.7142857142857
75.4113099098206 83.7142857142857
78.7237940311432 83.8095238095238
81.5368499755859 83.8095238095238
84.4678694725037 84
87.2067606925964 84
89.4838761806488 84
92.3313529968262 84
95.0961734294891 83.8095238095238
97.9719235897064 84.0952380952381
100.770885181427 84.2857142857143
103.611705446243 84.2857142857143
106.524723958969 84.1904761904762
109.359414434433 84.0952380952381
112.22438788414 84
115.024213314056 84.2857142857143
117.835305929184 84.2857142857143
120.667639017105 84.3809523809524
123.451947259903 84.3809523809524
126.368460321426 84.3809523809524
129.306394577026 84.3809523809524
131.019387340546 84.3809523809524
133.787702131271 84.4761904761905
136.674323987961 84.3809523809524
138.920525407791 84.3809523809524
142.192518997192 84.6666666666667
145.058482837677 84.6666666666667
147.244238710403 84.5714285714286
150.601484966278 84.6666666666667
153.402342987061 84.7619047619048
156.194509220123 84.6666666666667
159.027980613709 84.8571428571428
161.93503985405 84.5714285714286
164.817477893829 84.7619047619047
167.594429206848 84.6666666666666
170.412575435638 84.6666666666666
172.683518028259 84.8571428571428
175.480949354172 84.8571428571428
177.686825704575 84.8571428571428
180.981029891968 84.9523809523809
183.669411993027 85.0476190476191
186.626335430145 84.8571428571428
189.454277515411 84.8571428571428
192.243817901611 84.7619047619048
195.032463741302 84.7619047619048
197.826151752472 84.7619047619048
200.680549669266 84.952380952381
203.437863492966 84.952380952381
206.279640102386 84.952380952381
209.111816215515 84.952380952381
212.008021259308 84.952380952381
214.802338981628 84.952380952381
217.534843635559 85.1428571428571
220.277496337891 85.3333333333333
223.13670372963 85.4285714285714
225.913457250595 85.4285714285714
228.766126966476 85.4285714285714
231.626363372803 85.4285714285714
234.518062829971 85.4285714285714
237.382259464264 85.4285714285714
240.191265201569 85.5238095238095
242.974007129669 85.4285714285714
245.759115076065 85.2380952380952
248.516789627075 85.4285714285714
251.581876850128 85.2380952380952
};
\addlegendentry{20\%}
\addplot [darkorange2551480, dashed, mark=*, mark size=1, mark options={solid}]
table {%
6.92584609985352 61.4285714285714
10.5151941776276 61.3333333333333
14.9225340366364 61.6190476190476
19.432484960556 61.4285714285714
23.8808061122894 61.7142857142857
28.2863722324371 62.0952380952381
32.7410010814667 62.6666666666667
37.1872175216675 63.2380952380952
41.6591190814972 66.2857142857143
46.0753615856171 71.3333333333333
50.5141795635223 76.1904761904762
55.0714073181152 79.5238095238095
59.4996929168701 82
63.9830903530121 82.9523809523809
68.4037806510925 83.6190476190476
72.8186039447784 84.4761904761905
77.372851228714 84.5714285714286
81.737136554718 84.3809523809524
86.2496625900269 84.2857142857143
90.7466389656067 84.2857142857143
95.3029866218567 84.1904761904762
99.7566875934601 84.1904761904762
104.216941642761 84.2857142857143
108.747588825226 84.1904761904762
113.138658905029 84.1904761904762
117.520470285416 84.2857142857143
121.822071170807 84.0952380952381
126.299953460693 83.8095238095238
130.839840459824 83.8095238095238
135.458820438385 83.7142857142857
139.948149776459 83.8095238095238
144.445042514801 83.9047619047619
148.903579235077 84.0952380952381
153.31237578392 84.0952380952381
157.801959228516 84.1904761904762
162.18702750206 84.3809523809524
166.630939292908 84.3809523809524
171.228766918182 84.4761904761905
175.681766700745 84.5714285714286
180.188776111603 84.8571428571428
184.772325611114 84.8571428571428
189.164321517944 84.7619047619048
193.70549864769 85.0476190476191
198.194571638107 85.2380952380952
202.677945423126 85.0476190476191
207.055950737 85.3333333333333
211.567883110046 85.0476190476191
215.983783245087 85.0476190476191
220.309953975678 85.0476190476191
224.766782188416 85.0476190476191
229.259447479248 84.952380952381
233.753171157837 84.952380952381
238.182022476196 84.952380952381
242.594867181778 85.1428571428571
247.101494312286 85.0476190476191
251.633205747604 85.2380952380952
};
\addlegendentry{30\%}
\addplot [yellow22825518, dashed, mark=*, mark size=1, mark options={solid}]
table {%
9.57371940612793 61.4285714285714
13.1079522132874 61.8095238095238
18.9799912452698 61.8095238095238
24.864520740509 61.7142857142857
30.5035231590271 61.4285714285714
36.3322012424469 60.9523809523809
42.1520233631134 61.7142857142857
47.8357749938965 65.2380952380952
53.5323059082031 68.9523809523809
59.1573795318604 73.5238095238095
64.7749289989471 76.8571428571428
70.4675362110138 80.5714285714286
76.1357436180115 82.4761904761905
82.0169853687286 83.2380952380952
87.7845330238342 83.5238095238095
90.0811986923218 83.8095238095238
99.3793427467346 84.1904761904762
105.092013454437 84.0952380952381
110.909537553787 84
116.442808151245 84.3809523809524
122.085288715363 84
127.719506072998 83.9047619047619
133.487216186523 83.5238095238095
139.401715803146 83.5238095238095
145.085897159576 83.4285714285714
150.842044115067 83.5238095238095
156.521224927902 83.1428571428571
162.231451511383 82.9523809523809
168.091255760193 82.6666666666667
173.735045862198 82.7619047619047
176.006289052963 82.7619047619047
185.002396869659 82.4761904761905
190.684195756912 82.7619047619048
196.440318346024 82.3809523809524
202.094352912903 82.3809523809524
207.750426006317 82.2857142857143
213.344737482071 82.3809523809524
219.161036205292 82.3809523809524
224.973760128021 82.0952380952381
230.65581946373 82
236.569512224197 82
242.276014280319 82.0952380952381
247.822927427292 81.9047619047619
253.552560138702 82
};
\addlegendentry{40\%}
\addplot [lightgreen124255121, dashed, mark=*, mark size=1, mark options={solid}]
table {%
11.4135353088379 61.4285714285714
15.5311346530914 61.2380952380952
22.4894075870514 61.6190476190476
29.4456149101257 61.6190476190476
36.4712969779968 61.2380952380952
43.3996119976044 62.5714285714286
50.5486529827118 64.6666666666667
57.4055013656616 67.2380952380952
64.4163696289063 73.1428571428571
71.3131937980652 77.7142857142857
78.3837872505188 80.2857142857143
85.3414153575897 82.6666666666667
92.3515394210815 83.9047619047619
99.1986594200134 84.4761904761905
106.142526197433 84.2857142857143
113.226497459412 83.9047619047619
120.25051817894 83.7142857142857
127.311418056488 83.9047619047619
134.648930692673 84
141.648421621323 84.2857142857143
148.70078663826 84.2857142857143
155.741444826126 84
162.620490646362 84
169.58930888176 83.9047619047619
176.591585969925 83.8095238095238
183.553468132019 83.5238095238095
190.580014848709 83.4285714285714
197.400158786774 83.4285714285714
204.451155567169 83.4285714285714
211.35576634407 83.3333333333333
218.166674613953 83.5238095238095
225.166023206711 83.2380952380952
232.062416553497 83.1428571428571
238.846868038177 83.047619047619
245.703594636917 83.1428571428571
252.495811891556 83.1428571428571
};
\addlegendentry{50\%}
\addplot [cyan21255225, dashed, mark=*, mark size=1, mark options={solid}]
table {%
12.502206325531 61.4285714285714
13.9938461780548 61.6190476190476
21.8628584384918 61.6190476190476
29.4199700832367 61.6190476190476
37.168030166626 61.4285714285714
44.7855299472809 62.0952380952381
52.3668079853058 65.1428571428572
60.1366874217987 70.4761904761905
68.0455518722534 73.1428571428571
75.6139791488647 78.2857142857143
83.3375648498535 81.3333333333333
91.2836173534393 84.0952380952381
98.9658998012543 84.5714285714286
106.752843809128 84.8571428571428
114.770917844772 84.952380952381
122.527950239182 84.8571428571428
130.141804409027 84.5714285714286
137.856242847443 84.4761904761905
145.501202344894 84
153.203642654419 83.7142857142857
161.185816955566 83.7142857142857
169.034679079056 83.8095238095238
176.768963241577 83.4285714285714
184.248024320602 83.3333333333333
192.098141384125 83.4285714285714
199.799188995361 83.2380952380952
207.515574073791 82.9523809523809
215.375749826431 82.9523809523809
223.242349767685 82.9523809523809
231.003547668457 83.1428571428572
238.773049736023 83.0476190476191
246.451361513138 82.7619047619047
254.053384065628 82.8571428571428
};
\addlegendentry{60\%}
\addplot [dodgerblue0128255, dashed, mark=*, mark size=1, mark options={solid}]
table {%
13.9964662075043 61.4285714285714
19.183938741684 61.5238095238095
27.8976254463196 61.4285714285714
36.407507610321 61.9047619047619
44.9960407733917 62.0952380952381
53.5448992729187 64
62.2820516586304 69.3333333333333
70.9445230484009 73.4285714285714
79.5700613021851 77.4285714285714
88.4025691986084 81.2380952380952
96.9813408374786 84
105.861007356644 84.1904761904762
114.605151224136 84.1904761904762
123.361117506027 84.3809523809524
132.156996059418 84.0952380952381
140.996408891678 84
149.584428596497 84.0952380952381
158.385422420502 83.8095238095238
167.226110839844 83.7142857142857
175.736115694046 83.5238095238095
184.136945199966 83.4285714285714
192.81523566246 83.5238095238095
201.673490333557 83.5238095238095
210.30283241272 83.3333333333333
219.013444948196 83.5238095238095
227.552308511734 83.3333333333333
236.277521705627 83.3333333333333
245.070337247848 83.1428571428572
253.787868070602 83.0476190476191
};
\addlegendentry{70\%}
\addplot [blue, dashed, mark=*, mark size=1, mark options={solid}]
table {%
14.4601809024811 61.4285714285714
16.2947095870972 61.4285714285714
25.593458366394 61.3333333333333
35.3110456943512 61.047619047619
45.1214558124542 61.6190476190476
54.4904423236847 63.1428571428571
64.0754709720612 66.0952380952381
73.269922208786 72.2857142857143
82.6146405220032 76.4761904761905
92.1212346076965 79.8095238095238
101.601522350311 82.4761904761905
111.306623458862 83.7142857142857
120.863529348373 84.1904761904762
130.321436452866 83.9047619047619
139.759170436859 83.9047619047619
149.467693424225 83.9047619047619
159.048106098175 83.8095238095238
168.587756443024 83.7142857142857
178.139263105392 83.5238095238095
187.792557525635 83.4285714285714
197.125254583359 83.2380952380952
206.697101211548 83.2380952380952
216.147174406052 83.2380952380952
225.589801216125 83.1428571428571
235.112144136429 82.8571428571428
244.606186389923 83.047619047619
253.971092796326 83.3333333333333
};
\addlegendentry{80\%}
\addplot [navy00127, dashed, mark=*, mark size=1, mark options={solid}]
table {%
14.7284090518951 61.4285714285714
16.7130358695984 61.3333333333333
26.852290725708 61.4285714285714
37.3170840740204 61.3333333333333
47.5842183113098 61.1428571428571
57.9010594367981 63.7142857142857
68.2653579235077 67.9047619047619
78.5202262401581 70.9523809523809
88.7467342853546 76.1904761904762
99.1922794342041 80.4761904761905
109.397260046005 83.5238095238095
119.729198551178 84.2857142857143
129.86603884697 84.4761904761905
140.067887639999 84.7619047619047
150.523748445511 84.6666666666666
160.799932003021 84.6666666666666
171.143469524384 84.3809523809524
181.311746358871 84.0952380952381
191.403445100784 83.7142857142857
201.77010974884 83.5238095238095
211.950331068039 83.4285714285714
222.369553995132 83.3333333333333
232.605009555817 83.1428571428571
242.918841075897 83.2380952380952
253.293868350983 83.047619047619
};
\addlegendentry{90\%}

\addplot [ultra thick, gray, dashed, mark=*, mark size=2, mark options={solid}]
table {%
14.2887599468231 61.4285714285714
24.6740455150604 61.5238095238095
35.141822719574 61.6190476190476
45.638599729538 62.1904761904762
56.2690583229065 64.4761904761905
66.8843245983124 66.7619047619048
77.7658805847168 71.1428571428571
88.6925711154938 76.4761904761905
99.2472709655762 80.2857142857143
109.508946180344 82.5714285714286
119.790633487701 83.1428571428571
130.337510061264 83.9047619047619
140.955986309052 83.7142857142857
151.609069061279 84
162.209997320175 84.0952380952381
172.600471687317 84.2857142857143
183.052301597595 84.1904761904762
193.526771879196 84
204.086597251892 83.8095238095238
214.457344055176 83.7142857142857
225.259791564941 83.7142857142857
235.826626968384 83.4285714285714
246.619962596893 83.4285714285714
257.192080831528 83.6190476190476
};
\addlegendentry{100\%}

\end{axis}

\end{tikzpicture}

%% file: plots/pgf_plots/mondial_religion_1ep.tex
\begin{tikzpicture}

\definecolor{blue}{RGB}{0,0,255}
\definecolor{cyan21255225}{RGB}{21,255,225}
\definecolor{darkgray176}{RGB}{176,176,176}
\definecolor{darkorange2551480}{RGB}{255,148,0}
\definecolor{dodgerblue0128255}{RGB}{0,128,255}
\definecolor{gray}{RGB}{128,128,128}
\definecolor{lightgreen124255121}{RGB}{124,255,121}
\definecolor{maroon12700}{RGB}{127,0,0}
\definecolor{navy00127}{RGB}{0,0,127}
\definecolor{red255290}{RGB}{255,29,0}
\definecolor{yellow22825518}{RGB}{228,255,18}

\begin{axis}[
tick align=outside,
tick pos=left,
x grid style={darkgray176},
xmin=2.03753677368164, xmax=271.564446582794,
xtick style={color=black},
y grid style={darkgray176},
ymin=59.5095238095238, ymax=87.0619047619047,
ytick style={color=black},
yticklabel style={rotate=90.0},
label style={font=\Huge},
xlabel={\textbf{Time (sec)}},
tick label style={font=\huge}
]
\addplot [ultra thick, gray, dashed, mark=*, mark size=2, mark options={solid}]
table {%
14.2887599468231 61.4285714285714
24.6740455150604 61.5238095238095
35.141822719574 61.6190476190476
45.638599729538 62.1904761904762
56.2690583229065 64.4761904761905
66.8843245983124 66.7619047619048
77.7658805847168 71.1428571428571
88.6925711154938 76.4761904761905
99.2472709655762 80.2857142857143
109.508946180344 82.5714285714286
119.790633487701 83.1428571428571
130.337510061264 83.9047619047619
140.955986309052 83.7142857142857
151.609069061279 84
162.209997320175 84.0952380952381
172.600471687317 84.2857142857143
183.052301597595 84.1904761904762
193.526771879196 84
204.086597251892 83.8095238095238
214.457344055176 83.7142857142857
225.259791564941 83.7142857142857
235.826626968384 83.4285714285714
246.619962596893 83.4285714285714
257.192080831528 83.6190476190476
};
\addplot [maroon12700, dashed, mark=*, mark size=1, mark options={solid}]
table {%
14.698076915741 61.4285714285714
15.2983225822449 61.4285714285714
17.7558226108551 61.7142857142857
20.1364030361176 61.8095238095238
22.5962238311768 61.8095238095238
24.97872838974 61.9047619047619
27.3767763614655 61.9047619047619
29.7644388198853 61.8095238095238
32.273389339447 61.8095238095238
34.6336396694183 62.0952380952381
36.9490369319916 62.5714285714286
39.3037325382233 63.5238095238095
41.7493743896484 64.3809523809524
44.0213900566101 66.7619047619048
46.3241057872772 69.5238095238095
48.7335057735443 72.1904761904762
51.1955483436584 75.5238095238095
53.5363167285919 78.7619047619048
55.9738416671753 80.9523809523809
58.392813539505 82.5714285714286
60.7431192874908 83.1428571428572
63.2809841632843 83.4285714285714
65.7106703281403 83.9047619047619
68.1409956455231 84.4761904761905
70.4786836624146 84.6666666666667
72.768545627594 84.4761904761905
75.1873031139374 84.6666666666667
77.5726396560669 84.5714285714286
79.8704181671143 84.7619047619048
81.8400074958801 84.8571428571428
84.7948585510254 84.952380952381
87.2892577171326 85.0476190476191
89.5895990371704 85.0476190476191
92.0208185195923 85.1428571428571
93.9601778507233 85.1428571428571
96.4577213764191 85.2380952380952
98.8982103347778 85.1428571428571
101.322038936615 85.2380952380952
103.587348175049 85.3333333333333
106.000684499741 85.3333333333333
108.343754673004 85.2380952380952
110.810262680054 85.1428571428571
113.180758190155 85.2380952380952
115.65325551033 85.0476190476191
118.089742708206 84.952380952381
120.504594898224 84.8571428571428
122.856002998352 85.0476190476191
125.204787874222 84.952380952381
127.63148021698 84.8571428571428
130.059339523315 84.952380952381
132.423733377457 85.0476190476191
133.82831401825 84.952380952381
136.158626651764 84.952380952381
138.572796535492 85.0476190476191
140.864538431168 85.0476190476191
143.217523622513 84.952380952381
145.621880149841 84.952380952381
147.54322681427 85.0476190476191
149.913923311234 84.952380952381
152.308683204651 84.8571428571428
155.240330791473 84.952380952381
157.6011531353 84.952380952381
159.839221286774 84.8571428571428
162.32038731575 84.8571428571428
164.673073959351 84.7619047619048
167.03406329155 84.7619047619048
169.460466957092 84.7619047619048
171.971063375473 84.6666666666667
173.955243444443 84.7619047619048
176.288305997849 84.7619047619048
178.779683542252 84.8571428571428
181.170362567902 84.952380952381
183.551928710937 84.8571428571428
185.92118473053 84.8571428571428
188.383582687378 84.8571428571428
190.7720911026 84.8571428571428
193.202798032761 84.952380952381
195.604696321487 84.952380952381
197.963367986679 85.1428571428571
200.378071689606 85.0476190476191
202.753305768967 84.952380952381
205.198731422424 85.0476190476191
207.75115237236 85.0476190476191
210.192956829071 85.1428571428571
212.555962085724 85.0476190476191
215.016266393662 85.2380952380952
217.450374937057 85.3333333333333
219.900466585159 85.1428571428571
222.262853765488 85.3333333333333
224.756720638275 85.2380952380952
227.117399454117 85.3333333333333
229.612944459915 85.2380952380952
231.921159410477 85.2380952380952
234.378510379791 85.2380952380952
236.806293487549 85.2380952380952
239.229286289215 85.1428571428571
241.600559568405 85.1428571428571
244.111671733856 85.2380952380952
246.592567825317 85.2380952380952
248.958328962326 85.2380952380952
251.315862131119 85.2380952380952
};
\addplot [red255290, dashed, mark=*, mark size=1, mark options={solid}]
table {%
15.9686493396759 61.4285714285714
17.1397330760956 61.5238095238095
20.5230354309082 61.6190476190476
23.7389409065247 61.5238095238095
26.9170216560364 61.4285714285714
30.0455723762512 61.5238095238095
32.4687358379364 61.8095238095238
34.41153383255 62.0952380952381
37.3985941886902 62.1904761904762
40.497519826889 62.6666666666667
43.7065583229065 63.7142857142857
46.1784689903259 65.4285714285714
49.3143317699432 68.0952380952381
52.5431471347809 69.6190476190476
54.6322735309601 71.2380952380952
59.026153087616 74.4761904761905
60.1889125823975 74.4761904761905
63.4609766960144 77.2380952380952
66.643652009964 80.2857142857143
69.7137184619904 82
72.8315605163574 83.4285714285714
75.3466693878174 84.0952380952381
78.5470782279968 84.1904761904762
80.430567741394 84.4761904761905
83.3861630439758 84.5714285714286
86.5550084114075 84.6666666666667
89.032096862793 84.5714285714286
92.2294733524323 84.5714285714286
94.2115564346313 84.5714285714286
98.5888791561127 84.8571428571428
101.872082853317 84.8571428571428
103.813532876968 84.7619047619048
106.347950983047 84.8571428571428
109.440932798386 84.952380952381
112.595522451401 84.952380952381
115.837976264954 84.8571428571428
119.045490694046 84.8571428571428
121.506799697876 84.8571428571428
123.344175195694 84.7619047619048
126.536345243454 84.8571428571428
129.72468457222 84.952380952381
132.829197454453 84.952380952381
135.321566200256 84.952380952381
138.554370975494 84.952380952381
141.857542467117 84.952380952381
145.103267049789 84.952380952381
146.963045024872 85.0476190476191
149.344386291504 85.1428571428571
152.524838876724 85.1428571428571
155.667241668701 85.1428571428571
158.95153875351 85.1428571428571
162.195778226852 85.1428571428571
164.623935222626 85.1428571428571
165.349306869507 85.1428571428571
169.75160908699 85.0476190476191
172.837770700455 85.2380952380952
174.744931316376 85.2380952380952
179.124488592148 85.2380952380952
181.55909538269 85.2380952380952
184.673590278625 85.2380952380952
187.878416824341 85.1428571428571
189.779911613464 85.1428571428571
192.91748046875 85.1428571428571
195.370917797089 85.1428571428571
198.5396900177 85.1428571428571
201.714744710922 85.1428571428571
204.884667730331 85.3333333333333
208.048439359665 85.6190476190476
209.204015445709 85.6190476190476
212.355313205719 85.4285714285714
215.617087745667 85.7142857142857
218.788016080856 85.4285714285714
221.995576620102 85.6190476190476
224.473592996597 85.6190476190476
227.63522400856 85.5238095238095
230.867818164825 85.6190476190476
232.777125835419 85.5238095238095
235.955681371689 85.6190476190476
238.44534163475 85.5238095238095
241.671637439728 85.6190476190476
244.781603145599 85.4285714285714
247.94198474884 85.3333333333333
251.208693885803 85.4285714285714
};
\addplot [darkorange2551480, dashed, mark=*, mark size=1, mark options={solid}]
table {%
17.5749535083771 61.4285714285714
18.4654597759247 61.6190476190476
22.7276178359985 61.8095238095238
26.8849370956421 61.5238095238095
31.1378845214844 61.6190476190476
35.3498503684998 61.4285714285714
39.5399863243103 61.8095238095238
43.6704662799835 62.5714285714286
47.9224657058716 63.6190476190476
51.9917791366577 66.2857142857143
56.1059422969818 71.3333333333333
60.1610990524292 76.2857142857143
64.3527618408203 80.0952380952381
68.610099363327 81.9047619047619
72.7452664375305 83.3333333333333
77.1047095775604 83.7142857142857
81.2648400306702 84.3809523809524
85.5092399597168 84.952380952381
89.6749018192291 85.7142857142857
93.9379183292389 85.8095238095238
98.0655705928802 85.7142857142857
102.228067016602 85.4285714285714
106.404744052887 85.2380952380952
110.469295597076 85.0476190476191
114.601793861389 84.952380952381
118.86492228508 84.952380952381
123.115173578262 84.8571428571428
127.144150018692 84.952380952381
131.220208215714 84.8571428571428
135.246356773376 84.8571428571428
138.528792905808 84.8571428571428
141.908692026138 84.952380952381
146.925852298737 84.952380952381
151.051691198349 84.952380952381
155.172726774216 84.8571428571428
159.286485290527 84.8571428571428
163.563646841049 84.6666666666667
167.742686319351 84.952380952381
171.954366445541 84.952380952381
176.161936378479 84.952380952381
180.400860214233 84.952380952381
184.684672212601 84.8571428571428
188.948341464996 84.8571428571428
193.165257072449 84.8571428571428
197.268475437164 84.6666666666667
201.509207963943 84.6666666666667
205.875157642365 84.6666666666667
210.041715192795 84.7619047619048
214.365529489517 84.6666666666667
218.415070438385 84.7619047619048
222.570722484589 84.6666666666667
226.72094373703 84.8571428571428
230.964309740067 84.7619047619048
235.141293954849 84.8571428571428
239.268080329895 84.8571428571428
243.382287931442 84.8571428571428
247.74474029541 84.7619047619048
251.92183175087 84.8571428571428
};
\addplot [yellow22825518, dashed, mark=*, mark size=1, mark options={solid}]
table {%
20.315625667572 61.4285714285714
21.3724802017212 61.6190476190476
26.8501186847687 61.9047619047619
32.3384045600891 61.8095238095238
37.6390995979309 61.2380952380952
43.0790048599243 60.8571428571429
48.4887998580933 62
53.9272475719452 65.4285714285714
59.2745299816132 68.5714285714286
64.7169683456421 73.3333333333333
70.1015565395355 76.4761904761905
75.481813621521 80.3809523809524
80.9638800621033 82.1904761904762
85.3398586273193 83.9047619047619
91.8392519950867 85.5238095238095
97.494935798645 85.7142857142857
102.986922740936 85.5238095238095
108.420447015762 85.4285714285714
113.855384635925 85.4285714285714
119.222969388962 85.4285714285714
124.727639102936 85.2380952380952
130.163810491562 85.4285714285714
135.611229944229 85.1428571428571
139.899918317795 84.6666666666666
145.396133089066 84.1904761904762
150.730261611938 84
155.109507083893 84
161.578715705872 83.8095238095238
167.204794692993 83.6190476190476
172.610722780228 83.4285714285714
178.076874494553 83.2380952380952
183.533099603653 83.2380952380952
189.074991178513 83.1428571428571
194.401624298096 83.047619047619
199.730063295364 82.8571428571428
205.197498035431 82.8571428571428
210.663104295731 82.8571428571428
216.099415922165 82.8571428571428
221.628730678558 82.9523809523809
227.043706083298 83.1428571428571
232.526694440842 82.9523809523809
237.964596796036 82.9523809523809
243.367413520813 82.9523809523809
248.738407278061 82.8571428571428
254.188113069534 82.7619047619047
};
\addplot [lightgreen124255121, dashed, mark=*, mark size=1, mark options={solid}]
table {%
22.1155278682709 61.4285714285714
23.4218801498413 61.6190476190476
30.1521808624268 61.3333333333333
36.7432933807373 61.3333333333333
43.4808303833008 61.1428571428571
50.0476371765137 60.8571428571429
56.6867337226868 64.5714285714286
63.2916943073273 68.9523809523809
69.92965259552 71.4285714285714
76.465305185318 75.5238095238095
83.2375886440277 78.9523809523809
89.9061935901642 81.7142857142857
96.6226331710815 83.4285714285714
103.146990776062 83.5238095238095
109.765581226349 83.5238095238095
116.525564527512 83.3333333333333
123.196469545364 83.5238095238095
129.91145772934 83.5238095238095
136.43214802742 83.8095238095238
140.493693780899 83.9047619047619
147.172504663467 83.8095238095238
153.915599298477 83.7142857142857
160.577753067017 83.7142857142857
167.226939439774 83.4285714285714
173.845785331726 83.4285714285714
180.555317306519 83.4285714285714
187.239912414551 83.1428571428571
193.842200994492 83.2380952380952
200.726867866516 83.1428571428571
207.612194013596 83.2380952380952
214.219128274918 83.1428571428571
220.949236965179 83.3333333333333
227.496819162369 83.2380952380952
234.00103969574 83.2380952380952
240.90187087059 83.047619047619
247.597198295593 82.8571428571428
254.348054790497 83.047619047619
};
\addplot [cyan21255225, dashed, mark=*, mark size=1, mark options={solid}]
table {%
23.4138660430908 61.4285714285714
29.3572049617767 61.5238095238095
36.7632818222046 61.6190476190476
44.4139524936676 61.047619047619
52.0028223991394 61.3333333333333
59.6010702610016 62.5714285714286
67.1767308235168 66.952380952381
74.4518922328949 70.5714285714286
82.0262671470642 77.3333333333333
89.6114675045013 79.7142857142857
97.170084810257 82.0952380952381
104.717012929916 83.9047619047619
112.159259366989 84.6666666666667
119.65551071167 84.1904761904762
127.251973485947 84.2857142857143
134.697405290604 84.2857142857143
142.363550424576 84.3809523809524
149.9572057724 84.2857142857143
157.442044830322 84
164.858349561691 84
172.380339431763 83.9047619047619
179.815113830566 83.7142857142857
187.229023551941 83.9047619047619
194.913473844528 83.6190476190476
202.340913486481 83.5238095238095
209.985002851486 83.5238095238095
217.626701116562 83.3333333333333
225.045229768753 83.4285714285714
232.551393413544 83.3333333333333
240.282469654083 83.2380952380952
247.763100719452 83.047619047619
255.427403211594 83.1428571428571
};
\addplot [dodgerblue0128255, dashed, mark=*, mark size=1, mark options={solid}]
table {%
24.767286157608 61.4285714285714
26.4942299365997 61.4285714285714
35.039031791687 61.6190476190476
43.804583978653 60.7619047619048
52.3825150966644 61.3333333333333
61.1048717975616 63.1428571428571
69.8326612472534 66.4761904761905
78.5995337486267 71.7142857142857
87.3775654792786 75.5238095238095
96.2064107894898 78.7619047619048
105.028326892853 82.1904761904762
113.886031675339 83.6190476190476
122.631420516968 84.2857142857143
131.571714925766 84.4761904761905
140.137595701218 84.3809523809524
148.968453025818 84.4761904761905
157.806707143784 84.6666666666667
166.45600605011 84.4761904761905
175.055436229706 84.4761904761905
183.755931282043 84.2857142857143
192.336297512054 84.1904761904762
201.195875787735 84
209.970942020416 84.0952380952381
218.830154037476 83.8095238095238
227.715101957321 83.9047619047619
236.514896535873 83.6190476190476
245.352146720886 83.6190476190476
254.000170707703 83.3333333333333
};
\addplot [blue, dashed, mark=*, mark size=1, mark options={solid}]
table {%
25.9834870815277 61.4285714285714
29.5794893264771 61.3333333333333
39.0074287891388 61.047619047619
48.8433278560638 61.8095238095238
58.5391793251038 62.8571428571429
68.3624243736267 64
78.0481202602386 68.7619047619048
87.7440629482269 73.5238095238095
97.3079568862915 77.7142857142857
106.988627290726 82.0952380952381
116.70828089714 84.1904761904762
126.220481109619 84.3809523809524
135.940368461609 84
145.727100324631 84.0952380952381
155.221948862076 84
164.926831197739 83.9047619047619
174.692549371719 83.8095238095238
184.396075153351 83.8095238095238
194.256896114349 83.6190476190476
204.030600643158 83.9047619047619
213.800581169128 83.8095238095238
223.494852018356 83.7142857142857
233.162691545486 83.5238095238095
242.856339073181 83.5238095238095
252.725468969345 83.3333333333333
};
\addplot [navy00127, dashed, mark=*, mark size=1, mark options={solid}]
table {%
26.4456541061401 61.4285714285714
37.0462847232819 61.6190476190476
47.7129239559174 61.5238095238095
58.4085873126984 62.5714285714286
69.1151517391205 64.0952380952381
79.5788269519806 67.2380952380953
90.2032649993896 70.4761904761905
100.763081455231 76.7619047619048
111.24517416954 81.2380952380952
122.030337285995 83.4285714285714
132.663840150833 83.9047619047619
143.003348588943 84.1904761904762
153.3164021492 84.3809523809524
153.3164021492 84.3809523809524
174.386693334579 84.5714285714286
184.956469345093 84.2857142857143
195.331844091415 84
206.325021886826 83.7142857142857
216.855684089661 83.5238095238095
227.630943584442 83.5238095238095
238.116107988358 83.4285714285714
248.754796791077 83.2380952380952
259.313223409653 83.1428571428571
};
\end{axis}

\end{tikzpicture}

%% file: plots/pgf_plots/mondial_religion_random.tex
\begin{tikzpicture}

\definecolor{blue00254}{RGB}{0,0,254}
\definecolor{darkgray176}{RGB}{176,176,176}
\definecolor{darkorange2551220}{RGB}{255,122,0}
\definecolor{deepskyblue0212255}{RGB}{0,212,255}
\definecolor{dodgerblue096255}{RGB}{0,96,255}
\definecolor{gold2552290}{RGB}{255,229,0}
\definecolor{gray}{RGB}{128,128,128}
\definecolor{greenyellow17025576}{RGB}{170,255,76}
\definecolor{maroon12700}{RGB}{127,0,0}
\definecolor{red254180}{RGB}{254,18,0}
\definecolor{turquoise76255170}{RGB}{76,255,170}

\begin{axis}[
tick align=outside,
tick pos=left,
x grid style={darkgray176},
xmin=-10.5480736446381, xmax=269.972555131912,
xtick style={color=black},
y grid style={darkgray176},
ymin=60.1857142857143, ymax=85.4333333333333,
ytick style={color=black},
yticklabel style={rotate=90.0},
label style={font=\Huge},
xlabel={\textbf{Time (sec)}},
tick label style={font=\huge}
]
\addplot [maroon12700, dashed, mark=*, mark size=1, mark options={solid}]
table {%
2.20286402702332 61.4285714285714
2.5560574054718 61.4285714285714
4.81719527244568 61.7142857142857
6.98703985214233 61.8095238095238
8.09621353149414 61.7142857142857
9.80806188583374 61.7142857142857
11.9219880580902 61.6190476190476
14.0913488864899 61.5238095238095
15.269357252121 61.4285714285714
16.9989255428314 61.4285714285714
18.6851957798004 61.5238095238095
20.8329288482666 61.8095238095238
21.9885646343231 61.7142857142857
24.2161321640015 61.7142857142857
26.4252452850342 62.0952380952381
27.5816652297974 62
29.2434527873993 62.2857142857143
31.3814942836761 62.7619047619048
33.0979339122772 62.9523809523809
34.3827078342438 63.047619047619
36.0464380741119 63.8095238095238
38.1289276599884 64.0952380952381
40.1656268596649 64.4761904761905
41.3765030384064 64.2857142857143
43.1224255084991 64
45.1950078964233 64
46.911291217804 63.9047619047619
48.144819688797 63.7142857142857
50.2481157779694 63.8095238095238
51.9357481956482 63.8095238095238
53.6217921257019 64
55.2157918453217 64.0952380952381
57.3468459606171 64
59.0664599895477 63.8095238095238
60.7067603588104 63.8095238095238
61.9825736522675 63.7142857142857
64.0690699100494 63.7142857142857
66.2006471633911 63.9047619047619
67.4450879573822 63.9047619047619
69.0152147769928 64.0952380952381
71.2237362861633 64.3809523809524
72.8422658920288 64.3809523809524
74.4284302711487 64.5714285714286
75.6778112888336 64.5714285714286
77.7234303474426 64.5714285714286
79.3838660240173 64.4761904761905
81.116321182251 64.5714285714286
82.8069970607758 64.5714285714286
84.9452337741852 64.2857142857143
86.5627013683319 64.3809523809524
88.158745765686 64.3809523809524
90.2878704547882 64.3809523809524
91.5691679477692 64.6666666666667
93.2599022865295 64.6666666666667
94.8796440124512 64.5714285714286
96.9761597633362 64.7619047619048
98.6745968341827 64.7619047619048
100.338742208481 64.9523809523809
101.993496084213 64.9523809523809
104.024143600464 64.9523809523809
105.274143838882 64.9523809523809
106.903352212906 64.9523809523809
108.538686084747 64.9523809523809
110.675727176666 64.9523809523809
112.327125787735 64.9523809523809
113.975758647919 65.047619047619
115.55315322876 65.047619047619
117.217192792892 65.047619047619
119.322666931152 65.047619047619
120.544356870651 64.9523809523809
122.757340717316 64.9523809523809
123.959217834473 64.9523809523809
126.085544872284 65.1428571428571
127.70125041008 65.047619047619
129.804258584976 65.1428571428571
131.032003116608 65.1428571428571
133.105868673325 65.1428571428571
134.349823093414 65.1428571428571
136.505779266357 65.047619047619
138.150311946869 65.2380952380952
139.824636268616 65.3333333333333
141.522894334793 65.2380952380952
143.139012670517 65.4285714285714
145.267298078537 65.4285714285714
146.862862491608 65.5238095238095
148.263938570023 65.5238095238095
149.854583311081 65.6190476190476
152.047924423218 65.7142857142857
153.693988180161 65.7142857142857
155.335053682327 65.6190476190476
157.017194080353 65.6190476190476
159.229451560974 65.2380952380952
160.881215381622 65.2380952380952
162.179206466675 65.2380952380952
163.79938454628 65.2380952380952
165.991610717773 65.1428571428571
167.641137552261 65.047619047619
169.286092615128 64.9523809523809
171.416089677811 64.9523809523809
173.077594470978 64.9523809523809
174.66575345993 64.9523809523809
176.000052547455 64.9523809523809
178.436469507217 64.9523809523809
179.757841014862 64.9523809523809
181.354567289352 65.047619047619
183.022372436523 64.9523809523809
185.117664957047 65.1428571428571
186.805748319626 65.047619047619
188.016576862335 65.047619047619
189.798623228073 65.047619047619
192.274497699738 65.2380952380952
193.554398345947 65.2380952380952
195.271407747269 65.2380952380952
196.907061624527 65.4285714285714
199.028919410706 65.5238095238095
200.657833433151 65.5238095238095
202.281101131439 65.5238095238095
203.587680149078 65.5238095238095
206.112809228897 65.4285714285714
207.315100383759 65.5238095238095
208.941495323181 65.5238095238095
211.086475849152 65.5238095238095
212.76278386116 65.4285714285714
213.942681360245 65.3333333333333
216.004522657394 65.4285714285714
217.805058097839 65.5238095238095
219.806217336655 65.5238095238095
221.003356266022 65.6190476190476
222.725706243515 65.6190476190476
224.848740530014 65.6190476190476
226.449073171616 65.6190476190476
227.650921773911 65.6190476190476
229.759567737579 65.6190476190476
231.854639434814 65.6190476190476
233.076473855972 65.6190476190476
235.21147646904 65.5238095238095
236.931221437454 65.5238095238095
238.708122777939 65.5238095238095
239.851449251175 65.4285714285714
241.970396232605 65.4285714285714
244.038555717468 65.4285714285714
245.689846229553 65.6190476190476
246.842672920227 65.6190476190476
248.91886806488 65.5238095238095
251.028347396851 65.6190476190476
};
\addplot [red254180, dashed, mark=*, mark size=1, mark options={solid}]
table {%
3.67874484062195 61.4285714285714
4.20755677223206 61.4285714285714
7.34230017662048 61.7142857142857
10.1954313278198 61.7142857142857
13.249497795105 61.7142857142857
15.5659972667694 61.6190476190476
18.4755847930908 61.9047619047619
20.2800256729126 61.8095238095238
23.812687921524 62.1904761904762
24.9022344112396 62.2857142857143
27.796724653244 62.9523809523809
30.8459454536438 64.1904761904762
33.911888551712 65.4285714285714
36.2371391296387 66.8571428571429
37.9990769863129 66.8571428571429
41.4451992511749 67.6190476190476
44.5767281532288 67.9047619047619
46.341162776947 68.1904761904762
48.7525456905365 68.4761904761905
51.6807912349701 68
53.914469575882 68.2857142857143
56.8310263633728 68.5714285714286
59.8435831546783 68.7619047619048
62.169130563736 68.5714285714286
65.0579323291779 68.4761904761905
66.3514638900757 68.5714285714286
69.2390008926392 68.4761904761905
71.5609379291534 68.3809523809524
74.0538522720337 68.7619047619048
77.5789571285248 68.8571428571428
80.4905513286591 68.9523809523809
82.4124357223511 69.0476190476191
85.8650266170502 69.2380952380952
87.056218957901 69.0476190476191
90.1274044036865 68.8571428571428
93.1014050960541 68.8571428571428
95.3029860496521 68.9523809523809
98.2444681167603 68.6666666666667
101.131577014923 68.7619047619048
103.46192278862 68.6666666666667
105.688878107071 68.8571428571428
107.573543071747 68.6666666666667
110.601211738586 68.6666666666667
113.471786165237 68.8571428571428
116.316774988174 69.2380952380952
118.605540037155 69.3333333333333
121.004152917862 69.5238095238095
123.960364341736 69.6190476190476
126.877605247498 69.5238095238095
128.044188261032 69.4285714285714
130.471744632721 69.5238095238095
133.960593414307 69.5238095238095
136.307574415207 69.5238095238095
138.026861906052 69.6190476190476
141.462241172791 69.6190476190476
144.495905065537 69.5238095238095
146.949105501175 69.6190476190476
149.294033050537 69.6190476190476
151.599306964874 69.9047619047619
153.45424118042 70
156.991566610336 70.1904761904762
159.914042520523 70.0952380952381
161.705145692825 70.0952380952381
165.123362064362 70
168.109235191345 69.9047619047619
169.261197042465 69.9047619047619
172.276213502884 69.8095238095238
174.584580755234 69.9047619047619
177.723731040955 69.9047619047619
180.681712293625 70.4761904761905
183.735848474503 70.4761904761905
186.065729904175 70.5714285714286
188.367624282837 70.6666666666667
190.132685518265 70.5714285714286
193.074275922775 70.7619047619048
195.362009382248 70.9523809523809
198.300484228134 70.7619047619048
201.228501701355 70.7619047619048
203.579786300659 70.8571428571428
206.497482967377 70.8571428571428
208.789296388626 71.0476190476191
210.584059810638 71.1428571428571
213.628666353226 71.5238095238095
216.652894115448 71.4285714285714
218.894453334808 71.5238095238095
221.359839296341 71.5238095238095
224.36432723999 71.3333333333333
227.372708082199 71.6190476190476
230.383541822433 71.4285714285714
231.559777593613 71.5238095238095
234.543445777893 71.8095238095238
236.957866096497 71.7142857142857
240.033679151535 71.4285714285714
242.390899658203 71.5238095238095
245.42188949585 71.3333333333333
248.339447164536 71.6190476190476
251.365196752548 71.4285714285714
};
\addplot [darkorange2551220, dashed, mark=*, mark size=1, mark options={solid}]
table {%
5.29536366462708 61.4285714285714
6.03780636787415 61.6190476190476
10.0103371143341 61.6190476190476
13.8571484088898 61.6190476190476
17.8142520904541 61.6190476190476
20.8548550605774 61.7142857142857
23.9361349582672 61.9047619047619
27.828705739975 62.1904761904762
31.6890737056732 62.4761904761905
35.5992658615112 63.4285714285714
38.5601714611053 64
41.6533357620239 65.047619047619
44.7249318122864 65.6190476190476
48.6919362068176 66.5714285714286
51.6419382572174 67.5238095238095
55.4281406879425 68.5714285714286
59.2718497276306 68.6666666666667
62.3839135169983 68.9523809523809
66.3388046741486 69.1428571428572
68.5782406806946 69.2380952380952
72.4765304088592 69.1428571428572
75.6814748764038 69.0476190476191
79.5352306365967 68.8571428571429
81.7150686264038 68.9523809523809
85.6619840621948 68.5714285714286
89.5064570903778 68.9523809523809
93.5388353347778 69.2380952380952
97.3542468070984 70
100.304812002182 70.4761904761905
103.472557353973 70.7619047619048
107.389168548584 70.7619047619048
111.300522470474 71.0476190476191
113.559470796585 71.4285714285714
117.526994228363 71.8095238095238
120.547348642349 72.2857142857143
124.490840959549 72.2857142857143
128.434368562698 72
130.69932923317 72.3809523809524
134.647332715988 72.0952380952381
138.501718950272 72.1904761904762
141.472593212128 72
145.279942178726 72.0952380952381
147.54248380661 72.1904761904762
151.554386615753 72.3809523809524
155.486897754669 72.6666666666667
158.560892963409 72.7619047619048
161.580827951431 72.9523809523809
165.498095989227 73.1428571428571
169.459468460083 73.0476190476191
173.352217817307 72.7619047619048
177.246036529541 72.6666666666667
179.377416944504 72.5714285714286
183.401493740082 72.5714285714286
186.56835436821 72.4761904761905
190.409612751007 72.1904761904762
192.670074033737 72.1904761904762
196.575399160385 72.4761904761905
199.684923744202 72.4761904761905
203.48676738739 72.5714285714286
207.39145026207 72.9523809523809
210.3661008358 72.9523809523809
214.322856664658 72.9523809523809
217.365810918808 73.047619047619
220.528563833237 72.8571428571428
223.519459819794 72.9523809523809
227.414934158325 72.6666666666667
231.428426837921 72.6666666666667
235.295291519165 72.6666666666667
238.334145784378 72.6666666666667
241.314185857773 72.7619047619047
245.210621070862 72.8571428571429
248.974774837494 72.7619047619048
252.886905193329 72.952380952381
};
\addplot [gold2552290, dashed, mark=*, mark size=1, mark options={solid}]
table {%
7.13324980735779 61.4285714285714
8.14458312988281 61.5238095238095
13.4950493335724 61.4285714285714
18.6382183074951 61.6190476190476
23.7520748615265 61.6190476190476
28.9152451515198 62.0952380952381
34.1060287952423 63.047619047619
39.2153942108154 65.1428571428572
44.3780384063721 67.6190476190476
49.4199508666992 69.5238095238095
54.6451024532318 71.0476190476191
59.7116995334625 71.7142857142857
65.0686508655548 73.0476190476191
70.1964127540588 73.1428571428571
75.221151971817 73.3333333333333
80.3087235450745 73.8095238095238
85.648721408844 74
89.8211145401001 74.2857142857143
94.9330551624298 74.2857142857143
100.087172317505 74.3809523809524
105.351909208298 75.0476190476191
110.550023317337 74.5714285714286
115.615069818497 74.6666666666667
120.892097759247 75.0476190476191
126.100480747223 75.4285714285714
131.350183200836 76
135.416867303848 76
140.611897516251 76.1904761904762
145.76192317009 76.1904761904762
149.861826705933 75.9047619047619
156.054418563843 76.0952380952381
161.239154100418 76.4761904761905
166.508921051025 76.6666666666667
171.585744333267 76.7619047619048
175.6440762043 76.8571428571429
180.880347394943 76.9523809523809
186.153267049789 76.8571428571429
191.414802360535 77.0476190476191
196.583431720734 77.3333333333333
201.814394235611 77.2380952380952
206.954569864273 77.4285714285714
212.15918545723 77.5238095238095
217.364687824249 77.6190476190476
222.544212341309 77.7142857142857
227.764798307419 77.7142857142857
233.087895393372 77.7142857142857
238.177415704727 77.7142857142857
243.371784877777 77.8095238095238
248.449631595612 78.0952380952381
253.484661388397 78.0952380952381
};
\addplot [greenyellow17025576, dashed, mark=*, mark size=1, mark options={solid}]
table {%
7.87784190177917 61.4285714285714
10.1780866146088 61.6190476190476
16.1934928417206 61.6190476190476
22.1922439575195 61.5238095238095
28.1608184814453 61.4285714285714
34.1921167373657 62.1904761904762
40.213250541687 63.047619047619
46.290804195404 64.2857142857143
52.3515053272247 66.0952380952381
58.4204647064209 69.0476190476191
64.417483997345 70.2857142857143
70.4305641651154 71.9047619047619
76.44318318367 73.0476190476191
82.423285150528 73.0476190476191
88.5573534488678 72.8571428571428
93.3995400905609 72.952380952381
99.3941611766815 73.4285714285714
105.324018955231 73.7142857142857
111.333005571365 74
117.068821763992 73.9047619047619
123.158693742752 73.7142857142857
129.284668731689 73.8095238095238
134.033761501312 73.9047619047619
137.769381856918 74.1904761904762
146.00122961998 74.3809523809524
151.786076068878 74.6666666666667
157.929083204269 74.3809523809524
163.90899939537 74.3809523809524
169.850817346573 74
174.403936767578 73.8095238095238
180.488977003098 73.5238095238095
184.25099105835 73.5238095238095
192.589888954163 73.8095238095238
198.499558019638 73.6190476190476
204.408600473404 73.9047619047619
210.490593242645 73.9047619047619
216.601633739471 74.1904761904762
222.689267730713 74.2857142857143
228.688927745819 74.2857142857143
234.744983768463 74.6666666666667
240.75641169548 74.4761904761905
246.721366024017 74.1904761904762
252.691428613663 74.5714285714286
};
\addplot [turquoise76255170, dashed, mark=*, mark size=1, mark options={solid}]
table {%
9.25400514602661 61.4285714285714
12.0080399513245 61.3333333333333
19.1549551010132 61.4285714285714
26.2433980941772 61.4285714285714
33.437696313858 61.3333333333333
40.6428973674774 62.2857142857143
47.8109988212585 64.7619047619048
55.0298783779144 65.9047619047619
61.9776961803436 69.0476190476191
69.1189252376556 71.6190476190476
76.333492231369 73.1428571428572
83.2490103244781 75.7142857142857
90.550492811203 75.5238095238095
97.6688824653625 76.1904761904762
104.934264707565 76.8571428571429
112.101764345169 77.4285714285714
119.304126262665 78
126.40150437355 77.9047619047619
133.303655004501 78.3809523809524
138.89040312767 78.4761904761905
146.183464717865 78.1904761904762
153.301200532913 78
160.255118274689 77.9047619047619
167.320825004578 77.6190476190476
174.410904121399 78
181.265344762802 77.9047619047619
188.240982198715 77.9047619047619
195.451939344406 78.0952380952381
202.603005838394 78
209.808545160294 78.0952380952381
217.146792840958 78.4761904761905
224.475642347336 79.4285714285714
231.456284046173 79.3333333333333
238.74966340065 79.4285714285714
246.090912389755 79.7142857142857
253.364640331268 79.5238095238095
};
\addplot [deepskyblue0212255, dashed, mark=*, mark size=1, mark options={solid}]
table {%
10.7787230014801 61.4285714285714
12.2879600524902 61.5238095238095
19.9785836219788 61.5238095238095
27.7072557449341 61.8095238095238
35.4797786712646 61.5238095238095
43.2144836425781 61.5238095238095
51.0227803230286 63.5238095238095
58.843785572052 66.7619047619048
66.5212213039398 71.5238095238095
74.3864619731903 75.5238095238095
82.3098615646362 77.0476190476191
89.9760604381561 79.0476190476191
97.6927713871002 80.9523809523809
105.547055482864 81.4285714285714
113.147517633438 82.1904761904762
121.035076236725 82.6666666666667
128.707287740707 83.047619047619
136.576956796646 83.3333333333333
144.181591033936 83.2380952380952
151.929181241989 83.047619047619
159.911274194717 82.8571428571428
167.708892440796 82.8571428571428
175.544827604294 82.7619047619047
183.312181568146 82.8571428571428
191.040288686752 82.8571428571428
198.770842647552 82.9523809523809
206.64641289711 82.9523809523809
214.521179389954 83.047619047619
222.314964056015 83.047619047619
229.962507724762 82.9523809523809
237.991020107269 83.047619047619
245.903901767731 83.1428571428571
253.790887880325 83.047619047619
};
\addplot [dodgerblue096255, dashed, mark=*, mark size=1, mark options={solid}]
table {%
12.1620532512665 61.4285714285714
15.5877549171448 61.6190476190476
24.3430017471313 61.6190476190476
32.981134223938 61.7142857142857
41.7203585147858 61.9047619047619
50.4697587966919 62.1904761904762
59.0254769325256 64.3809523809524
67.8258729934692 66.952380952381
76.5373094081879 71.6190476190476
84.9746689796448 75.8095238095238
93.7207908153534 79.1428571428572
102.487405157089 81.0476190476191
111.380782413483 82.4761904761905
120.106691598892 82.9523809523809
128.81602473259 82.5714285714286
137.1761926651 83.047619047619
144.035069942474 83.1428571428571
152.794446563721 83.2380952380952
161.235832929611 83.1428571428571
169.993304491043 83.2380952380952
178.827638101578 83.1428571428571
187.332048988342 83.2380952380952
196.014358091354 83.2380952380952
204.732096862793 83.5238095238095
213.238833284378 83.8095238095238
222.136850500107 83.7142857142857
230.986657762527 83.5238095238095
239.630443143845 83.9047619047619
248.424101781845 83.9047619047619
257.221617460251 84.0952380952381
};
\addplot [blue00254, dashed, mark=*, mark size=1, mark options={solid}]
table {%
13.4387179851532 61.4285714285714
15.3117681503296 61.6190476190476
24.7643397808075 61.6190476190476
34.3521013259888 61.3333333333333
43.7902143478394 61.9047619047619
53.3942825317383 62.5714285714286
62.9538483142853 65.1428571428572
72.8070870399475 67.0476190476191
82.436506319046 70.9523809523809
92.0432461738586 74
101.539973497391 76.6666666666667
111.028438949585 79.2380952380952
120.572354221344 80.6666666666667
130.081339359283 81.4285714285714
139.483454704285 82.6666666666667
148.901809263229 83.047619047619
158.415180253983 83.047619047619
167.900707197189 83.1428571428571
177.138372802734 82.8571428571428
186.528070497513 82.6666666666667
196.104494285583 82.4761904761905
205.709280967712 82.2857142857143
215.256691169739 82.3809523809524
224.686488676071 83.1428571428571
234.197661066055 83.2380952380952
243.804293012619 83.047619047619
253.512127065659 83.047619047619
};
\addplot [ultra thick, gray, dashed, mark=*, mark size=2, mark options={solid}]
table {%
14.2887599468231 61.4285714285714
24.6740455150604 61.5238095238095
35.141822719574 61.6190476190476
45.638599729538 62.1904761904762
56.2690583229065 64.4761904761905
66.8843245983124 66.7619047619048
77.7658805847168 71.1428571428571
88.6925711154938 76.4761904761905
99.2472709655762 80.2857142857143
109.508946180344 82.5714285714286
119.790633487701 83.1428571428571
130.337510061264 83.9047619047619
140.955986309052 83.7142857142857
151.609069061279 84
162.209997320175 84.0952380952381
172.600471687317 84.2857142857143
183.052301597595 84.1904761904762
193.526771879196 84
204.086597251892 83.8095238095238
214.457344055176 83.7142857142857
225.259791564941 83.7142857142857
235.826626968384 83.4285714285714
246.619962596893 83.4285714285714
257.192080831528 83.6190476190476
};
\end{axis}

\end{tikzpicture}

%% file: plots/short_pgf_plots_for_cikm/mondial_religion_k_var_short.tex
\begin{tikzpicture}

\definecolor{blue}{RGB}{0,0,255}
\definecolor{cyan21255225}{RGB}{21,255,225}
\definecolor{darkgray176}{RGB}{176,176,176}
\definecolor{darkorange2551480}{RGB}{255,148,0}
\definecolor{dodgerblue0128255}{RGB}{0,128,255}
\definecolor{gray}{RGB}{128,128,128}
\definecolor{lightgray204}{RGB}{204,204,204}
\definecolor{lightgreen124255121}{RGB}{124,255,121}
\definecolor{maroon12700}{RGB}{127,0,0}
\definecolor{navy00127}{RGB}{0,0,127}
\definecolor{red255290}{RGB}{255,29,0}
\definecolor{yellow22825518}{RGB}{228,255,18}

\begin{axis}[
yscale=\cikmshortenscaley,
legend cell align={left},
legend columns=2, 
legend style={
  fill opacity=0.8,
  draw opacity=1,
  text opacity=1,
  at={(0.97,0.03)},
  anchor=south east,
  draw=lightgray204,
  font= \LARGE,
  /tikz/column 2/.style={column sep=5pt,}
},
tick align=outside,
tick pos=left,
x grid style={darkgray176},
xmin=-9.49949369192124, xmax=269.891679618359,
xtick style={color=black},
y grid style={darkgray176},
ymin=59.7238095238095, ymax=86.7523809523809,
ytick style={color=black},
yticklabel style={rotate=90.0},
tick label style={font=\huge}
]

\addplot [maroon12700, dashed, mark=*, mark size=1, mark options={solid}]
table {%
3.20010509490967 61.4285714285714
3.54124207496643 61.4285714285714
5.61002097129822 61.6190476190476
7.54065475463867 61.5238095238095
9.43085970878601 61.7142857142857
10.8351606369019 61.8095238095238
12.7455938339233 61.8095238095238
14.6470879077911 61.7142857142857
16.5664762020111 61.7142857142857
17.9700488090515 61.4285714285714
19.8871760368347 61.5238095238095
21.877098274231 61.7142857142857
23.7437811851501 62.0952380952381
25.1904420852661 62.5714285714286
27.0815311431885 63.8095238095238
28.9699927330017 65.1428571428572
30.8776287078857 66.3809523809524
31.9371017456055 67.1428571428572
33.8280087947845 68.5714285714286
34.9592000484467 69.8095238095238
36.8460198402405 72.2857142857143
38.2621229171753 74.3809523809524
40.1294054508209 76.2857142857143
42.0015071392059 77.7142857142857
43.859371805191 78.4761904761905
45.3107107639313 78.9523809523809
47.2096750736237 79.6190476190476
49.0738535404205 79.6190476190476
50.9947299480438 80.3809523809524
52.4271988391876 80.4761904761905
54.3406090736389 80.4761904761905
56.2781112194061 80.8571428571429
58.1644431591034 80.8571428571429
59.2299629688263 80.952380952381
61.141882276535 80.952380952381
63.0120406150818 80.8571428571428
64.9232442378998 81.0476190476191
65.6415172100067 81.1428571428571
67.5749418735504 81.4285714285714
69.4717048168182 81.5238095238095
71.4279217243195 81.6190476190476
72.8913562774658 81.6190476190476
74.7667907714844 81.6190476190476
76.6311904907227 81.7142857142857
78.5480662822723 81.8095238095238
79.9855391025543 81.9047619047619
81.9133882522583 81.9047619047619
83.8275811672211 81.7142857142857
85.6988813877106 82.0952380952381
86.830659198761 82
88.725977563858 82.0952380952381
90.5986225605011 82.0952380952381
92.4963326931 81.9047619047619
93.9224049091339 81.9047619047619
95.7779573917389 81.9047619047619
96.9180153369904 82
98.7893461704254 82.2857142857143
100.231992483139 82.2857142857143
102.19532251358 82.3809523809524
104.113616085052 82.3809523809524
106.024155092239 82.3809523809524
107.463712310791 82.3809523809524
109.421467590332 82.3809523809524
111.328037548065 82.3809523809524
113.196814489365 82.3809523809524
114.243938016891 82.3809523809524
116.189148759842 82.4761904761905
118.051090955734 82.5714285714286
120.01628575325 82.6666666666667
121.496128416061 82.6666666666667
123.365881919861 82.6666666666667
125.226609134674 82.6666666666667
127.164329195023 82.7619047619048
127.84545750618 82.7619047619048
129.715088319778 82.6666666666667
131.553492164612 82.6666666666667
133.417690324783 82.6666666666667
134.821751117706 82.6666666666667
136.687398958206 82.6666666666667
138.569668245316 82.6666666666667
140.477064561844 82.6666666666667
141.527513122559 82.6666666666667
143.397725486755 82.6666666666667
145.331127786636 82.6666666666667
147.255424690247 82.6666666666667
148.359981775284 82.6666666666667
150.621307468414 82.6666666666667
152.587673854828 82.6666666666667
154.516018867493 82.6666666666667
155.896261835098 82.5714285714286
157.768652439117 82.5714285714286
158.90481672287 82.6666666666667
160.789396858215 82.6666666666667
161.852890443802 82.5714285714286
164.040152215958 82.4761904761905
165.928826379776 82.4761904761905
167.792389297485 82.4761904761905
168.869053173065 82.4761904761905
170.760714578629 82.3809523809524
172.621715784073 82.3809523809524
174.446911478043 82.3809523809524
175.510928726196 82.4761904761905
177.720569038391 82.4761904761905
179.600006961823 82.4761904761905
181.518051862717 82.4761904761905
182.967929506302 82.4761904761905
184.91376748085 82.4761904761905
186.79220957756 82.5714285714286
188.730482053757 82.4761904761905
189.100766420364 82.4761904761905
191.322060823441 82.3809523809524
193.248282814026 82.3809523809524
195.105732488632 82.3809523809524
196.145779943466 82.3809523809524
198.035275030136 82.3809523809524
199.898157310486 82.4761904761905
201.783017683029 82.4761904761905
202.885470676422 82.4761904761905
205.103067970276 82.4761904761905
207.009948825836 82.2857142857143
208.879050779343 82.2857142857143
210.345342588425 82.2857142857143
212.246946001053 82.2857142857143
214.103614854813 82.2857142857143
215.993298721313 82.1904761904762
217.073727798462 82.2857142857143
219.289979600906 82.3809523809524
220.515578556061 82.3809523809524
222.450581216812 82.3809523809524
223.521905422211 82.3809523809524
225.376316261291 82.2857142857143
227.221725654602 82.2857142857143
229.131285047531 82.2857142857143
230.548656892776 82.2857142857143
232.405953359604 82.2857142857143
234.316825580597 82.2857142857143
236.256581306458 82.2857142857143
237.718823194504 82.2857142857143
239.551573991776 82.1904761904762
241.490371227264 82.1904761904762
243.419442129135 82.1904761904762
244.89688038826 82.1904761904762
246.761911678314 82.1904761904762
248.7189930439 82.0952380952381
250.635789203644 82.0952380952381
};
\addlegendentry{10\%}
\addplot [red255290, dashed, mark=*, mark size=1, mark options={solid}]
table {%
4.89611015319824 61.4285714285714
5.50591926574707 61.4285714285714
8.40583052635193 61.6190476190476
11.2301819801331 61.8095238095238
14.0538328647614 61.9047619047619
16.8491012096405 61.7142857142857
19.6522707462311 62.0952380952381
22.4260688304901 62.3809523809524
25.2263686180115 62.952380952381
28.1268455505371 64.1904761904762
30.9646714687347 66.5714285714286
33.7516958236694 69.4285714285714
36.5469467639923 72.8571428571428
38.7606802463531 74.9523809523809
42.1619637012482 79.6190476190476
44.9626482486725 81.1428571428571
47.7161621570587 81.6190476190476
50.5871609687805 82.3809523809524
53.4217401027679 83.047619047619
56.2359489440918 83.047619047619
58.9948250293732 83.2380952380952
61.8181144714355 83.5238095238095
64.6322612285614 83.3333333333333
67.4561078548431 83.5238095238095
70.3708894729614 83.5238095238095
73.1669690132141 83.7142857142857
75.4113099098206 83.7142857142857
78.7237940311432 83.8095238095238
81.5368499755859 83.8095238095238
84.4678694725037 84
87.2067606925964 84
89.4838761806488 84
92.3313529968262 84
95.0961734294891 83.8095238095238
97.9719235897064 84.0952380952381
100.770885181427 84.2857142857143
103.611705446243 84.2857142857143
106.524723958969 84.1904761904762
109.359414434433 84.0952380952381
112.22438788414 84
115.024213314056 84.2857142857143
117.835305929184 84.2857142857143
120.667639017105 84.3809523809524
123.451947259903 84.3809523809524
126.368460321426 84.3809523809524
129.306394577026 84.3809523809524
131.019387340546 84.3809523809524
133.787702131271 84.4761904761905
136.674323987961 84.3809523809524
138.920525407791 84.3809523809524
142.192518997192 84.6666666666667
145.058482837677 84.6666666666667
147.244238710403 84.5714285714286
150.601484966278 84.6666666666667
153.402342987061 84.7619047619048
156.194509220123 84.6666666666667
159.027980613709 84.8571428571428
161.93503985405 84.5714285714286
164.817477893829 84.7619047619047
167.594429206848 84.6666666666666
170.412575435638 84.6666666666666
172.683518028259 84.8571428571428
175.480949354172 84.8571428571428
177.686825704575 84.8571428571428
180.981029891968 84.9523809523809
183.669411993027 85.0476190476191
186.626335430145 84.8571428571428
189.454277515411 84.8571428571428
192.243817901611 84.7619047619048
195.032463741302 84.7619047619048
197.826151752472 84.7619047619048
200.680549669266 84.952380952381
203.437863492966 84.952380952381
206.279640102386 84.952380952381
209.111816215515 84.952380952381
212.008021259308 84.952380952381
214.802338981628 84.952380952381
217.534843635559 85.1428571428571
220.277496337891 85.3333333333333
223.13670372963 85.4285714285714
225.913457250595 85.4285714285714
228.766126966476 85.4285714285714
231.626363372803 85.4285714285714
234.518062829971 85.4285714285714
237.382259464264 85.4285714285714
240.191265201569 85.5238095238095
242.974007129669 85.4285714285714
245.759115076065 85.2380952380952
248.516789627075 85.4285714285714
251.581876850128 85.2380952380952
};
\addlegendentry{20\%}
\addplot [darkorange2551480, dashed, mark=*, mark size=1, mark options={solid}]
table {%
6.92584609985352 61.4285714285714
10.5151941776276 61.3333333333333
14.9225340366364 61.6190476190476
19.432484960556 61.4285714285714
23.8808061122894 61.7142857142857
28.2863722324371 62.0952380952381
32.7410010814667 62.6666666666667
37.1872175216675 63.2380952380952
41.6591190814972 66.2857142857143
46.0753615856171 71.3333333333333
50.5141795635223 76.1904761904762
55.0714073181152 79.5238095238095
59.4996929168701 82
63.9830903530121 82.9523809523809
68.4037806510925 83.6190476190476
72.8186039447784 84.4761904761905
77.372851228714 84.5714285714286
81.737136554718 84.3809523809524
86.2496625900269 84.2857142857143
90.7466389656067 84.2857142857143
95.3029866218567 84.1904761904762
99.7566875934601 84.1904761904762
104.216941642761 84.2857142857143
108.747588825226 84.1904761904762
113.138658905029 84.1904761904762
117.520470285416 84.2857142857143
121.822071170807 84.0952380952381
126.299953460693 83.8095238095238
130.839840459824 83.8095238095238
135.458820438385 83.7142857142857
139.948149776459 83.8095238095238
144.445042514801 83.9047619047619
148.903579235077 84.0952380952381
153.31237578392 84.0952380952381
157.801959228516 84.1904761904762
162.18702750206 84.3809523809524
166.630939292908 84.3809523809524
171.228766918182 84.4761904761905
175.681766700745 84.5714285714286
180.188776111603 84.8571428571428
184.772325611114 84.8571428571428
189.164321517944 84.7619047619048
193.70549864769 85.0476190476191
198.194571638107 85.2380952380952
202.677945423126 85.0476190476191
207.055950737 85.3333333333333
211.567883110046 85.0476190476191
215.983783245087 85.0476190476191
220.309953975678 85.0476190476191
224.766782188416 85.0476190476191
229.259447479248 84.952380952381
233.753171157837 84.952380952381
238.182022476196 84.952380952381
242.594867181778 85.1428571428571
247.101494312286 85.0476190476191
251.633205747604 85.2380952380952
};
\addlegendentry{30\%}
\addplot [yellow22825518, dashed, mark=*, mark size=1, mark options={solid}]
table {%
9.57371940612793 61.4285714285714
13.1079522132874 61.8095238095238
18.9799912452698 61.8095238095238
24.864520740509 61.7142857142857
30.5035231590271 61.4285714285714
36.3322012424469 60.9523809523809
42.1520233631134 61.7142857142857
47.8357749938965 65.2380952380952
53.5323059082031 68.9523809523809
59.1573795318604 73.5238095238095
64.7749289989471 76.8571428571428
70.4675362110138 80.5714285714286
76.1357436180115 82.4761904761905
82.0169853687286 83.2380952380952
87.7845330238342 83.5238095238095
90.0811986923218 83.8095238095238
99.3793427467346 84.1904761904762
105.092013454437 84.0952380952381
110.909537553787 84
116.442808151245 84.3809523809524
122.085288715363 84
127.719506072998 83.9047619047619
133.487216186523 83.5238095238095
139.401715803146 83.5238095238095
145.085897159576 83.4285714285714
150.842044115067 83.5238095238095
156.521224927902 83.1428571428571
162.231451511383 82.9523809523809
168.091255760193 82.6666666666667
173.735045862198 82.7619047619047
176.006289052963 82.7619047619047
185.002396869659 82.4761904761905
190.684195756912 82.7619047619048
196.440318346024 82.3809523809524
202.094352912903 82.3809523809524
207.750426006317 82.2857142857143
213.344737482071 82.3809523809524
219.161036205292 82.3809523809524
224.973760128021 82.0952380952381
230.65581946373 82
236.569512224197 82
242.276014280319 82.0952380952381
247.822927427292 81.9047619047619
253.552560138702 82
};
\addlegendentry{40\%}
\addplot [lightgreen124255121, dashed, mark=*, mark size=1, mark options={solid}]
table {%
11.4135353088379 61.4285714285714
15.5311346530914 61.2380952380952
22.4894075870514 61.6190476190476
29.4456149101257 61.6190476190476
36.4712969779968 61.2380952380952
43.3996119976044 62.5714285714286
50.5486529827118 64.6666666666667
57.4055013656616 67.2380952380952
64.4163696289063 73.1428571428571
71.3131937980652 77.7142857142857
78.3837872505188 80.2857142857143
85.3414153575897 82.6666666666667
92.3515394210815 83.9047619047619
99.1986594200134 84.4761904761905
106.142526197433 84.2857142857143
113.226497459412 83.9047619047619
120.25051817894 83.7142857142857
127.311418056488 83.9047619047619
134.648930692673 84
141.648421621323 84.2857142857143
148.70078663826 84.2857142857143
155.741444826126 84
162.620490646362 84
169.58930888176 83.9047619047619
176.591585969925 83.8095238095238
183.553468132019 83.5238095238095
190.580014848709 83.4285714285714
197.400158786774 83.4285714285714
204.451155567169 83.4285714285714
211.35576634407 83.3333333333333
218.166674613953 83.5238095238095
225.166023206711 83.2380952380952
232.062416553497 83.1428571428571
238.846868038177 83.047619047619
245.703594636917 83.1428571428571
252.495811891556 83.1428571428571
};
\addlegendentry{50\%}
\addplot [cyan21255225, dashed, mark=*, mark size=1, mark options={solid}]
table {%
12.502206325531 61.4285714285714
13.9938461780548 61.6190476190476
21.8628584384918 61.6190476190476
29.4199700832367 61.6190476190476
37.168030166626 61.4285714285714
44.7855299472809 62.0952380952381
52.3668079853058 65.1428571428572
60.1366874217987 70.4761904761905
68.0455518722534 73.1428571428571
75.6139791488647 78.2857142857143
83.3375648498535 81.3333333333333
91.2836173534393 84.0952380952381
98.9658998012543 84.5714285714286
106.752843809128 84.8571428571428
114.770917844772 84.952380952381
122.527950239182 84.8571428571428
130.141804409027 84.5714285714286
137.856242847443 84.4761904761905
145.501202344894 84
153.203642654419 83.7142857142857
161.185816955566 83.7142857142857
169.034679079056 83.8095238095238
176.768963241577 83.4285714285714
184.248024320602 83.3333333333333
192.098141384125 83.4285714285714
199.799188995361 83.2380952380952
207.515574073791 82.9523809523809
215.375749826431 82.9523809523809
223.242349767685 82.9523809523809
231.003547668457 83.1428571428572
238.773049736023 83.0476190476191
246.451361513138 82.7619047619047
254.053384065628 82.8571428571428
};
\addlegendentry{60\%}
\addplot [dodgerblue0128255, dashed, mark=*, mark size=1, mark options={solid}]
table {%
13.9964662075043 61.4285714285714
19.183938741684 61.5238095238095
27.8976254463196 61.4285714285714
36.407507610321 61.9047619047619
44.9960407733917 62.0952380952381
53.5448992729187 64
62.2820516586304 69.3333333333333
70.9445230484009 73.4285714285714
79.5700613021851 77.4285714285714
88.4025691986084 81.2380952380952
96.9813408374786 84
105.861007356644 84.1904761904762
114.605151224136 84.1904761904762
123.361117506027 84.3809523809524
132.156996059418 84.0952380952381
140.996408891678 84
149.584428596497 84.0952380952381
158.385422420502 83.8095238095238
167.226110839844 83.7142857142857
175.736115694046 83.5238095238095
184.136945199966 83.4285714285714
192.81523566246 83.5238095238095
201.673490333557 83.5238095238095
210.30283241272 83.3333333333333
219.013444948196 83.5238095238095
227.552308511734 83.3333333333333
236.277521705627 83.3333333333333
245.070337247848 83.1428571428572
253.787868070602 83.0476190476191
};
\addlegendentry{70\%}
\addplot [blue, dashed, mark=*, mark size=1, mark options={solid}]
table {%
14.4601809024811 61.4285714285714
16.2947095870972 61.4285714285714
25.593458366394 61.3333333333333
35.3110456943512 61.047619047619
45.1214558124542 61.6190476190476
54.4904423236847 63.1428571428571
64.0754709720612 66.0952380952381
73.269922208786 72.2857142857143
82.6146405220032 76.4761904761905
92.1212346076965 79.8095238095238
101.601522350311 82.4761904761905
111.306623458862 83.7142857142857
120.863529348373 84.1904761904762
130.321436452866 83.9047619047619
139.759170436859 83.9047619047619
149.467693424225 83.9047619047619
159.048106098175 83.8095238095238
168.587756443024 83.7142857142857
178.139263105392 83.5238095238095
187.792557525635 83.4285714285714
197.125254583359 83.2380952380952
206.697101211548 83.2380952380952
216.147174406052 83.2380952380952
225.589801216125 83.1428571428571
235.112144136429 82.8571428571428
244.606186389923 83.047619047619
253.971092796326 83.3333333333333
};
\addlegendentry{80\%}
\addplot [navy00127, dashed, mark=*, mark size=1, mark options={solid}]
table {%
14.7284090518951 61.4285714285714
16.7130358695984 61.3333333333333
26.852290725708 61.4285714285714
37.3170840740204 61.3333333333333
47.5842183113098 61.1428571428571
57.9010594367981 63.7142857142857
68.2653579235077 67.9047619047619
78.5202262401581 70.9523809523809
88.7467342853546 76.1904761904762
99.1922794342041 80.4761904761905
109.397260046005 83.5238095238095
119.729198551178 84.2857142857143
129.86603884697 84.4761904761905
140.067887639999 84.7619047619047
150.523748445511 84.6666666666666
160.799932003021 84.6666666666666
171.143469524384 84.3809523809524
181.311746358871 84.0952380952381
191.403445100784 83.7142857142857
201.77010974884 83.5238095238095
211.950331068039 83.4285714285714
222.369553995132 83.3333333333333
232.605009555817 83.1428571428571
242.918841075897 83.2380952380952
253.293868350983 83.047619047619
};
\addlegendentry{90\%}

\addplot [ultra thick, gray, dashed, mark=*, mark size=2, mark options={solid}]
table {%
14.2887599468231 61.4285714285714
24.6740455150604 61.5238095238095
35.141822719574 61.6190476190476
45.638599729538 62.1904761904762
56.2690583229065 64.4761904761905
66.8843245983124 66.7619047619048
77.7658805847168 71.1428571428571
88.6925711154938 76.4761904761905
99.2472709655762 80.2857142857143
109.508946180344 82.5714285714286
119.790633487701 83.1428571428571
130.337510061264 83.9047619047619
140.955986309052 83.7142857142857
151.609069061279 84
162.209997320175 84.0952380952381
172.600471687317 84.2857142857143
183.052301597595 84.1904761904762
193.526771879196 84
204.086597251892 83.8095238095238
214.457344055176 83.7142857142857
225.259791564941 83.7142857142857
235.826626968384 83.4285714285714
246.619962596893 83.4285714285714
257.192080831528 83.6190476190476
};
\addlegendentry{100\%}

\end{axis}

\end{tikzpicture}

%% file: plots/short_pgf_plots_for_cikm/genes_k_var_short.tex
\begin{tikzpicture}

\definecolor{blue}{RGB}{0,0,255}
\definecolor{cyan21255225}{RGB}{21,255,225}
\definecolor{darkgray176}{RGB}{176,176,176}
\definecolor{darkorange2551480}{RGB}{255,148,0}
\definecolor{dodgerblue0128255}{RGB}{0,128,255}
\definecolor{gray}{RGB}{128,128,128}
\definecolor{lightgreen124255121}{RGB}{124,255,121}
\definecolor{maroon12700}{RGB}{127,0,0}
\definecolor{navy00127}{RGB}{0,0,127}
\definecolor{red255290}{RGB}{255,29,0}
\definecolor{yellow22825518}{RGB}{228,255,18}

\begin{axis}[
yscale=\cikmshortenscaley,
tick align=outside,
tick pos=left,
x grid style={darkgray176},
xmin=-14.0328826880455, xmax=397.140934417248,
xtick style={color=black},
y grid style={darkgray176},
ymin=39.7767441860465, ymax=100.967441860465,
ytick style={color=black},
yticklabel style={rotate=90.0},
tick label style={font=\huge}
]
\addplot [semithick, red, dashed]
table {%
-14.0328826880455 89.1232558139535
397.140934417248 89.1232558139535
};
\addplot [ultra thick, gray, dashed, mark=*, mark size=2, mark options={solid}]
table {%
18.7418682098389 42.8372093023256
34.6578793048859 43
50.8279933929443 60.6744186046512
66.9683317661285 68.8604651162791
83.4322623729706 77.7209302325581
99.9514236450195 80.7674418604651
116.412863349915 84.1162790697674
132.916092252731 87.1860465116279
149.050437068939 88.7906976744186
165.219602060318 90.2325581395349
181.240503978729 90.7674418604651
197.671772670746 91.3488372093023
213.76312084198 91.5581395348837
213.76312084198 91.5581395348837
247.107247304916 92.3255813953488
263.459987735748 92.6744186046512
279.685011863708 92.7906976744186
296.458326721191 92.906976744186
313.223739433289 93.1860465116279
329.267318487167 93.2790697674419
345.646366643906 93.4883720930233
362.122578382492 93.6279069767442
378.451215457916 93.8139534883721
};
\addplot [maroon12700, dashed, mark=*, mark size=1, mark options={solid}]
table {%
4.65683627128601 42.8372093023256
5.37564210891724 42.8139534883721
8.91585721969605 42.8372093023256
12.4967307567596 42.8604651162791
15.9560532569885 42.8837209302325
19.551619386673 43.0930232558139
23.0817546367645 42.9302325581395
26.6246088027954 42.8604651162791
30.1970772743225 42.8837209302326
33.7174451828003 42.9302325581395
37.2591628551483 42.953488372093
40.814814043045 43.046511627907
44.3660357475281 43.0232558139535
47.9197825431824 43.0930232558139
51.4746498584747 43.1860465116279
55.0717413902283 43.1395348837209
58.6753057003021 43.0697674418605
62.1731298446655 43.0232558139535
65.6773921489716 42.9767441860465
69.1741401195526 43.0232558139535
72.7188216209412 42.953488372093
76.2816942691803 42.9302325581395
79.9515723705292 42.906976744186
83.5253832817078 42.9302325581395
87.0392501354218 42.906976744186
90.6432150363922 42.9302325581395
94.2452067375183 42.906976744186
97.85221824646 42.906976744186
101.323241043091 42.9302325581395
104.896925735474 42.9302325581395
108.417454004288 42.953488372093
111.898036813736 42.9767441860465
115.492794942856 42.953488372093
119.063938570023 42.953488372093
122.594718790054 42.906976744186
126.185467386246 42.8604651162791
128.262730312347 42.906976744186
131.819737052917 42.9302325581395
135.352217721939 42.906976744186
139.057763338089 42.8604651162791
142.556604480743 42.906976744186
146.071690750122 42.8604651162791
149.706871986389 42.8604651162791
153.451170778275 42.8372093023256
157.015243673325 42.8604651162791
160.587218236923 42.8372093023256
164.185776901245 42.7906976744186
167.747001647949 42.8139534883721
171.219807004929 42.8372093023256
174.822407150269 42.7906976744186
178.27899055481 42.7674418604651
181.816708230972 42.7441860465116
185.359631633759 42.7674418604651
188.922224235535 42.7906976744186
191.744671964645 42.7209302325581
195.269308423996 42.7674418604651
198.83927898407 42.6976744186046
202.472204637527 42.7209302325581
205.982473993301 42.6976744186046
209.514427852631 42.7441860465116
213.068478393555 42.7441860465116
216.612233924866 42.7441860465116
220.245177173615 42.6976744186046
223.819584178925 42.6744186046512
227.341230487823 42.6976744186046
230.87093834877 42.6976744186046
234.380801534653 42.6976744186046
237.93546833992 42.6046511627907
241.455200338364 42.6046511627907
244.989575481415 42.6046511627907
248.477643156052 42.6279069767442
250.643164587021 42.5581395348837
254.180692768097 42.6511627906977
257.737179899216 42.6511627906977
261.278067207336 42.6511627906977
264.736353397369 42.5813953488372
268.291567468643 42.6511627906977
271.880365753174 42.6046511627907
275.366758537292 42.6046511627907
278.936541700363 42.6279069767442
282.509337854385 42.6744186046512
286.101557779312 42.6279069767442
289.669760608673 42.6744186046512
293.162963056564 42.6511627906977
296.718877315521 42.6511627906977
300.332331085205 42.6976744186046
303.897773694992 42.6279069767442
307.447214746475 42.6279069767442
310.955476951599 42.6279069767442
314.468853092194 42.6511627906977
318.006241703033 42.6046511627907
321.483834218979 42.6511627906977
325.052871179581 42.6046511627907
328.637032651901 42.6511627906977
332.166634941101 42.6511627906977
335.807176160812 42.5813953488372
339.340653181076 42.6511627906977
342.966777467728 42.6511627906977
346.55522351265 42.6744186046512
350.102083492279 42.7441860465116
353.705636024475 42.7906976744186
357.173271512985 42.8139534883721
360.7540122509 42.8139534883721
364.287714147568 42.7441860465116
367.819392967224 42.7674418604651
371.293722724915 42.8372093023256
};
\addplot [red255290, dashed, mark=*, mark size=1, mark options={solid}]
table {%
5.8619592666626 42.8372093023256
8.83043675422669 42.8372093023256
13.4851257801056 42.953488372093
18.2548281669617 42.8139534883721
22.8845983505249 43.0697674418605
27.610197019577 43.046511627907
32.4035225391388 42.906976744186
37.1483335494995 42.9767441860465
41.8503079414368 42.953488372093
46.5348022460937 43.0232558139535
51.1765160560608 42.906976744186
55.7661699295044 42.9302325581395
60.4162755012512 42.9767441860465
65.2413229942322 42.9767441860465
69.9641365528107 42.9767441860465
74.665619468689 42.9302325581395
79.4792321205139 42.953488372093
84.1686374664307 42.953488372093
88.8887523651123 42.953488372093
93.5945940494537 43
98.3115624427795 43
100.149202537537 43
107.847269201279 43.046511627907
112.504111671448 43
117.196786260605 43.046511627907
121.898981952667 43.0232558139535
126.654943418503 43.0232558139535
131.268739509583 43
135.966292572021 43.0232558139535
140.734322404861 43.0232558139535
145.385327100754 43.046511627907
150.083221912384 43.0232558139535
154.832230901718 42.9767441860465
159.578937339783 43.0232558139535
164.371395874023 43.046511627907
169.034504890442 43.046511627907
173.791693115234 43.0232558139535
178.619537830353 42.9302325581395
183.284975099564 42.9767441860465
188.020134305954 42.906976744186
192.773446083069 42.9302325581395
197.530314588547 42.906976744186
199.454279088974 42.906976744186
206.991030931473 42.8372093023256
211.747831726074 42.8604651162791
216.570797872543 42.8139534883721
221.350337505341 42.8372093023256
223.29087562561 42.8139534883721
230.770955324173 42.7674418604651
235.409031772614 42.8139534883721
240.145981407166 42.7441860465116
244.911959981918 42.7674418604651
249.545217990875 42.7674418604651
254.283104991913 42.7906976744186
259.094871330261 42.7674418604651
263.896987056732 42.7674418604651
268.650044107437 42.6744186046512
273.263445138931 42.7441860465116
277.873847484589 42.7209302325581
282.575718069077 42.7674418604651
287.43803153038 42.7441860465116
292.239601612091 42.7674418604651
296.866827964783 42.6976744186046
301.614987516403 42.7209302325581
306.281884479523 42.6976744186046
310.951313257217 42.6976744186046
315.695355272293 42.7674418604651
320.416744709015 42.8139534883721
325.180294704437 42.8372093023256
329.846116876602 42.7441860465116
334.59952545166 42.7674418604651
339.293500423431 42.7674418604651
344.025556850433 42.7441860465116
348.844466352463 42.7441860465116
353.576533746719 42.7906976744186
358.250085544586 42.7674418604651
362.891864776611 42.7441860465116
367.611008501053 42.7906976744186
372.298185968399 42.7674418604651
};
\addplot [darkorange2551480, dashed, mark=*, mark size=1, mark options={solid}]
table {%
7.77462520599365 42.8372093023256
10.3594027996063 42.8372093023256
17.0372488498688 44.046511627907
23.6163023948669 50.3023255813953
30.5647964000702 58.6279069767442
37.1894445896149 66.093023255814
43.7793123245239 70.4418604651163
50.426499414444 74.3720930232558
57.2948344707489 76.6279069767442
64.0319117546082 78.1162790697674
70.7514793872833 79.7906976744186
77.4866584300995 81.3255813953488
84.1124987125397 83.3023255813954
91.0373836040497 85.6744186046512
97.7558959007263 87.3488372093023
104.498023176193 89.3488372093023
111.217102861404 91.046511627907
117.962502384186 92.1395348837209
124.576713180542 92.9302325581396
131.256309652328 93.6046511627907
138.036812067032 94.1627906976744
144.685187721252 94.8372093023256
151.476530122757 95.3023255813954
158.025437402725 95.6744186046512
164.564543676376 96.1162790697674
171.235788345337 96.4651162790698
177.908112955093 96.8372093023256
184.817539691925 97.0232558139535
191.605309295654 97.2093023255814
198.254035568237 97.3255813953488
205.066015100479 97.4883720930233
211.92867231369 97.5813953488372
218.698678779602 97.6511627906977
225.515848731995 97.7441860465116
232.231773948669 97.7441860465116
238.957235527039 97.8139534883721
245.502389526367 97.8139534883721
252.208363819122 97.8372093023256
258.937342977524 97.9069767441861
265.718951702118 97.9069767441861
272.650874090195 97.9302325581395
279.35715637207 97.9069767441861
286.229867362976 97.953488372093
293.087826156616 98
299.715262126923 98
306.462542247772 97.9767441860465
313.180337810516 98.0232558139535
319.794782066345 98
326.666536712646 97.9767441860465
333.430054664612 98.046511627907
340.202891206741 98.0232558139535
346.966923999786 98.0232558139535
353.612877941132 98.0232558139535
360.458914375305 98.046511627907
367.062369346619 98.046511627907
373.500980949402 98.0697674418605
};
\addplot [yellow22825518, dashed, mark=*, mark size=1, mark options={solid}]
table {%
10.40737657547 42.8372093023256
17.3273522853851 42.7441860465116
25.8241305351257 46.1627906976744
34.6712962627411 59.9069767441861
43.4067992210388 65.7674418604651
52.1807224750519 71.4883720930233
60.912023973465 75.6511627906977
69.5744720935822 77.6976744186046
78.3085381507874 79.4883720930233
86.9004789829254 81.6511627906977
95.6131331920624 83.5581395348837
104.238210248947 86.0697674418604
112.978817415237 87.6046511627907
121.688319826126 89.4418604651163
130.231570529938 90.8837209302326
138.825943422317 92.093023255814
147.63374710083 92.6976744186047
156.290187168121 93.8139534883721
165.108127307892 94.3023255813954
173.69300160408 95.046511627907
182.359554576874 95.3720930232558
190.964942979813 95.9767441860465
199.609686183929 96.2558139534884
201.315172386169 96.3488372093023
216.645715999603 96.8604651162791
225.219941282272 97.093023255814
233.830698204041 97.1860465116279
242.429631519318 97.4418604651163
251.18292016983 97.6046511627907
259.96048283577 97.7441860465116
268.723565626144 97.7674418604651
270.412981414795 97.8139534883721
285.801660728455 97.8837209302326
294.406510734558 97.9767441860465
303.145503473282 97.953488372093
311.684198999405 98.0232558139535
320.370187568665 98.046511627907
329.206597328186 98.0697674418604
338.016325139999 98.1395348837209
346.597557878494 98.1860465116279
355.155255889893 98.1860465116279
363.874028539658 98.1627906976744
372.466823911667 98.1860465116279
};
\addplot [lightgreen124255121, dashed, mark=*, mark size=1, mark options={solid}]
table {%
11.6167194843292 42.8372093023256
21.4230082035065 42.7209302325581
31.1771524429321 47.0697674418605
40.9508226394653 62.8837209302326
50.72467045784 67.5348837209302
60.6231926441193 74.1395348837209
70.459054851532 77.8604651162791
80.2118445873261 79.2093023255814
90.2599890232086 81.2558139534884
99.9358073234558 83.8604651162791
109.803356361389 85.8604651162791
119.704927062988 87.7674418604651
129.339247512817 89.5116279069767
139.3801217556 90.6279069767442
149.253162240982 91.7674418604651
159.16281619072 92.5581395348837
168.793564367294 93.2790697674419
178.570427846909 93.7674418604651
188.584294223785 94.093023255814
198.47213177681 94.6744186046512
208.519085884094 94.953488372093
218.44115447998 95.1860465116279
228.100399160385 95.5116279069768
237.983974170685 95.8139534883721
247.827902650833 96.046511627907
257.645413303375 96.2790697674419
267.340274095535 96.4883720930233
277.045525836945 96.6976744186046
286.991406965256 96.906976744186
296.618041324615 97
306.453501224518 97.2558139534884
316.341722917557 97.3720930232558
326.171481990814 97.5348837209302
335.904342508316 97.6744186046512
345.487187862396 97.8372093023256
355.089465332031 97.953488372093
365.051227140427 98
374.886465454102 98.046511627907
};
\addplot [cyan21255225, dashed, mark=*, mark size=1, mark options={solid}]
table {%
12.8278096199036 42.8372093023256
23.9056611537933 42.7209302325581
34.9654524326324 51.0232558139535
45.8735581874847 64.953488372093
56.8881539821625 70.906976744186
67.9489386081696 77.4418604651163
79.2024352550507 79.5116279069767
90.3599349021912 81.8604651162791
101.474826955795 84.4418604651163
112.739647722244 86.9767441860465
123.46970076561 89.0232558139535
134.505759572983 90.6511627906977
145.39651517868 91.7441860465116
156.315843629837 92.5116279069768
167.191132307053 92.9767441860465
178.139766407013 93.3255813953488
189.267197561264 93.6744186046512
200.245384550095 94.1395348837209
211.256330966949 94.3488372093023
222.113379573822 94.6279069767442
233.155984592438 94.8837209302326
244.123579359055 95.1162790697674
255.0471326828 95.5116279069768
265.943395090103 95.7906976744186
276.951868247986 96
288.071962785721 96.2790697674419
299.112960720062 96.5581395348837
310.233203363419 96.6511627906977
321.367013168335 96.8604651162791
332.439546966553 97.0232558139535
343.47875289917 97.1395348837209
354.42607755661 97.1627906976744
365.546039295197 97.2093023255814
376.482745170593 97.2558139534884
};
\addplot [dodgerblue0128255, dashed, mark=*, mark size=1, mark options={solid}]
table {%
14.0578108787537 42.8372093023256
16.4970048427582 42.7674418604651
28.8455300807953 44.8604651162791
41.5622821807861 56.1627906976744
53.974516916275 68.7209302325581
66.1024549484253 75.6046511627907
78.7087979793549 79.5348837209302
91.0352302551269 82.5116279069767
103.605426263809 85.6279069767442
116.197887086868 88.0930232558139
128.872345733643 89.7674418604651
141.310158967972 90.8837209302326
153.756464242935 91.6279069767442
166.112067699432 92.093023255814
178.533482265472 92.5348837209302
191.277584028244 93.1860465116279
203.78862156868 93.5813953488372
216.1148042202 93.8372093023256
228.694708538055 94.093023255814
241.420179891586 94.4186046511628
254.206185913086 94.6744186046512
266.621324396133 95.0232558139535
279.122977304459 95.2093023255814
291.650282478333 95.3720930232558
304.013698863983 95.7209302325581
316.240296220779 95.9302325581395
328.683960151672 96.0697674418605
341.199859571457 96.3720930232558
353.537830448151 96.5348837209302
365.783430528641 96.6046511627907
377.970103502274 96.6511627906977
};
\addplot [blue, dashed, mark=*, mark size=1, mark options={solid}]
table {%
16.2069649219513 42.8372093023256
19.1170386791229 42.8139534883721
33.4171431064606 45.5116279069767
47.5830177307129 57.953488372093
61.7885466575623 68.3488372093023
75.9906610965729 76.4651162790698
90.5197649478912 81.5813953488372
104.660490226746 84.906976744186
118.999130249023 87.7674418604651
133.264083862305 89.2093023255814
147.397602033615 90.3023255813954
161.62582244873 91.1395348837209
176.016419315338 91.7674418604651
190.228950023651 92.0697674418605
204.404938268662 92.3953488372093
218.615711641312 92.5581395348837
232.972533464432 92.5813953488372
247.361030054092 92.953488372093
261.514213466644 93.093023255814
276.089876413345 93.3255813953488
290.378377485275 93.4418604651163
304.786791658402 93.6511627906977
318.911885595322 93.8604651162791
333.294992542267 94.1395348837209
347.604949045181 94.3023255813954
361.80433049202 94.3720930232558
376.183323574066 94.4418604651163
};
\addplot [navy00127, dashed, mark=*, mark size=1, mark options={solid}]
table {%
19.5081378936768 42.8372093023256
25.5806555747986 42.7209302325581
40.711599445343 48.5116279069767
55.7829661846161 62.046511627907
71.1365592956543 71.1627906976744
86.6871074199676 78.906976744186
102.120712900162 82.3953488372093
117.229339170456 85.2790697674419
132.718991947174 87.7906976744186
148.240753793716 89.093023255814
163.750403213501 90.2790697674419
178.749170875549 91.1395348837209
194.18596663475 91.6744186046512
209.468977069855 91.9767441860465
224.935243225098 92.3023255813953
240.210892868042 92.5581395348837
255.259968519211 92.6976744186046
270.607349967957 92.8837209302326
286.050008821487 93.0232558139535
301.236810064316 93.1627906976744
316.243044519424 93.3953488372093
331.427279663086 93.6046511627907
346.685082769394 93.8837209302326
362.011796092987 93.953488372093
377.346285772324 94.2093023255814
};
\end{axis}

\end{tikzpicture}

%% file: plots/short_pgf_plots_for_cikm/mondial_religion_1ep_short.tex
\begin{tikzpicture}

\definecolor{blue}{RGB}{0,0,255}
\definecolor{cyan21255225}{RGB}{21,255,225}
\definecolor{darkgray176}{RGB}{176,176,176}
\definecolor{darkorange2551480}{RGB}{255,148,0}
\definecolor{dodgerblue0128255}{RGB}{0,128,255}
\definecolor{gray}{RGB}{128,128,128}
\definecolor{lightgreen124255121}{RGB}{124,255,121}
\definecolor{maroon12700}{RGB}{127,0,0}
\definecolor{navy00127}{RGB}{0,0,127}
\definecolor{red255290}{RGB}{255,29,0}
\definecolor{yellow22825518}{RGB}{228,255,18}

\begin{axis}[
yscale=\cikmshortenscaley,
tick align=outside,
tick pos=left,
x grid style={darkgray176},
xmin=2.03753677368164, xmax=271.564446582794,
xtick style={color=black},
y grid style={darkgray176},
ymin=59.5095238095238, ymax=87.0619047619047,
ytick style={color=black},
yticklabel style={rotate=90.0},
tick label style={font=\huge}
]
\addplot [ultra thick, gray, dashed, mark=*, mark size=2, mark options={solid}]
table {%
14.2887599468231 61.4285714285714
24.6740455150604 61.5238095238095
35.141822719574 61.6190476190476
45.638599729538 62.1904761904762
56.2690583229065 64.4761904761905
66.8843245983124 66.7619047619048
77.7658805847168 71.1428571428571
88.6925711154938 76.4761904761905
99.2472709655762 80.2857142857143
109.508946180344 82.5714285714286
119.790633487701 83.1428571428571
130.337510061264 83.9047619047619
140.955986309052 83.7142857142857
151.609069061279 84
162.209997320175 84.0952380952381
172.600471687317 84.2857142857143
183.052301597595 84.1904761904762
193.526771879196 84
204.086597251892 83.8095238095238
214.457344055176 83.7142857142857
225.259791564941 83.7142857142857
235.826626968384 83.4285714285714
246.619962596893 83.4285714285714
257.192080831528 83.6190476190476
};
\addplot [maroon12700, dashed, mark=*, mark size=1, mark options={solid}]
table {%
14.698076915741 61.4285714285714
15.2983225822449 61.4285714285714
17.7558226108551 61.7142857142857
20.1364030361176 61.8095238095238
22.5962238311768 61.8095238095238
24.97872838974 61.9047619047619
27.3767763614655 61.9047619047619
29.7644388198853 61.8095238095238
32.273389339447 61.8095238095238
34.6336396694183 62.0952380952381
36.9490369319916 62.5714285714286
39.3037325382233 63.5238095238095
41.7493743896484 64.3809523809524
44.0213900566101 66.7619047619048
46.3241057872772 69.5238095238095
48.7335057735443 72.1904761904762
51.1955483436584 75.5238095238095
53.5363167285919 78.7619047619048
55.9738416671753 80.9523809523809
58.392813539505 82.5714285714286
60.7431192874908 83.1428571428572
63.2809841632843 83.4285714285714
65.7106703281403 83.9047619047619
68.1409956455231 84.4761904761905
70.4786836624146 84.6666666666667
72.768545627594 84.4761904761905
75.1873031139374 84.6666666666667
77.5726396560669 84.5714285714286
79.8704181671143 84.7619047619048
81.8400074958801 84.8571428571428
84.7948585510254 84.952380952381
87.2892577171326 85.0476190476191
89.5895990371704 85.0476190476191
92.0208185195923 85.1428571428571
93.9601778507233 85.1428571428571
96.4577213764191 85.2380952380952
98.8982103347778 85.1428571428571
101.322038936615 85.2380952380952
103.587348175049 85.3333333333333
106.000684499741 85.3333333333333
108.343754673004 85.2380952380952
110.810262680054 85.1428571428571
113.180758190155 85.2380952380952
115.65325551033 85.0476190476191
118.089742708206 84.952380952381
120.504594898224 84.8571428571428
122.856002998352 85.0476190476191
125.204787874222 84.952380952381
127.63148021698 84.8571428571428
130.059339523315 84.952380952381
132.423733377457 85.0476190476191
133.82831401825 84.952380952381
136.158626651764 84.952380952381
138.572796535492 85.0476190476191
140.864538431168 85.0476190476191
143.217523622513 84.952380952381
145.621880149841 84.952380952381
147.54322681427 85.0476190476191
149.913923311234 84.952380952381
152.308683204651 84.8571428571428
155.240330791473 84.952380952381
157.6011531353 84.952380952381
159.839221286774 84.8571428571428
162.32038731575 84.8571428571428
164.673073959351 84.7619047619048
167.03406329155 84.7619047619048
169.460466957092 84.7619047619048
171.971063375473 84.6666666666667
173.955243444443 84.7619047619048
176.288305997849 84.7619047619048
178.779683542252 84.8571428571428
181.170362567902 84.952380952381
183.551928710937 84.8571428571428
185.92118473053 84.8571428571428
188.383582687378 84.8571428571428
190.7720911026 84.8571428571428
193.202798032761 84.952380952381
195.604696321487 84.952380952381
197.963367986679 85.1428571428571
200.378071689606 85.0476190476191
202.753305768967 84.952380952381
205.198731422424 85.0476190476191
207.75115237236 85.0476190476191
210.192956829071 85.1428571428571
212.555962085724 85.0476190476191
215.016266393662 85.2380952380952
217.450374937057 85.3333333333333
219.900466585159 85.1428571428571
222.262853765488 85.3333333333333
224.756720638275 85.2380952380952
227.117399454117 85.3333333333333
229.612944459915 85.2380952380952
231.921159410477 85.2380952380952
234.378510379791 85.2380952380952
236.806293487549 85.2380952380952
239.229286289215 85.1428571428571
241.600559568405 85.1428571428571
244.111671733856 85.2380952380952
246.592567825317 85.2380952380952
248.958328962326 85.2380952380952
251.315862131119 85.2380952380952
};
\addplot [red255290, dashed, mark=*, mark size=1, mark options={solid}]
table {%
15.9686493396759 61.4285714285714
17.1397330760956 61.5238095238095
20.5230354309082 61.6190476190476
23.7389409065247 61.5238095238095
26.9170216560364 61.4285714285714
30.0455723762512 61.5238095238095
32.4687358379364 61.8095238095238
34.41153383255 62.0952380952381
37.3985941886902 62.1904761904762
40.497519826889 62.6666666666667
43.7065583229065 63.7142857142857
46.1784689903259 65.4285714285714
49.3143317699432 68.0952380952381
52.5431471347809 69.6190476190476
54.6322735309601 71.2380952380952
59.026153087616 74.4761904761905
60.1889125823975 74.4761904761905
63.4609766960144 77.2380952380952
66.643652009964 80.2857142857143
69.7137184619904 82
72.8315605163574 83.4285714285714
75.3466693878174 84.0952380952381
78.5470782279968 84.1904761904762
80.430567741394 84.4761904761905
83.3861630439758 84.5714285714286
86.5550084114075 84.6666666666667
89.032096862793 84.5714285714286
92.2294733524323 84.5714285714286
94.2115564346313 84.5714285714286
98.5888791561127 84.8571428571428
101.872082853317 84.8571428571428
103.813532876968 84.7619047619048
106.347950983047 84.8571428571428
109.440932798386 84.952380952381
112.595522451401 84.952380952381
115.837976264954 84.8571428571428
119.045490694046 84.8571428571428
121.506799697876 84.8571428571428
123.344175195694 84.7619047619048
126.536345243454 84.8571428571428
129.72468457222 84.952380952381
132.829197454453 84.952380952381
135.321566200256 84.952380952381
138.554370975494 84.952380952381
141.857542467117 84.952380952381
145.103267049789 84.952380952381
146.963045024872 85.0476190476191
149.344386291504 85.1428571428571
152.524838876724 85.1428571428571
155.667241668701 85.1428571428571
158.95153875351 85.1428571428571
162.195778226852 85.1428571428571
164.623935222626 85.1428571428571
165.349306869507 85.1428571428571
169.75160908699 85.0476190476191
172.837770700455 85.2380952380952
174.744931316376 85.2380952380952
179.124488592148 85.2380952380952
181.55909538269 85.2380952380952
184.673590278625 85.2380952380952
187.878416824341 85.1428571428571
189.779911613464 85.1428571428571
192.91748046875 85.1428571428571
195.370917797089 85.1428571428571
198.5396900177 85.1428571428571
201.714744710922 85.1428571428571
204.884667730331 85.3333333333333
208.048439359665 85.6190476190476
209.204015445709 85.6190476190476
212.355313205719 85.4285714285714
215.617087745667 85.7142857142857
218.788016080856 85.4285714285714
221.995576620102 85.6190476190476
224.473592996597 85.6190476190476
227.63522400856 85.5238095238095
230.867818164825 85.6190476190476
232.777125835419 85.5238095238095
235.955681371689 85.6190476190476
238.44534163475 85.5238095238095
241.671637439728 85.6190476190476
244.781603145599 85.4285714285714
247.94198474884 85.3333333333333
251.208693885803 85.4285714285714
};
\addplot [darkorange2551480, dashed, mark=*, mark size=1, mark options={solid}]
table {%
17.5749535083771 61.4285714285714
18.4654597759247 61.6190476190476
22.7276178359985 61.8095238095238
26.8849370956421 61.5238095238095
31.1378845214844 61.6190476190476
35.3498503684998 61.4285714285714
39.5399863243103 61.8095238095238
43.6704662799835 62.5714285714286
47.9224657058716 63.6190476190476
51.9917791366577 66.2857142857143
56.1059422969818 71.3333333333333
60.1610990524292 76.2857142857143
64.3527618408203 80.0952380952381
68.610099363327 81.9047619047619
72.7452664375305 83.3333333333333
77.1047095775604 83.7142857142857
81.2648400306702 84.3809523809524
85.5092399597168 84.952380952381
89.6749018192291 85.7142857142857
93.9379183292389 85.8095238095238
98.0655705928802 85.7142857142857
102.228067016602 85.4285714285714
106.404744052887 85.2380952380952
110.469295597076 85.0476190476191
114.601793861389 84.952380952381
118.86492228508 84.952380952381
123.115173578262 84.8571428571428
127.144150018692 84.952380952381
131.220208215714 84.8571428571428
135.246356773376 84.8571428571428
138.528792905808 84.8571428571428
141.908692026138 84.952380952381
146.925852298737 84.952380952381
151.051691198349 84.952380952381
155.172726774216 84.8571428571428
159.286485290527 84.8571428571428
163.563646841049 84.6666666666667
167.742686319351 84.952380952381
171.954366445541 84.952380952381
176.161936378479 84.952380952381
180.400860214233 84.952380952381
184.684672212601 84.8571428571428
188.948341464996 84.8571428571428
193.165257072449 84.8571428571428
197.268475437164 84.6666666666667
201.509207963943 84.6666666666667
205.875157642365 84.6666666666667
210.041715192795 84.7619047619048
214.365529489517 84.6666666666667
218.415070438385 84.7619047619048
222.570722484589 84.6666666666667
226.72094373703 84.8571428571428
230.964309740067 84.7619047619048
235.141293954849 84.8571428571428
239.268080329895 84.8571428571428
243.382287931442 84.8571428571428
247.74474029541 84.7619047619048
251.92183175087 84.8571428571428
};
\addplot [yellow22825518, dashed, mark=*, mark size=1, mark options={solid}]
table {%
20.315625667572 61.4285714285714
21.3724802017212 61.6190476190476
26.8501186847687 61.9047619047619
32.3384045600891 61.8095238095238
37.6390995979309 61.2380952380952
43.0790048599243 60.8571428571429
48.4887998580933 62
53.9272475719452 65.4285714285714
59.2745299816132 68.5714285714286
64.7169683456421 73.3333333333333
70.1015565395355 76.4761904761905
75.481813621521 80.3809523809524
80.9638800621033 82.1904761904762
85.3398586273193 83.9047619047619
91.8392519950867 85.5238095238095
97.494935798645 85.7142857142857
102.986922740936 85.5238095238095
108.420447015762 85.4285714285714
113.855384635925 85.4285714285714
119.222969388962 85.4285714285714
124.727639102936 85.2380952380952
130.163810491562 85.4285714285714
135.611229944229 85.1428571428571
139.899918317795 84.6666666666666
145.396133089066 84.1904761904762
150.730261611938 84
155.109507083893 84
161.578715705872 83.8095238095238
167.204794692993 83.6190476190476
172.610722780228 83.4285714285714
178.076874494553 83.2380952380952
183.533099603653 83.2380952380952
189.074991178513 83.1428571428571
194.401624298096 83.047619047619
199.730063295364 82.8571428571428
205.197498035431 82.8571428571428
210.663104295731 82.8571428571428
216.099415922165 82.8571428571428
221.628730678558 82.9523809523809
227.043706083298 83.1428571428571
232.526694440842 82.9523809523809
237.964596796036 82.9523809523809
243.367413520813 82.9523809523809
248.738407278061 82.8571428571428
254.188113069534 82.7619047619047
};
\addplot [lightgreen124255121, dashed, mark=*, mark size=1, mark options={solid}]
table {%
22.1155278682709 61.4285714285714
23.4218801498413 61.6190476190476
30.1521808624268 61.3333333333333
36.7432933807373 61.3333333333333
43.4808303833008 61.1428571428571
50.0476371765137 60.8571428571429
56.6867337226868 64.5714285714286
63.2916943073273 68.9523809523809
69.92965259552 71.4285714285714
76.465305185318 75.5238095238095
83.2375886440277 78.9523809523809
89.9061935901642 81.7142857142857
96.6226331710815 83.4285714285714
103.146990776062 83.5238095238095
109.765581226349 83.5238095238095
116.525564527512 83.3333333333333
123.196469545364 83.5238095238095
129.91145772934 83.5238095238095
136.43214802742 83.8095238095238
140.493693780899 83.9047619047619
147.172504663467 83.8095238095238
153.915599298477 83.7142857142857
160.577753067017 83.7142857142857
167.226939439774 83.4285714285714
173.845785331726 83.4285714285714
180.555317306519 83.4285714285714
187.239912414551 83.1428571428571
193.842200994492 83.2380952380952
200.726867866516 83.1428571428571
207.612194013596 83.2380952380952
214.219128274918 83.1428571428571
220.949236965179 83.3333333333333
227.496819162369 83.2380952380952
234.00103969574 83.2380952380952
240.90187087059 83.047619047619
247.597198295593 82.8571428571428
254.348054790497 83.047619047619
};
\addplot [cyan21255225, dashed, mark=*, mark size=1, mark options={solid}]
table {%
23.4138660430908 61.4285714285714
29.3572049617767 61.5238095238095
36.7632818222046 61.6190476190476
44.4139524936676 61.047619047619
52.0028223991394 61.3333333333333
59.6010702610016 62.5714285714286
67.1767308235168 66.952380952381
74.4518922328949 70.5714285714286
82.0262671470642 77.3333333333333
89.6114675045013 79.7142857142857
97.170084810257 82.0952380952381
104.717012929916 83.9047619047619
112.159259366989 84.6666666666667
119.65551071167 84.1904761904762
127.251973485947 84.2857142857143
134.697405290604 84.2857142857143
142.363550424576 84.3809523809524
149.9572057724 84.2857142857143
157.442044830322 84
164.858349561691 84
172.380339431763 83.9047619047619
179.815113830566 83.7142857142857
187.229023551941 83.9047619047619
194.913473844528 83.6190476190476
202.340913486481 83.5238095238095
209.985002851486 83.5238095238095
217.626701116562 83.3333333333333
225.045229768753 83.4285714285714
232.551393413544 83.3333333333333
240.282469654083 83.2380952380952
247.763100719452 83.047619047619
255.427403211594 83.1428571428571
};
\addplot [dodgerblue0128255, dashed, mark=*, mark size=1, mark options={solid}]
table {%
24.767286157608 61.4285714285714
26.4942299365997 61.4285714285714
35.039031791687 61.6190476190476
43.804583978653 60.7619047619048
52.3825150966644 61.3333333333333
61.1048717975616 63.1428571428571
69.8326612472534 66.4761904761905
78.5995337486267 71.7142857142857
87.3775654792786 75.5238095238095
96.2064107894898 78.7619047619048
105.028326892853 82.1904761904762
113.886031675339 83.6190476190476
122.631420516968 84.2857142857143
131.571714925766 84.4761904761905
140.137595701218 84.3809523809524
148.968453025818 84.4761904761905
157.806707143784 84.6666666666667
166.45600605011 84.4761904761905
175.055436229706 84.4761904761905
183.755931282043 84.2857142857143
192.336297512054 84.1904761904762
201.195875787735 84
209.970942020416 84.0952380952381
218.830154037476 83.8095238095238
227.715101957321 83.9047619047619
236.514896535873 83.6190476190476
245.352146720886 83.6190476190476
254.000170707703 83.3333333333333
};
\addplot [blue, dashed, mark=*, mark size=1, mark options={solid}]
table {%
25.9834870815277 61.4285714285714
29.5794893264771 61.3333333333333
39.0074287891388 61.047619047619
48.8433278560638 61.8095238095238
58.5391793251038 62.8571428571429
68.3624243736267 64
78.0481202602386 68.7619047619048
87.7440629482269 73.5238095238095
97.3079568862915 77.7142857142857
106.988627290726 82.0952380952381
116.70828089714 84.1904761904762
126.220481109619 84.3809523809524
135.940368461609 84
145.727100324631 84.0952380952381
155.221948862076 84
164.926831197739 83.9047619047619
174.692549371719 83.8095238095238
184.396075153351 83.8095238095238
194.256896114349 83.6190476190476
204.030600643158 83.9047619047619
213.800581169128 83.8095238095238
223.494852018356 83.7142857142857
233.162691545486 83.5238095238095
242.856339073181 83.5238095238095
252.725468969345 83.3333333333333
};
\addplot [navy00127, dashed, mark=*, mark size=1, mark options={solid}]
table {%
26.4456541061401 61.4285714285714
37.0462847232819 61.6190476190476
47.7129239559174 61.5238095238095
58.4085873126984 62.5714285714286
69.1151517391205 64.0952380952381
79.5788269519806 67.2380952380953
90.2032649993896 70.4761904761905
100.763081455231 76.7619047619048
111.24517416954 81.2380952380952
122.030337285995 83.4285714285714
132.663840150833 83.9047619047619
143.003348588943 84.1904761904762
153.3164021492 84.3809523809524
153.3164021492 84.3809523809524
174.386693334579 84.5714285714286
184.956469345093 84.2857142857143
195.331844091415 84
206.325021886826 83.7142857142857
216.855684089661 83.5238095238095
227.630943584442 83.5238095238095
238.116107988358 83.4285714285714
248.754796791077 83.2380952380952
259.313223409653 83.1428571428571
};
\end{axis}

\end{tikzpicture}

%% file: plots/short_pgf_plots_for_cikm/genes_1ep_short.tex
\begin{tikzpicture}

\definecolor{blue}{RGB}{0,0,255}
\definecolor{cyan21255225}{RGB}{21,255,225}
\definecolor{darkgray176}{RGB}{176,176,176}
\definecolor{darkorange2551480}{RGB}{255,148,0}
\definecolor{dodgerblue0128255}{RGB}{0,128,255}
\definecolor{gray}{RGB}{128,128,128}
\definecolor{lightgreen124255121}{RGB}{124,255,121}
\definecolor{maroon12700}{RGB}{127,0,0}
\definecolor{navy00127}{RGB}{0,0,127}
\definecolor{red255290}{RGB}{255,29,0}
\definecolor{yellow22825518}{RGB}{228,255,18}

\begin{axis}[
%
yscale=\cikmshortenscaley,
tick align=outside,
tick pos=left,
x grid style={darkgray176},
xmin=0.750904121398925, xmax=396.552114067078,
xtick style={color=black},
y grid style={darkgray176},
ymin=39.9023255813953, ymax=100.888372093023,
ytick style={color=black},
yticklabel style={rotate=90.0},
tick label style={font=\huge}
]
\addplot [ultra thick, gray, dashed, mark=*, mark size=2, mark options={solid}]
table {%
18.7418682098389 42.8372093023256
34.6578793048859 43
50.8279933929443 60.6744186046512
66.9683317661285 68.8604651162791
83.4322623729706 77.7209302325581
99.9514236450195 80.7674418604651
116.412863349915 84.1162790697674
132.916092252731 87.1860465116279
149.050437068939 88.7906976744186
165.219602060318 90.2325581395349
181.240503978729 90.7674418604651
197.671772670746 91.3488372093023
213.76312084198 91.5581395348837
213.76312084198 91.5581395348837
247.107247304916 92.3255813953488
263.459987735748 92.6744186046512
279.685011863708 92.7906976744186
296.458326721191 92.906976744186
313.223739433289 93.1860465116279
329.267318487167 93.2790697674419
345.646366643906 93.4883720930233
362.122578382492 93.6279069767442
378.451215457916 93.8139534883721
};
\addplot [maroon12700, dashed, mark=*, mark size=1, mark options={solid}]
table {%
19.839794588089 42.8372093023256
22.695743560791 42.8372093023256
26.140824508667 42.8372093023256
29.6172603607178 42.8604651162791
33.1739074230194 42.906976744186
36.5923642635345 43
40.2112493515015 43.1627906976744
43.5286998748779 43.1627906976744
47.1236826896668 43.2093023255814
50.6501331806183 43.1860465116279
54.181188249588 43.1395348837209
57.7141266822815 43.1860465116279
61.4184632301331 43.2325581395349
64.8179613590241 43.2093023255814
68.1556833744049 43.2093023255814
71.5325571537018 43.2325581395349
74.9547048568726 43.2790697674419
78.4973690986633 43.3023255813954
82.005387210846 43.2790697674419
85.4736286640167 43.2558139534884
88.9436983108521 43.2558139534884
92.4008068561554 43.2790697674419
95.8825394630432 43.2790697674419
99.2770111560822 43.3488372093023
102.775295352936 43.3488372093023
106.258724164963 43.3720930232558
109.835197734833 43.3720930232558
113.182870197296 43.3023255813953
116.74207406044 43.3255813953488
120.157164096832 43.3023255813953
123.531704521179 43.3023255813953
124.211724281311 43.3023255813953
130.573982858658 43.2790697674419
134.063445043564 43.2558139534884
137.629412317276 43.2790697674419
141.012291955948 43.2790697674419
144.515233135223 43.2325581395349
148.021944713593 43.2093023255814
151.652167701721 43.2325581395349
155.16945438385 43.2325581395349
158.684369754791 43.2093023255814
162.306138706207 43.1162790697674
165.846583271027 43.1162790697674
169.344696712494 43.1395348837209
172.848862314224 43.1860465116279
176.319840335846 43.2093023255814
179.633705663681 43.1860465116279
183.051858663559 43.2093023255814
186.4593272686 43.2790697674419
189.947561788559 43.2558139534884
193.432874631882 43.2325581395349
196.963741779327 43.2093023255814
200.398335552216 43.2325581395349
203.893950891495 43.2790697674419
207.540527200699 43.2093023255814
211.081966257095 43.2093023255814
214.636774682999 43.2093023255814
218.044996023178 43.2558139534884
221.57808804512 43.2093023255814
225.069985675812 43.1627906976744
228.582210302353 43.1627906976744
231.959030056 43.1860465116279
232.637424373627 43.1627906976744
239.06225605011 43.2093023255814
242.475042152405 43.1627906976744
246.118146800995 43.1395348837209
249.620413064957 43.1627906976744
252.975922298431 43.2325581395349
256.391177415848 43.2093023255814
259.921007966995 43.1860465116279
263.527158498764 43.2325581395349
267.116750669479 43.2325581395349
270.749297904968 43.1395348837209
274.11722612381 43.1627906976744
277.585965251923 43.1162790697674
281.04747800827 43.0930232558139
284.579418706894 43.1162790697674
288.106624746323 43.046511627907
291.662030553818 43
295.08782954216 42.9767441860465
298.554833745956 43
302.082434654236 43
305.807660865784 42.9302325581395
309.420103883743 42.9767441860465
313.030579662323 42.9302325581395
316.524045085907 42.9302325581395
319.987852954864 42.906976744186
323.530163192749 42.8837209302325
327.029419565201 42.8604651162791
330.574849891663 42.906976744186
333.971498918533 42.8837209302325
337.546960830688 42.9302325581395
341.145388269424 42.9767441860465
344.716472911835 42.953488372093
348.287282657623 42.9302325581395
351.762807321548 42.8837209302325
355.309735155106 42.8372093023256
358.74199719429 42.8604651162791
362.309473609924 42.8604651162791
365.712177848816 42.8372093023256
369.192029953003 42.8604651162791
372.597342777252 42.8372093023256
};
\addplot [red255290, dashed, mark=*, mark size=1, mark options={solid}]
table {%
21.1624392032623 42.8372093023256
22.1191129684448 42.8604651162791
26.7930511951447 42.8837209302326
31.495298576355 42.8372093023256
36.1612645149231 42.953488372093
39.8364887237549 42.9302325581395
45.5284746646881 42.8837209302326
50.1638140678406 42.8139534883721
54.799205160141 42.7441860465116
59.4528113365173 42.6976744186046
63.274574804306 42.8837209302326
68.7310170650482 42.9767441860465
73.4314738750458 42.9302325581395
78.1291780948639 42.9302325581395
82.7930714130402 42.9302325581395
87.425718164444 43
92.0786502838135 43
96.7106272697449 43.0232558139535
101.428462028503 43
106.145950841904 43
109.840386390686 42.9302325581395
115.515019130707 42.9302325581395
120.257042598724 42.906976744186
124.959957170486 42.8837209302326
129.660809803009 42.8372093023256
134.381789779663 42.8837209302325
139.058727455139 42.906976744186
143.82794919014 42.953488372093
148.56606259346 42.953488372093
153.152679872513 42.8837209302325
157.925745916367 42.9302325581395
162.54202914238 42.9302325581395
167.16804728508 42.953488372093
171.848989200592 42.9302325581395
176.580663490295 43
181.356801462173 42.9767441860465
186.061311912537 42.906976744186
190.803492307663 42.906976744186
195.5181016922 42.8837209302326
198.397731494904 42.906976744186
203.259736442566 42.8837209302326
208.710161209106 42.8604651162791
213.401228380203 42.906976744186
218.09307808876 42.906976744186
222.765782833099 42.9302325581395
227.406156349182 42.8372093023256
231.93954744339 42.9302325581395
236.645421695709 42.8837209302326
241.249228048325 42.8837209302326
245.890829801559 42.906976744186
250.615105056763 42.8604651162791
255.290913963318 42.8372093023256
259.997003793716 42.8372093023256
264.664121246338 42.7906976744186
269.231747293472 42.8139534883721
274.004154825211 42.7906976744186
278.745925045013 42.7441860465116
283.427136611938 42.7209302325581
288.008933782578 42.6976744186046
292.710395002365 42.6976744186046
297.435066223145 42.7674418604651
302.143979454041 42.6976744186046
306.751320314407 42.6976744186046
311.452672433853 42.6976744186046
316.046512174606 42.7674418604651
320.762447071075 42.7209302325581
325.348928880692 42.6976744186046
330.115785741806 42.7441860465116
334.805218791962 42.7209302325581
339.523593187332 42.7441860465116
344.164177274704 42.7906976744186
348.807863235474 42.8139534883721
353.369349002838 42.8139534883721
358.029272842407 42.8837209302325
362.732591342926 42.8604651162791
367.486305856705 42.8837209302325
372.189680814743 42.9302325581395
};
\addplot [darkorange2551480, dashed, mark=*, mark size=1, mark options={solid}]
table {%
23.2494355678558 42.8372093023256
24.6247612476349 42.8139534883721
31.688454914093 43.0697674418605
38.7999643325806 46.3255813953488
45.5842258453369 58.046511627907
52.6717593669891 65.4186046511628
59.4605291366577 69.3255813953488
66.5535765171051 73.8837209302326
73.4266484260559 75.8837209302326
80.4255704402924 77.5581395348837
87.2797524929047 79.0232558139535
94.3607317447662 81.046511627907
101.28918337822 83.5348837209302
108.223071718216 85.7674418604651
113.709525680542 87.1627906976744
120.675100851059 88.6046511627907
127.683014583588 90.3488372093023
134.6981985569 91.4418604651163
141.929196023941 92.3023255813954
144.791168260574 92.7209302325581
151.725949192047 93.5116279069767
158.576783466339 94.0930232558139
165.444598293304 94.6744186046512
172.264963245392 95.3488372093023
179.365720272064 95.8372093023256
186.254044055939 96.3255813953488
193.238543987274 96.6279069767442
200.147317695618 96.906976744186
205.694056367874 97.1860465116279
212.749541091919 97.2558139534884
219.809916687012 97.4883720930232
226.875987434387 97.5581395348837
233.630612373352 97.6511627906977
240.506379747391 97.7674418604651
247.517276620865 97.7906976744186
254.556716299057 97.8372093023256
261.595568275452 97.860465116279
264.391700983047 97.8837209302326
271.376812410355 98
278.408281898499 97.9767441860465
285.361020803452 98.0232558139535
290.917532205582 98.0232558139535
297.954526567459 97.9767441860465
304.899319505692 98
312.014249753952 98.046511627907
319.016141843796 98.046511627907
325.873650836945 98.0697674418604
332.810785293579 98.0930232558139
339.741583681107 98.0697674418604
346.572565937042 98.0232558139535
353.387074804306 98.0930232558139
360.36987285614 98.046511627907
367.343076944351 98
374.325678443909 98.1162790697674
};
\addplot [yellow22825518, dashed, mark=*, mark size=1, mark options={solid}]
table {%
25.5751091003418 42.8372093023256
32.6352754116058 42.6744186046512
41.414736032486 47.3720930232558
50.1587089061737 59.6279069767442
58.924261713028 65.8139534883721
67.8646119594574 71.906976744186
76.7183005809784 75.7209302325581
85.658029127121 77.3953488372093
94.5503008842468 79.1627906976744
103.408923721313 81.0232558139535
112.431806135178 83.3953488372093
121.302973127365 86.0232558139535
130.075528383255 87.8139534883721
138.916523456573 89.4883720930233
147.776653575897 90.9302325581396
156.663361692429 91.9069767441861
165.351282262802 93.1162790697674
174.166581344604 93.9767441860465
182.95574297905 94.8372093023256
191.668877506256 95.3488372093023
200.540460586548 95.7441860465116
209.048454618454 96.2790697674419
217.874329328537 96.5581395348837
226.574269151688 96.906976744186
235.62039141655 97.1627906976744
244.31933016777 97.3023255813954
253.044112920761 97.3953488372093
261.648012924194 97.4651162790698
270.268446016312 97.6279069767442
278.926194953918 97.7674418604651
287.469729423523 97.8139534883721
296.200722789764 97.8837209302326
305.138183736801 97.9767441860465
313.898196840286 98
322.64971909523 98
331.51318230629 98
340.261712169647 98.0232558139535
349.075561618805 97.9767441860465
358.003320884705 98
366.658710575104 98.0232558139535
375.39258351326 98.046511627907
};
\addplot [lightgreen124255121, dashed, mark=*, mark size=1, mark options={solid}]
table {%
27.0188595294952 42.8372093023256
28.9375638961792 42.7674418604651
38.589860534668 43.3953488372093
48.3599937438965 50.953488372093
58.1882607460022 64.6511627906977
67.9366462230682 68.7906976744186
77.8537021160126 74.6744186046512
87.7641509056091 78.4418604651163
97.7038267612457 80
107.59081363678 82.046511627907
117.463397789001 84.5581395348837
127.378373813629 86.5581395348837
137.233432912827 88.4418604651163
146.905843114853 90.093023255814
156.678805541992 91.2093023255814
166.719143199921 92.2790697674419
176.601769542694 93.0697674418605
186.606962013245 93.7441860465116
196.556894540787 94.2325581395349
206.473974609375 94.5813953488372
216.355820512772 94.8837209302326
226.382388544083 95.2558139534884
236.305454540253 95.5813953488372
245.924613237381 95.8604651162791
255.748436641693 96.2790697674418
265.697275209427 96.5116279069768
275.646078443527 96.6744186046512
285.56091337204 96.8604651162791
295.377829122543 97.0232558139535
305.236872291565 97.0930232558139
315.118064212799 97.2093023255814
324.935134077072 97.3255813953488
334.945643615723 97.3488372093023
344.433732175827 97.4186046511628
354.280577850342 97.5116279069767
364.183051204681 97.6279069767442
374.104853916168 97.6744186046512
};
\addplot [cyan21255225, dashed, mark=*, mark size=1, mark options={solid}]
table {%
28.5189386367798 42.8372093023256
35.0869171619415 42.8372093023256
46.1204494476318 47.4186046511628
56.7810942173004 59.6046511627907
67.7864831924439 67.9767441860465
78.8079293251038 73.8139534883721
89.7747990131378 78.6976744186047
100.6577709198 81.2790697674419
111.745693016052 83.6976744186046
122.623010921478 86.1860465116279
133.713986539841 88.6511627906977
144.931122970581 90
156.04733543396 91
167.066186189651 91.8372093023256
177.929360628128 92.5813953488372
188.936727428436 93.0697674418605
200.054201841354 93.3488372093023
211.079790592194 93.7209302325581
222.060373735428 94.046511627907
233.153118896484 94.3488372093023
244.310883903503 94.5348837209303
255.27470369339 94.9069767441861
266.401179790497 95.1860465116279
277.441615104675 95.5348837209302
288.375249290466 95.8604651162791
299.662703752518 96.093023255814
310.81183180809 96.2093023255814
321.707679605484 96.3720930232558
332.616712522507 96.5348837209302
343.537173748016 96.6976744186047
354.730137586594 96.9767441860465
365.643888902664 97
376.828820085526 97.046511627907
};
\addplot [dodgerblue0128255, dashed, mark=*, mark size=1, mark options={solid}]
table {%
29.5265080928802 42.8372093023256
39.387158203125 42.7441860465116
51.9450975894928 50.7906976744186
64.5559819221497 63.7209302325582
77.2169281959534 72.7906976744186
90.0200599193573 78.3953488372093
102.569763660431 82.0697674418604
115.037800645828 85.2558139534884
127.189331245422 87.8604651162791
139.890969276428 89.953488372093
152.144149065018 91.0232558139535
165.161415433884 91.7209302325581
177.900431251526 92.4186046511628
190.234754276276 93.1395348837209
202.930086803436 93.5116279069768
215.539416122437 94.046511627907
228.186646556854 94.4418604651163
240.967716026306 94.8139534883721
253.519105815887 94.9767441860465
265.921100759506 95.2325581395349
278.324674320221 95.4651162790698
290.947765922546 95.7209302325581
303.367177724838 95.9767441860465
315.960980367661 96.1395348837209
328.456915378571 96.3255813953488
340.9063808918 96.3720930232558
353.386622571945 96.4883720930233
366.053413009644 96.7209302325582
378.522457504272 96.8604651162791
};
\addplot [blue, dashed, mark=*, mark size=1, mark options={solid}]
table {%
31.6754893302917 42.8372093023256
37.1994411945343 42.7441860465116
50.9459073543549 47.8139534883721
65.0815687656403 60.8372093023256
79.332327413559 70.6744186046512
93.7815219402313 77.3488372093023
107.636353445053 81.1860465116279
121.567006111145 84.8604651162791
135.696373558044 87.2325581395349
150.027084684372 89.1162790697675
163.966705131531 90.3953488372093
177.983021068573 91.1162790697675
192.375245952606 91.8139534883721
206.447863674164 92.2790697674419
220.531658411026 92.4418604651163
229.05703997612 92.5813953488372
248.590930366516 93
262.865787649155 93.3023255813954
276.851116323471 93.3023255813954
291.143612003326 93.5581395348838
305.163221311569 93.6744186046512
319.334906387329 93.8372093023256
333.346614360809 94
347.724591732025 94.3720930232558
361.675174474716 94.4651162790698
375.746467590332 94.5581395348837
};
\addplot [navy00127, dashed, mark=*, mark size=1, mark options={solid}]
table {%
34.408545589447 42.8372093023256
43.4481165409088 42.7906976744186
58.3664643764496 51.3953488372093
73.6694990634918 64.6511627906977
89.0846886634827 73.4651162790698
104.326226854324 79.8372093023256
119.404794454575 82.7441860465116
134.601596212387 85.3720930232558
149.725624275208 87.7209302325581
164.673254394531 89.3953488372093
180.043390035629 90.5813953488372
195.613752698898 91.2325581395349
210.634994411469 91.5581395348837
225.747588396072 91.8372093023256
241.040662097931 92.093023255814
256.214940834045 92.2790697674419
271.644025754929 92.4651162790698
286.859267234802 92.6511627906977
302.125139904022 92.7674418604651
317.108119297028 92.953488372093
332.299588346481 93.1162790697675
347.792752122879 93.4651162790698
363.27049908638 93.6279069767442
378.561149978638 93.8604651162791
};
\end{axis}

\end{tikzpicture}

%% file: plots/short_pgf_plots_for_cikm/mondial_religion_random_short.tex
\begin{tikzpicture}

\definecolor{blue00254}{RGB}{0,0,254}
\definecolor{darkgray176}{RGB}{176,176,176}
\definecolor{darkorange2551220}{RGB}{255,122,0}
\definecolor{deepskyblue0212255}{RGB}{0,212,255}
\definecolor{dodgerblue096255}{RGB}{0,96,255}
\definecolor{gold2552290}{RGB}{255,229,0}
\definecolor{gray}{RGB}{128,128,128}
\definecolor{greenyellow17025576}{RGB}{170,255,76}
\definecolor{maroon12700}{RGB}{127,0,0}
\definecolor{red254180}{RGB}{254,18,0}
\definecolor{turquoise76255170}{RGB}{76,255,170}

\begin{axis}[
yscale=\cikmshortenscaley,
tick align=outside,
tick pos=left,
x grid style={darkgray176},
xmin=-10.5480736446381, xmax=269.972555131912,
xtick style={color=black},
y grid style={darkgray176},
ymin=60.1857142857143, ymax=85.4333333333333,
ytick style={color=black},
yticklabel style={rotate=90.0},
tick label style={font=\huge}
]
\addplot [maroon12700, dashed, mark=*, mark size=1, mark options={solid}]
table {%
2.20286402702332 61.4285714285714
2.5560574054718 61.4285714285714
4.81719527244568 61.7142857142857
6.98703985214233 61.8095238095238
8.09621353149414 61.7142857142857
9.80806188583374 61.7142857142857
11.9219880580902 61.6190476190476
14.0913488864899 61.5238095238095
15.269357252121 61.4285714285714
16.9989255428314 61.4285714285714
18.6851957798004 61.5238095238095
20.8329288482666 61.8095238095238
21.9885646343231 61.7142857142857
24.2161321640015 61.7142857142857
26.4252452850342 62.0952380952381
27.5816652297974 62
29.2434527873993 62.2857142857143
31.3814942836761 62.7619047619048
33.0979339122772 62.9523809523809
34.3827078342438 63.047619047619
36.0464380741119 63.8095238095238
38.1289276599884 64.0952380952381
40.1656268596649 64.4761904761905
41.3765030384064 64.2857142857143
43.1224255084991 64
45.1950078964233 64
46.911291217804 63.9047619047619
48.144819688797 63.7142857142857
50.2481157779694 63.8095238095238
51.9357481956482 63.8095238095238
53.6217921257019 64
55.2157918453217 64.0952380952381
57.3468459606171 64
59.0664599895477 63.8095238095238
60.7067603588104 63.8095238095238
61.9825736522675 63.7142857142857
64.0690699100494 63.7142857142857
66.2006471633911 63.9047619047619
67.4450879573822 63.9047619047619
69.0152147769928 64.0952380952381
71.2237362861633 64.3809523809524
72.8422658920288 64.3809523809524
74.4284302711487 64.5714285714286
75.6778112888336 64.5714285714286
77.7234303474426 64.5714285714286
79.3838660240173 64.4761904761905
81.116321182251 64.5714285714286
82.8069970607758 64.5714285714286
84.9452337741852 64.2857142857143
86.5627013683319 64.3809523809524
88.158745765686 64.3809523809524
90.2878704547882 64.3809523809524
91.5691679477692 64.6666666666667
93.2599022865295 64.6666666666667
94.8796440124512 64.5714285714286
96.9761597633362 64.7619047619048
98.6745968341827 64.7619047619048
100.338742208481 64.9523809523809
101.993496084213 64.9523809523809
104.024143600464 64.9523809523809
105.274143838882 64.9523809523809
106.903352212906 64.9523809523809
108.538686084747 64.9523809523809
110.675727176666 64.9523809523809
112.327125787735 64.9523809523809
113.975758647919 65.047619047619
115.55315322876 65.047619047619
117.217192792892 65.047619047619
119.322666931152 65.047619047619
120.544356870651 64.9523809523809
122.757340717316 64.9523809523809
123.959217834473 64.9523809523809
126.085544872284 65.1428571428571
127.70125041008 65.047619047619
129.804258584976 65.1428571428571
131.032003116608 65.1428571428571
133.105868673325 65.1428571428571
134.349823093414 65.1428571428571
136.505779266357 65.047619047619
138.150311946869 65.2380952380952
139.824636268616 65.3333333333333
141.522894334793 65.2380952380952
143.139012670517 65.4285714285714
145.267298078537 65.4285714285714
146.862862491608 65.5238095238095
148.263938570023 65.5238095238095
149.854583311081 65.6190476190476
152.047924423218 65.7142857142857
153.693988180161 65.7142857142857
155.335053682327 65.6190476190476
157.017194080353 65.6190476190476
159.229451560974 65.2380952380952
160.881215381622 65.2380952380952
162.179206466675 65.2380952380952
163.79938454628 65.2380952380952
165.991610717773 65.1428571428571
167.641137552261 65.047619047619
169.286092615128 64.9523809523809
171.416089677811 64.9523809523809
173.077594470978 64.9523809523809
174.66575345993 64.9523809523809
176.000052547455 64.9523809523809
178.436469507217 64.9523809523809
179.757841014862 64.9523809523809
181.354567289352 65.047619047619
183.022372436523 64.9523809523809
185.117664957047 65.1428571428571
186.805748319626 65.047619047619
188.016576862335 65.047619047619
189.798623228073 65.047619047619
192.274497699738 65.2380952380952
193.554398345947 65.2380952380952
195.271407747269 65.2380952380952
196.907061624527 65.4285714285714
199.028919410706 65.5238095238095
200.657833433151 65.5238095238095
202.281101131439 65.5238095238095
203.587680149078 65.5238095238095
206.112809228897 65.4285714285714
207.315100383759 65.5238095238095
208.941495323181 65.5238095238095
211.086475849152 65.5238095238095
212.76278386116 65.4285714285714
213.942681360245 65.3333333333333
216.004522657394 65.4285714285714
217.805058097839 65.5238095238095
219.806217336655 65.5238095238095
221.003356266022 65.6190476190476
222.725706243515 65.6190476190476
224.848740530014 65.6190476190476
226.449073171616 65.6190476190476
227.650921773911 65.6190476190476
229.759567737579 65.6190476190476
231.854639434814 65.6190476190476
233.076473855972 65.6190476190476
235.21147646904 65.5238095238095
236.931221437454 65.5238095238095
238.708122777939 65.5238095238095
239.851449251175 65.4285714285714
241.970396232605 65.4285714285714
244.038555717468 65.4285714285714
245.689846229553 65.6190476190476
246.842672920227 65.6190476190476
248.91886806488 65.5238095238095
251.028347396851 65.6190476190476
};
\addplot [red254180, dashed, mark=*, mark size=1, mark options={solid}]
table {%
3.67874484062195 61.4285714285714
4.20755677223206 61.4285714285714
7.34230017662048 61.7142857142857
10.1954313278198 61.7142857142857
13.249497795105 61.7142857142857
15.5659972667694 61.6190476190476
18.4755847930908 61.9047619047619
20.2800256729126 61.8095238095238
23.812687921524 62.1904761904762
24.9022344112396 62.2857142857143
27.796724653244 62.9523809523809
30.8459454536438 64.1904761904762
33.911888551712 65.4285714285714
36.2371391296387 66.8571428571429
37.9990769863129 66.8571428571429
41.4451992511749 67.6190476190476
44.5767281532288 67.9047619047619
46.341162776947 68.1904761904762
48.7525456905365 68.4761904761905
51.6807912349701 68
53.914469575882 68.2857142857143
56.8310263633728 68.5714285714286
59.8435831546783 68.7619047619048
62.169130563736 68.5714285714286
65.0579323291779 68.4761904761905
66.3514638900757 68.5714285714286
69.2390008926392 68.4761904761905
71.5609379291534 68.3809523809524
74.0538522720337 68.7619047619048
77.5789571285248 68.8571428571428
80.4905513286591 68.9523809523809
82.4124357223511 69.0476190476191
85.8650266170502 69.2380952380952
87.056218957901 69.0476190476191
90.1274044036865 68.8571428571428
93.1014050960541 68.8571428571428
95.3029860496521 68.9523809523809
98.2444681167603 68.6666666666667
101.131577014923 68.7619047619048
103.46192278862 68.6666666666667
105.688878107071 68.8571428571428
107.573543071747 68.6666666666667
110.601211738586 68.6666666666667
113.471786165237 68.8571428571428
116.316774988174 69.2380952380952
118.605540037155 69.3333333333333
121.004152917862 69.5238095238095
123.960364341736 69.6190476190476
126.877605247498 69.5238095238095
128.044188261032 69.4285714285714
130.471744632721 69.5238095238095
133.960593414307 69.5238095238095
136.307574415207 69.5238095238095
138.026861906052 69.6190476190476
141.462241172791 69.6190476190476
144.495905065537 69.5238095238095
146.949105501175 69.6190476190476
149.294033050537 69.6190476190476
151.599306964874 69.9047619047619
153.45424118042 70
156.991566610336 70.1904761904762
159.914042520523 70.0952380952381
161.705145692825 70.0952380952381
165.123362064362 70
168.109235191345 69.9047619047619
169.261197042465 69.9047619047619
172.276213502884 69.8095238095238
174.584580755234 69.9047619047619
177.723731040955 69.9047619047619
180.681712293625 70.4761904761905
183.735848474503 70.4761904761905
186.065729904175 70.5714285714286
188.367624282837 70.6666666666667
190.132685518265 70.5714285714286
193.074275922775 70.7619047619048
195.362009382248 70.9523809523809
198.300484228134 70.7619047619048
201.228501701355 70.7619047619048
203.579786300659 70.8571428571428
206.497482967377 70.8571428571428
208.789296388626 71.0476190476191
210.584059810638 71.1428571428571
213.628666353226 71.5238095238095
216.652894115448 71.4285714285714
218.894453334808 71.5238095238095
221.359839296341 71.5238095238095
224.36432723999 71.3333333333333
227.372708082199 71.6190476190476
230.383541822433 71.4285714285714
231.559777593613 71.5238095238095
234.543445777893 71.8095238095238
236.957866096497 71.7142857142857
240.033679151535 71.4285714285714
242.390899658203 71.5238095238095
245.42188949585 71.3333333333333
248.339447164536 71.6190476190476
251.365196752548 71.4285714285714
};
\addplot [darkorange2551220, dashed, mark=*, mark size=1, mark options={solid}]
table {%
5.29536366462708 61.4285714285714
6.03780636787415 61.6190476190476
10.0103371143341 61.6190476190476
13.8571484088898 61.6190476190476
17.8142520904541 61.6190476190476
20.8548550605774 61.7142857142857
23.9361349582672 61.9047619047619
27.828705739975 62.1904761904762
31.6890737056732 62.4761904761905
35.5992658615112 63.4285714285714
38.5601714611053 64
41.6533357620239 65.047619047619
44.7249318122864 65.6190476190476
48.6919362068176 66.5714285714286
51.6419382572174 67.5238095238095
55.4281406879425 68.5714285714286
59.2718497276306 68.6666666666667
62.3839135169983 68.9523809523809
66.3388046741486 69.1428571428572
68.5782406806946 69.2380952380952
72.4765304088592 69.1428571428572
75.6814748764038 69.0476190476191
79.5352306365967 68.8571428571429
81.7150686264038 68.9523809523809
85.6619840621948 68.5714285714286
89.5064570903778 68.9523809523809
93.5388353347778 69.2380952380952
97.3542468070984 70
100.304812002182 70.4761904761905
103.472557353973 70.7619047619048
107.389168548584 70.7619047619048
111.300522470474 71.0476190476191
113.559470796585 71.4285714285714
117.526994228363 71.8095238095238
120.547348642349 72.2857142857143
124.490840959549 72.2857142857143
128.434368562698 72
130.69932923317 72.3809523809524
134.647332715988 72.0952380952381
138.501718950272 72.1904761904762
141.472593212128 72
145.279942178726 72.0952380952381
147.54248380661 72.1904761904762
151.554386615753 72.3809523809524
155.486897754669 72.6666666666667
158.560892963409 72.7619047619048
161.580827951431 72.9523809523809
165.498095989227 73.1428571428571
169.459468460083 73.0476190476191
173.352217817307 72.7619047619048
177.246036529541 72.6666666666667
179.377416944504 72.5714285714286
183.401493740082 72.5714285714286
186.56835436821 72.4761904761905
190.409612751007 72.1904761904762
192.670074033737 72.1904761904762
196.575399160385 72.4761904761905
199.684923744202 72.4761904761905
203.48676738739 72.5714285714286
207.39145026207 72.9523809523809
210.3661008358 72.9523809523809
214.322856664658 72.9523809523809
217.365810918808 73.047619047619
220.528563833237 72.8571428571428
223.519459819794 72.9523809523809
227.414934158325 72.6666666666667
231.428426837921 72.6666666666667
235.295291519165 72.6666666666667
238.334145784378 72.6666666666667
241.314185857773 72.7619047619047
245.210621070862 72.8571428571429
248.974774837494 72.7619047619048
252.886905193329 72.952380952381
};
\addplot [gold2552290, dashed, mark=*, mark size=1, mark options={solid}]
table {%
7.13324980735779 61.4285714285714
8.14458312988281 61.5238095238095
13.4950493335724 61.4285714285714
18.6382183074951 61.6190476190476
23.7520748615265 61.6190476190476
28.9152451515198 62.0952380952381
34.1060287952423 63.047619047619
39.2153942108154 65.1428571428572
44.3780384063721 67.6190476190476
49.4199508666992 69.5238095238095
54.6451024532318 71.0476190476191
59.7116995334625 71.7142857142857
65.0686508655548 73.0476190476191
70.1964127540588 73.1428571428571
75.221151971817 73.3333333333333
80.3087235450745 73.8095238095238
85.648721408844 74
89.8211145401001 74.2857142857143
94.9330551624298 74.2857142857143
100.087172317505 74.3809523809524
105.351909208298 75.0476190476191
110.550023317337 74.5714285714286
115.615069818497 74.6666666666667
120.892097759247 75.0476190476191
126.100480747223 75.4285714285714
131.350183200836 76
135.416867303848 76
140.611897516251 76.1904761904762
145.76192317009 76.1904761904762
149.861826705933 75.9047619047619
156.054418563843 76.0952380952381
161.239154100418 76.4761904761905
166.508921051025 76.6666666666667
171.585744333267 76.7619047619048
175.6440762043 76.8571428571429
180.880347394943 76.9523809523809
186.153267049789 76.8571428571429
191.414802360535 77.0476190476191
196.583431720734 77.3333333333333
201.814394235611 77.2380952380952
206.954569864273 77.4285714285714
212.15918545723 77.5238095238095
217.364687824249 77.6190476190476
222.544212341309 77.7142857142857
227.764798307419 77.7142857142857
233.087895393372 77.7142857142857
238.177415704727 77.7142857142857
243.371784877777 77.8095238095238
248.449631595612 78.0952380952381
253.484661388397 78.0952380952381
};
\addplot [greenyellow17025576, dashed, mark=*, mark size=1, mark options={solid}]
table {%
7.87784190177917 61.4285714285714
10.1780866146088 61.6190476190476
16.1934928417206 61.6190476190476
22.1922439575195 61.5238095238095
28.1608184814453 61.4285714285714
34.1921167373657 62.1904761904762
40.213250541687 63.047619047619
46.290804195404 64.2857142857143
52.3515053272247 66.0952380952381
58.4204647064209 69.0476190476191
64.417483997345 70.2857142857143
70.4305641651154 71.9047619047619
76.44318318367 73.0476190476191
82.423285150528 73.0476190476191
88.5573534488678 72.8571428571428
93.3995400905609 72.952380952381
99.3941611766815 73.4285714285714
105.324018955231 73.7142857142857
111.333005571365 74
117.068821763992 73.9047619047619
123.158693742752 73.7142857142857
129.284668731689 73.8095238095238
134.033761501312 73.9047619047619
137.769381856918 74.1904761904762
146.00122961998 74.3809523809524
151.786076068878 74.6666666666667
157.929083204269 74.3809523809524
163.90899939537 74.3809523809524
169.850817346573 74
174.403936767578 73.8095238095238
180.488977003098 73.5238095238095
184.25099105835 73.5238095238095
192.589888954163 73.8095238095238
198.499558019638 73.6190476190476
204.408600473404 73.9047619047619
210.490593242645 73.9047619047619
216.601633739471 74.1904761904762
222.689267730713 74.2857142857143
228.688927745819 74.2857142857143
234.744983768463 74.6666666666667
240.75641169548 74.4761904761905
246.721366024017 74.1904761904762
252.691428613663 74.5714285714286
};
\addplot [turquoise76255170, dashed, mark=*, mark size=1, mark options={solid}]
table {%
9.25400514602661 61.4285714285714
12.0080399513245 61.3333333333333
19.1549551010132 61.4285714285714
26.2433980941772 61.4285714285714
33.437696313858 61.3333333333333
40.6428973674774 62.2857142857143
47.8109988212585 64.7619047619048
55.0298783779144 65.9047619047619
61.9776961803436 69.0476190476191
69.1189252376556 71.6190476190476
76.333492231369 73.1428571428572
83.2490103244781 75.7142857142857
90.550492811203 75.5238095238095
97.6688824653625 76.1904761904762
104.934264707565 76.8571428571429
112.101764345169 77.4285714285714
119.304126262665 78
126.40150437355 77.9047619047619
133.303655004501 78.3809523809524
138.89040312767 78.4761904761905
146.183464717865 78.1904761904762
153.301200532913 78
160.255118274689 77.9047619047619
167.320825004578 77.6190476190476
174.410904121399 78
181.265344762802 77.9047619047619
188.240982198715 77.9047619047619
195.451939344406 78.0952380952381
202.603005838394 78
209.808545160294 78.0952380952381
217.146792840958 78.4761904761905
224.475642347336 79.4285714285714
231.456284046173 79.3333333333333
238.74966340065 79.4285714285714
246.090912389755 79.7142857142857
253.364640331268 79.5238095238095
};
\addplot [deepskyblue0212255, dashed, mark=*, mark size=1, mark options={solid}]
table {%
10.7787230014801 61.4285714285714
12.2879600524902 61.5238095238095
19.9785836219788 61.5238095238095
27.7072557449341 61.8095238095238
35.4797786712646 61.5238095238095
43.2144836425781 61.5238095238095
51.0227803230286 63.5238095238095
58.843785572052 66.7619047619048
66.5212213039398 71.5238095238095
74.3864619731903 75.5238095238095
82.3098615646362 77.0476190476191
89.9760604381561 79.0476190476191
97.6927713871002 80.9523809523809
105.547055482864 81.4285714285714
113.147517633438 82.1904761904762
121.035076236725 82.6666666666667
128.707287740707 83.047619047619
136.576956796646 83.3333333333333
144.181591033936 83.2380952380952
151.929181241989 83.047619047619
159.911274194717 82.8571428571428
167.708892440796 82.8571428571428
175.544827604294 82.7619047619047
183.312181568146 82.8571428571428
191.040288686752 82.8571428571428
198.770842647552 82.9523809523809
206.64641289711 82.9523809523809
214.521179389954 83.047619047619
222.314964056015 83.047619047619
229.962507724762 82.9523809523809
237.991020107269 83.047619047619
245.903901767731 83.1428571428571
253.790887880325 83.047619047619
};
\addplot [dodgerblue096255, dashed, mark=*, mark size=1, mark options={solid}]
table {%
12.1620532512665 61.4285714285714
15.5877549171448 61.6190476190476
24.3430017471313 61.6190476190476
32.981134223938 61.7142857142857
41.7203585147858 61.9047619047619
50.4697587966919 62.1904761904762
59.0254769325256 64.3809523809524
67.8258729934692 66.952380952381
76.5373094081879 71.6190476190476
84.9746689796448 75.8095238095238
93.7207908153534 79.1428571428572
102.487405157089 81.0476190476191
111.380782413483 82.4761904761905
120.106691598892 82.9523809523809
128.81602473259 82.5714285714286
137.1761926651 83.047619047619
144.035069942474 83.1428571428571
152.794446563721 83.2380952380952
161.235832929611 83.1428571428571
169.993304491043 83.2380952380952
178.827638101578 83.1428571428571
187.332048988342 83.2380952380952
196.014358091354 83.2380952380952
204.732096862793 83.5238095238095
213.238833284378 83.8095238095238
222.136850500107 83.7142857142857
230.986657762527 83.5238095238095
239.630443143845 83.9047619047619
248.424101781845 83.9047619047619
257.221617460251 84.0952380952381
};
\addplot [blue00254, dashed, mark=*, mark size=1, mark options={solid}]
table {%
13.4387179851532 61.4285714285714
15.3117681503296 61.6190476190476
24.7643397808075 61.6190476190476
34.3521013259888 61.3333333333333
43.7902143478394 61.9047619047619
53.3942825317383 62.5714285714286
62.9538483142853 65.1428571428572
72.8070870399475 67.0476190476191
82.436506319046 70.9523809523809
92.0432461738586 74
101.539973497391 76.6666666666667
111.028438949585 79.2380952380952
120.572354221344 80.6666666666667
130.081339359283 81.4285714285714
139.483454704285 82.6666666666667
148.901809263229 83.047619047619
158.415180253983 83.047619047619
167.900707197189 83.1428571428571
177.138372802734 82.8571428571428
186.528070497513 82.6666666666667
196.104494285583 82.4761904761905
205.709280967712 82.2857142857143
215.256691169739 82.3809523809524
224.686488676071 83.1428571428571
234.197661066055 83.2380952380952
243.804293012619 83.047619047619
253.512127065659 83.047619047619
};
\addplot [ultra thick, gray, dashed, mark=*, mark size=2, mark options={solid}]
table {%
14.2887599468231 61.4285714285714
24.6740455150604 61.5238095238095
35.141822719574 61.6190476190476
45.638599729538 62.1904761904762
56.2690583229065 64.4761904761905
66.8843245983124 66.7619047619048
77.7658805847168 71.1428571428571
88.6925711154938 76.4761904761905
99.2472709655762 80.2857142857143
109.508946180344 82.5714285714286
119.790633487701 83.1428571428571
130.337510061264 83.9047619047619
140.955986309052 83.7142857142857
151.609069061279 84
162.209997320175 84.0952380952381
172.600471687317 84.2857142857143
183.052301597595 84.1904761904762
193.526771879196 84
204.086597251892 83.8095238095238
214.457344055176 83.7142857142857
225.259791564941 83.7142857142857
235.826626968384 83.4285714285714
246.619962596893 83.4285714285714
257.192080831528 83.6190476190476
};
\end{axis}

\end{tikzpicture}

%% file: plots/short_pgf_plots_for_cikm/genes_random_short.tex
\begin{tikzpicture}

\definecolor{blue00254}{RGB}{0,0,254}
\definecolor{darkgray176}{RGB}{176,176,176}
\definecolor{darkorange2551220}{RGB}{255,122,0}
\definecolor{deepskyblue0212255}{RGB}{0,212,255}
\definecolor{dodgerblue096255}{RGB}{0,96,255}
\definecolor{gold2552290}{RGB}{255,229,0}
\definecolor{gray}{RGB}{128,128,128}
\definecolor{greenyellow17025576}{RGB}{170,255,76}
\definecolor{maroon12700}{RGB}{127,0,0}
\definecolor{red254180}{RGB}{254,18,0}
\definecolor{turquoise76255170}{RGB}{76,255,170}

\begin{axis}[
yscale=\cikmshortenscaley,
tick align=outside,
tick pos=left,
x grid style={darkgray176},
xmin=-16.2479414820671, xmax=398.137098157406,
xtick style={color=black},
y grid style={darkgray176},
ymin=40.0127906976744, ymax=99.5918604651163,
ytick style={color=black},
yticklabel style={rotate=90.0},
tick label style={font=\huge}
]
\addplot [maroon12700, dashed, mark=*, mark size=1, mark options={solid}]
table {%
2.58774213790894 42.8372093023256
3.64498281478882 42.8372093023256
6.65758190155029 42.8372093023256
9.66098575592041 42.8372093023256
12.7042511463165 42.8604651162791
15.0632170200348 42.8837209302325
17.3694919586182 42.906976744186
20.3698163032532 42.953488372093
23.3625504970551 43.0232558139535
26.4224194526672 43.0232558139535
28.7044408321381 43.046511627907
30.97461977005 43.046511627907
33.9751330375671 43.046511627907
36.9684103965759 43.0697674418605
40.0352935791016 43.1395348837209
42.3032980918884 43.1860465116279
45.2747196674347 43.2325581395349
47.6180274009705 43.2790697674419
50.7034152507782 43.3488372093023
52.9804457187653 43.4186046511628
55.9396831035614 43.4186046511628
58.9492003917694 43.5116279069767
61.3004956245422 43.5581395348837
64.3602819442749 43.7209302325581
66.6317977905273 43.7441860465116
69.6478846549988 43.8604651162791
72.6586328983307 43.9302325581395
75.682994556427 44.046511627907
77.3960463047028 44.1162790697674
79.6655840396881 44.1395348837209
82.706064081192 44.2325581395349
85.72474360466 44.2558139534884
88.7378216266632 44.2558139534884
91.0521395683289 44.1860465116279
93.3226320266724 44.1162790697674
96.2821929454803 44.1860465116279
99.3078223705292 44.1627906976744
102.451764154434 44.2093023255814
104.829879426956 44.2790697674419
107.090959453583 44.3255813953488
110.166672420502 44.3488372093023
113.060292053223 44.4186046511628
116.059549665451 44.4883720930233
118.309755897522 44.4651162790698
120.67734246254 44.6744186046512
123.603910827637 44.6279069767442
126.613899803162 44.6279069767442
129.677856588364 44.6511627906977
132.052360200882 44.6744186046512
135.132677316666 44.7209302325581
137.490012598038 44.6744186046512
140.477062177658 44.6744186046512
142.790997982025 44.6976744186046
145.706832170486 44.6744186046512
148.644187927246 44.7209302325581
150.428758239746 44.6976744186046
153.375205612183 44.8139534883721
155.601852321625 44.8372093023256
158.658036088943 44.953488372093
160.538588237762 45.0232558139535
164.658755493164 44.9767441860465
166.876250362396 44.953488372093
169.112927436829 45
172.15791888237 45
175.133863115311 44.906976744186
178.096272468567 44.906976744186
180.407681179047 45.0232558139535
182.747528648376 44.9767441860465
185.695875549316 45
188.660253953934 44.953488372093
191.630499887466 44.9302325581395
194.051074123383 44.9767441860465
196.328502082825 45.093023255814
199.198698568344 45.0697674418605
202.212542486191 45.1627906976744
205.187919473648 45.1395348837209
207.541822004318 45.1860465116279
209.817450618744 45.2325581395349
212.914886236191 45.2325581395349
215.935506343842 45.3023255813953
218.924695301056 45.3255813953488
221.333780431747 45.3023255813953
224.284006977081 45.3255813953488
226.083477735519 45.3488372093023
229.1890604496 45.3255813953488
232.142998075485 45.3023255813953
234.505363464355 45.2790697674419
237.427761173248 45.2093023255814
239.728816080093 45.1627906976744
242.646643686295 45.0930232558139
244.950968551636 45.1395348837209
247.907439184189 45.0930232558139
250.981411457062 45.0930232558139
253.981238222122 45.0930232558139
256.326464653015 45.0697674418605
258.655218982697 45.046511627907
261.645161008835 45.1162790697674
264.59021821022 44.9767441860465
267.558640384674 45
269.929532194138 45.046511627907
272.244231939316 45.0697674418605
275.205262041092 45.0697674418605
278.353735256195 45.2325581395349
281.479959154129 45.1627906976744
283.80009765625 45.1860465116279
286.105051183701 45.2558139534884
289.126295232773 45.1860465116279
292.189613389969 45.2558139534884
295.167724084854 45.2093023255814
297.485000085831 45.2093023255814
299.236741542816 45.2325581395349
302.271394777298 45.1395348837209
305.261975765228 45.2093023255814
308.329113769531 45.2325581395349
310.668033123016 45.2325581395349
313.663542079926 45.1395348837209
316.018733930588 45.1162790697674
318.945441436768 45.0930232558139
320.93944568634 45.1162790697674
324.390257787704 45.1162790697674
327.37824587822 45.1395348837209
329.773576545715 45.1627906976744
332.784627532959 45.1395348837209
335.240753030777 45.2093023255814
338.334758234024 45.1162790697674
341.36905207634 45.1395348837209
344.408077049255 45.1860465116279
346.763212108612 45.1627906976744
349.098349142075 45.2558139534884
352.169553661346 45.2790697674419
355.051040744782 45.2558139534884
358.07454624176 45.2790697674418
360.35591173172 45.2558139534884
362.630840682983 45.3023255813953
365.715178632736 45.2790697674419
368.823685121536 45.3953488372093
371.84010720253 45.4418604651163
};
\addplot [red254180, dashed, mark=*, mark size=1, mark options={solid}]
table {%
4.31169328689575 42.8372093023256
4.31169328689575 42.8372093023256
8.790345287323 42.8372093023256
14.0260390758514 43
16.6623085021973 42.9302325581395
21.9681106567383 44.4651162790698
25.6406538486481 47.0232558139535
30.1031998634338 48.046511627907
31.9441672801971 48.1860465116279
36.2950868606567 49.4651162790698
40.8391095161438 50.3255813953488
44.2206384181976 50.4883720930233
47.7593249320984 50.8372093023256
52.2320488452911 51.4651162790698
56.6488260746002 52.2790697674419
60.197688627243 52.4418604651163
62.9905045509338 52.8372093023256
68.2740552902222 53.2790697674419
69.893551158905 53.3488372093023
74.3738564968109 53.8372093023256
78.9175070285797 54.4186046511628
83.3931062698364 55.1627906976744
86.7473275184631 55.3720930232558
90.4086434364319 55.5581395348837
93.9096077442169 55.6976744186046
97.3201712608337 55.6976744186046
101.788083934784 55.7441860465116
104.670553350449 55.8604651162791
109.935978794098 56.1627906976744
112.414102697372 56.2558139534884
116.955051422119 56.3953488372093
120.493772602081 56.6511627906977
123.166401672363 56.6744186046512
128.486864280701 56.9302325581395
132.922188043594 57.2558139534884
136.542898368835 57.3720930232558
139.133341693878 57.3720930232558
143.645131731033 57.5348837209302
147.264328193665 57.6046511627907
150.6314848423 57.6976744186046
155.019815063477 57.8139534883721
158.591538381577 57.7906976744186
163.164888715744 57.7209302325581
166.527259922028 57.8372093023256
170.972443294525 57.9302325581395
174.603324317932 57.9302325581395
177.170546150208 57.953488372093
180.629717826843 58.1162790697674
185.035067081451 58.1162790697674
189.506503582001 58.1860465116279
192.973140859604 58.3255813953488
197.377596998215 58.2790697674419
200.908441925049 58.3255813953488
203.526812553406 58.3255813953488
206.393932342529 58.2325581395349
210.919243431091 58.3255813953488
216.160555076599 58.2558139534884
219.661838722229 58.2558139534884
223.347539901733 58.2790697674419
227.803232049942 58.3720930232558
231.302397012711 58.3720930232558
234.860852193832 58.3720930232558
239.310798072815 58.4883720930232
242.936102581024 58.3255813953488
244.698364496231 58.3255813953488
249.950592422485 58.3488372093023
254.408914804459 58.2325581395349
257.750686168671 58.2325581395349
261.289889764786 58.2790697674419
265.73624124527 58.2790697674418
269.268650960922 58.3255813953488
272.592414808273 58.2790697674418
276.241024208069 58.3488372093023
280.669624757767 58.3023255813954
284.071642208099 58.2325581395349
287.664269685745 58.1395348837209
291.204878425598 58.1162790697674
295.725933456421 58.0930232558139
299.117264795303 58.2093023255814
303.583247661591 58.1627906976744
308.085458278656 58.2093023255814
309.706330490112 58.2093023255814
314.089363956451 58.1162790697674
317.653959274292 58.1162790697674
322.119074678421 58.2558139534884
325.565437936783 58.2325581395349
329.973973321915 58.2325581395349
333.43696770668 58.2558139534884
336.905438804626 58.3023255813953
341.343982553482 58.2325581395349
345.04150800705 58.2325581395349
348.570084810257 58.2093023255814
352.097009325027 58.1395348837209
355.585381889343 58.1162790697674
360.10271730423 58.1395348837209
363.547417211533 58.0697674418605
367.875452756882 58.093023255814
372.390941524506 58.1162790697674
};
\addplot [darkorange2551220, dashed, mark=*, mark size=1, mark options={solid}]
table {%
6.51910810470581 42.8372093023256
7.59494256973267 42.8372093023256
13.6851147651672 42.953488372093
19.813981962204 44.0232558139535
25.6875759601593 48.953488372093
30.4115665912628 50.9767441860465
36.5832704544067 52.9069767441861
41.3642321586609 55.3953488372093
47.2997756958008 56.7441860465116
53.2122789382935 57.9302325581395
57.9082548618317 58.5581395348837
62.8889686584473 59.6511627906977
68.8552669048309 60.4651162790698
73.5144555568695 61.4418604651163
78.1426040172577 62.093023255814
84.2341351509094 62.7906976744186
90.2226552009583 63.4418604651163
96.2571826934814 64.3488372093023
102.173807239532 65.3255813953488
105.595280885696 65.6046511627907
111.545814085007 66.3023255813954
117.575326251984 66.6976744186047
122.611799192429 67
129.611014175415 67.5813953488372
133.036630630493 67.8837209302326
139.110080385208 68
143.823471212387 68.1860465116279
149.981896543503 68.3953488372093
154.581298971176 68.5581395348837
160.530321788788 68.4883720930233
166.488547420502 68.5581395348837
172.495368909836 68.7209302325581
177.355223655701 68.9767441860465
181.987901258469 68.8837209302326
188.041476631165 68.9767441860465
192.819005632401 69.2325581395349
198.862381982803 69.3255813953488
203.633615970612 69.1860465116279
208.476520681381 69.2790697674419
214.606227874756 69.2558139534884
220.722453832626 69.2790697674419
226.810481929779 69.2790697674419
231.465577125549 69.2790697674419
237.464864253998 69.1627906976744
241.185885572433 69.1162790697674
248.371618175507 69.2325581395349
254.387973642349 69.2790697674419
257.853048086166 69.3023255813954
263.831109333038 69.3023255813954
269.920245981216 69.2558139534884
276.05698390007 69.3255813953488
279.453358745575 69.2790697674419
285.50254406929 69.1860465116279
291.464717769623 69.2093023255814
297.318070554733 69.2790697674419
303.392170000076 69.3255813953488
307.995660400391 69.3488372093023
312.802172374725 69.3255813953488
317.598172569275 69.3488372093023
323.630449914932 69.4186046511628
328.197847604752 69.3953488372093
334.092807102203 69.4418604651163
340.099442243576 69.6046511627907
343.862395429611 69.6279069767442
350.91166434288 69.5813953488372
355.566873979568 69.7209302325581
361.661051607132 69.7209302325581
367.612751865387 69.8372093023256
373.555086135864 69.7674418604651
};
\addplot [gold2552290, dashed, mark=*, mark size=1, mark options={solid}]
table {%
8.50979175567627 42.8372093023256
11.23190741539 42.7209302325581
18.7447402000427 43.6744186046512
26.3842206954956 50.046511627907
33.9271439552307 59.6279069767442
41.4747193813324 64.3023255813954
47.3051511287689 68.3255813953488
53.0441835403442 70.7906976744186
60.4205767154694 73
68.0285848140716 74.6511627906977
75.6040594577789 76.7441860465116
81.3027666091919 78
88.9091298103333 79.4651162790698
96.3846137523651 81.3488372093023
102.141714811325 82.6976744186046
109.621267986298 83.6511627906977
117.088499069214 84.7209302325581
122.791181468964 85.4651162790698
130.210542678833 85.9069767441861
137.741696166992 86.2093023255814
145.158782291412 86.7441860465116
151.071596288681 86.8837209302326
156.830593109131 87.1627906976744
164.221179628372 87.4186046511628
171.540661334991 87.6046511627907
178.960459899902 87.8372093023256
186.385471343994 88.2093023255814
193.839096069336 88.3255813953488
197.992307424545 88.4186046511628
205.607534313202 88.6046511627907
213.352861785889 88.7906976744186
220.835208320618 88.7906976744186
228.433389234543 88.8372093023256
234.346502733231 88.8604651162791
240.226408529282 88.8837209302326
247.806152439117 88.906976744186
255.137610435486 89.0232558139535
262.717171049118 89
268.62842502594 89.046511627907
276.344967794418 89.1162790697674
283.823961591721 89.0697674418605
289.724031543732 89.0930232558139
297.086873435974 89.2093023255814
304.429228448868 89.1395348837209
310.350488853455 89.1860465116279
317.877024841309 89.0930232558139
325.235259246826 89.1860465116279
332.849966001511 89.1627906976744
338.770459938049 89.1162790697674
344.489685201645 89.1395348837209
352.064213275909 89.2093023255814
359.677355384827 89.2325581395349
367.235768079758 89.2093023255814
374.575281190872 89.2325581395349
};
\addplot [greenyellow17025576, dashed, mark=*, mark size=1, mark options={solid}]
table {%
10.2594799041748 42.8372093023256
13.7322043418884 42.7674418604651
22.8115735054016 45.6511627906977
31.6335004329681 53.5348837209302
40.5786554813385 63.4883720930233
49.5313712120056 67.7441860465116
58.7491654872894 72.8139534883721
67.7819414138794 75.2325581395349
76.9264497756958 77.4883720930232
86.079205083847 79.2790697674419
95.0152848243713 81.3953488372093
102.028309202194 82.6511627906977
110.931581878662 84.1162790697674
119.999487686157 84.8604651162791
129.192391014099 85.6511627906977
136.306196308136 86.0232558139535
145.164340877533 86.5581395348837
154.186092185974 86.9069767441861
163.321058797836 87.3720930232558
172.367743682861 87.8604651162791
181.207101345062 88.3023255813954
190.086663532257 88.5813953488372
197.171494436264 88.8139534883721
202.565614557266 88.953488372093
215.18542971611 89.3023255813954
224.267865514755 89.3023255813954
233.182832241058 89.6976744186047
242.172231674194 89.8837209302326
251.137376308441 90.1627906976744
258.355054283142 90.1860465116279
267.370028495789 90.3255813953488
272.73792347908 90.2790697674419
284.969505214691 90.3720930232558
291.924668645859 90.3023255813954
300.778187131882 90.3720930232558
309.815341043472 90.3953488372093
318.767434215546 90.5581395348837
327.877022790909 90.5348837209302
336.772830581665 90.5116279069767
345.854109334946 90.5116279069767
354.906508255005 90.6046511627907
364.019502019882 90.4651162790698
373.131201553345 90.4883720930232
};
\addplot [turquoise76255170, dashed, mark=*, mark size=1, mark options={solid}]
table {%
12.0210280895233 42.8372093023256
16.2015858650208 42.8604651162791
27.1489161014557 45.6744186046512
37.8110353469849 54.8139534883721
48.6037202835083 65.953488372093
59.370680141449 74.046511627907
70.1103183269501 79.953488372093
80.9176777362823 82.4418604651163
91.5143751621246 84.9069767441861
102.522165346146 87.0697674418605
113.208915996552 89.3953488372093
124.217329454422 90.6046511627907
135.128449106216 91.3488372093023
145.768268489838 91.9767441860465
156.587456130981 92.3023255813954
167.393191576004 92.5813953488372
178.112707996368 92.906976744186
189.023208332062 93.3255813953488
199.573137712479 93.6279069767442
203.634694719315 93.7674418604651
214.446725225449 93.953488372093
225.127882051468 94.3023255813954
236.066826295853 94.7209302325582
247.011710548401 94.8837209302326
258.107643985748 95.093023255814
269.031563520432 95.2093023255814
279.677290296555 95.4186046511628
290.383556890488 95.6279069767442
301.287638998032 95.906976744186
312.244945907593 96.0697674418605
322.754330205917 96.1860465116279
333.528727769852 96.3255813953488
344.277459192276 96.4651162790698
354.933006858826 96.6511627906977
365.47378783226 96.7209302325581
376.293748235703 96.8837209302326
};
\addplot [deepskyblue0212255, dashed, mark=*, mark size=1, mark options={solid}]
table {%
14.1317253112793 42.8372093023256
16.4813459396362 42.9302325581395
28.4612557411194 45.0697674418605
40.8850030422211 54.7906976744186
52.8570613861084 67.1627906976744
64.8925700187683 74.3023255813954
76.9088493347168 79.6511627906977
89.0400539875031 82.953488372093
100.839324808121 85.7906976744186
113.097235822678 88.2325581395349
125.002539348602 90
136.701075935364 90.953488372093
146.393066596985 91.7209302325581
158.796195745468 92.4186046511628
170.941131496429 92.6046511627907
182.89168047905 93.1627906976744
194.988002443314 93.2093023255814
199.768937397003 93.4418604651163
212.123265266418 93.5581395348837
224.317221879959 93.7209302325581
236.564394903183 94.046511627907
248.746862459183 94.3023255813954
260.505202007294 94.7209302325581
270.123868227005 94.8837209302326
282.293724012375 95.2325581395349
294.423954772949 95.3023255813953
306.410960102081 95.3953488372093
318.351197052002 95.6511627906977
330.363576984406 95.8604651162791
342.551256227493 96.1395348837209
354.892612504959 96.1860465116279
367.188641881943 96.3953488372093
379.026525783539 96.6511627906977
};
\addplot [dodgerblue096255, dashed, mark=*, mark size=1, mark options={solid}]
table {%
15.6186882972717 42.8372093023256
18.3695953845978 42.8372093023256
31.8037138462067 45.0930232558139
45.3939785957336 55.7209302325582
58.8721303462982 69.0697674418605
72.2492671966553 76.3720930232558
85.7724052429199 80.5348837209303
99.1154541015625 83.7674418604651
112.914837026596 86.4418604651163
127.014417505264 88.7674418604651
140.299557065964 90.5813953488372
153.457708692551 91.5116279069768
167.044228458405 92.2790697674419
180.277895402908 92.7906976744186
193.655517482758 93.0697674418605
204.278023147583 93.3255813953488
218.102642202377 93.6744186046512
231.230952072144 93.8837209302326
244.536957073212 94.0930232558139
257.928685188293 94.1860465116279
271.307204675674 94.3023255813954
284.767873191833 94.3953488372093
298.107170581818 94.6046511627907
311.483020067215 94.8139534883721
325.1706407547 94.9302325581395
338.601550626755 95.2325581395349
352.200178956985 95.5116279069768
365.538585329056 95.6046511627907
379.30141453743 95.906976744186
};
\addplot [blue00254, dashed, mark=*, mark size=1, mark options={solid}]
table {%
17.5018193721771 42.8372093023256
20.3336221694946 42.8372093023256
35.1589440822601 45.4883720930233
50.1571625709534 58.8837209302326
64.844313287735 69.8372093023256
79.8479808807373 78.2790697674419
94.9318290710449 82.0697674418604
110.028542280197 85.6046511627907
124.789586877823 88.1627906976744
139.798922109604 89.6976744186047
154.779183673859 90.8837209302325
169.743154478073 91.6744186046512
184.546446561813 92.1627906976744
199.318388652802 92.4418604651163
214.39347114563 92.6744186046512
226.604488134384 92.8372093023256
244.461021757126 93
259.059178161621 93.2790697674419
273.91031794548 93.4418604651163
288.811427164078 93.6046511627907
303.677708530426 93.9069767441861
318.653675031662 94.1162790697674
333.550012636185 94.4186046511628
348.366209888458 94.5581395348837
363.153579187393 94.6511627906977
377.866205739975 94.8837209302326
};
\addplot [ultra thick, gray, dashed, mark=*, mark size=2, mark options={solid}]
table {%
18.7418682098389 42.8372093023256
34.6578793048859 43
50.8279933929443 60.6744186046512
66.9683317661285 68.8604651162791
83.4322623729706 77.7209302325581
99.9514236450195 80.7674418604651
116.412863349915 84.1162790697674
132.916092252731 87.1860465116279
149.050437068939 88.7906976744186
165.219602060318 90.2325581395349
181.240503978729 90.7674418604651
197.671772670746 91.3488372093023
213.76312084198 91.5581395348837
213.76312084198 91.5581395348837
247.107247304916 92.3255813953488
263.459987735748 92.6744186046512
279.685011863708 92.7906976744186
296.458326721191 92.906976744186
313.223739433289 93.1860465116279
329.267318487167 93.2790697674419
345.646366643906 93.4883720930233
362.122578382492 93.6279069767442
378.451215457916 93.8139534883721
};
\end{axis}

\end{tikzpicture}

%% file: table-of-winners.tex
{\begin{table*}[t]
\small
\caption {The best training time for each scheme selection strategy and task. The table contains the shortest time it takes to reach the threshold quality $\alpha^*$ over all tested removal ratios $r$ (i.e. $t^{*}(\mathcal{T})$). The last row (ALL) refers to the embedding time of the original \forward run, with all schemes, as a baseline.
The best (lowest) train time for each task is printed in bold. 
\label{tab:table_of_winners}}
\small
\vskip-0.5em
\begin{tabular}{rccccccccc}
\toprule
{} &       Mutagenesis &           World &         Hepatitis &             Genes & M.-Religion & M.-Infant & M.-Continent & M.-GDP & M.-Inflation \\
  \midrule
  \kvar    &       \textbf{20.59} &     \textbf{154.32} &      \textbf{29.31} &     107.18 &                   \textbf{42.75} &                               \textbf{42.17} &                    54.39 &                   \textbf{56.84} &                       \textbf{49.46} \\
  {}    &    (+-2.67) &  (+-18.71) &   (+-6.52) &   (+-5.73) &                (+-2.63) &                            (+-4.61) &                 (+-1.58) &                (+-8.20) &                    (+-6.47) \\
  \midrule
\onlineschemeelimination   &       41.89 &     \textbf{150.38} &      60.88 &     115.91 &                   61.76 &                               61.96 &                    69.44 &                   \textbf{56.55} &                       65.31 \\
  {}  &    (+-2.68) &  (+-11.09) &  (+-33.53) &   (+-4.15) &                (+-2.31) &                            (+-2.48) &                 (+-2.88) &                (+-9.86) &                    (+-4.80) \\
  \midrule
  \random   &       39.09 &     268.97 &      69.98 &     115.39 &                   88.22 &                               76.16 &                   113.15 &                   71.02 &                        82.5 \\
  {}   &    (+-7.29) &  (+-81.67) &  (+-16.33) &   (+-5.83) &               (+-14.12) &                            (+-0.71) &                 (+-7.07) &                (+-6.00) &                   (+-35.24) \\
  \midrule
  \oneepoch      &       41.57 &      191.2 &      58.68 &     126.46 &                   55.06 &                               56.03 &                    67.16 &                   65.63 &                       65.08 \\
  {}      &    (+-1.21) &  (+-31.45) &  (+-15.50) &   (+-7.12) &                (+-1.20) &                            (+-3.36) &                 (+-1.37) &                (+-4.88) &                    (+-6.58) \\
  \midrule
  All       &       55.89 &      550.65 &      58.36 &     161.9 &                   101.32 &                               76.16 &                    113.14 &                   69.03 &                       95.37 \\
  &  (+-8.67) &  (+-22.92) &   (+-9.3) &   (+-6.74) &                (+-8.53) &                            (+-0.71) &                 (+-7.06) &                (+-8.77) &                   (+-8.61) \\
\bottomrule
\end{tabular}
\end{table*}}

%% file: plots/pgf_winners_croped/winners_genes.tex
\begin{tikzpicture}

\definecolor{brown}{RGB}{165,42,42}
\definecolor{cyan}{RGB}{0,255,255}
\definecolor{darkgray176}{RGB}{176,176,176}
\definecolor{gray}{RGB}{128,128,128}
\definecolor{pink}{RGB}{255,192,203}
\definecolor{yellow}{RGB}{255,255,0}

\begin{axis}[
tick align=outside,
tick pos=left,
x grid style={darkgray176},
xmin=-3.17090572834015, xmax=209.227615971565,
xtick style={color=black},
y grid style={darkgray176},
ymin=39.9953488372093, ymax=100.46976744186,
ytick style={color=black},
yticklabel style={rotate=90.0},
tick label style={font=\huge}
]
\addplot [brown, dashed, mark=*, mark size=1, mark options={solid}]
table {%
23.2494355678558 42.8372093023256
24.6247612476349 42.8139534883721
31.688454914093 43.0697674418605
38.7999643325806 46.3255813953488
45.5842258453369 58.046511627907
52.6717593669891 65.4186046511628
59.4605291366577 69.3255813953488
66.5535765171051 73.8837209302326
73.4266484260559 75.8837209302326
80.4255704402924 77.5581395348837
87.2797524929047 79.0232558139535
94.3607317447662 81.046511627907
101.28918337822 83.5348837209302
108.223071718216 85.7674418604651
113.709525680542 87.1627906976744
120.675100851059 88.6046511627907
127.683014583588 90.3488372093023
134.6981985569 91.4418604651163
141.929196023941 92.3023255813954
144.791168260574 92.7209302325581
151.725949192047 93.5116279069767
158.576783466339 94.0930232558139
165.444598293304 94.6744186046512
172.264963245392 95.3488372093023
179.365720272064 95.8372093023256
186.254044055939 96.3255813953488
193.238543987274 96.6279069767442
};
\addplot [pink, dashed, mark=*, mark size=1, mark options={solid}]
table {%
11.7192026615143 42.8372093023256
22.5542466640472 42.7441860465116
33.8371977329254 49.7906976744186
44.7859028816223 64.9069767441861
55.8318242549896 69.5581395348837
66.8345057010651 76.4651162790698
78.0249980926514 79.046511627907
89.1814641952515 81.7209302325581
100.343766403198 84.3720930232558
111.430643415451 87.1860465116279
122.098607587814 89.1627906976744
133.165959072113 90.3953488372093
144.139920282364 91.5581395348837
154.971298885345 92.3023255813954
165.981677436829 92.7674418604651
177.162645339966 93.1627906976744
188.341822147369 93.6976744186047
199.448613643646 93.9767441860465
};
\addplot [green, dashed, mark=*, mark size=1, mark options={solid}]
table {%
18.622926902771 43
21.9658090591431 43.0697674418605
37.7216928958893 46.4883720930233
49.6956450939178 60.6976744186046
57.8922268390656 68.1395348837209
65.8411688327789 74.6744186046512
73.8553527832031 77.2790697674419
82.0740615844727 79.093023255814
90.222373342514 81.2093023255814
98.2225781440735 84.1627906976744
106.153576803207 87.0232558139535
114.145459651947 89.5348837209303
122.34808216095 91.0697674418605
130.470261573792 92.5581395348837
138.61401219368 93.6279069767442
145.164632225037 94.3488372093023
154.766314220428 95.2790697674419
163.001870727539 95.5348837209302
171.019965982437 95.8139534883721
179.12060751915 96.2558139534884
187.313134098053 96.7674418604651
195.444982528687 97.1395348837209
};
\addplot [red, dashed, mark=*, mark size=1, mark options={solid}]
table {%
7.77462520599365 42.8372093023256
10.3594027996063 42.8372093023256
17.0372488498688 44.046511627907
23.6163023948669 50.3023255813953
30.5647964000702 58.6279069767442
37.1894445896149 66.093023255814
43.7793123245239 70.4418604651163
50.426499414444 74.3720930232558
57.2948344707489 76.6279069767442
64.0319117546082 78.1162790697674
70.7514793872833 79.7906976744186
77.4866584300995 81.3255813953488
84.1124987125397 83.3023255813954
91.0373836040497 85.6744186046512
97.7558959007263 87.3488372093023
104.498023176193 89.3488372093023
111.217102861404 91.046511627907
117.962502384186 92.1395348837209
124.576713180542 92.9302325581396
131.256309652328 93.6046511627907
138.036812067032 94.1627906976744
144.685187721252 94.8372093023256
151.476530122757 95.3023255813954
158.025437402725 95.6744186046512
164.564543676376 96.1162790697674
171.235788345337 96.4651162790698
177.908112955093 96.8372093023256
184.817539691925 97.0232558139535
191.605309295654 97.2093023255814
198.254035568237 97.3255813953488
};
\addplot [yellow, dashed, mark=*, mark size=1, mark options={solid}]
table {%
6.48357253074646 42.8372093023256
7.7639434337616 42.8139534883721
13.8575928688049 43.3255813953488
19.8965342998505 45.7209302325581
26.0380805969238 54.6511627906977
32.1302185058594 65.2790697674419
38.1156848430634 69.5348837209302
44.4157537937164 74.7674418604651
50.5324304580688 78.2790697674419
56.5397579193115 79.6279069767442
62.6747064590454 81.3255813953488
68.9167706489563 82.953488372093
75.0054110050201 84.6279069767442
79.8116587162018 86.1860465116279
85.8359529018402 88.0697674418605
91.8297462940216 90
97.7462100982666 91.1627906976744
103.707015657425 92.1860465116279
109.891026544571 93.2790697674419
116.259363555908 94.2558139534884
122.295182991028 95.046511627907
128.469789361954 95.5348837209302
133.242115879059 95.8837209302326
139.281222438812 96.2325581395349
145.426985454559 96.5348837209302
151.36715092659 96.7906976744186
156.107701253891 96.8372093023256
162.297390317917 96.9767441860465
168.391027927399 97.1627906976744
174.618225193024 97.3255813953488
180.633572864532 97.5348837209302
186.787736654282 97.6046511627907
191.574842882156 97.6511627906977
197.779824447632 97.7209302325581
};
\addplot [cyan, dashed, mark=*, mark size=1, mark options={solid}]
table {%
12.0210280895233 42.8372093023256
16.2015858650208 42.8604651162791
27.1489161014557 45.6744186046512
37.8110353469849 54.8139534883721
48.6037202835083 65.953488372093
59.370680141449 74.046511627907
70.1103183269501 79.953488372093
80.9176777362823 82.4418604651163
91.5143751621246 84.9069767441861
102.522165346146 87.0697674418605
113.208915996552 89.3953488372093
124.217329454422 90.6046511627907
135.128449106216 91.3488372093023
145.768268489838 91.9767441860465
156.587456130981 92.3023255813954
167.393191576004 92.5813953488372
178.112707996368 92.906976744186
189.023208332062 93.3255813953488
199.573137712479 93.6279069767442
};
\addplot [blue, dashed, mark=*, mark size=1, mark options={solid}]
table {%
37.6385472297668 42.8372093023256
39.649892616272 42.7441860465116
50.15989112854 43.8139534883721
60.8234440326691 53.8837209302326
71.6672357559204 65.7674418604651
82.5544308185577 70.2325581395349
93.0585238933563 76.6976744186046
103.485283231735 79.6976744186046
114.061302375793 82.1860465116279
124.754217100143 85.0232558139535
135.230596017838 87.3488372093023
146.06155667305 89.0232558139535
150.06518330574 89.3953488372093
160.757141828537 90.9767441860465
171.336586809158 92.0697674418605
182.299304962158 92.953488372093
192.786563682556 93.2558139534884
};
\addplot [ultra thick, gray, dashed, mark=*, mark size=2, mark options={solid}]
table {%
18.7418682098389 42.8372093023256
34.6578793048859 43
50.8279933929443 60.6744186046512
66.9683317661285 68.8604651162791
83.4322623729706 77.7209302325581
99.9514236450195 80.7674418604651
116.412863349915 84.1162790697674
132.916092252731 87.1860465116279
149.050437068939 88.7906976744186
165.219602060318 90.2325581395349
181.240503978729 90.7674418604651
197.671772670746 91.3488372093023
};
\addplot [semithick, red, dashed]
table {%
-3.17090572834015 86.7813953488372
209.227615971565 86.7813953488372
};
\end{axis}

\end{tikzpicture}

%% file: plots/pgf_winners_croped/winners_world_B.tex
\begin{tikzpicture}

\definecolor{brown}{RGB}{165,42,42}
\definecolor{cyan}{RGB}{0,255,255}
\definecolor{darkgray176}{RGB}{176,176,176}
\definecolor{gray}{RGB}{128,128,128}
\definecolor{pink}{RGB}{255,192,203}
\definecolor{yellow}{RGB}{255,255,0}

\begin{axis}[
tick align=outside,
tick pos=left,
x grid style={darkgray176},
xmin=-30.210105381012, xmax=732.751528606415,
xtick style={color=black},
y grid style={darkgray176},
ymin=18.6875, ymax=91.5625,
ytick style={color=black},
yticklabel style={rotate=90.0},
tick label style={font=\huge}
]
\addplot [brown, dashed, mark=*, mark size=1, mark options={solid}]
table {%
25.0860235214233 22.5833333333333
26.2827126979828 22.5
32.2812690258026 22.0833333333333
38.2539871692657 23.5
44.2793148994446 27
50.22057056427 35.3333333333333
56.2605782985687 42.0833333333333
62.2110053539276 45.25
68.0360562801361 49
74.0648292064667 50.8333333333333
80.0442442893982 53.3333333333333
86.0116858959198 54.25
91.9268402576447 55.25
97.7491015911102 56.0833333333333
103.646101427078 57.4166666666667
108.256164550781 58.5833333333333
115.366025209427 60.1666666666667
121.267319488525 61.5833333333333
127.132529783249 63.0833333333333
133.275859308243 64.1666666666667
139.166446495056 64.9166666666667
145.222410821915 65.6666666666667
150.991731977463 66.8333333333333
157.071754741669 67.1666666666667
162.998249292374 68.8333333333333
168.790479373932 69.8333333333333
174.724611759186 71.1666666666667
180.722168684006 72.3333333333333
186.479514312744 73.1666666666667
192.268340206146 74.1666666666667
197.023870849609 74.8333333333333
204.195869016647 75.5833333333333
210.103495073318 76.25
216.054578018188 76.4166666666667
221.853789997101 76.5833333333333
227.677316331863 76.9166666666667
233.583371210098 77
239.304552268982 77.3333333333333
245.337725305557 77.1666666666667
251.340889167786 77.8333333333333
257.327148294449 78.0833333333333
263.422853517532 78.5833333333333
269.408509683609 78.75
275.447693634033 78.9166666666667
281.139996623993 79.1666666666667
286.859344530106 79.0833333333333
292.734325885773 79.3333333333333
298.859671401978 79.25
304.824724960327 79.3333333333333
310.916499567032 79.5833333333333
316.872930192947 79.5833333333333
322.846194934845 79.6666666666667
328.64347743988 79.6666666666667
334.612163639069 79.8333333333333
340.340123605728 79.9166666666667
346.264596605301 80
352.279837608337 80.25
358.12956237793 80.6666666666667
364.035776996613 80.6666666666667
367.638718795776 80.9166666666666
372.301554059982 80.9166666666666
379.329895019531 81
385.241554403305 81
391.289697551727 81.1666666666667
397.071071386337 81.4166666666667
403.153685522079 81.6666666666667
409.138433361053 81.5833333333333
414.967230463028 82.0833333333333
420.960290765762 82.1666666666667
426.783482408524 82.1666666666667
432.766364049911 82.75
438.642784404755 82.5833333333333
444.435585451126 82.75
450.090340471268 82.75
455.947971582413 82.6666666666667
461.746206665039 82.6666666666667
467.583166742325 82.8333333333333
473.563024377823 82.75
479.521649074554 83
485.471543741226 83.25
491.211081504822 83.25
497.332694339752 83.4166666666667
503.2024269104 83.8333333333333
509.152420568466 83.9166666666667
515.011170482636 84
520.865822887421 83.9166666666667
526.807621669769 83.9166666666667
532.787357187271 84
538.787224578857 83.8333333333333
544.525101184845 83.9166666666667
550.587138462067 84.0833333333333
556.222374582291 83.8333333333333
562.030862665176 84
568.052035713196 83.9166666666667
573.916428613663 84.1666666666667
579.86898188591 84.25
585.782248973846 84.3333333333333
591.55517449379 84.5
597.247019052505 84.5
603.061942958832 84.5833333333333
608.962184524536 84.5833333333333
614.70754737854 84.9166666666667
620.527933216095 84.9166666666667
626.510812711716 84.9166666666667
632.378077459335 85.0833333333333
638.121665859222 85.25
644.082363843918 85.1666666666667
649.776740789413 85.25
655.796411514282 85.25
661.591683292389 85.6666666666667
667.420389556885 85.75
673.392203617096 85.75
679.22435541153 85.75
685.1305311203 85.9166666666667
691.136224412918 86
697.129099082947 86
};
\addplot [pink, dashed, mark=*, mark size=1, mark options={solid}]
table {%
9.11594152450562 22.5833333333333
10.56304063797 22.4166666666667
17.8171504020691 22.25
25.1104745388031 24.5
32.5365001678467 28.25
39.8687160015106 33.8333333333333
47.1873913288116 39.6666666666667
53.0876314640045 45.4166666666667
61.6670958995819 50.0833333333333
69.1351158618927 52.5
76.4046672821045 54.75
83.5515907764435 57
90.9636759757996 57.4166666666667
98.2954662799835 59.5
104.226324129105 60.3333333333333
113.066499710083 62.4166666666667
120.280933141708 64.25
127.544965791702 66.75
134.815542697906 67.5
142.158148574829 68.25
149.486611366272 68.6666666666667
156.665077257156 69.5833333333333
163.912523889542 70.4166666666667
171.308596038818 71.1666666666667
178.517246675491 71.9166666666667
184.238875102997 72.6666666666667
193.071205806732 73.3333333333333
200.534725856781 73.8333333333333
206.434026813507 74.4166666666667
215.300890207291 75
222.599689531326 75.6666666666667
228.565152072906 76
237.349562597275 76.1666666666667
244.579954433441 76.6666666666667
251.938164234161 76.9166666666667
259.151642894745 77.25
266.410074615478 77.5
273.701488494873 77.5833333333333
281.066662979126 77.75
288.408547115326 77.75
295.681350183487 77.6666666666667
302.97221827507 77.8333333333333
310.446417713165 78
317.680783319473 78.3333333333333
325.03112912178 78.4166666666667
332.1916680336 78.75
339.436156845093 78.75
346.814049863815 78.75
353.989941358566 79
358.352842950821 79
364.225965118408 78.9166666666667
373.132987737656 79.25
380.474234819412 79.1666666666667
386.386105489731 79.4166666666667
395.154476499558 79.5833333333333
402.576778268814 79.6666666666667
408.426223707199 79.5833333333333
417.211919116974 79.5
424.499564933777 79.75
430.375469112396 79.75
439.080480241775 79.5
446.32309718132 79.5
452.182417631149 79.75
460.889898061752 79.75
468.123783397675 79.9166666666667
475.202279520035 79.8333333333333
482.515830278397 79.8333333333333
489.889284944534 80.1666666666667
497.263785409927 80.1666666666667
504.580767393112 80.5
511.910187721252 80.1666666666667
519.236450910568 80.25
526.592895698547 79.9166666666667
533.847836828232 80.0833333333334
541.060649585724 80.4166666666667
548.349960803986 80.4166666666667
555.781271362305 80.5833333333334
563.069402885437 80.8333333333333
570.368483257294 80.8333333333333
577.607534074783 81
584.963909816742 81.4166666666667
592.104468822479 81.5833333333333
599.349423694611 81.9166666666667
606.722956037521 82.75
613.883631706238 82.8333333333333
621.226952695847 83.4166666666667
628.479061889648 83.75
635.773095655441 84.1666666666667
643.00127620697 84.1666666666667
650.375952863693 84.4166666666667
657.552937316894 84.6666666666667
664.860231876373 84.75
672.114423179626 85
679.376511287689 85.25
686.866709041595 85.4166666666667
694.116657066345 85.25
};
\addplot [green, dashed, mark=*, mark size=1, mark options={solid}]
table {%
18.616015625 22.75
28.6291922569275 23.3333333333333
42.9614906311035 24.75
52.3213272094727 32.5833333333333
59.3847906112671 41.9166666666667
65.9561538696289 48
72.7841572761536 51.6666666666667
79.6930151462555 54.5
86.5083760738373 56.3333333333333
93.1417127609253 58.8333333333333
99.7257121562958 61.1666666666667
106.517902565002 63.6666666666667
113.317599105835 64.75
120.250095176697 66.1666666666667
127.160383987427 67.25
134.082486629486 68.3333333333333
140.857656288147 70.5833333333333
147.711095428467 71.5833333333333
154.515029239655 73
161.32924618721 74.75
168.168316841125 75.5833333333333
174.874844455719 76.4166666666667
181.688160514832 77
188.550011777878 77.5833333333333
195.284852266312 78.0833333333333
202.181763935089 78.4166666666667
208.754256248474 78.9166666666667
211.438436460495 79.0833333333333
222.532293558121 79.5
229.123419284821 79.6666666666667
235.902329158783 79.5833333333333
242.705471897125 79.8333333333333
249.219847488403 79.9166666666667
255.891364479065 80.25
262.71932349205 80.3333333333333
269.642860889435 80.3333333333333
276.645918750763 80.5833333333333
283.395501041412 80.6666666666667
290.388944101334 81.0833333333333
297.117770576477 81
303.96763586998 81.1666666666667
310.875057315826 81.3333333333333
317.78729300499 81.25
324.521592760086 81.3333333333333
331.47936706543 81.5833333333333
338.35534658432 82.0833333333333
345.311339044571 82.3333333333333
352.261207771301 82.1666666666667
359.153538751602 82
366.109471273422 81.9166666666667
373.042904806137 82.3333333333333
380.076124191284 82.5833333333333
387.017407178879 82.75
393.858662843704 82.75
396.659104347229 83
407.793112468719 83.1666666666667
414.614626932144 83.1666666666667
421.317696619034 83.4166666666667
428.301153469086 83.6666666666667
431.018435049057 83.6666666666667
441.887230062485 83.5
448.699633789062 83.8333333333333
455.689566278458 84
462.400917625427 83.75
469.249961566925 83.9166666666667
476.020994520187 84
482.867783784866 84.1666666666667
489.710268211365 83.9166666666667
496.68376083374 84.1666666666667
503.64286646843 84.0833333333334
510.494350290298 84.4166666666667
517.539237499237 84.6666666666667
524.476916790009 84.8333333333333
531.259484529495 84.6666666666667
538.256646680832 84.9166666666667
545.093531417847 85
551.990100383759 84.9166666666667
558.844481086731 85.0833333333333
565.821234989166 84.8333333333333
572.547137928009 85
579.362403106689 85
586.240674257278 84.9166666666667
593.154061174393 85.0833333333333
599.952108049393 85.25
606.795005083084 85.5
613.433232545853 85.4166666666667
620.333096504211 85.5
627.454693365097 85.75
634.381126356125 85.6666666666667
641.349507904053 85.8333333333333
648.158539676666 85.9166666666667
655.180845975876 85.9166666666667
662.161024713516 86.25
669.164602994919 86.25
676.240614414215 86.5833333333333
682.961290550232 86.75
689.870697832108 87
696.729036426544 87
};
\addplot [red, dashed, mark=*, mark size=1, mark options={solid}]
table {%
9.23638291358948 22.5833333333333
10.3861789226532 22.75
16.4110993385315 22.25
22.2442367076874 23.1666666666667
28.2719284057617 26.1666666666667
34.1454751491547 34
39.9273540973663 40.9166666666667
45.805556678772 46.1666666666667
51.7629457950592 49.9166666666667
57.6226449012756 52.6666666666667
63.5718671798706 54
69.2679532527924 56.5
75.0211925029755 57.5833333333333
80.8889626026154 57.5833333333333
86.7500386238098 57.5
92.6538736820221 59.1666666666667
98.6739389419556 60.9166666666667
104.461622333527 62.0833333333333
110.376719665527 63.9166666666667
116.109654664993 64.8333333333333
122.128563499451 66.9166666666667
126.886749792099 68.1666666666667
133.058447408676 69.4166666666667
139.045617198944 70.5
145.006178045273 71.6666666666667
149.694328784943 72.25
155.484511184692 72.9166666666667
161.241547060013 73.4166666666667
167.106673669815 73.8333333333333
173.034643173218 74.4166666666667
179.140009069443 74.8333333333333
182.710550498962 75.3333333333333
188.550299596786 76.25
194.331913375854 76.75
200.103748321533 77.4166666666667
205.897148561478 77.6666666666667
211.794255971909 77.5833333333333
217.573823261261 77.75
223.45588183403 77.9166666666667
229.364922904968 78.1666666666667
235.315810871124 78.5
240.182314443588 78.6666666666667
246.054366350174 78.8333333333333
251.920932388306 79
257.675008773804 79.3333333333333
263.422937202454 79.8333333333333
269.344264173508 80
275.316494894028 80.0833333333333
281.192018890381 80.4166666666667
285.805142641068 80.5
291.578530311584 80.8333333333333
297.404376745224 81
303.207215547562 80.9166666666667
309.242569351196 81
314.941020584106 81
320.937270116806 81.25
326.796694517136 81.3333333333333
332.758875989914 81.5
338.52130279541 81.6666666666667
344.395524311066 81.4166666666667
350.134671497345 81.75
356.21048784256 81.6666666666667
358.574853563309 81.6666666666667
364.637186050415 81.75
370.57351307869 81.6666666666667
376.593037319183 81.5833333333333
382.551786661148 81.5833333333333
388.576428747177 81.6666666666667
394.501066684723 81.8333333333333
400.53444442749 82.1666666666667
406.4561814785 82.1666666666667
412.501690006256 82.3333333333333
418.43266043663 82.5
424.471207809448 82.5
429.263992261887 82.6666666666667
435.196634817123 82.6666666666667
441.258301496506 82.8333333333333
447.236821460724 82.9166666666667
453.083085823059 83.0833333333333
458.961817026138 82.8333333333333
464.820489645004 83.1666666666667
470.728022956848 83.5
475.349216651916 83.5
481.355702543259 83.9166666666667
487.352149915695 83.8333333333333
493.29750790596 84.0833333333333
499.262848520279 84.3333333333333
505.137950134277 84.1666666666667
511.157253170013 84.25
516.936588144302 84.5
522.989297771454 84.5833333333333
529.082339096069 85.0833333333333
532.659819698334 85.1666666666667
538.743068170548 85.3333333333333
544.729449319839 85.25
550.562538528442 85.5
556.774528884888 85.4166666666667
562.785152959824 85.5833333333333
567.342396020889 85.75
573.269977664948 85.8333333333333
579.030414056778 86.25
585.018435907364 86.1666666666667
589.809393405914 86.1666666666667
595.873262643814 86.4166666666667
601.663264989853 86.3333333333333
607.588901996613 86.0833333333333
613.481321048737 86.25
619.489391183853 86.4166666666667
625.520484876633 86.4166666666667
631.30504989624 86.3333333333333
637.320953416824 86.3333333333333
643.182402658462 86.8333333333333
649.12017364502 86.8333333333333
655.081592798233 86.9166666666667
661.07085351944 86.8333333333333
667.047111034393 86.9166666666667
673.141823959351 87.1666666666667
679.079108667374 87.5833333333333
685.096891450882 87.75
691.02840590477 87.5833333333333
697.006950426102 88
};
\addplot [yellow, dashed, mark=*, mark size=1, mark options={solid}]
table {%
4.4699688911438 22.5833333333333
5.27130947113037 22.25
9.75041561126709 22.25
14.1467833995819 22.5833333333333
18.5265609741211 23.4166666666667
22.840101480484 25
27.2001683712006 27.9166666666667
31.4956392765045 31.5833333333333
35.8896826267242 35.8333333333333
40.1019978523254 42.0833333333333
44.4530199050903 46
48.7432887554169 48.75
53.1708775997162 50.4166666666667
56.7337131500244 52.5833333333333
61.9246577739716 53.8333333333333
66.1870774745941 55.4166666666667
70.6062957763672 56.1666666666667
75.0627590656281 57.4166666666667
79.3864991664886 58.4166666666667
82.999379825592 59.1666666666667
87.3781530380249 59.75
90.9187253952026 60.1666666666667
96.214068365097 61.5
100.498280763626 61.8333333333333
103.967177200317 62.0833333333333
108.240904474258 63.5833333333333
111.84866733551 64
117.028347015381 64.4166666666667
120.567056274414 65
124.858860254288 65.75
129.218354320526 65.9166666666667
132.781913852692 66.5
138.010064649582 67.8333333333333
142.340133523941 68.6666666666667
146.650178098679 68.8333333333333
151.023005771637 69.5
155.397209739685 70.0833333333333
158.84445309639 70.3333333333333
163.172280550003 70.5833333333333
167.667069816589 71.4166666666667
171.945327854156 72
176.223181295395 72.3333333333333
178.850650930405 72.4166666666667
184.052212190628 72.8333333333333
188.48330039978 73.0833333333333
192.851354932785 73.5833333333333
197.131633996964 74.1666666666667
201.535893917084 74.6666666666667
204.975197410583 74.75
209.437919807434 74.8333333333333
214.045346403122 75.25
218.404870557785 75.6666666666667
221.967694187164 75.8333333333333
227.245981931686 76.25
231.584314775467 76.3333333333333
235.959222507477 76.6666666666667
238.534332084656 76.6666666666667
243.050458526611 76.9166666666667
247.342018318176 77.0833333333333
251.737379693985 77.1666666666667
256.309774827957 77.6666666666667
260.803671455383 77.8333333333333
264.400333881378 77.8333333333333
269.816293764114 77.9166666666667
274.249887561798 78.0833333333333
278.489660644531 77.9166666666667
282.815541601181 78.0833333333333
287.251055288315 78.3333333333333
291.654502296448 78.25
296.013011169434 78.3333333333333
300.469627094269 78.4166666666667
303.974075269699 78.5
308.476353311539 78.5833333333333
313.00851817131 78.6666666666667
316.527601289749 78.75
320.935880327225 78.75
325.429632997513 79.0833333333333
329.762343549728 79.25
334.171634864807 79.25
338.713622951508 79.5833333333333
343.057870721817 79.75
347.443507957459 79.75
351.874941396713 79.9166666666667
354.565466785431 79.9166666666667
358.039425230026 80
363.359320259094 80.25
367.771149778366 80.4166666666667
371.321776580811 80.5833333333333
376.620911836624 80.8333333333333
380.995489549637 81.0833333333333
385.300623559952 81.0833333333333
389.679277801514 81.4166666666667
393.189489269257 81.5
397.589614200592 81.6666666666667
401.98997092247 81.6666666666667
405.515289592743 81.9166666666667
409.959539031982 81.9166666666667
414.285161066055 81.9166666666667
418.634602499008 82
423.041706943512 82.1666666666667
427.662180328369 82.0833333333333
431.260691595078 82.25
436.500091743469 82.5
440.937622451782 82.5
444.395221948624 82.6666666666667
449.752142047882 82.8333333333333
454.299606180191 83.0833333333333
458.690525007248 83.25
463.116887569427 83.75
467.541650438309 83.75
471.845095682144 83.9166666666667
474.49598274231 83.9166666666667
479.001118326187 83.9166666666667
483.376789283752 83.9166666666667
487.717223501205 84.25
492.090374565125 84.5
496.531180667877 84.5833333333333
500.929527711868 84.8333333333333
504.40210852623 85
508.812271595001 85.0833333333333
513.229584407806 85.1666666666667
516.691639566421 85.1666666666667
522.004472541809 85.1666666666667
526.451979207993 85.0833333333333
529.234886360169 85.3333333333333
534.422003078461 85.5
538.762857151032 85.6666666666667
543.123995637894 85.5833333333334
547.591242837906 85.6666666666667
551.077024555206 85.6666666666667
555.448881340027 85.8333333333333
559.84564948082 85.9166666666667
564.327604055405 86
568.68019695282 86
573.091469526291 85.9166666666667
577.510302829742 86.25
581.864729070663 86.0833333333333
586.339353704452 86.25
589.860335540772 86.4166666666667
594.273645877838 86.4166666666667
598.649187231064 86.5
603.093046808243 86.5833333333333
606.649367332458 86.5
611.018252277374 86.75
615.380814313889 86.8333333333333
619.687491178513 86.8333333333333
623.997539567947 86.9166666666666
627.579664754868 86.9166666666666
631.850350904465 87
636.237300252914 86.9166666666667
640.70318031311 87
645.222400283813 86.9166666666667
649.595023059845 87.0833333333333
653.948827838898 87.0833333333333
658.272344255447 87.1666666666667
662.748894357681 87.3333333333333
667.280250406265 87.4166666666667
671.673124265671 87.6666666666667
676.199498176575 87.6666666666667
680.589657306671 87.75
685.010057830811 88.0833333333333
689.418587970734 88.0833333333333
693.800903463364 88.1666666666667
698.071454334259 88.25
};
\addplot [cyan, dashed, mark=*, mark size=1, mark options={solid}]
table {%
6.43122534751892 22.5833333333333
8.81283097267151 22.5833333333333
14.8769606113434 22.25
20.7850884914398 23.5
26.6747678279877 25.0833333333333
32.4928243637085 29.8333333333333
38.1971713066101 35.6666666666667
44.2153155326843 39.1666666666667
50.2671238899231 42.4166666666667
56.1138815879822 44.75
61.8188449859619 46.3333333333333
67.6348478794098 47.9166666666667
73.4401064395905 50
79.1734086513519 51.8333333333333
85.0431874275207 53.0833333333333
91.0016731262207 53.5833333333333
96.9193133354187 54.0833333333333
102.812760066986 54.5833333333333
108.449950504303 55.4166666666667
114.15920791626 55.5
119.913034200668 56.9166666666667
125.679399251938 57.1666666666667
131.463982200623 58.5
137.28494849205 59.9166666666667
142.969402885437 60.25
148.816598320007 61.5833333333333
154.654702997208 62.6666666666667
160.443565320969 63.5
166.320088672638 64.3333333333333
172.350032281876 64.9166666666667
178.147874116898 65.9166666666667
183.945937728882 66.4166666666667
189.901603269577 67.25
195.872093963623 67.5833333333333
201.660679483414 68.4166666666667
207.469095945358 69.25
213.324212551117 69.5833333333333
219.121399259567 70
225.06913523674 70.4166666666667
230.864901590347 71
236.839551067352 71.4166666666667
241.490161800385 72
247.243415021896 72.3333333333333
252.951020383835 72.75
258.851562023163 72.9166666666667
264.630669260025 73
270.329252719879 73
276.138264894485 73.0833333333333
282.05242061615 73.3333333333333
287.906253671646 73.4166666666667
293.757434082031 73.4166666666667
299.532314968109 73.75
305.26839466095 73.8333333333333
310.999406814575 74
316.75797533989 73.5833333333333
322.560655021667 73.9166666666667
328.342462396622 73.9166666666667
334.250786113739 74.25
339.985710048676 74.5
345.865006971359 74.5
351.572902059555 74.4166666666667
357.337659502029 74.3333333333333
363.168154478073 74.6666666666667
368.946554279327 74.8333333333333
374.646411085129 75.25
380.415846347809 75.25
386.142128181458 75.8333333333333
392.096011447907 75.5833333333333
397.952522516251 75.9166666666667
403.672364330292 76.1666666666667
409.509442234039 76.0833333333333
415.318934011459 76.1666666666667
421.182525777817 75.8333333333333
426.943093967438 76.0833333333333
432.764059495926 76.1666666666667
438.52746424675 76.0833333333333
444.290608119965 76.1666666666667
449.939287090301 76.1666666666667
455.696005773544 76.3333333333333
461.395187091827 76.4166666666667
467.240144252777 76.5
473.03946928978 76.3333333333333
477.610955810547 76.4166666666667
483.400771379471 76.6666666666667
489.190783882141 76.75
495.018560981751 77.0833333333333
500.725539827347 77.5833333333333
506.546478319168 78
512.264865779877 78.0833333333333
518.066817903519 78.3333333333333
523.796079730988 78.6666666666667
529.425601959229 78.9166666666667
535.185656023026 78.9166666666667
541.053680086136 79
546.831066322327 79
552.609439277649 79.3333333333333
558.388147497177 79.4166666666667
564.158620500565 79.1666666666667
569.952137231827 79.4166666666667
575.676239442825 79.5
581.532207107544 79.5833333333333
587.435890674591 79.6666666666667
593.310384321213 79.6666666666667
599.118904924393 80
604.837247085571 80.0833333333333
610.667086219788 80.0833333333333
616.371988153458 80.1666666666667
622.029833078384 80.0833333333333
627.837475395203 80
633.546877002716 79.75
639.242747926712 79.8333333333333
644.940567970276 79.9166666666667
650.728256988525 79.9166666666667
656.498630094528 80
662.136303091049 79.9166666666667
667.97618765831 80.0833333333334
673.921423625946 80.0833333333333
679.715238285065 80.3333333333333
685.589077615738 80.5
691.474015045166 80.5833333333333
697.179818534851 80.5
};
\addplot [blue, dashed, mark=*, mark size=1, mark options={solid}]
table {%
34.5126860618591 22.5833333333333
36.0354284286499 22.5833333333333
43.470697927475 22
50.7254657745361 25.5
58.054380273819 32.3333333333333
65.4503344535828 41.6666666666667
72.6425598144531 47.9166666666667
80.1283818721771 50.5
87.4301087379456 54.1666666666667
94.850963640213 55.9166666666667
102.237096071243 55.8333333333333
109.708394670486 56.8333333333333
117.201426267624 58.5833333333333
123.258878040314 59.4166666666667
132.179966020584 60.3333333333333
139.459653234482 61.5833333333333
147.062865924835 62.25
154.413259029388 63.4166666666667
161.909748458862 64.9166666666667
169.277540636063 65.8333333333333
176.753540182114 67.0833333333333
184.106998252869 68
191.685158252716 68.6666666666667
199.053951215744 69.6666666666667
206.490519809723 70.8333333333333
213.769937944412 71.4166666666667
219.776741790771 72.1666666666667
228.610340356827 74
235.980994510651 74.25
243.51540145874 75.4166666666667
250.999583721161 75.75
258.325998306274 76.4166666666667
265.663250255585 76.5
273.060679340363 76.9166666666667
280.364753818512 77.1666666666667
287.764552354813 77.25
295.273175621033 77.0833333333333
302.737619304657 77
310.200894451141 77.0833333333333
317.677342367172 77.25
325.155071735382 77.75
332.595004653931 77.75
339.978720092773 77.8333333333333
347.453957748413 78.1666666666667
354.817855072021 78.25
362.335796642303 78.25
369.875693130493 78.3333333333333
375.85436797142 78.5
383.320063304901 78.3333333333333
390.90283446312 78.5833333333333
398.359758424759 78.4166666666667
405.832607460022 78.5833333333333
411.875697231293 78.5
420.802732276917 78.9166666666667
428.343592739105 78.6666666666667
435.784527635574 78.75
443.212068557739 78.8333333333333
450.744319581985 78.8333333333333
458.278686571121 79.0833333333333
465.571330499649 79.3333333333333
472.93037276268 79.3333333333333
478.863255500793 79.5
487.943277740478 79.9166666666667
495.475908899307 79.8333333333333
502.8315117836 80.1666666666667
510.294158267975 80.1666666666667
517.712148189545 80.25
525.19252371788 80.6666666666667
532.52270321846 80.5833333333333
540.101017284393 81
547.548204421997 81.0833333333333
554.936985254288 81.4166666666667
562.357444190979 81.5833333333333
569.75916800499 81.4166666666667
577.187279462814 81.5833333333333
584.50418639183 81.4166666666667
591.918533992767 81.5833333333333
599.403176259995 81.3333333333333
606.829646062851 81.3333333333333
614.236388444901 81.3333333333333
621.675781059265 81.0833333333333
629.126983356476 81.4166666666667
636.640923595428 81.4166666666667
644.167530584335 81.6666666666667
651.625846099854 81.4166666666667
659.106616210938 81.5
666.46586689949 81.6666666666667
673.961219739914 81.8333333333333
681.370941019058 82.0833333333333
688.857014274597 82.1666666666667
696.343899106979 82.25
};
\addplot [ultra thick, gray, dashed, mark=*, mark size=2, mark options={solid}]
table {%
18.9488789558411 22.5833333333333
31.7672966003418 22.75
48.3186616897583 27.4166666666667
64.4016299247742 41.3333333333333
81.1966495990753 49.5
97.7700888633728 52.5
114.098706007004 54.5833333333333
130.189612293243 54.25
146.626890611649 54.5833333333333
163.143731021881 55.5
179.388672447205 56.5833333333333
196.012619447708 58.5
212.693633317947 59.3333333333333
229.751964426041 60.0833333333333
246.220080947876 61
262.863930845261 62.25
279.444280099869 63
295.982031106949 64.1666666666667
312.397791624069 65.3333333333333
328.796055650711 65.6666666666667
344.894921255112 65.75
361.428588104248 65.75
377.78165230751 66.25
381.040011978149 66.4166666666667
410.330419874191 67.6666666666667
426.958624744415 67.8333333333333
443.264547204971 68.4166666666667
459.484428548813 68.4166666666667
475.715395784378 68.8333333333333
491.818427228928 69.5
507.853320646286 70.75
511.230825233459 70.5833333333333
540.720626878738 71.3333333333333
557.041469621658 72.6666666666667
573.255614471436 72.9166666666667
589.647432184219 74.3333333333333
606.61689324379 74.5833333333333
622.908334302902 74.9166666666667
639.928363752365 75.3333333333333
656.150574111938 75.4166666666667
673.077827501297 75.75
690.339850091934 75.5
};
\addplot [semithick, red, dashed]
table {%
-30.210105381012 71.725
732.751528606415 71.725
};
\end{axis}

\end{tikzpicture}

%% file: plots/pgf_winners_croped/winners_mondial_original_target.tex
\begin{tikzpicture}

\definecolor{brown}{RGB}{165,42,42}
\definecolor{cyan}{RGB}{0,255,255}
\definecolor{darkgray176}{RGB}{176,176,176}
\definecolor{gray}{RGB}{128,128,128}
\definecolor{pink}{RGB}{255,192,203}
\definecolor{yellow}{RGB}{255,255,0}

\begin{axis}[
tick align=outside,
tick pos=left,
x grid style={darkgray176},
xmin=-2.89310127019882, xmax=157.190448291302,
xtick style={color=black},
y grid style={darkgray176},
ymin=59.7809523809524, ymax=87.647619047619,
ytick style={color=black},
yticklabel style={rotate=90.0},
tick label style={font=\huge}
]
\addplot [brown, dashed, mark=*, mark size=1, mark options={solid}]
table {%
14.698076915741 61.4285714285714
15.2983225822449 61.4285714285714
17.7558226108551 61.7142857142857
20.1364030361176 61.8095238095238
22.5962238311768 61.8095238095238
24.97872838974 61.9047619047619
27.3767763614655 61.9047619047619
29.7644388198853 61.8095238095238
32.273389339447 61.8095238095238
34.6336396694183 62.0952380952381
36.9490369319916 62.5714285714286
39.3037325382233 63.5238095238095
41.7493743896484 64.3809523809524
44.0213900566101 66.7619047619048
46.3241057872772 69.5238095238095
48.7335057735443 72.1904761904762
51.1955483436584 75.5238095238095
53.5363167285919 78.7619047619048
55.9738416671753 80.9523809523809
58.392813539505 82.5714285714286
60.7431192874908 83.1428571428572
63.2809841632843 83.4285714285714
65.7106703281403 83.9047619047619
68.1409956455231 84.4761904761905
70.4786836624146 84.6666666666667
72.768545627594 84.4761904761905
75.1873031139374 84.6666666666667
77.5726396560669 84.5714285714286
79.8704181671143 84.7619047619048
81.8400074958801 84.8571428571428
84.7948585510254 84.952380952381
87.2892577171326 85.0476190476191
89.5895990371704 85.0476190476191
92.0208185195923 85.1428571428571
93.9601778507233 85.1428571428571
96.4577213764191 85.2380952380952
98.8982103347778 85.1428571428571
101.322038936615 85.2380952380952
103.587348175049 85.3333333333333
106.000684499741 85.3333333333333
108.343754673004 85.2380952380952
110.810262680054 85.1428571428571
113.180758190155 85.2380952380952
115.65325551033 85.0476190476191
118.089742708206 84.952380952381
120.504594898224 84.8571428571428
122.856002998352 85.0476190476191
125.204787874222 84.952380952381
127.63148021698 84.8571428571428
130.059339523315 84.952380952381
132.423733377457 85.0476190476191
133.82831401825 84.952380952381
136.158626651764 84.952380952381
138.572796535492 85.0476190476191
140.864538431168 85.0476190476191
143.217523622513 84.952380952381
145.621880149841 84.952380952381
147.54322681427 85.0476190476191
149.913923311234 84.952380952381
};
\addplot [pink, dashed, mark=*, mark size=1, mark options={solid}]
table {%
4.38342370986939 61.4285714285714
5.07939982414246 61.6190476190476
8.50708723068237 61.7142857142857
11.8203262329102 61.7142857142857
15.1883370399475 61.8095238095238
18.6384330749512 61.9047619047619
22.0633954048157 62.1904761904762
25.5087637901306 62.4761904761905
29.020857667923 64.3809523809524
32.3905697822571 66.952380952381
35.7948574066162 70.3809523809524
39.21466588974 74.4761904761905
42.6438771247864 77.7142857142857
46.0749265193939 78.9523809523809
49.4663778305054 80.8571428571428
52.7439102649689 83.4285714285714
56.0476069450378 84
59.3630633831024 84.1904761904762
62.7361399173737 84.3809523809524
66.0841264247894 84
69.4412368297577 84.5714285714286
72.8874227046967 84.952380952381
76.4359043121338 85.2380952380952
79.9356790542602 85.1428571428571
83.5161196708679 85.2380952380952
87.0053724765778 85.4285714285714
90.3885465145111 85.6190476190476
93.8035314083099 85.8095238095238
97.1755195140839 85.8095238095238
100.529595947266 85.8095238095238
103.919578266144 85.7142857142857
107.344441270828 85.9047619047619
110.90401134491 85.8095238095238
114.404598283768 85.9047619047619
117.838589525223 86.1904761904762
121.310479402542 86.0952380952381
124.651069307327 86
127.942993736267 86.1904761904762
129.979146814346 86.2857142857143
133.39539937973 86.2857142857143
136.932654523849 86.1904761904762
140.304484701157 86.3809523809524
143.73984003067 86.2857142857143
147.143090057373 86.3809523809524
};
\addplot [green, dashed, mark=*, mark size=1, mark options={solid}]
table {%
14.7743490695953 62.0952380952381
16.8809562206268 61.9047619047619
27.4371603965759 61.5238095238095
36.6257665157318 61.2380952380952
43.7424241542816 61.8095238095238
48.7166043758392 63.8095238095238
51.3360242366791 65.3333333333333
54.5290847301483 70.1904761904762
57.6805480480194 74.3809523809524
60.1854357242584 77.1428571428572
62.0622883796692 79.9047619047619
65.2614570140839 82.1904761904762
67.8293035030365 83.047619047619
71.1079682350159 84
74.3532142162323 83.9047619047619
77.6000157833099 84.0952380952381
80.8087503910065 83.9047619047619
83.3527614593506 84
85.7857280731201 83.8095238095238
88.9393623352051 83.8095238095238
90.8921231746674 83.9047619047619
94.1479386806488 83.7142857142857
97.2286239624023 83.9047619047619
100.374445867538 83.9047619047619
102.784418296814 83.9047619047619
106.130931520462 83.9047619047619
108.703183507919 84
111.855281448364 84.1904761904762
115.055754137039 84.1904761904762
116.986456727982 84.0952380952381
119.53282828331 84.1904761904762
122.849647760391 84.0952380952381
126.110135984421 84.0952380952381
129.451090478897 84.0952380952381
132.156960296631 84.0952380952381
135.396771669388 84.0952380952381
138.02777929306 84.1904761904762
141.179642963409 84.2857142857143
144.423562955856 84.2857142857143
146.342431163788 84.3809523809524
149.534579706192 84.5714285714286
};
\addplot [red, dashed, mark=*, mark size=1, mark options={solid}]
table {%
4.89611015319824 61.4285714285714
5.50591926574707 61.4285714285714
8.40583052635193 61.6190476190476
11.2301819801331 61.8095238095238
14.0538328647614 61.9047619047619
16.8491012096405 61.7142857142857
19.6522707462311 62.0952380952381
22.4260688304901 62.3809523809524
25.2263686180115 62.952380952381
28.1268455505371 64.1904761904762
30.9646714687347 66.5714285714286
33.7516958236694 69.4285714285714
36.5469467639923 72.8571428571428
38.7606802463531 74.9523809523809
42.1619637012482 79.6190476190476
44.9626482486725 81.1428571428571
47.7161621570587 81.6190476190476
50.5871609687805 82.3809523809524
53.4217401027679 83.047619047619
56.2359489440918 83.047619047619
58.9948250293732 83.2380952380952
61.8181144714355 83.5238095238095
64.6322612285614 83.3333333333333
67.4561078548431 83.5238095238095
70.3708894729614 83.5238095238095
73.1669690132141 83.7142857142857
75.4113099098206 83.7142857142857
78.7237940311432 83.8095238095238
81.5368499755859 83.8095238095238
84.4678694725037 84
87.2067606925964 84
89.4838761806488 84
92.3313529968262 84
95.0961734294891 83.8095238095238
97.9719235897064 84.0952380952381
100.770885181427 84.2857142857143
103.611705446243 84.2857142857143
106.524723958969 84.1904761904762
109.359414434433 84.0952380952381
112.22438788414 84
115.024213314056 84.2857142857143
117.835305929184 84.2857142857143
120.667639017105 84.3809523809524
123.451947259903 84.3809523809524
126.368460321426 84.3809523809524
129.306394577026 84.3809523809524
131.019387340546 84.3809523809524
133.787702131271 84.4761904761905
136.674323987961 84.3809523809524
138.920525407791 84.3809523809524
142.192518997192 84.6666666666667
145.058482837677 84.6666666666667
147.244238710403 84.5714285714286
};
\addplot [yellow, dashed, mark=*, mark size=1, mark options={solid}]
table {%
10.6568693161011 61.4285714285714
15.2724944591522 61.5238095238095
23.0260461807251 61.4285714285714
30.5410176277161 61.047619047619
38.3033139228821 61.2380952380952
46.1019373893738 61.4285714285714
53.8768954277039 64.0952380952381
61.7169619083405 66.952380952381
69.4299517631531 71.0476190476191
77.2554444789887 74.0952380952381
84.908372926712 78.0952380952381
92.4887240409851 80.0952380952381
99.9627792835236 82.1904761904762
107.500917387009 81.9047619047619
115.126453638077 82.0952380952381
122.834245634079 82.1904761904762
130.545607423782 82.3809523809524
138.038800477982 82.4761904761905
145.815336990356 82.8571428571428
};
\addplot [cyan, dashed, mark=*, mark size=1, mark options={solid}]
table {%
10.7787230014801 61.4285714285714
12.2879600524902 61.5238095238095
19.9785836219788 61.5238095238095
27.7072557449341 61.8095238095238
35.4797786712646 61.5238095238095
43.2144836425781 61.5238095238095
51.0227803230286 63.5238095238095
58.843785572052 66.7619047619048
66.5212213039398 71.5238095238095
74.3864619731903 75.5238095238095
82.3098615646362 77.0476190476191
89.9760604381561 79.0476190476191
97.6927713871002 80.9523809523809
105.547055482864 81.4285714285714
113.147517633438 82.1904761904762
121.035076236725 82.6666666666667
128.707287740707 83.047619047619
136.576956796646 83.3333333333333
144.181591033936 83.2380952380952
};
\addplot [blue, dashed, mark=*, mark size=1, mark options={solid}]
table {%
22.4422422885895 61.4285714285714
23.1068072319031 61.4285714285714
26.5935059070587 61.8095238095238
30.0307149887085 61.7142857142857
33.5481232643127 61.6190476190476
37.0906091690063 61.7142857142857
40.6109576702118 62
44.0632231712341 62.1904761904762
46.6681141853333 62.5714285714286
50.1966708183289 63.1428571428571
53.7058256626129 65.6190476190476
56.4100612163544 67.2380952380952
59.8274386405945 69.9047619047619
62.6368016242981 73.2380952380952
66.0807146549225 77.047619047619
68.1518216133118 78.7619047619047
71.6110654354095 81.4285714285714
75.0902753829956 84.3809523809524
78.6514372825623 84.0952380952381
82.2027966499329 84.952380952381
85.6321295261383 85.3333333333333
88.4728714466095 84.952380952381
91.1019326686859 85.1428571428571
94.5895817756653 85.0476190476191
98.1331709384918 85.1428571428571
100.856713867187 85.0476190476191
104.318062400818 85.0476190476191
107.839714050293 85.1428571428571
111.376835346222 85.3333333333333
114.910464286804 85.3333333333333
116.861245727539 85.2380952380952
120.344224405289 85.3333333333333
123.091249036789 85.4285714285714
126.566146469116 85.3333333333333
130.053457307816 85.4285714285714
133.460772848129 85.5238095238095
136.898417663574 85.4285714285714
138.85296254158 85.3333333333333
142.283022928238 85.1428571428571
145.814593076706 84.952380952381
149.406064558029 84.8571428571428
};
\addplot [ultra thick, gray, dashed, mark=*, mark size=2, mark options={solid}]
table {%
14.2887599468231 61.4285714285714
24.6740455150604 61.5238095238095
35.141822719574 61.6190476190476
45.638599729538 62.1904761904762
56.2690583229065 64.4761904761905
66.8843245983124 66.7619047619048
77.7658805847168 71.1428571428571
88.6925711154938 76.4761904761905
99.2472709655762 80.2857142857143
109.508946180344 82.5714285714286
119.790633487701 83.1428571428571
130.337510061264 83.9047619047619
140.955986309052 83.7142857142857
};
\addplot [semithick, red, dashed]
table {%
-2.89310127019883 79.5285714285714
157.190448291302 79.5285714285714
};
\end{axis}

\end{tikzpicture}

%% file: plots/pgf_winners_croped/winners_mondial_target_continent.tex
\begin{tikzpicture}

\definecolor{brown}{RGB}{165,42,42}
\definecolor{cyan}{RGB}{0,255,255}
\definecolor{darkgray176}{RGB}{176,176,176}
\definecolor{gray}{RGB}{128,128,128}
\definecolor{pink}{RGB}{255,192,203}
\definecolor{yellow}{RGB}{255,255,0}

\begin{axis}[
tick align=outside,
tick pos=left,
x grid style={darkgray176},
xmin=-2.28420758247376, xmax=156.97147269249,
xtick style={color=black},
y grid style={darkgray176},
ymin=15.328, ymax=102.272,
ytick style={color=black},
yticklabel style={rotate=90.0},
tick label style={font=\huge}
]
\addplot [brown, dashed, mark=*, mark size=1, mark options={solid}]
table {%
16.0457027435303 20.32
16.5588525772095 20.56
19.1919611930847 20.48
21.7057517051697 20.08
24.2451385974884 21.28
26.7204097747803 23.36
29.4276232242584 24.4
31.8888906478882 25.6
34.3004273414612 28.72
36.7917958259583 31.6
39.338597869873 35.92
41.8118765354156 40.72
44.3177455425262 45.6
46.7829558372498 51.2
49.1874317646027 57.92
51.6641351699829 63.52
54.1059942722321 67.84
56.4818259716034 74.32
58.9168274402618 81.04
61.3491666793823 85.68
63.8715246677399 89.6
66.4025282382965 92.24
68.7765811920166 94.24
71.1960085391998 96
73.6583811283112 97.04
75.1459679603577 97.12
77.5304139614105 97.44
79.5055543422699 97.44
82.43299036026 97.52
84.9147801876068 97.6
87.4206475257874 97.84
89.8881799697876 98
92.2671276092529 98
94.7002780437469 98
96.05814909935 97.92
98.469016456604 97.92
100.97275686264 97.92
103.442506122589 97.92
105.830712461472 97.92
108.307972145081 97.92
110.839192008972 97.92
113.392295980453 97.92
115.904011201859 97.92
118.422290372849 97.92
120.929263734818 98
123.395779895782 98
125.887817287445 98.08
128.348697280884 98.08
130.750486516953 98.08
133.152153158188 98
134.68532705307 98
137.171514463425 98.08
139.687248706818 98.08
142.158250379562 98.08
144.177545976639 98.08
147.187125682831 98.08
149.643684959412 98.08
};
\addplot [pink, dashed, mark=*, mark size=1, mark options={solid}]
table {%
4.95468697547913 20.32
5.72803592681885 20.64
9.45080542564392 20.8
12.9820202827454 22.48
16.6986379623413 25.12
20.3211130619049 28
23.9314215660095 32.48
27.6061813831329 38.88
31.2222077846527 47.6
34.7731305122376 58.56
38.3720645904541 71.76
41.9794923782349 80.8
45.5591248989105 87.44
49.2888522624969 91.6
52.917386341095 94.96
56.5868987083435 96.48
60.148024892807 97.36
63.7748331546783 97.68
67.3344802379608 97.68
70.8304974555969 97.52
74.4759945392609 97.52
78.1373114585876 97.52
81.7882863044739 97.76
85.3784690856934 97.76
89.0453692913055 97.76
92.6339940547943 97.76
96.190891456604 97.76
99.8153057098389 97.76
103.431419897079 97.76
106.965034914017 97.76
110.61224489212 97.76
114.222929048538 97.76
117.797448730469 97.76
121.42561211586 97.76
125.10311923027 97.76
128.809581613541 97.76
131.717138576508 97.76
135.325381088257 97.76
138.890767288208 97.76
142.470072603226 97.76
146.117287063599 97.76
149.732578134537 97.76
};
\addplot [green, dashed, mark=*, mark size=1, mark options={solid}]
table {%
17.2849167346954 19.28
27.7692975521088 21.92
39.4866205692291 27.76
48.7170464038849 38.72
55.2717237949371 49.12
59.2238833904266 61.04
62.6052206516266 76.8
66.1008305072784 88.4
69.4030287265778 94.16
72.8322463989258 97.12
76.3478938579559 97.68
79.6836457252502 97.68
83.0237297534943 97.6
86.3228056430817 97.6
89.6728494167328 97.6
93.1236606121063 97.6
96.5816916942596 97.6
100.016261386871 97.6
103.479615449905 97.6
106.864049577713 97.6
110.193656587601 97.6
113.680002784729 97.6
117.016611385345 97.6
120.305480003357 97.6
123.784982395172 97.6
127.125139760971 97.6
130.68228597641 97.6
134.192456531525 97.6
137.629749822617 97.6
141.12303519249 97.6
144.553245353699 97.6
147.840136003494 97.6
};
\addplot [red, dashed, mark=*, mark size=1, mark options={solid}]
table {%
5.43065748214722 20.32
6.83576049804688 20.08
10.3112974643707 20.96
13.6377604484558 22.72
17.0658325195312 24.8
20.5052733421326 29.28
23.8604970932007 33.28
27.1789124965668 39.6
30.4543503761291 46.08
33.7647562503815 52.08
37.2089438438416 60.88
40.5671126842499 70.16
43.9990500450134 77.52
47.4502488613129 84.72
50.9401096343994 90.72
54.395011472702 94.24
57.7957909107208 95.84
61.2764418125153 96.8
64.6769004821777 97.52
68.2030175209045 97.84
71.565771150589 98
75.0689106464386 98
78.4550751209259 98.08
81.9290331840515 98.24
85.4697267055512 98.32
88.8005656242371 98.32
92.1271332263946 98.32
95.6355189800262 98.32
99.0784399032593 98.24
102.430794000626 98.32
105.775479888916 98.32
109.167886829376 98.24
112.525441360474 98.24
115.950814437866 98.24
119.400351238251 98.24
122.746820449829 98.16
126.166454982758 98
129.583533620834 97.84
132.354285001755 98.24
134.356541109085 98.16
139.117497873306 97.92
142.504558801651 97.92
145.877117586136 97.92
147.933216571808 97.92
};
\addplot [yellow, dashed, mark=*, mark size=1, mark options={solid}]
table {%
16.3265501499176 20.32
26.0961140632629 21.68
38.1845696926117 27.44
50.1657813072205 40.48
62.4123040676117 52.8
74.3807766914368 64.8
86.4856204986572 78.24
98.5762437820435 88.56
110.811388587952 92.8
122.898545265198 96
134.841468095779 97.44
147.019548559189 97.6
};
\addplot [cyan, dashed, mark=*, mark size=1, mark options={solid}]
table {%
16.3265501499176 20.32
26.0961140632629 21.68
38.1845696926117 27.44
50.1657813072205 40.48
62.4123040676117 52.8
74.3807766914368 64.8
86.4856204986572 78.24
98.5762437820435 88.56
110.811388587952 92.8
122.898545265198 96
134.841468095779 97.44
147.019548559189 97.6
};
\addplot [blue, dashed, mark=*, mark size=1, mark options={solid}]
table {%
26.3901982784271 20.32
27.1678026199341 20.24
31.3112878799438 20.32
35.3226613998413 22.56
39.3966839313507 25.6
43.3823961734772 30.72
47.3890415668488 37.68
50.5601827144623 42.72
53.6481104850769 47.76
57.6903801441193 54.32
61.7450476646423 63.44
65.6967744827271 71.6
69.6584103107452 80.4
72.8342720031738 86.64
75.9957699298859 90.88
79.9195597648621 93.6
83.1636034488678 95.68
87.224892950058 97.04
91.3109025478363 97.52
95.4486396312714 97.6
99.5260828018188 97.6
102.705163049698 97.6
106.75606341362 97.6
110.878221940994 97.6
113.965167045593 97.6
117.190441513062 97.6
121.142134475708 97.6
125.288538646698 97.6
128.423936796188 97.6
132.470950269699 97.6
136.458878040314 97.6
140.388941049576 97.6
143.502544975281 97.6
147.371001243591 97.6
};
\addplot [ultra thick, gray, dashed, mark=*, mark size=2, mark options={solid}]
table {%
16.3265501499176 20.32
26.0961140632629 21.68
38.1845696926117 27.44
50.1657813072205 40.48
62.4123040676117 52.8
74.3807766914368 64.8
86.4856204986572 78.24
98.5762437820435 88.56
110.811388587952 92.8
122.898545265198 96
134.841468095779 97.44
147.019548559189 97.6
};
\addplot [semithick, red, dashed]
table {%
-2.28420758247375 92.72
156.97147269249 92.72
};
\end{axis}

\end{tikzpicture}

%% file: plots/pgf_winners_croped/winners_mondial_target_GDP_g8e3.tex
\begin{tikzpicture}

\definecolor{brown}{RGB}{165,42,42}
\definecolor{cyan}{RGB}{0,255,255}
\definecolor{darkgray176}{RGB}{176,176,176}
\definecolor{gray}{RGB}{128,128,128}
\definecolor{pink}{RGB}{255,192,203}
\definecolor{yellow}{RGB}{255,255,0}

\begin{axis}[
tick align=outside,
tick pos=left,
x grid style={darkgray176},
xmin=6.74914447546005, xmax=104.133963901997,
xtick style={color=black},
y grid style={darkgray176},
ymin=45.3, ymax=88.2,
ytick style={color=black},
yticklabel style={rotate=90.0},
tick label style={font=\huge}
]
\addplot [brown, dashed, mark=*, mark size=1, mark options={solid}]
table {%
20.1620898246765 47.8333333333333
22.958353805542 47.25
27.5522701740265 49
31.9825407981873 55.25
36.5511929988861 66.0833333333333
41.0834279537201 72.5
45.7600992679596 75.9166666666667
50.3011875629425 79.3333333333333
54.7337348461151 81.1666666666667
59.2172321796417 82.4166666666667
63.7688039302826 82.9166666666667
68.2338673114777 83
72.7304885387421 83.6666666666667
77.211897945404 84.1666666666667
81.7004569530487 84.5
86.2228432655334 84.6666666666667
90.8062027454376 85.0833333333333
95.3158818721771 85.0833333333333
98.9223797798157 85.1666666666667
};
\addplot [pink, dashed, mark=*, mark size=1, mark options={solid}]
table {%
13.51762342453 47.8333333333333
18.9815626621246 49.25
28.1259922981262 59.0833333333333
37.1728739261627 71.9166666666667
46.1924201488495 76.8333333333333
54.9828945159912 80.1666666666667
64.0090117454529 81.5
72.8825735569 83.3333333333333
82.1637422084808 85.25
91.3576706886292 85.9166666666667
};
\addplot [green, dashed, mark=*, mark size=1, mark options={solid}]
table {%
16.6713171482086 53.5
26.547794675827 56.5833333333333
33.4697833061218 66.25
38.9301347255707 75.8333333333333
44.4797628879547 81.4166666666667
49.9156012058258 82.8333333333333
55.4991955280304 83.9166666666667
60.9445429325104 84.25
66.357009267807 84.75
71.8968421936035 85.0833333333333
77.3633548736572 85.0833333333333
82.8441922187805 85.1666666666667
88.4280145645142 85.4166666666667
93.9500205993652 85.75
99.4238965511322 86.25
};
\addplot [red, dashed, mark=*, mark size=1, mark options={solid}]
table {%
11.1757271766663 47.8333333333333
17.5256522655487 48.75
23.9249308109283 54.6666666666667
30.2630819320679 70.1666666666667
36.5364116668701 78.6666666666667
42.797188615799 81.3333333333333
48.9969023227692 81.75
55.4302190303803 83
61.7837102413177 84.0833333333333
68.0301961898804 84.3333333333333
74.3938725948334 84.9166666666667
80.7219450473785 85.5833333333333
87.1043240070343 86
93.3989288330078 85.9166666666667
99.7073812007904 86.0833333333333
};
\addplot [yellow, dashed, mark=*, mark size=1, mark options={solid}]
table {%
15.5687262058258 47.8333333333333
27.1517079353333 48.8333333333333
38.7435367584228 65.4166666666667
50.2853672981262 77.5833333333333
61.6614611148834 81.0833333333333
73.487383556366 83.3333333333333
84.7370540618897 84.5833333333333
96.2248195171356 85
};
\addplot [cyan, dashed, mark=*, mark size=1, mark options={solid}]
table {%
14.091329574585 47.8333333333333
19.853950214386 48.4166666666667
29.5747041702271 54.5
39.2458741664886 69.9166666666667
49.1407929897308 77.8333333333333
59.1485999107361 80.3333333333333
69.1158522605896 83.4166666666667
79.0260347366333 84.25
88.8669773101807 85.3333333333333
98.6963552474976 85.8333333333333
};
\addplot [blue, dashed, mark=*, mark size=1, mark options={solid}]
table {%
16.0969293117523 47.8333333333333
27.9411900043488 50.5833333333333
39.8930039405823 69.3333333333333
52.1040120124817 78.6666666666667
64.237485408783 82.5833333333333
76.4451250076294 84.1666666666667
88.5167634487152 85.3333333333333
};
\addplot [ultra thick, gray, dashed, mark=*, mark size=2, mark options={solid}]
table {%
16.0969293117523 47.8333333333333
27.9411900043488 50.5833333333333
39.8930039405823 69.3333333333333
52.1040120124817 78.6666666666667
64.237485408783 82.5833333333333
76.4451250076294 84.1666666666667
88.5167634487152 85.3333333333333
};
\addplot [semithick, red, dashed]
table {%
6.74914447546005 81.0666666666667
104.133963901997 81.0666666666667
};
\end{axis}

\end{tikzpicture}

%% file: plots/pgf_winners/legend_winners.tex
\begin{tikzpicture}

\definecolor{cyan}{RGB}{0,255,255}
\definecolor{darkgray176}{RGB}{176,176,176}
\definecolor{gray}{RGB}{128,128,128}
\definecolor{green}{RGB}{0,128,0}
\definecolor{lightgray204}{RGB}{204,204,204}
\definecolor{pink}{RGB}{255,192,203}
\definecolor{yellow}{RGB}{255,255,0}

\begin{axis}[
hide axis,
legend cell align={left},
legend style={
  draw opacity=1,
  text opacity=1,
  font=\huge,
  draw=lightgray204
}
,
tick align=outside,
tick pos=left,
x grid style={darkgray176},
xmin=-7.62244165658951, xmax=267.859734280109,
xtick style={color=black},
y grid style={darkgray176},
ymin=58.5708333333333, ymax=84.5125,
ytick style={color=black},
yticklabel style={rotate=90.0}
]
\addlegendimage{brown, dashed, mark=*, mark size=1, mark options={solid}}
\addlegendentry{ 1ep}

\addlegendimage{pink, dashed, mark=*, mark size=1, mark options={solid}}
\addlegendentry{ MI}

\addlegendimage{green, dashed, mark=*, mark size=1, mark options={solid}}

\addlegendentry{O.S. Elimination}
\addlegendimage{red, dashed, mark=*, mark size=1, mark options={solid}}

\addlegendentry{ k\_var}
\addlegendimage{yellow, dashed, mark=*, mark size=1, mark options={solid}}

\addlegendentry{ len}
\addlegendimage{cyan, dashed, mark=*, mark size=1, mark options={solid}}

\addlegendentry{ random}
\addlegendimage{blue, dashed, mark=*, mark size=1, mark options={solid}}

\addlegendentry{ sampling}
\addlegendimage{ultra thick, gray, dashed, mark=*, mark size=2, mark options={solid}}

\addlegendentry{all schemes}
\addlegendimage{semithick, red, dashed}
\addlegendentry{$a^*$}
\end{axis}

\end{tikzpicture}

%% file: plots/pgf_winners_croped/winners_mondial_target_infant_mortality_g40.tex
\begin{tikzpicture}

\definecolor{brown}{RGB}{165,42,42}
\definecolor{cyan}{RGB}{0,255,255}
\definecolor{darkgray176}{RGB}{176,176,176}
\definecolor{gray}{RGB}{128,128,128}
\definecolor{pink}{RGB}{255,192,203}
\definecolor{yellow}{RGB}{255,255,0}

\begin{axis}[
tick align=outside,
tick pos=left,
x grid style={darkgray176},
xmin=-2.35323530912399, xmax=157.206400983334,
xtick style={color=black},
y grid style={darkgray176},
ymin=58.5708333333333, ymax=84.5125,
ytick style={color=black},
yticklabel style={rotate=90.0},
tick label style={font=\huge}
]
\addplot [brown, dashed, mark=*, mark size=1, mark options={solid}]
table {%
16.9311137199402 60.1666666666667
17.5473432540894 60.1666666666667
21.0419445991516 60.1666666666667
24.196910905838 60
27.6179584980011 60.5
30.9072341918945 61.25
34.3182094097137 61.8333333333333
36.9760506629944 63.4166666666667
38.8984811306 65
41.4755094051361 66.25
44.6926735877991 67.9166666666667
47.9265513420105 71.0833333333333
51.3511735916138 74
54.6504467964172 76.5833333333333
57.1821298122406 78.4166666666667
59.2172025203705 79.1666666666667
62.584023475647 80.25
65.167192029953 81
68.5038274765015 81.4166666666667
71.781912279129 81.5833333333333
75.0944358348846 81.5
77.7154151439667 81.5
81.0031798839569 81.5
82.8999683380127 81.5
86.1096590518951 81.6666666666667
88.723354434967 81.5833333333333
91.9731688976288 81.5833333333333
95.1721151351929 81.5833333333333
97.7529295444488 81.4166666666667
100.999123096466 81.4166666666667
102.935169267654 81.4166666666667
106.117507886887 81.4166666666667
109.422446537018 81.5833333333333
111.930992412567 81.5
115.087894201279 81.5
117.605214881897 81.5833333333333
120.786475658417 81.75
124.056494808197 81.75
125.955122375488 81.8333333333333
129.112744617462 81.75
132.436576652527 81.75
135.877964496613 81.9166666666667
137.868985128403 82
141.359275770187 82.0833333333333
144.51007566452 82.1666666666667
146.427950954437 82.1666666666667
149.703204917908 82.0833333333333
};
\addplot [pink, dashed, mark=*, mark size=1, mark options={solid}]
table {%
4.89947543144226 60.1666666666667
7.82394027709961 59.9166666666667
11.4347266674042 60.6666666666667
15.0307157039642 60.75
18.6346575260162 60.8333333333333
22.2457696437836 61.5
25.8876314163208 63.3333333333333
29.504533290863 65.4166666666667
33.0884732723236 67.9166666666667
36.7062056064606 71.5
40.2990034580231 74.5833333333333
43.8098185062408 76.8333333333333
47.3854161262512 78.25
50.8895783424377 78.9166666666667
54.554872751236 79.25
58.1781638622284 79.3333333333333
61.7881991386414 79.3333333333333
65.504129743576 79.3333333333333
69.0712954521179 79.4166666666667
72.614026927948 79.4166666666667
76.215799665451 79.1666666666667
79.826846408844 79
83.3981410503388 79.1666666666667
86.9649083614349 79.0833333333333
90.4987591266632 79
94.1034366130829 78.9166666666667
97.5771925449371 79
101.195344638824 79.0833333333333
104.782456064224 79.25
108.343536281586 79.25
111.859301710129 79.3333333333333
115.524883651733 79.25
119.135671186447 79.1666666666667
122.712864637375 79.25
126.205410337448 79.1666666666667
129.756781721115 79.3333333333333
133.340742778778 79.25
136.90165605545 79.25
140.485796546936 79.3333333333333
144.083957481384 79.4166666666667
147.614830350876 79.5833333333333
};
\addplot [green, dashed, mark=*, mark size=1, mark options={solid}]
table {%
16.5046418190002 60.1666666666667
29.0877882003784 60.4166666666667
38.6320990085602 61.5
44.3746705532074 64.3333333333333
49.6878423213959 69.75
55.32249584198 75.0833333333333
60.8728928089142 79.25
66.3684053421021 81
71.7497013568878 81.5
77.1544960975647 81.5833333333333
82.7464256286621 81.6666666666667
88.1459220409393 81.4166666666667
93.5720414638519 81.4166666666667
99.055349445343 81
104.573083543777 81
109.994650936127 80.8333333333333
115.601798009872 80.75
121.113242578506 80.8333333333333
126.686620950699 80.75
132.101806259155 80.75
137.531160449982 80.8333333333333
142.911577510834 80.9166666666667
148.179688358307 80.75
};
\addplot [red, dashed, mark=*, mark size=1, mark options={solid}]
table {%
5.23998208045959 60.1666666666667
5.91258769035339 60.25
9.36176590919495 60.1666666666667
12.6594692230225 60.1666666666667
15.9485745429993 61.1666666666667
18.5350309371948 61.5833333333333
22.5097081661224 62
25.7764139175415 63.9166666666667
29.1999598026276 65.8333333333333
32.5980147838593 68.5
35.3771455287933 71.8333333333333
39.5041326999664 74.75
42.9203379631042 77.25
46.1678334236145 79.9166666666667
49.5356514930725 81.3333333333333
52.8748941898346 82.1666666666667
56.2104884624481 82.25
59.456716966629 82
62.7116649150848 82.1666666666667
66.0603768348694 82.0833333333333
68.0979384422302 82.1666666666667
72.115732049942 82.25
75.5818877696991 82.25
78.9533288002014 82.0833333333333
82.4278914928436 82.0833333333333
85.819783782959 82.1666666666667
89.1357683181763 82.25
92.4349367141724 82.1666666666667
95.7310111522675 82.0833333333333
99.0262969493866 82.0833333333333
102.392909240723 82.0833333333333
105.700537109375 82.0833333333333
109.159113454819 82.0833333333333
112.439932250977 82.1666666666667
115.867274332047 82.0833333333333
119.223071575165 81.6666666666667
122.597341871262 81.5833333333333
126.033034467697 81.5833333333333
129.425231981277 81.6666666666667
129.425231981277 81.6666666666667
132.778923940659 81.5833333333333
136.826019573212 81.5833333333333
139.9821808815 81.6666666666667
143.233758974075 81.75
146.53347864151 81.75
149.953690242767 81.75
};
\addplot [yellow, dashed, mark=*, mark size=1, mark options={solid}]
table {%
13.7118401527405 60.1666666666667
15.5053819656372 60.3333333333333
25.0091074466705 59.75
34.5313910007477 60.75
43.9841123104095 63.5
53.7577264785767 68.5
63.5738349914551 73.6666666666667
73.3087660312653 78
82.9421723365784 80.0833333333333
92.5964328765869 81.25
98.3820118427277 81.75
107.940003108978 81.75
117.668544626236 82.0833333333333
127.353395462036 81.5833333333333
137.07154583931 81.25
142.922348594666 81.25
};
\addplot [cyan, dashed, mark=*, mark size=1, mark options={solid}]
table {%
9.53164157867432 60.1666666666667
12.0962210178375 60.4166666666667
18.848298740387 60.0833333333333
25.4440735816956 60.8333333333333
32.0769765377045 62.9166666666667
38.6793867588043 65.0833333333333
45.2872186660767 69.0833333333333
51.9085192203522 73
58.5197752952576 74.8333333333333
65.1686027050018 76.9166666666667
72.0227611541748 78.0833333333333
78.5361822128296 78.8333333333333
85.0855539798737 78.75
91.7537227153778 78.4166666666667
95.6750045776367 78.4166666666667
102.45827870369 78.5
109.050459384918 78.5
115.834570884705 78.8333333333333
122.429646015167 78.9166666666667
129.067892360687 78.8333333333333
135.681305932999 78.5
142.388558244705 78.5
148.990706825256 78.4166666666667
};
\addplot [blue, dashed, mark=*, mark size=1, mark options={solid}]
table {%
23.7089948177338 60.1666666666667
24.4252932548523 59.8333333333333
28.3206722259521 60.25
31.9707099437714 60.25
35.6526594638824 60.5
39.351797246933 61.75
43.2221056938171 63.25
46.2062661647797 64.9166666666667
49.9202579498291 68.0833333333333
53.7464544296265 71.4166666666667
57.5396409511566 75.4166666666667
61.2918229103088 77.25
65.0196395874023 78.9166666666667
68.7134562015533 79.8333333333333
70.8874855995178 80.3333333333333
74.6422992229462 80.9166666666667
78.4474907875061 81.5
82.2463329792023 81.4166666666667
85.9304672718048 81.0833333333333
89.6146360397339 81.25
92.4602353096008 81
96.2872835636139 80.9166666666667
100.050167131424 80.9166666666667
102.985034990311 80.9166666666667
106.690055751801 81
110.482716321945 81
114.31133723259 81
116.483137655258 81.0833333333333
120.188880109787 81.1666666666667
123.984361314774 81.3333333333333
127.71478562355 81.4166666666667
131.358526325226 81.4166666666667
134.996283578873 81.5
138.739706754684 81.5833333333333
140.9629986763 81.6666666666667
144.735026836395 81.75
148.532567882538 81.6666666666667
};
\addplot [ultra thick, gray, dashed, mark=*, mark size=2, mark options={solid}]
table {%
16.5984631061554 60.1666666666667
28.1814455032349 60
40.315362071991 61.1666666666667
52.0262512683868 67.6666666666667
64.1670199871063 74.75
76.1628958702087 78.4166666666667
88.6180658817291 81
100.585505867004 83
112.604430341721 83.3333333333333
124.459515380859 83.3333333333333
136.194041395187 82.5833333333333
148.042492675781 82.9166666666667
};
\addplot [semithick, red, dashed]
table {%
-2.353235309124 78.7708333333333
157.206400983334 78.7708333333333
};
\end{axis}

\end{tikzpicture}

%% file: plots/pgf_winners_croped/winners_mondial_target_Inflation_g6.tex
\begin{tikzpicture}

\definecolor{brown}{RGB}{165,42,42}
\definecolor{cyan}{RGB}{0,255,255}
\definecolor{darkgray176}{RGB}{176,176,176}
\definecolor{gray}{RGB}{128,128,128}
\definecolor{pink}{RGB}{255,192,203}
\definecolor{yellow}{RGB}{255,255,0}

\begin{axis}[
tick align=outside,
tick pos=left,
x grid style={darkgray176},
xmin=-1.87144912242889, xmax=155.966447768211,
xtick style={color=black},
y grid style={darkgray176},
ymin=44.3125, ymax=77.7708333333333,
ytick style={color=black},
yticklabel style={rotate=90.0},
tick label style={font=\huge}
]
\addplot [brown, dashed, mark=*, mark size=1, mark options={solid}]
table {%
17.0484127521515 49.5
17.678400182724 49.5833333333333
21.1552993297577 48.5833333333333
24.58697514534 49.25
27.9035368442535 50.4166666666667
31.2801049232483 52.75
34.595677614212 55.3333333333333
37.2445429801941 57.8333333333333
39.24343085289 58.9166666666667
41.9623391151428 60.1666666666667
45.4937628269196 62.6666666666667
48.7932200431824 63.6666666666667
52.1999771595001 65.8333333333333
55.5300091266632 66.8333333333333
58.1123378276825 68
61.5818552970886 68.9166666666667
63.5145415306091 69
66.8969338417053 70.4166666666667
69.497583770752 70.6666666666667
72.7670844078064 71.8333333333333
75.5706390857697 71.9166666666667
78.4234882831574 72.0833333333333
82.6190340995789 72.6666666666667
86.0198445796967 72.8333333333333
87.9534823417664 73.0833333333333
91.3659903526306 73.25
94.7125775337219 73.25
96.5501856803894 73.5
100.049115419388 73.75
103.339069461823 73.9166666666667
106.650040245056 74.4166666666667
110.158252620697 74.5
112.207782030106 74.5833333333333
114.859696769714 74.4166666666667
118.379635429382 74.6666666666667
121.77370634079 74.6666666666667
124.303344345093 74.6666666666667
127.556022977829 74.5833333333333
130.993736696243 74.8333333333333
134.498653268814 74.9166666666667
135.810876750946 74.9166666666667
139.267338657379 74.9166666666667
142.048422002792 75.1666666666667
146.069470310211 75
148.791997909546 75.1666666666667
};
\addplot [pink, dashed, mark=*, mark size=1, mark options={solid}]
table {%
8.03201594352722 49.5
11.286297416687 50.25
16.8694111824036 53.8333333333333
22.4089104175568 57.8333333333333
27.8659633159637 60
33.2957955360413 61.6666666666667
38.8361692905426 63
44.1853216171265 63.9166666666667
49.6256896972656 66.8333333333333
55.1190024375916 68.0833333333333
60.5993985176086 69.4166666666667
66.0761244773865 69.9166666666667
71.4706439971924 69.9166666666667
73.6158396244049 70.4166666666667
82.3617008686066 71.0833333333333
87.9967645645142 70.8333333333333
93.4778291702271 70.6666666666667
98.9969967365265 70.25
104.460803461075 70
109.871499538422 70.1666666666667
115.334781599045 69.9166666666667
120.779654359817 69.75
126.26823797226 70
131.788343524933 70
137.335003995895 69.9166666666666
142.755234956741 70.1666666666667
144.877880239487 70.3333333333333
};
\addplot [green, dashed, mark=*, mark size=1, mark options={solid}]
table {%
16.8090879917145 45.8333333333333
19.3154926776886 46.25
31.2869473934174 49.5
41.5973016262054 56.9166666666667
49.4720220088959 60.4166666666667
55.0922466754913 64.3333333333333
58.5668571472168 67.0833333333333
62.0043425559998 69
65.4812146663666 70.75
68.7608491897583 71.3333333333333
72.0314132213593 72.9166666666667
75.4525308132172 73.25
78.8532417297363 74.1666666666667
82.3257374763489 75
85.7591735363007 75.3333333333333
89.0638714790344 75.5
92.2953790187836 75.6666666666667
95.0668358802795 76.0833333333333
98.5202805519104 76.25
101.750113010406 76.1666666666667
105.213312482834 75.8333333333333
108.722652482986 76
112.104950284958 76
115.383316516876 76
118.756726789474 76.0833333333333
121.998012876511 76.1666666666667
125.485525751114 76.0833333333333
128.953210020065 76.1666666666667
132.299895763397 76.0833333333333
135.614204883575 75.8333333333333
139.066542387009 75.6666666666667
142.306833744049 75.6666666666667
145.799219274521 75.9166666666667
148.425099468231 75.9166666666667
};
\addplot [red, dashed, mark=*, mark size=1, mark options={solid}]
table {%
5.30300073623657 49.5
5.99628658294678 49.3333333333333
9.39646363258362 49.3333333333333
12.7230198383331 50.4166666666667
15.950887966156 51.3333333333333
19.2629868030548 53.9166666666667
22.7026097297668 56.4166666666667
26.0595188617706 58.4166666666667
29.3302160739899 60.75
32.6518438339233 62.9166666666667
35.9916444778442 64.25
39.435019826889 66.3333333333333
42.7560686588287 68.25
46.0728822231293 70.1666666666667
49.3758982181549 70.5833333333333
52.8352555751801 71.4166666666667
56.0687204837799 71.8333333333333
59.3745008468628 71.9166666666667
62.7489873409271 72.1666666666667
66.0633807182312 72.3333333333333
69.4712064266205 72.5833333333333
72.7828215122223 73.1666666666667
76.128889465332 73.6666666666667
79.3918385028839 73.9166666666667
82.8301580429077 73.9166666666667
86.1322573184967 73.8333333333333
89.4371671199799 73.6666666666667
92.7493282318115 74
96.1304249286652 74
99.4978145122528 74.0833333333333
102.881577301025 74.3333333333333
106.1473965168 74.3333333333333
109.427856636047 74.4166666666667
112.701091480255 74.6666666666667
115.9862637043 74.8333333333333
119.340169143677 74.9166666666667
122.662667036057 74.9166666666667
126.155270862579 75
129.471201086044 75
132.095922327042 74.9166666666667
135.423255586624 74.75
138.736149024963 74.9166666666667
142.037830352783 74.9166666666667
145.356272125244 75.1666666666667
148.674462366104 75.3333333333333
};
\addplot [yellow, dashed, mark=*, mark size=1, mark options={solid}]
table {%
14.4962070465088 49.5
16.5527016639709 50
26.7339658737183 52.9166666666667
37.098665189743 57.4166666666667
47.4647608280182 64.4166666666667
57.9842515945435 68.3333333333333
68.2724118232727 70
78.4825753688812 71.6666666666667
89.0674791812897 71.9166666666667
99.4664085388184 72.6666666666667
109.720269298553 73.0833333333333
120.210882568359 72.6666666666667
130.604937028885 73
141.006517267227 72.75
};
\addplot [cyan, dashed, mark=*, mark size=1, mark options={solid}]
table {%
10.6587616920471 49.5
12.2859515190125 49.8333333333333
20.5026773452759 51.9166666666667
28.4421624183655 55.5833333333333
36.1806312084198 59.3333333333333
44.1086795330048 62.75
52.1526405334473 65.6666666666667
60.2531779766083 68.5
68.2391900062561 68.6666666666667
76.1782367706299 69.75
83.8999354362488 70.1666666666667
91.9557888507843 70.8333333333333
99.9097412586212 70.8333333333333
107.879487800598 71.25
116.15420999527 71.5
124.140296792984 71.8333333333333
132.097077322006 71.9166666666667
135.273308372498 71.6666666666667
143.189809656143 72.4166666666667
};
\addplot [blue, dashed, mark=*, mark size=1, mark options={solid}]
table {%
16.3372807025909 49.5
28.363972234726 51
40.0075924396515 56
52.3726006984711 62.1666666666667
64.3468844890594 66.6666666666667
76.40316157341 70
88.4611751079559 72.25
100.633054685593 72.3333333333333
112.51878900528 72.5
124.411121320724 72.5833333333333
136.452869653702 72.5
148.293310546875 72
};
\addplot [ultra thick, gray, dashed, mark=*, mark size=2, mark options={solid}]
table {%
16.3372807025909 49.5
28.363972234726 51
40.0075924396515 56
52.3726006984711 62.1666666666667
64.3468844890594 66.6666666666667
76.40316157341 70
88.4611751079559 72.25
100.633054685593 72.3333333333333
112.51878900528 72.5
124.411121320724 72.5833333333333
136.452869653702 72.5
148.293310546875 72
};
\addplot [semithick, red, dashed]
table {%
-1.8714491224289 68.4
155.966447768211 68.4
};
\end{axis}

\end{tikzpicture}

%% file: plots/pgf_winners_croped/winners_mutagenesis.tex
\begin{tikzpicture}

\definecolor{brown}{RGB}{165,42,42}
\definecolor{cyan}{RGB}{0,255,255}
\definecolor{darkgray176}{RGB}{176,176,176}
\definecolor{gray}{RGB}{128,128,128}
\definecolor{pink}{RGB}{255,192,203}
\definecolor{yellow}{RGB}{255,255,0}

\begin{axis}[
tick align=outside,
tick pos=left,
x grid style={darkgray176},
xmin=-3.9757631278038, xmax=156.844786307812,
xtick style={color=black},
y grid style={darkgray176},
ymin=66.1421052631579, ymax=90.8052631578947,
ytick style={color=black},
yticklabel style={rotate=90.0},
tick label style={font=\huge}
]
\addplot [brown, dashed, mark=*, mark size=1, mark options={solid}]
table {%
17.8435227394104 68.3157894736842
20.8619500637054 68.4210526315789
25.7261642932892 68
30.7112481117249 71.5789473684211
35.6877571582794 79.3684210526316
40.570689868927 84.1052631578947
45.6081754207611 86.4210526315789
50.4778173446655 86
55.4049418926239 86.1052631578947
60.3781939029694 86
65.3354783058167 86
70.3442277908325 86
75.4156158924103 86
80.4047678947449 86
85.1199974536896 86
90.004772233963 86
95.0359864234924 85.8947368421052
99.8649735927582 85.8947368421052
104.919453144073 86
109.914064741135 86
114.927567100525 86
119.848970746994 86.1052631578947
124.77239408493 86.1052631578947
129.73989315033 86.2105263157895
134.827441263199 86.2105263157895
139.784896659851 86.2105263157895
144.650943374634 86.2105263157895
149.534761333466 86.2105263157895
};
\addplot [pink, dashed, mark=*, mark size=1, mark options={solid}]
table {%
16.2485374450684 68.3157894736842
26.7728936195374 67.2631578947368
39.9227077960968 78.4210526315789
53.1387888431549 83.3684210526316
66.2761762142181 85.0526315789473
79.5224836826324 85.8947368421053
92.7710679531097 86.7368421052632
105.786900615692 87.2631578947368
118.914722919464 88.5263157894737
132.124685573578 89.4736842105263
145.290973472595 89.6842105263158
};
\addplot [green, dashed, mark=*, mark size=1, mark options={solid}]
table {%
16.2464845657349 68.4210526315789
18.8792244911194 68.3157894736842
30.7518632411957 69.7894736842105
36.2902421474457 73.3684210526316
41.8483083248138 85.0526315789474
47.2733262062073 85.8947368421052
52.7428381919861 85.7894736842105
58.2820515155792 85.7894736842105
63.925176525116 85.8947368421052
69.3708214759827 85.7894736842105
74.8056708812714 85.7894736842105
80.3383496284485 85.7894736842105
85.7936067581177 85.7894736842105
91.3210443973541 85.6842105263158
96.8219730854034 85.578947368421
101.373147821426 85.6842105263158
107.875172758102 85.8947368421053
113.484284973145 85.8947368421053
119.000917291641 85.8947368421053
124.503808546066 85.8947368421053
130.061574316025 86.1052631578947
135.668582391739 86.1052631578947
141.122426176071 86.1052631578947
146.762848329544 86.1052631578947
};
\addplot [red, dashed, mark=*, mark size=1, mark options={solid}]
table {%
3.33426184654236 68.3157894736842
3.88238430023193 68.2105263157895
6.39587259292603 68
8.89383835792541 67.8947368421052
11.4023426532745 69.0526315789474
13.8319504261017 73.0526315789474
16.1827096939087 78.4210526315789
18.5930037021637 82.7368421052632
21.0595015048981 85.578947368421
23.536563539505 86
26.0210731983185 86.2105263157895
28.4517224311829 86.2105263157895
30.9464452266693 86.1052631578947
33.3789485454559 86
35.8100881099701 86
38.3434246063232 85.8947368421053
40.2786363124847 86
42.7166600704193 85.8947368421053
44.6684238433838 86
47.1017467021942 86.1052631578947
49.6161536693573 86.1052631578947
52.0862219810486 86.1052631578947
54.5723384857178 86.1052631578947
57.0482008457184 86.1052631578947
59.4973804473877 86.1052631578947
61.968210029602 86.1052631578947
64.3882242679596 86.1052631578947
66.8639847278595 86.1052631578947
69.3820900440216 86.1052631578947
71.8847407817841 86.1052631578947
74.3994134902954 86.1052631578947
76.8149490833282 86.1052631578947
78.853942155838 86.1052631578947
81.2729687213898 86.1052631578947
83.6311048030853 86
86.1132632732391 86
88.1197888374329 86
90.7120322227478 85.8947368421052
93.199295425415 85.8947368421052
95.6464065074921 85.8947368421052
98.183863067627 85.8947368421052
100.647318744659 85.8947368421052
103.181474256516 85.8947368421052
105.669119787216 85.8947368421052
108.10304069519 85.8947368421052
110.509303092957 85.8947368421052
113.058263778687 85.8947368421052
114.963843870163 85.8947368421052
117.464518880844 85.7894736842105
119.935337400436 85.8947368421052
122.418598842621 85.8947368421052
124.862433719635 85.8947368421052
127.238402366638 85.8947368421052
129.712123060226 85.8947368421052
131.715967655182 85.8947368421052
134.308370256424 85.8947368421052
136.754314279556 85.8947368421052
139.213976478577 85.7894736842105
141.638230991364 85.7894736842105
144.141536664963 85.7894736842105
146.613330316544 85.7894736842105
149.133766174316 85.7894736842105
};
\addplot [yellow, dashed, mark=*, mark size=1, mark options={solid}]
table {%
4.04078688621521 68.3157894736842
4.75276780128479 68.3157894736842
8.46524605751038 68.4210526315789
12.0108994483948 68.2105263157895
15.584344959259 67.2631578947368
19.1459563732147 68.1052631578947
22.6501215457916 74.7368421052632
26.2488862991333 81.4736842105263
29.7696659088135 84.4210526315789
33.3124840259552 85.4736842105263
36.861393737793 86.1052631578947
40.426593208313 85.8947368421053
44.0291914463043 86
47.5588137626648 85.7894736842105
51.1096040725708 85.8947368421053
54.6879282951355 85.7894736842105
58.2784836292267 86.2105263157895
61.8440107345581 86.6315789473684
65.4201407432556 87.0526315789474
68.945779132843 87.1578947368421
72.4809255123138 86.8421052631579
76.1223058223724 87.0526315789474
79.5997244358063 86.9473684210526
82.437482881546 87.0526315789474
86.7279588699341 87.0526315789474
90.4053864002228 87.0526315789474
94.0856507778168 87.3684210526316
97.6924868106842 87.2631578947368
101.297245407104 87.3684210526316
104.131272792816 87.2631578947368
107.718541431427 86.9473684210526
110.604003286362 86.8421052631579
114.857481718063 86.8421052631579
118.363312911987 86.9473684210526
121.972118425369 87.1578947368421
125.567019701004 87.2631578947368
129.101438617706 87.4736842105263
132.684987688065 87.3684210526316
136.195730543137 87.2631578947368
139.800640058517 87.2631578947368
143.347228479385 87.2631578947368
146.916685962677 87.2631578947368
};
\addplot [cyan, dashed, mark=*, mark size=1, mark options={solid}]
table {%
5.52530031204224 68.3157894736842
6.56632614135742 68.1052631578947
11.7525291919708 68
16.7976236820221 68.1052631578947
21.7729344367981 69.4736842105263
26.8304495811462 75.4736842105263
31.9155782222748 82.3157894736842
36.94086561203 84.6315789473684
42.0730394363403 85.3684210526316
47.3129106521606 85.3684210526316
52.4020988941193 86
57.579217004776 86.6315789473684
62.6462994575501 87.0526315789474
67.6945642948151 86.9473684210526
72.6737117767334 87.0526315789474
77.6249974727631 87.2631578947368
82.6714374542236 87.4736842105263
87.896532869339 87.6842105263158
92.9235910892487 87.578947368421
98.0616010665894 87.6842105263158
102.961553430557 87.8947368421053
108.133250951767 88.1052631578947
113.129405021667 88.3157894736842
118.213858127594 88.3157894736842
123.304725170135 88.2105263157895
128.28464140892 88.4210526315789
133.199989557266 88.4210526315789
138.249615240097 88.4210526315789
143.274423885345 88.7368421052632
148.333761692047 88.7368421052632
};
\addplot [blue, dashed, mark=*, mark size=1, mark options={solid}]
table {%
24.828892660141 68.3157894736842
25.8876240253448 68.3157894736842
30.8873748779297 68.5263157894737
35.8114795207977 69.0526315789474
40.7850778102875 73.7894736842105
45.8545447349548 82.1052631578947
50.7823562145233 86.5263157894737
55.7062371253967 86.3157894736842
60.6690821170807 86.3157894736842
65.5658272266388 86.2105263157895
70.6683065414429 86.2105263157895
75.5946927070618 86.2105263157895
80.5545369148254 86.1052631578947
85.4467895030975 86.1052631578947
90.4287147045136 86.1052631578947
95.3941514492035 86
100.424858474731 85.8947368421053
105.387567138672 85.8947368421053
110.382889795303 86
115.245321893692 86
120.163519716263 86
125.126777601242 86.2105263157895
129.933956384659 86.2105263157895
135.095337581635 86.3157894736842
140.090336561203 86.3157894736842
145.045413684845 86.3157894736842
};
\addplot [ultra thick, gray, dashed, mark=*, mark size=2, mark options={solid}]
table {%
16.2485374450684 68.3157894736842
26.7728936195374 67.2631578947368
39.9227077960968 78.4210526315789
53.1387888431549 83.3684210526316
66.2761762142181 85.0526315789473
79.5224836826324 85.8947368421053
92.7710679531097 86.7368421052632
105.786900615692 87.2631578947368
118.914722919464 88.5263157894737
132.124685573578 89.4736842105263
145.290973472595 89.6842105263158
};
\addplot [semithick, red, dashed]
table {%
-3.9757631278038 85.2
156.844786307812 85.2
};
\end{axis}

\end{tikzpicture}

%% file: plots/pgf_winners_croped/winners_hepatitis.tex
\begin{tikzpicture}

\definecolor{brown}{RGB}{165,42,42}
\definecolor{cyan}{RGB}{0,255,255}
\definecolor{darkgray176}{RGB}{176,176,176}
\definecolor{gray}{RGB}{128,128,128}
\definecolor{pink}{RGB}{255,192,203}
\definecolor{yellow}{RGB}{255,255,0}

\begin{axis}[
tick align=outside,
tick pos=left,
x grid style={darkgray176},
xmin=-2.56601620912552, xmax=114.655741655827,
xtick style={color=black},
y grid style={darkgray176},
ymin=55.34, ymax=84.82,
ytick style={color=black},
yticklabel style={rotate=90.0},
tick label style={font=\huge}
]
\addplot [brown, dashed, mark=*, mark size=1, mark options={solid}]
table {%
17.6276167869568 57.28
24.0328241348267 58.08
30.6071630477905 64.84
36.9731293678284 73.24
43.3918805122375 76.8
49.6762267112732 78.48
56.0587526798248 79.68
62.5581286430359 79.36
68.8544138908386 80
75.2558041572571 80
81.6834528923035 80.24
88.0334574222565 80.84
94.3848372936249 80.84
100.573726081848 81.12
107.025639295578 81.64
};
\addplot [pink, dashed, mark=*, mark size=1, mark options={solid}]
table {%
8.54640083312988 57.28
10.1865703105927 57.16
19.1313881874084 59.64
27.8905177593231 66.48
36.679985332489 74.92
45.5086735725403 77.84
54.4403559207916 79.12
63.1706827640533 79.68
71.7446753501892 80.6
80.2960828781128 80.88
89.1698167800903 81.52
97.7015100002289 81.96
106.265043020248 82.08
};
\addplot [green, dashed, mark=*, mark size=1, mark options={solid}]
table {%
8.40771589279175 57.08
15.5008587837219 58.16
23.4411792755127 62.84
28.9206766605377 72.2
34.1117887020111 76.16
39.5853971004486 77.92
44.9175058364868 78.48
50.3198857307434 78.8
55.6768052577972 79.08
61.0231036186218 79
66.4487933158875 79.36
71.7385346889496 79.36
77.0594984531403 79.36
79.2349946022034 79.4
87.7665338993072 79.6
93.161488866806 79.64
98.5341997623444 79.92
103.923720026016 80.12
109.327479934692 80.24
};
\addplot [red, dashed, mark=*, mark size=1, mark options={solid}]
table {%
2.76224551200867 57.28
5.46926102638245 56.88
7.93621716499329 56.68
10.3843283176422 58.52
12.8692348480225 61.68
15.3687616348267 67.76
17.8511716365814 73.68
20.3605574131012 76.32
22.9009051322937 77.28
25.3300085067749 78.48
27.8582464694977 79.4
30.4030945777893 79.96
32.860059928894 80.16
35.4308568000793 80.6
37.8195768356323 80.6
40.2623320102692 80.64
42.7870958328247 80.8
45.3761504173279 80.64
47.9606598854065 80.68
50.5243131160736 80.84
53.0452958106995 80.92
55.5166260242462 81.08
57.9397608757019 81.36
60.3353009223938 81.4
62.7926379680634 81.56
65.382465171814 81.48
67.9280975818634 81.68
70.3826522827148 81.56
72.7879244327545 81.48
72.7879244327545 81.48
77.6806331634521 81.52
80.2285726547241 81.6
82.6799172878265 81.72
85.3347559928894 81.64
87.8873690128326 81.64
90.325442647934 81.72
92.77712225914 81.76
95.2588743209839 81.68
97.8167747020721 81.8
100.384043073654 81.8
102.88360042572 81.8
105.338499689102 81.76
107.828908777237 81.92
};
\addplot [yellow, dashed, mark=*, mark size=1, mark options={solid}]
table {%
7.6194821357727 57.28
11.8258083820343 57.72
22.4960074424744 63.76
32.9443266868591 71.16
43.6395750999451 76.2
54.3510541439056 78.16
64.7576148509979 79.12
75.3845237731934 80.12
86.0281160354614 81.36
96.9171225547791 82.36
107.420509910583 83.04
};
\addplot [cyan, dashed, mark=*, mark size=1, mark options={solid}]
table {%
7.00538191795349 57.28
14.5443735122681 58.08
23.8886922836304 65.44
33.5698641300201 71.72
43.0742495059967 75.24
52.594345998764 77.88
62.2020195007324 80
72.0795325756073 80.76
81.5922877788544 81.56
91.3791663646698 81.96
101.035784769058 82.76
};
\addplot [blue, dashed, mark=*, mark size=1, mark options={solid}]
table {%
8.21237759590149 57.28
20.2504953861237 59
32.1498851299286 72.04
44.2220183849335 77.04
56.102493429184 79.96
68.0249538421631 81.68
79.8677460670471 82.72
91.6624037265778 83.36
103.409123373032 83.48
};
\addplot [ultra thick, gray, dashed, mark=*, mark size=2, mark options={solid}]
table {%
8.21237759590149 57.28
20.2504953861237 59
32.1498851299286 72.04
44.2220183849335 77.04
56.102493429184 79.96
68.0249538421631 81.68
79.8677460670471 82.72
91.6624037265778 83.36
103.409123373032 83.48
};
\addplot [semithick, red, dashed]
table {%
-2.56601620912552 79.306
114.655741655827 79.306
};
\end{axis}

\end{tikzpicture}

%% file: plots/dynamic/GenesDynamic.tex
\begin{tikzpicture}[xscale=0.9]
\begin{axis}[
    title={Genes},
    width=0.275\textwidth, 
    height=3.3cm, 
    xlabel={Ratio of new data (\%)},
    ylabel={Accuracy (\%)},
    xmin=5, xmax=95,
    ymin=0, ymax=105,
    xtick={20,40,60,80},
    ytick={0,20,40,60,80,100},
    xtick pos=left,
    ytick pos=left,
    y label style={at={(axis description cs:0.12,0.5)},anchor=south},
    legend style={at={(0,0)},anchor=south west, font=\small},
    ymajorgrids=true,
    grid style=dashed,
]

\addplot+[
    color=gray,
    mark=square,
    mark options={solid,fill=gray},
    ]
    coordinates {
    (10,96)(20,94.50)(30,92.95)(40,92)(50,89)(60,83.44)(70,75.37)(80,64.7)(90,50.75)
    };

\addplot[
    color=red,
    mark=o,
    ]
    coordinates {
    (10,99.49)(20,96.97)(30,96.71)(40,96.22)(50,95.26)(60,94.11)(70,92.59)(80,88.87)(90,76.66)
    };

\addplot+[
    color=blue,
    mark=triangle,
    mark options={solid,fill=blue},
    ]
    coordinates {
    (10,99.53)(20,98.60)(30,98.45)(40,98.49)(50,98.37)(60,97.87)(70,96.11)(80,94.13)(90,78.45)
    };

\addplot[
    color=black,
    ]
    coordinates {
    (5,42.5)(95,42.5)
    };
    
\legend{N2V, FW, FW-KV}
    
\end{axis}
\end{tikzpicture}

%% file: plots/dynamic/WorldDynamic.tex
\begin{tikzpicture}
\begin{axis}[
    title={World},
    width=0.275\textwidth, 
    height=3.3cm, 
    xlabel={Ratio of new data (\%)},
    xmin=5, xmax=95,
    ymin=0, ymax=105,
    xtick={20,40,60,80},
    ytick={0,20,40,60,80,100},
    xtick pos=left,
    ytick pos=left,
    y label style={at={(axis description cs:0.12,0.5)},anchor=south},
    legend style={at={(1.05,0.5)},anchor=west},
    ymajorgrids=true,
    grid style=dashed,
]

\addplot[
    color=gray,
    mark=square,
    ]
    coordinates {
    (10,94.58)(20,92.08)(30,89.58)(40,90.31)(50,88.08)(60,88.96)(70,83.69)(80,76.46)(90,24)
    };
    
\addplot[
    color=red,
    mark=o,
    ]
    coordinates {
    (10,77.08)(20,75.83)(30,74.44)(40,73.44)(50,69.17)(60,65.9)(70,61.01)(80,51.35)(90,44.17)
    };

\addplot[
    color=blue,
    mark=triangle,
    ]
    coordinates {
    (10,80.00)(20,84.17)(30,81.67)(40,77.71)(50,76.17)(60,72.50)(70,65.12)(80,57.60)(90,38.70)
    };
    
\addplot[
    color=black,
    ]
    coordinates {
    (5,24.2)(95,24.2)
    };
    
    
\end{axis}
\end{tikzpicture}

%% file: plots/dynamic/MondialDynamic.tex
\begin{tikzpicture}
\begin{axis}[
    title={Mondial-Religion},
    width=0.275\textwidth, 
    height=3.3cm, 
    xlabel={Ratio of new data (\%)},
    xmin=5, xmax=95,
    ymin=0, ymax=105,
    xtick={20,40,60,80},
    ytick={0,20,40,60,80,100},
    xtick pos=left,
    ytick pos=left,
    y label style={at={(axis description cs:0.12,0.5)},anchor=south},
    legend style={at={(1.05,0.5)},anchor=west},
    ymajorgrids=true,
    grid style=dashed,
]

\addplot[
    color=gray,
    mark=square,
    ]
    coordinates {
    (10,76.67)(20,80.95)(30,84.35)(40,81.08)(50,80.49)(60,79.11)(70,77.03)(80,76.06)(90,61.40)
    };
    
\addplot[
    color=red,
    mark=o,
    ]
    coordinates {
    (10,80.47)(20,84.28)(30,87.1)(40,83.37)(50,82.43)(60,82.50)(70,82.48)(80,82.12)(90,78.76)
    };

\addplot[
    color=blue,
    mark=triangle,
    ]
    coordinates {
    (10,83.81)(20,82.93)(30,81.94)(40,84.63)(50,82.55)(60,82.60)(70,80.84)(80,79.76)(90,70.11)
    };
    
\addplot[
    color=black,
    ]
    coordinates {
    (5,54.8)(95,54.8)
    };
    
    
\end{axis}
\end{tikzpicture}

%% file: plots/dynamic/MutagenesisDynamic.tex
\begin{tikzpicture}
\begin{axis}[
    title={Mutagenesis},
    width=0.275\textwidth, 
    height=3.3cm, 
    xlabel={Ratio of new data (\%)},
    xmin=5, xmax=95,
    ymin=0, ymax=105,
    xtick={20,40,60,80},
    ytick={0,20,40,60,80,100},
    xtick pos=left,
    ytick pos=left,
    y label style={at={(axis description cs:0.12,0.5)},anchor=south},
    legend style={at={(1.05,0.5)},anchor=west},
    ymajorgrids=true,
    grid style=dashed,
]

\addplot[
    color=gray,
    mark=square,
    ]
    coordinates {
    (10,87.89)(20,87.63)(30,87.02)(40,87.63)(50,85.53)(60,80.97)(70,78.79)(80,70.99)(90,68.18)
    };
    
\addplot[
    color=red,
    mark=o,
    ]
    coordinates {
    (10,89.47)(20,87.11)(30,87.54)(40,89.34)(50,89.04)(60,85.4)(70,84.77)(80,84.30)(90,81.59)
    };

\addplot[
    color=blue,
    mark=triangle,
    ]
    coordinates {
    (10,85.26)(20,82.63)(30,81.40)(40,81.05)(50,80.00)(60,78.76)(70,80.30)(80,76.03)(90,67.65)
    };

\addplot[
    color=black,
    ]
    coordinates {
    (5,66.4)(95,66.4)
    };
    
    
\end{axis}
\end{tikzpicture}

%% file: section_conclusions.tex
\section{Concluding Remarks}\label{sec:conclusions}
Walking through connected data items is the basis of sequence-based embeddings like \wordtovec and \nodetovec. 
Database sequences are also associated with meta-data, namely the walk scheme.
The premise of this work was the conjecture that the walk schemes have significant semantic value, as they can guide the embedding algorithm to a small subset of informative sequences, thus dramatically improving efficiency for a mild sacrifice of quality. We studied the problem of selecting walk schemes in the context of \forward. We considered different strategies of three types: \forward-less, light training, and online scheme
elimination. We conducted an experimental study that measured the benefit of each strategy, compared between them and tested how well they preserved the main strength of \forward---extensibility to newly inserted tuples in dynamic settings.
Our study has confirmed our conjecture and showed that we can considerably accelerate \forward with negligent loss of quality.
Moreover, restricting the embedding phase to the right walk schemes can even \e{improve} the quality 
on downstream classification tasks. 
The kernel-variance strategy typically outperforms the rest, and we recommend this one to be used alongside \forward.

The idea of directing the embedding algorithm to beneficial walk schemes goes beyond \forward and is applicable to every database embedding we are aware of. Our experience with \embdi~\cite{DBLP:conf/sigmod/CappuzzoPT20} has indicated that \embdi uses a large number of walk schemes with tiny pairwise differences and a heavy-tailed distribution of a number of instances. Thus we need to use ways of abstracting (clustering) walk schemes. We also plan to expand the scope of our work to data integration tasks (e.g., entity and schema matching) as done with other database embedding algorithms~\cite{DBLP:conf/sigmod/MudgalLRDPKDAR18,DBLP:conf/sigmod/CappuzzoPT20}. In the case of \forward, there is a need to devise \e{alignment} between embeddings of different databases, since \forward makes no attempt to produce similar embeddings to matching entities of different databases.